\documentclass[acmtog,nonacm]{acmart} 
\setcopyright{none}

\usepackage{mysiggraph}

\definecolor{dblue}{rgb}{0.0,0.0,0.5}
\definecolor{dgreen}{rgb}{0.0,0.5,0.0}
\definecolor{dred}{rgb}{0.6,0.0,0.0}
\definecolor{dorange}{rgb}{0.6,0.25,0.0}
\definecolor{dyellow}{rgb}{0.5,0.5,0.0}

\newcommand{\ignorethis}[1]{}

\newcommand\norm[1]{\lVert#1\rVert}

\renewcommand{\textrightarrow}{$\rightarrow$}

\begin{document}
\title{DreamPBR: Text-driven Generation of High-resolution SVBRDF with Multi-modal Guidance}

\author{Linxuan Xin}
\affiliation{%
\institution{Peking University}
\city{Shenzhen}
\country{China}}
\email{linxuanxin@stu.pku.edu.cn}

\author{Zheng Zhang}
\affiliation{%
\institution{Huawei Cloud Computing Technologies Co., Ltd.}
\city{Hangzhou}
\country{China}}
\email{zhangzheng119@huawei.com}

\author{Jinfu Wei}
\affiliation{%
\institution{Tsinghua University}
\city{Shenzhen}
\country{China}}
\email{weijf22@mails.tsinghua.edu.cn}

\author{Wei Gao}
\affiliation{%
\institution{School of Electronic and Computer Engineering, Shenzhen Graduate Schoool, Peking University}
\city{Shenzhen}
\country{China}}
\email{gaowei262@pku.edu.cn}

\author{Duan Gao}
\affiliation{%
\institution{Huawei Cloud Computing Technologies Co., Ltd.}
\city{Shenzhen}
\country{China}}
\email{gaoduan0306@gmail.com}
\authornote{corresponding author.}

\renewcommand\shortauthors{Xin, L. et al.}

% !TeX root = ../main.tex
\begin{abstract}

Prior material creation methods had limitations in producing diverse results mainly because reconstruction-based methods relied on real-world measurements and generation-based methods were trained on relatively small material datasets. 
To address these challenges, we propose DreamPBR, a novel diffusion-based generative framework designed to create spatially-varying appearance properties guided by text and multi-modal controls, providing high controllability and diversity in material generation. 
The key to achieving diverse and high-quality PBR material generation lies in integrating the capabilities of recent large-scale vision-language models trained on billions of text-image pairs, along with material priors derived from hundreds of PBR material samples.
We utilize a novel material Latent Diffusion Model (LDM) to establish the mapping between albedo maps and the corresponding latent space. The latent representation is then decoded into full SVBRDF parameter maps using a rendering-aware PBR decoder. Our method supports tileable generation through convolution with circular padding.
Furthermore, we introduce a multi-modal guidance module, which includes pixel-aligned guidance, style image guidance, and 3D shape guidance, to enhance the control capabilities of the material LDM. 
We demonstrate the effectiveness of DreamPBR in material creation, showcasing its versatility and user-friendliness on a wide range of controllable generation and editing applications.

\end{abstract}

    

% !TeX root = ../../main.tex

\newcommand{\teaserColumnFigWidth}{3.0cm}

\begin{teaserfigure}
   \centering
\includegraphics[width=\linewidth]{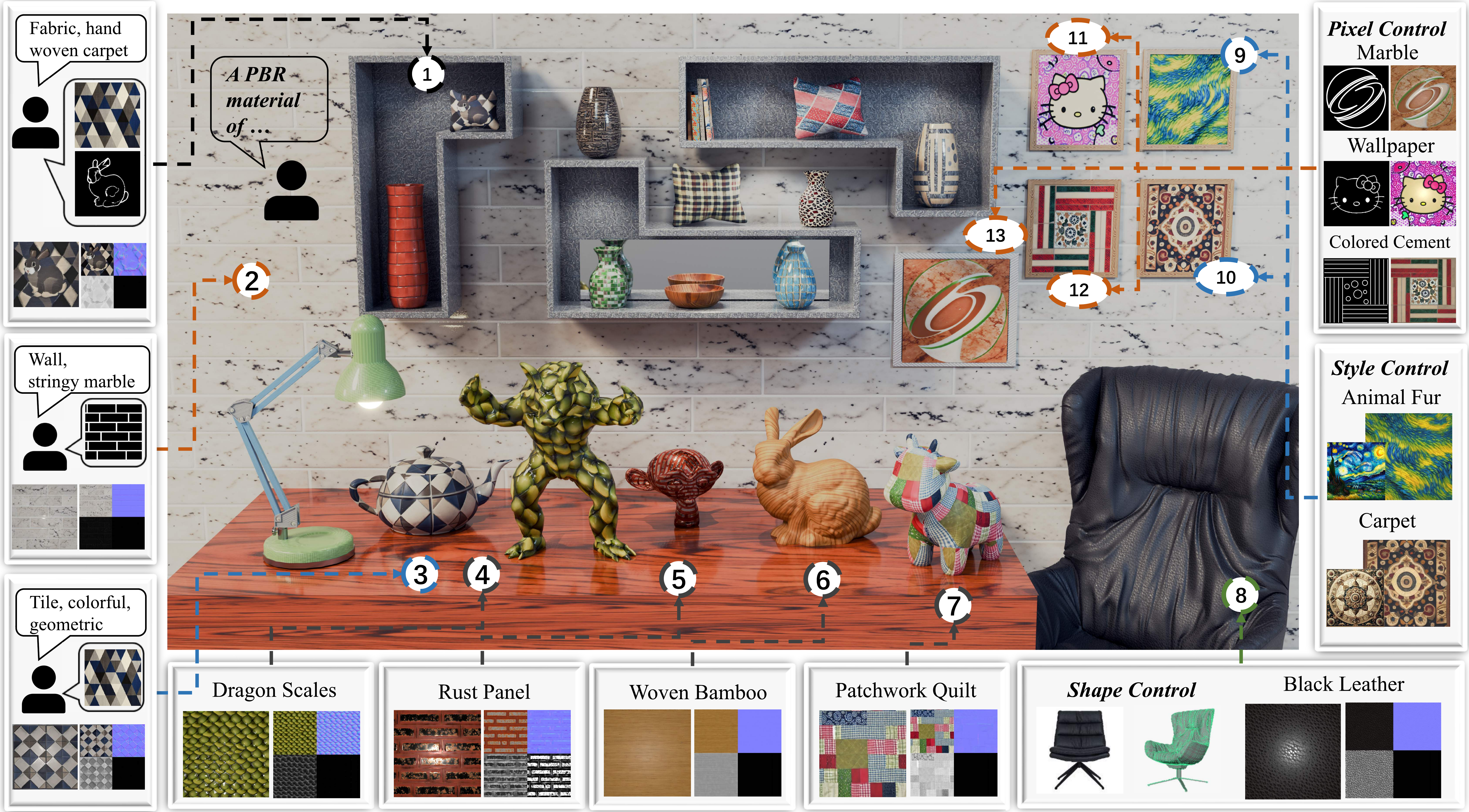}
\caption{
    DreamPBR, an innovative material generation framework, enables personalized creation with multi-modal controls. We present various controls such as text descriptions (4, 5, 6, 7), binary images (2, 9, 10), RGB images (3, 11, 12, 13), segmented geometry (8), and their combination (1) in this figure. The high-quality and tileable textures from DreamPBR show high applicability in different objects.
    }
    \label{fig:teaser}
    \Description{fig:teaser}
\end{teaserfigure}

\maketitle

%%%%%%%%%%%%%%%%%%%%%%%%%%%%%%%%%%%%%%%%%%%%%%%%%%%%%%%%%%%%%%%%%%%
% !TeX root = ../main.tex
\section{Introduction}
\label{sec:intro}

High-quality materials are crucial for achieving photorealistic rendering. Despite advancements in appearance modeling over the past few decades, material creation remains a challenging research area. The material generation approaches can be categorized into reconstruction-based methods and generation-based methods. 
Reconstruction\-/~based methods use one or many input photographs to estimate surface reflectance properties either through optimization-based inverse rendering \cite{gao2019deep,guo2020materialgan, Hu2019novel} or deep neural network inference \cite{deschaintre2018single,guo2023ultra}. However, the scope of these methods is constrained to real-world photographs, limiting their ability to create imaginative and creative materials.

Recent approaches have explored material generation \cite{guo2020materialgan,zhou2022tilegen} using Generative Adversarial Networks (GANs) \cite{goodfellow2014generative}.
However, these methods are typically train\-ed on hundreds to thousands of materials, which pales in comparison to the billions of images used in large-scale Language-Image generative models. The dataset capacity restricted their generating diversity. 
Furthermore, GAN-based methods also had training challenges including unstable training, mode collapse, and scalability issues with large datasets.
On the other hand, diffusion models \cite{ho2020denoising, rombach2021highresolution} have shown significant advancements, exhibiting advantages in scalability and diversity.
Recent advances \cite{poole2022dreamfusion,wang2023prolificdreamer} leverage 2D diffusion models before generating 3D content. However, these methods mainly focus on implicit representation or textured mesh, lacking the capability to disentangle physically based material and illumination.

To address these challenges, we introduce DreamPBR, a novel generative framework for creating high-resolution spatially-varying bidirectional reflectance distribution functions (SVBRDFs) conditioned with text inputs and a variety of multi-modal guidance.
The main advantages of our method lie in generating diversity and controllability. 
Our method can generate semantically correct and detailed materials based on various textual prompts, ranging from highly structured materials with stationary patterns to imaginative materials with flexible content, such as a Hello Kitty carpet (as shown in \Cref{fig:teaser}).

The key idea of our method is to integrate pretrained 2D text-to-image diffusion models \cite{rombach2021highresolution} with material priors to generate high-fidelity and diverse materials.
While 2D text-to-image Latent Diffusion Models (LDM) excel in generating natural images, they had challenges in producing spatially-varying physically-based material maps due to the large domain gaps between natural images and materials. Consequently, adapting pretrained 2D diffusion models into the material domain, while preserving both quality and diversity, is a non-trivial research task.
We introduce a novel material LDM which is learned by a two-stage strategy to address this challenge. 
In the initial stage, we observed albedo map is a specialized RGB image and stores spatially-varying surface reflectance by RGB pixel values.
We transfer the pretrained LDM from the text-to-image domain to the text-to-albedo domain using fine-tuning, which can be regarded as the distillation from a large source domain (natural images) to a relatively small target domain (albedo texture maps) by leveraging the target domain priors.
In the subsequent stage, we leverage a PBR decoder to reconstruct SVBRDFs from the latent space of albedo maps learned in the former stage. 
The reasons that we employ a decoder-only architecture for SVBRDFs generation are: 
1. The generated SVBRDF parameter maps exhibit strong correlations since they share a common latent representation as the starting point for decoding.
2. The decoder module does not compromise generating diversity, as we keep the denoising UNet frozen during the training of the PBR decoder.
Additionally, we introduced a highlight-aware decoder for the albedo map to further enhance regularization.

We introduce a multi-modal guidance module designed to serve as the conditioning mechanism for our material LDM, enabling a wide variety of controls for user-friendly material creation. Specifically, this guidance module includes three key components: 
\textbf{Pixel Control} allows pixel-aligned guidance from inputs like sketches or inpainting masks.
\textbf{Style Control} extracts style features from reference images and employs them to guide the generation process.
\textbf{Shape Control} enables automatic material generation for a given 3D object with segmentations with an optional 2D exemplar image for reference. 
Importantly, our framework supports the concurrent use of multiple guidances seamlessly.

We have trained our DreamPBR method on a publicly available SVBRDF dataset, comprising over 700 high-resolution (2$k$) SVBRDFs.
Thanks to the convolutional backbone of LDM, seamless tileable material generation can be supported by utilizing circular padding in all convolutional operators. 

To summarize, our main contributions are as follows:
\begin{itemize}
\item We introduce a novel generative framework for high-quality material generation under text and multi-modal guidances that combine pretrained 2D diffusion model and material domain priors efficiently;
\item We present a rendering-aware decoder module that learns the mapping from a shared latent space to SVBRDFs; 
\item Our multi-model guidance module offers rich user-friendly controllability, enabling users to manipulate the generation process effectively;
\item We propose an image-to-image editing scheme that facilitates material editing tasks such as stylization, inpainting, and seamless texture synthesis.
\end{itemize}
% !TeX root = ../main.tex

\section{Related Work}
\label{sec:related}

\subsection{Material estimation}

Material estimation approaches aim to acquire material data from real-world measurements under varying viewpoints and lighting conditions. We specifically focus on recent material estimation methods that utilize lightweight capture setups using consumer cameras. For a more comprehensive overview of general appearance modeling, please refer to surveys \cite{DONG201959,Weinmann2015,Guarnera2016brdf}.

Methods have been developed to leverage multiple images or video sequences captured by a handheld camera to estimate appearance properties. Due to the limitations of lightweight setups, most approaches still rely on regularization such as handcrafted heuristics for diffuse/specular separation \cite{Riviere2016mobile, Palma2012}, linear combinations of basis BRDFs \cite{Hui2017reflectance}, and sparsity assumption for incident lighting \cite{dong2014afm}. Another class of methods focuses on reducing the number of input images by leveraging material priors such as stationary materials ~\cite{Aittala2015twoshot,Aittala2016texture}, homogeneous or piece-wise materials ~\cite{Xu2016minimal}, and spatially sparse materials ~\cite{Zhou2016sparse}. 

In recent years, deep learning-based methods have shown significant progress in recovering SVBRDFs from single image \cite{li2017modeling,deschaintre2018single,Li2018materials,guo2021highlight,guo2023ultra,Henzler2021generative}. These methods employ deep convolutional neural network to predict plausible SVBRDFs from in-the-wild input images in a feed-forward manner. ~\citet{DADDB19} extended a single-image-based solution to multiple images by latent space max-pooling.
More recent work by \citet{gao2019deep} introduced a deep inverse rendering pipeline that enables appearance estimation from an arbitrary number of input images. 
In procedural material modeling, 
~\citet{Hu2019novel,Shi2020:ToG,Hu2022node} proposed to optimize material parameters with fixed node graphs to match input images. ~\citet{Hu2022inverse} introduced a new pipeline that eliminates the need for predefined node graphs. 
Most recently, ~\citet{sartor2023matfusion} proposed a diffusion-based model to estimate the material properties from a single photograph.

The methods mentioned above rely on captured photographs to reconstruct material and cannot produce non-real-world materials. In contrast, our approach can generate diverse and creative SVBRDFs using natural language inputs.

\subsection{Generative models}

\subsubsection*{Image generation} Generative Adversarial Networks (GANs)~\cite{goodfellow2014generative} have demonstrated remarkable capabilities in producing high-fidelity images. Subsequent research has focused on GAN improvements such as training stability \cite{kodali2017convergence,karras2018progressive}, attribute disentanglement \cite{karras2019stylebased}, conditional controllability \cite{li2021collaging, park2019SPADE}, and generation quality \cite{Karras2019stylegan2,Karras2021}. GAN can be used in various applications including text-to-image synthesis \cite{reed2016generative,reed2016learning,zhu2019dmgan}, image-to-image translation \cite{isola2018imagetoimage,zhu2020unpaired}, 
video generation \cite{tulyakov2017mocogan}, and even 3D shape generation \cite{li2019synthesizing}.

Recent advancements in text-to-image generation have been main\-ly driven by diffusion models (DMs) \cite{sohl2015deep,ho2020denoising,ramesh2022hierarchical}. Later advancements \cite{song2020denoising,song2021scorebased,liu2022pseudo} have explored efficient sampling strategies to significantly reduce the number of required sampling steps, thereby improving image generation performance. ~\citet{rombach2021highresolution} proposed Latent Diffusion Model (LDM) to perform the denoising process in learned compact latent space, enabling high-reso\-lution image synthesis and efficient image manipulation.

\subsubsection*{Controllable generation}
Integrating multi-modal controllability into a text-to-image diffusion model is crucial for creation applications. 
Recent research \cite{zavadski2023controlnetxs,zhang2023adding,ye2023ip-adapter,hu2022lora,mou2023t2i} has focused on lightweight multi-modal controllability without the requirements of extensive data and high computational power.
\citet{hu2022lora} introduces a fine-tuning strategy using low-rank matrices, enabling domain-specific adaptation. \citet{zhang2023adding} proposed ControlNet, adding spatial conditioning to diffusion models for precise generation control. \citet{ye2023ip-adapter} presented a lightweight framework enhancing diffusion models with image prompts using a decoupled cross-attention mechanism. 

\subsubsection*{Material generation}
\citet{guo2020materialgan} proposed an unconditional MaterialGAN for synthesizing SVBRDFs from random noise. The learned latent space facilitates efficient material estimation in inverse renderings. \citet{zhou2022tilegen} developed a StyleGAN2-based model, conditioned by spatial structure and material category, for tileable material synthesis. These GAN-based methods show advantages in generating high-resolution and visually compelling materials. However, their diversity is constrained by the training instability of GANs and the limited range of training datasets. 
In procedural material generation, \citet{guerrero2022matformer} first introduced a transformer-based autoregressive model. Later work by \citet{Hu_2023} proposed a multi-model node graph generation architecture for creating high-quality procedural materials, guided by both text and image inputs. 
While procedural representations are compact and resolution-independent, they are limited to stationary patterns and cannot create arbitrary styles. 

In concurrent work, \citet{vecchio2023controlmat} introduced ControlMat, a diffusion-based material generative model, capable of generating tileable materials using text and a single photograph as input. This model was trained on a synthetic material dataset comprising $126,000$ samples, derived from $8,615$ raw material graphs. 
While quite large in the material domain, this dataset is relatively small compared to the billions of text-image pairs used in text-to-image diffusion model training. This scale discrepancy leads to constrained diversity. Furthermore, this work only supports guidance of text and single photograph, limiting the scenarios range. 

In contrast, our method significantly enhances material generation diversity through the efficient integration of pretrained diffusion models with material priors. We also provide a variety of user-friendly controls for guiding the generation process, expanding the scope and flexibility of applications.

\subsection{Text-to-3D Generation}

Transitioning 2D text-to-image approaches to 3D generation presents significant challenges, mainly due to lacking large-scale labeled 3D datasets. Recent approaches \cite{poole2022dreamfusion,wang2023prolificdreamer, lin2023magic3d, tang2023dreamgaussian} have explored text-to-3D generation without the dependency of 3D data. \cite{poole2022dreamfusion} integrates Score Distillation Sampling (SDS) with text-to-image diffusion models. \citet{wang2023prolificdreamer} further improved the quality and diversity by introducing Variational Score Distillation (VSD). The development of large-scale 3D datasets \cite{deitke2023objaversexl} enabled direct learning from 3D data \cite{liu2023zero1to3,shi2023zero123plus}. However, current 3D generation methods mainly focus on geometry modeling and fail to produce high-quality, disentangled materials.

\citet{photoshape2018} introduced a neural method to assign materials from a predefined set to different parts of a 3D shape. Extending this, \citet{TMT} employs a translation network to establish the correspondence between 2D exemplar image and 3D shape. This allows for extracting material cues from 2D images and selecting optimal materials from a candidate pool using a perceptual metric. However, these methods are constrained by the variety of their predefined material assets and lack the ability to transfer complex spatial patterns from 2D exemplars to 3D shapes.
In contrast, our generative model can produce diverse materials and effectively transfer spatial structures from 2D exemplar images to 3D models, showcasing a significant advancement in material assignment.

% !TeX root = ../main.tex

\section{Method}
\label{sec:method}
\subsection{Overview}

% !TeX root = ../../main.tex

\begin{figure*}
    \centering
    \includegraphics[width=\linewidth]{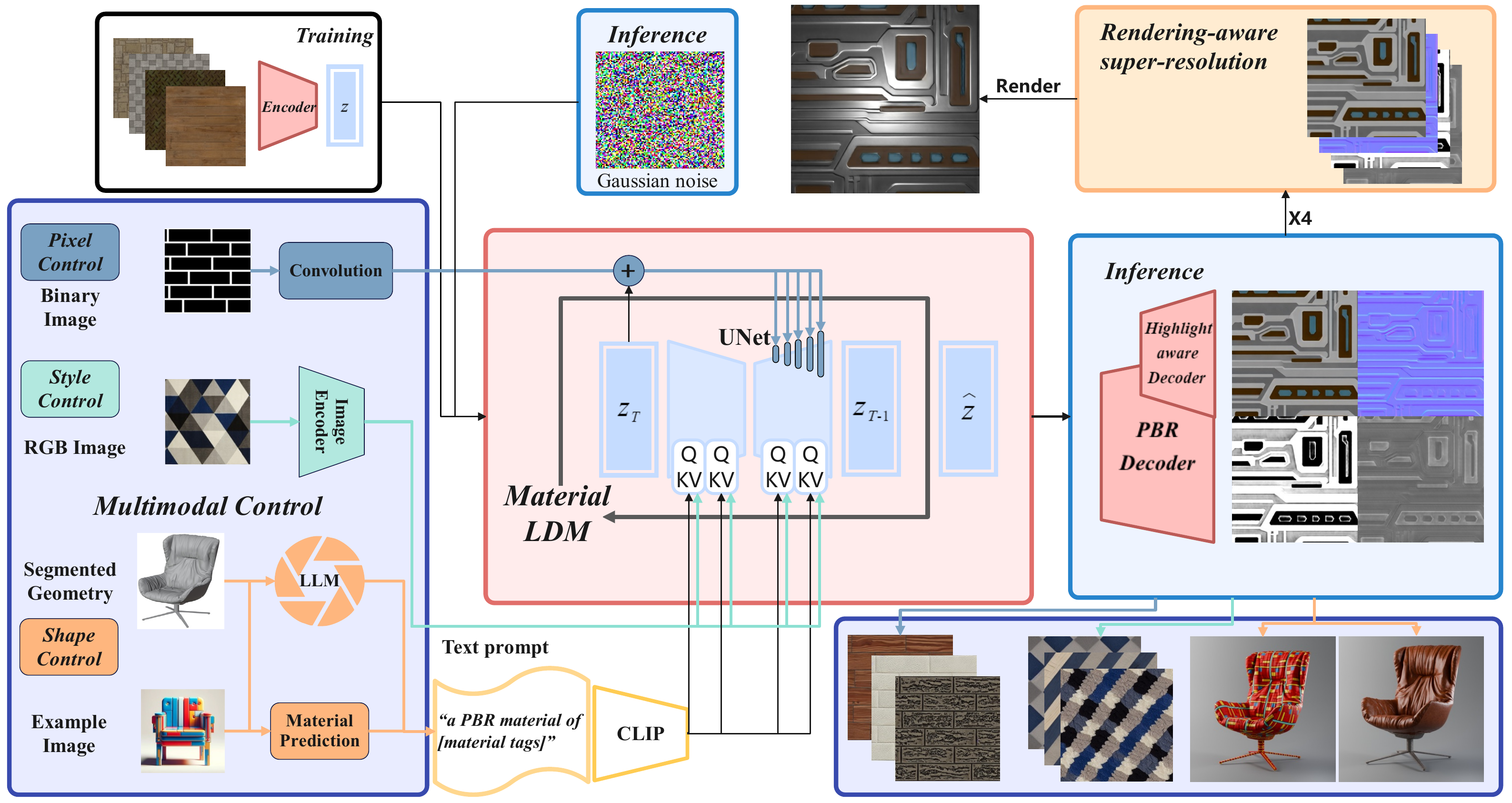}
    \caption{Overview of DreamPBR: The denoising UNet in our Material LDM is trained with only albedo textures (upper left) and a PBR Decoder with Highlight Aware Decoder is used to transform albedo textures to other physically-based textures (middle right). In the blue box on the left, we present three individual control modules: Pixel Control, Style Control, and Shape Control, whose results under controls are shown on the lower right. Besides, an additional Rendering-aware-super-resolution module is given for higher-quality textures (upper right).}
    \label{fig:overview}
    \Description{}
\end{figure*}

\paragraph*{Preliminaries} The goal of our method is to generate spatially-varying materials which are represented by the Cook-Torrance microfacet BRDF model with GGX normal distribution function ~\cite{Walter2007Microfacet}. Specifically, we use metallic-based PBR workflow and represent surface reflectance properties as albedo map $\mathcal{P}$, normal map $\mathcal{N}$, roughness map $\mathcal{R}$, and metallic map $\mathcal{M}$.

DreamPBR is a Latent Diffusion Model (LDM)-based generative framework capable of producing diverse, high-quality SVBRDF maps under text and multi-modal guidance, as illustrated in \autoref{fig:overview}.

The core generative module of our framework is the material Latent Diffusion Model (material LDM), which takes textual description $T$ as inputs to encode high-dimensional surface reflectance properties into a compact latent representation $z$. 
This representation effectively compresses complex material data and guides the SVBRDF decoder in reconstructing detailed SVBRDF maps (i.e. albe\-do, normal, roughness, and metallic) $S = \{\mathcal{P}, \mathcal{N}, \mathcal{R}, \mathcal{M}\}$. 
Our critical observation is that while pre-trained text-to-image diffusion models can capture a wide range of natural images that fulfill the diversity needs of material generation, their flexibility often leads to less plausible materials due to the absence of material priors. 
Instead of training material LDM from scratch with limited material data, we opted to fine-tune a pre-trained text-to-image diffusion model with target material data. This strategy effectively tailors the model from a broad image domain to a specific material domain, ensuring both diversity and authenticity of output.

Our text-to-material framework seamlessly integrates three types of control modules to enhance material generation capabilities.
First, we introduce the Pixel Control module $G_{P}$ that takes pixel-aligned inputs (e.g. sketches, masks), utilizing the ControlNet architecture \cite{zhang2023adding}. It adds conditional controls into diffusion models, providing spatial guidance for material generation. 
Second, we use Style Control module $G_{I}$ to extract image features from the input image prompt, which are then utilized to adapt material LDM via cross attention.  
Third, we propose a Shape control module $G_{S}$ to generate SVBRDF maps automatically for a given segmented 3D shape. This module can leverage large language models to generate text prompts corresponding to different parts of the input shape. It also supports taking a 2D photo exemplar as additional input, enabling the generation of material maps for each segmented part, guided by the segmented 2D image.
In the rest of the current section, we will dive into the key components of our framework. \Cref{ssec:method:ldm} introduce our core text-to-material module that enables tileable, diverse material generation. Next, \Cref{ssec:method:decoder} describes the SVBRDF decoder, responsible for reconstructing high-resolution SVBRDF maps from a unified latent space. Finally, \Cref{ssec:method:control} discusses the Multi-modal control module, providing image and 3D control capabilities to the diffusion model.

\subsection{Physically-based material diffusion}
\label{ssec:method:ldm}
Our material LDM transforms text features $\tau(T)$, extracted by CLIP's text encoder $\tau(\cdot)$ \cite{radford2021learning} from user prompts $T$, into a latent representation $z$ of SVBRDF maps $S$. The latent space is characterized by a Variational Autoencoder (VAE) architecture $\mathcal{E}$ ~\cite{Kingma2014,rezende2014stochastic}. 
Specifically, for an albedo map $\mathcal{P} \in \mathbb{R}^{H \times W \times C}, C=3$, the map is compressed into latent space $z = \mathcal{E}(\mathcal{P}) \in \mathbb{R}^{h \times w \times c} $. Consistent with ~\citet{rombach2021highresolution}, we adopt the parameters $c=4,h=H/8, w=W/8$.

The core component of diffusion model is the denoising U-Net module \cite{ronneberger2015unet} which is conditioned on timestep $t$. Following Denoising Diffusion Probabilistic Models (DDPM) \cite{ho2020denoising}, our model employs a deterministic forward diffusion process $q(z_t|z_{t-1})$ to transform latent vectors $z$ towards an isotropic Gaussian distribution. The U-Net network is specifically trained to reverse the diffusion process $q(z_{t-1}|z_{t})$, iteratively denoising the Gaussian noise back into latent vectors. 
Adopting the strategy proposed by \citet{rombach2021highresolution}, we incorporates the text feature $\tau(T) \in \mathbb{R}^{M \times d_\tau}$ into the intermediate layer of UNet through a cross-attention mechanism $\operatorname{Attention}(Q, K, V)=\operatorname{softmax}\left(\frac{Q K^T}{\sqrt{d}}\right) \cdot V $, where $Q=W_Q^i \cdot \varphi_i\left(z_t\right), K=W_K^i \cdot \tau(T), V=W_V^i \tau(T)$, $\varphi_i\left(z_t\right) \in \mathbb{R}^{N \times d_\epsilon^i}$ represents an intermediate representation of the UNet $\epsilon_\theta$, and $W_V^i \in \mathbb{R}^{d \times d_\epsilon^i}$, $W_Q^i \in \mathbb{R}^{d \times d_\tau^i}, W_K^i \in \mathbb{R}^{d \times d_\tau^i}$ are learnable projection matrices.

Our material LDM is fine-tuned on text-material pairs via:

\begin{equation}\label{eq:PBRLDM}
\mathcal{L}_{ldm}:=\mathbb{E}_{\mathcal{E}(\mathcal{P}),T, \epsilon \sim N(0,1), t}\left[\left\|\epsilon-\epsilon_\theta\left(z_t, t,\tau_\theta(T)\right)\right\|_2^2\right] \text {. }
\end{equation}

\subsubsection*{Seamless tileable texture synthesis} Creating tileable texture maps is critical in material generation, involving meeting two requirements: a) maintenance of consistent spatial patterns and visual appearance, and b) the ability to tile textures without visible artifacts like seams and blocks.

While zero padding is the standard practice in CNNs, we found that circular padding is particularly effective for seamless content generation. We employ circular padding in all convolutional layers of our generative model for two main reasons:

\begin{enumerate}
\item Continuity across boundaries. Unlike classic padding methods such as zero padding, which may introduce artificial edges, circular padding ensures boundary continuity. It wraps image content around both horizontal and vertical boundaries, providing a seamless transition when tiling. 
\item Pattern preservation. Circular padding mainly affects the boundary area of the image, leaving the central area and overall texture patterns unchanged. 
\end{enumerate}

Our tileable generation algorithm can serve two purposes: firstly, it can inherently produce tileable material maps without additional post-processing. Secondly, it can transform a non-tileable texture into a tileable version through an image-to-image generation pipe\-line, maintaining visual similarity with the original.

\subsection{Render-aware SVBRDF decoder}
\label{ssec:method:decoder}

The SVBRDF decoder, denoted as $\mathcal{D}=\{{\mathcal{D}_P, \mathcal{D}_S}\}$, decodes the unified latent representation $z$ into SVBRDFs $S \coloneqq \{\mathcal{P}, \mathcal{N}, \mathcal{R}, \mathcal{M}\} = \mathcal{D}(z) $. Here, $\mathcal{P},\mathcal{N} \in \mathbb{R}^{H \times W \times 3}$, $\mathcal{M},\mathcal{R} \in \mathbb{R}^{H \times W \times 1}$. In our implementation, we set $H=W=512$.
Specifically, we utilize separate decoder networks: $\mathcal{D}_P(z)$ for the albedo map $ \mathcal{P}$, and $\mathcal{D}_S(z)$ for other property maps $\{\mathcal{N}, \mathcal{R}, \mathcal{M} \}$. These decoder networks follow the decoder architecture in VAE proposed by \cite{Kingma2014,rezende2014stochastic}, and are initialized with the weights from a pre-trained VAE decoder. 

\subsubsection*{Training of PBR decoder}

The training loss function for our PBR decoder $\mathcal{D}_S$ comprises the following terms:
\begin{equation}
 \label{eq:losstotal}
 \begin{aligned}
 \mathcal{L}_{\text {PBR}} = \mathcal{L}_{map} + \mathcal{L}_{perp} + \mathcal{L}_{gan} + \mathcal{L}_{reg} +  \mathcal{L}_{render},
 \end{aligned}
\end{equation}
\begin{equation}
 \label{eq:lossrender}
 \begin{aligned}
 \mathcal{L}_{\text{render}}(x, y) = \norm{log(x+0.01),log(y+0.01)}_1,
 \end{aligned}
\end{equation}
where $\mathcal{L}_{map}$ is $L_1$ loss on the material property maps, $\mathcal{L}_{perp}$ is perceptual loss based on LPIPS \cite{zhang2018perceptual}, $\mathcal{L}_{gan}$ is the generative adversarial loss, $\mathcal{L}_{reg}$ is the Kullback-Leibler divergence penalty, and $\mathcal{L}_{render}$ is $L1$ log rendering loss applied to the rendered images. 

For the rendering loss, we adopt the sampling scheme proposed by \citet{DADDB18} to render nine images per material map. The images include three images rendered with independently distant light and view directions, and six images using near-field mirrored view and lighting directions. The rendering loss yields desirable SVBRDF reconstructions, achieved by encouraging the training process to focus on minimizing errors in crucial material parameters rather than treating them with equal importance.

\subsubsection*{Highlight-aware albedo decoder}
\label{ssec:method:highlight}
As previously mentioned in \Cref{ssec:method:ldm}, our material LDM training utilizes the standard VAE decoder to map the latent space to the albedo map.
While effective in generating plausible RGB images, this decoder tends to produce images with strong highlights, especially for shiny materials such as leather and metal.

To address this, we introduce a highlight-aware albedo decoder $\mathcal{D}_P$, which is finetuned on a synthetic shaded-to-albedo dataset, ensuring robust regularization to effectively minimize highlight artifacts in albedo maps. For each material sample in our SVBRDF dataset, we simulate various lighting conditions and viewpoints by randomly positioning point lights and cameras parallel to the material plane and then rendering SVBRDFs to reference shaded images by a physically-based renderer. 

During training, the default VAE image encoder maps the shaded images into latent space, which are then decoded back to image space by our specialized albedo decoder. The training process for this decoder follows the original VAE loss function \cite{Kingma2014}.

\subsubsection*{Material super-resolution}
High-resolution material maps are essential for achieving photorealistic renderings. However, due to the memory and performance constraints, current diffusion models typically generate images at a resolution of $512 \times 512$, which falls short of high-quality production rendering. 

We introduce a material super-resolution module comprising four super-resolution networks $SR$, each following the Real-ESRGAN architecture\cite{wang2021realesrgan}. These super-resolution networks, denoted as $SR_{P}, SR_{N}, SR_{R}, SR_{M}$, are designed to augment the resolution of different SVBRDF property maps to $2,048 \times 2,048$.

We fine-tune the Real-ESRGAN with material data, which is trai\-ned on purely synthetic data, to more effectively capture the high-frequ\-ency details of materials. We incorporate a rendering loss (similar to \Cref{eq:lossrender}) into the training of the super-resolution module to ensure that the generated details contribute to high-frequency shading effects rather than visual artifacts. We should note that special care must be taken for normal maps during augmentation involving flipping and rotation. The directions stored in a normal map must be adjusted consistently with the map orientation to ensure consistent knowledge about surface normals.

\subsection{Multi-model control}
\label{ssec:method:control}
We propose three control modules for DreamPBR: Pixel Control, Style Control, and Shape Control. These modules are designed to be decoupled, allowing for flexible combinations of multiple controls.

\subsubsection{Pixel Control}
\label{sssec:method:PixelControl}
Spatial property guidance is widely used in material creation by artists. Our Pixel Control module $G_P$ takes spatial control maps $I_P$ as input, utilizing the ControlNet architecture \cite{zhang2023adding}, to guide the generation of spatially-consistent SVBRDFs. It supports controlled generation under sketch guidance and allows for image-to-image material inpainting with a binary mask.

Our material LDM, as described in \Cref{ssec:method:ldm}, is adapted in the material domain, enabling plausible material generation controlled by pre-trained ControlNet checkpoints, which are trained with 2D supervision. However, we found that fine-tuning pre-trained ControlNet with material data significantly improves both the controllability and the quality of generated materials. Specifically, we initialize our ControlNet using the ControlNet 1.1 Scribble checkpoint and fine-tune it on our SVBRDFs dataset. To generate the sketch guidance, we employ Pidinet \cite{su2021pixel} for extracting sketches from albedo maps. 

\subsubsection{Style Control}

The Style Control module $G_I$ takes image pro\-mpt $I_S$ as input and extracts the style characteristics to guide material generation. Inspired by \citet{ye2023ip-adapter}, image prompts $I_s$ are first encoded into image features by CLIP's image encoder, and then embedded into material LDM using a decoupled cross-attention adaptation module. Multimodal material generation can be achieved by accompanying the image prompt with a text prompt.

Style Control module can effectively capture the appearance properties and structural information from the input images, to generate realistic and coherent material maps. This functionality is particularly useful in scenarios where materials need to be created based on specific exemplar images, which is a frequent requirement in the material design industry.
The interaction of the Style Control module with the Shape Control module will be detailed in \Cref{ssec:method:shape}.

\subsubsection{Shape Control}
\label{ssec:method:shape}
The Shape Control module $G_S$ takes a segmented 3D model $O_s=\{O, s\}$ ($s$ denotes the geometry segmentation) and an optional photo exemplar $I_o$ as input and automatically generates high-quality material maps for each segmentation.
When provided with only a segmented 3D model and a basic text prompt, we leverage large language models(LLMs) such as ChatGPT \cite{openai2023gpt4}, to enrich the text descriptions for each segmentation. For instance, given a 3D chair model, the language model can generate diverse text descriptions for each part like seat, leg, and armrest, each featuring varied design styles. Furthermore, integration with existing Pixel Control and Style Control modules supports enhanced SVBRDF generation, ensuring superior quality and detailed material characteristics.

Our model integrates the material transfer pipeline TMT \cite{TMT} to automatically assign diverse generated materials to 3D shapes based on an image exemplar.
The TMT pipeline involves two stages: firstly, translating color from exemplar image to the projection of 3D shape and vice versa for segmentation results; secondly, assigning materials to projected parts using a material classifier network, based on the translated image. 
Unlike \citet{TMT}, we do not rely on predefined material collections in material assignment. Instead, we use predicted material labels of TMT directly as text prompts and translated images as image prompts in the Style Control module, allowing high-quality SVBRDF generation for each part.
The proposed algorithm offers two significant advantages over traditional material transfer models: it expands material diversity beyond limited predefined material collections and transfers not only color and category information but also comprehensive material attributes including styles and spatial structures from 2D exemplar to 3D shapes, leveraging the capabilities of our Style Control module.

% !TeX root = ../main.tex

\section{Results}

\subsection{Implementation Details}

\newcommand{\PBRColumnWidth}{1.55cm} 
\newcommand{\TypeColumnWidth}{1.2cm} 
\begin{figure*}[htbp]
\centering 

\bgroup

\def\arraystretch{0.1} 
\setlength\tabcolsep{1pt} 
\begin{tabular}{
    m{0.5cm}
    m{\PBRColumnWidth}
    @{\hspace{0.5pt}}m{\PBRColumnWidth}
    m{\PBRColumnWidth}
    @{\hspace{0.5pt}}m{\PBRColumnWidth}
    m{\PBRColumnWidth}
    @{\hspace{0.5pt}}m{\PBRColumnWidth}
    m{\PBRColumnWidth}
    @{\hspace{0.5pt}}m{\PBRColumnWidth}
    m{\PBRColumnWidth}
    @{\hspace{0.5pt}}m{\PBRColumnWidth}
    }
    
\multicolumn{1}{c}{Type} & \multicolumn{1}{c}{Render} & \multicolumn{1}{c}{SVBRDF} & \multicolumn{1}{c}{Render} & \multicolumn{1}{c}{SVBRDF} & \multicolumn{1}{c}{Render} & \multicolumn{1}{c}{SVBRDF} & \multicolumn{1}{c}{Render} & \multicolumn{1}{c}{SVBRDF} & \multicolumn{1}{c}{Render} & \multicolumn{1}{c}{SVBRDF} \\

\raisebox{+0.7\height}{\rotatebox[origin=c]{90}{\small\textbf{Brick}}} &
\includegraphics[width=\PBRColumnWidth]{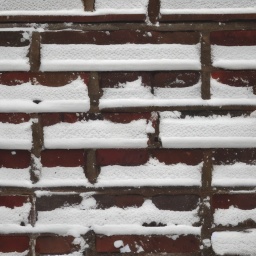} &
\includegraphics[width=\PBRColumnWidth]{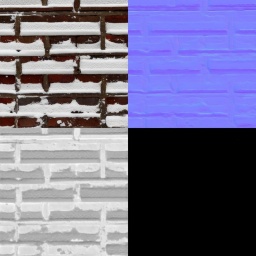} &
\includegraphics[width=\PBRColumnWidth]{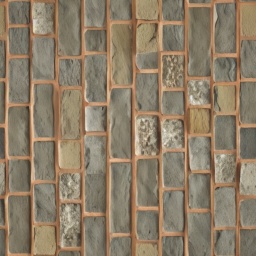} &
\includegraphics[width=\PBRColumnWidth]{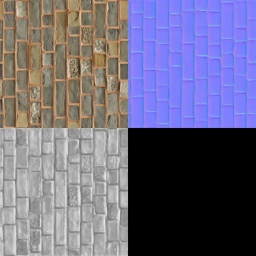} &
\includegraphics[width=\PBRColumnWidth]{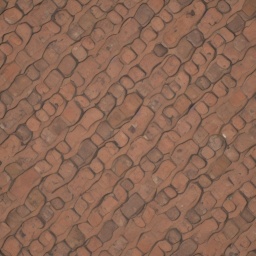} &
\includegraphics[width=\PBRColumnWidth]{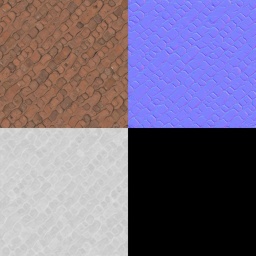} &
\includegraphics[width=\PBRColumnWidth]{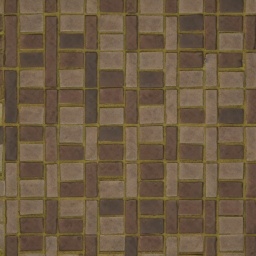} &
\includegraphics[width=\PBRColumnWidth]{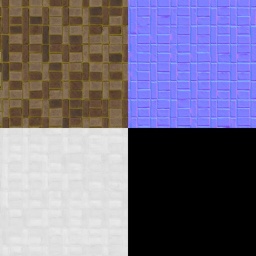} &
\includegraphics[width=\PBRColumnWidth]{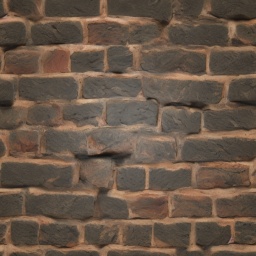} &
\includegraphics[width=\PBRColumnWidth]{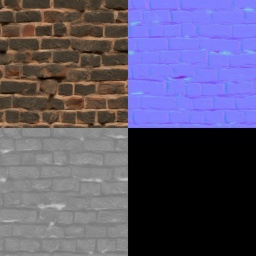} \\

&
\multicolumn{2}{c}{
  \begin{minipage}{3.1cm}
    \centering
    \scriptsize\textit{snow-covered bricks, winter, outdoor, house} 
\end{minipage} 
} &
\multicolumn{2}{c}{
  \begin{minipage}{3.1cm}
    \centering
    \scriptsize\textit{coastal barrier bricks, sea-salt resistant, outdoor, barrier} 
\end{minipage} 
} &
\multicolumn{2}{c}{
  \begin{minipage}{3.1cm}
    \centering
    \scriptsize\textit{stenciled brick floor, paving, terracotta, scratched} 
\end{minipage} 
} &
\multicolumn{2}{c}{
  \begin{minipage}{3.1cm}
    \centering
    \scriptsize\textit{narrow bricks, walls} 
\end{minipage} 
} &
\multicolumn{2}{c}{
  \begin{minipage}{3.1cm}
    \centering
    \scriptsize\textit{blackened fireplace bricks, charred} 
\end{minipage} 
} \\
\noalign{\smallskip}

\raisebox{+0.7\height}{\rotatebox[origin=c]{90}{\small\textbf{Fabric}}} &
\includegraphics[width=\PBRColumnWidth]{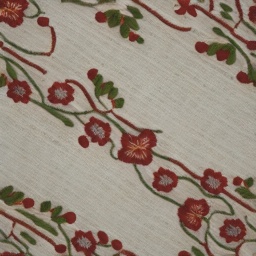} &
\includegraphics[width=\PBRColumnWidth]{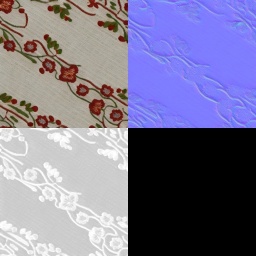} &
\includegraphics[width=\PBRColumnWidth]{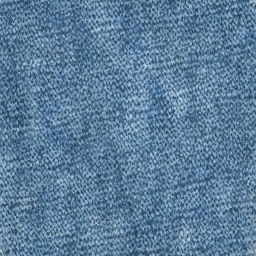} &
\includegraphics[width=\PBRColumnWidth]{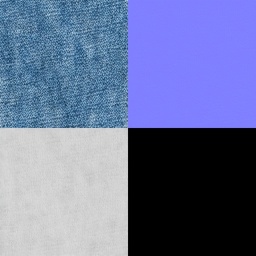} &

\includegraphics[width=\PBRColumnWidth]{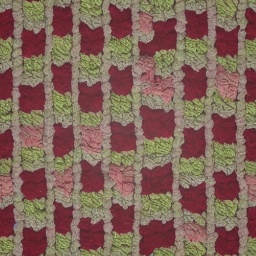} &
\includegraphics[width=\PBRColumnWidth]{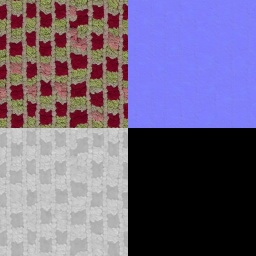} &
\includegraphics[width=\PBRColumnWidth]{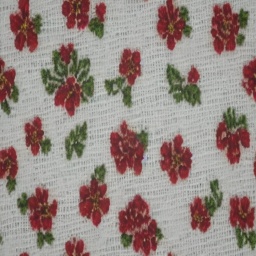} &
\includegraphics[width=\PBRColumnWidth]{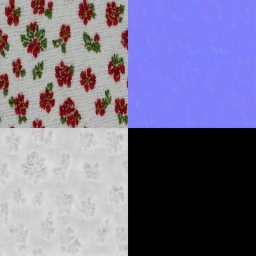} &
\includegraphics[width=\PBRColumnWidth]{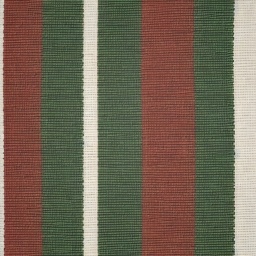} &
\includegraphics[width=\PBRColumnWidth]{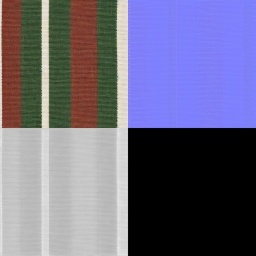} \\

&
\multicolumn{2}{c}{
  \begin{minipage}{3.1cm}
    \centering
    \scriptsize\textit{tablecloth, delicate} 
\end{minipage} 
} &
\multicolumn{2}{c}{
  \begin{minipage}{3.1cm}
    \centering
    \scriptsize\textit{denim jacket texture, clothing} 
\end{minipage} 
} &
\multicolumn{2}{c}{
  \begin{minipage}{3.1cm}
    \centering
    \scriptsize\textit{hand woven carpet, artisan, carpet} 
\end{minipage} 
} &
\multicolumn{2}{c}{
  \begin{minipage}{3.1cm}
    \centering
    \scriptsize\textit{floral cotton dress, clothing} 
\end{minipage} 
} &
\multicolumn{2}{c}{
  \begin{minipage}{3.1cm}
    \centering
    \scriptsize\textit{backpack fabric, sturdy} 
\end{minipage} 
} \\
\noalign{\smallskip}

\raisebox{+0.7\height}{\rotatebox[origin=c]{90}{\small\textbf{Ground}}} &
\includegraphics[width=\PBRColumnWidth]{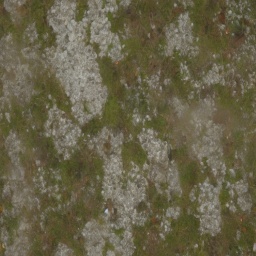} &
\includegraphics[width=\PBRColumnWidth]{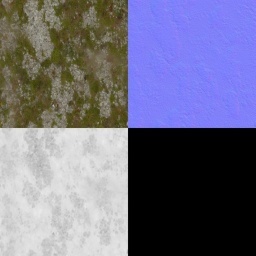} &
\includegraphics[width=\PBRColumnWidth]{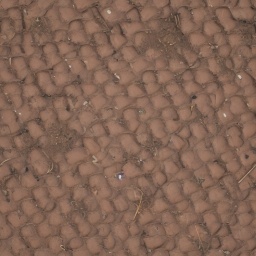} &
\includegraphics[width=\PBRColumnWidth]{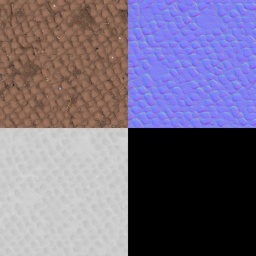} &
\includegraphics[width=\PBRColumnWidth]{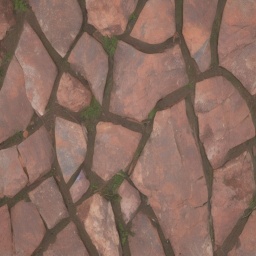} &
\includegraphics[width=\PBRColumnWidth]{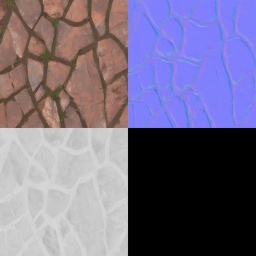} &
\includegraphics[width=\PBRColumnWidth]{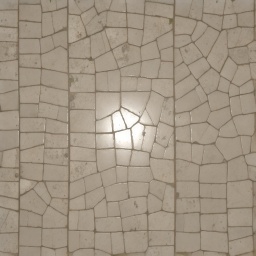} &
\includegraphics[width=\PBRColumnWidth]{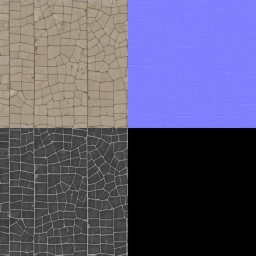} &
\includegraphics[width=\PBRColumnWidth]{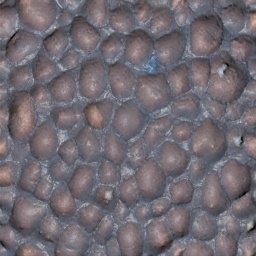} &
\includegraphics[width=\PBRColumnWidth]{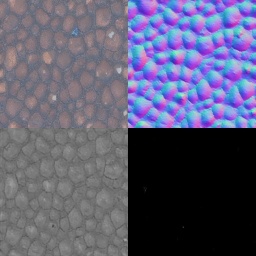} \\

&
\multicolumn{2}{c}{
  \begin{minipage}{3.1cm}
    \centering
    \scriptsize\textit{ice glazed slippery, outdoor, winter} 
\end{minipage} 
} &
\multicolumn{2}{c}{
  \begin{minipage}{3.1cm}
    \centering
    \scriptsize\textit{aerial mud, road, tracks} 
\end{minipage} 
} &
\multicolumn{2}{c}{
  \begin{minipage}{3.1cm}
    \centering
    \scriptsize\textit{dry rocky ground} 
\end{minipage} 
} &
\multicolumn{2}{c}{
  \begin{minipage}{3.1cm}
    \centering
    \scriptsize\textit{marble floor, polished, indoor} 
\end{minipage} 
} &
\multicolumn{2}{c}{
  \begin{minipage}{3.1cm}
    \centering
    \scriptsize\textit{stone ground} 
\end{minipage} 
} \\
\noalign{\smallskip}

\raisebox{+0.7\height}{\rotatebox[origin=c]{90}{\small\textbf{Leather}}} &
\includegraphics[width=\PBRColumnWidth]{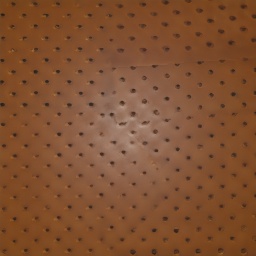} &
\includegraphics[width=\PBRColumnWidth]{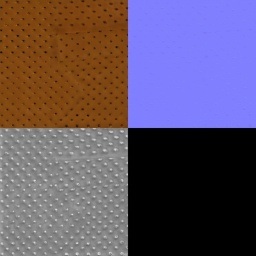} &
\includegraphics[width=\PBRColumnWidth]{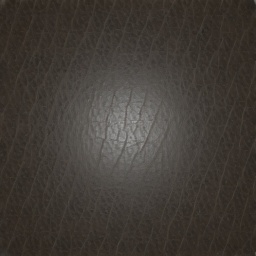} &
\includegraphics[width=\PBRColumnWidth]{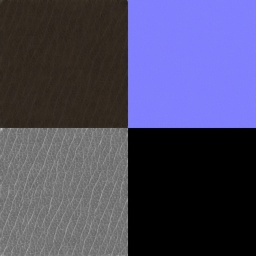} &
\includegraphics[width=\PBRColumnWidth]{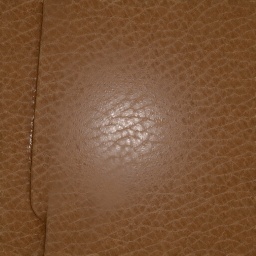} &
\includegraphics[width=\PBRColumnWidth]{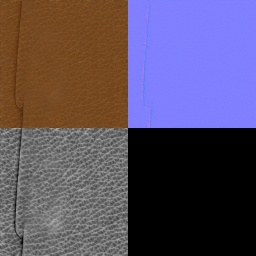} &
\includegraphics[width=\PBRColumnWidth]{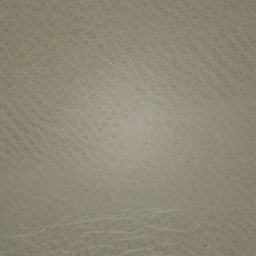} &
\includegraphics[width=\PBRColumnWidth]{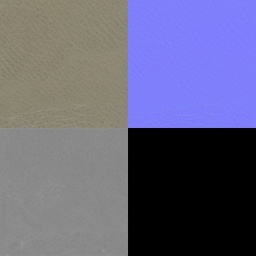} &
\includegraphics[width=\PBRColumnWidth]{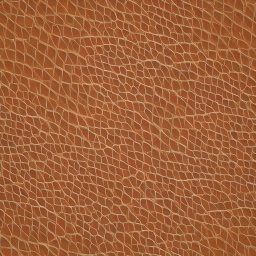} &
\includegraphics[width=\PBRColumnWidth]{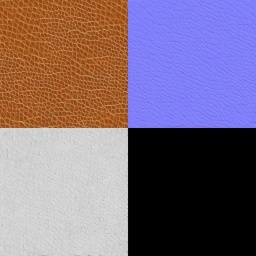} \\

&
\multicolumn{2}{c}{
  \begin{minipage}{3.1cm}
    \centering
    \scriptsize\textit{perforated leather, breathable} 
\end{minipage} 
} &
\multicolumn{2}{c}{
  \begin{minipage}{3.1cm}
    \centering
    \scriptsize\textit{black leather} 
\end{minipage} 
} &
\multicolumn{2}{c}{
  \begin{minipage}{3.1cm}
    \centering
    \scriptsize\textit{decoration, indoor} 
\end{minipage} 
} &
\multicolumn{2}{c}{
  \begin{minipage}{3.1cm}
    \centering
    \scriptsize\textit{leather white, smooth} 
\end{minipage} 
} &
\multicolumn{2}{c}{
  \begin{minipage}{3.1cm}
    \centering
    \scriptsize\textit{reptile skin leather, textured} 
\end{minipage} 
} \\
\noalign{\smallskip}

\raisebox{+0.7\height}{\rotatebox[origin=c]{90}{\small\textbf{Metal}}} &
\includegraphics[width=\PBRColumnWidth]{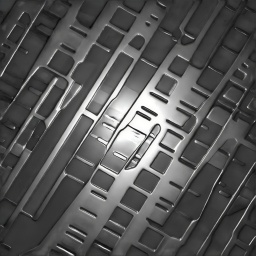} &
\includegraphics[width=\PBRColumnWidth]{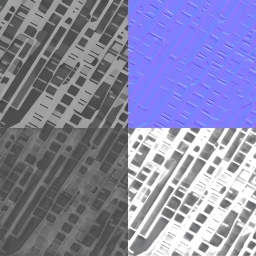} &
\includegraphics[width=\PBRColumnWidth]{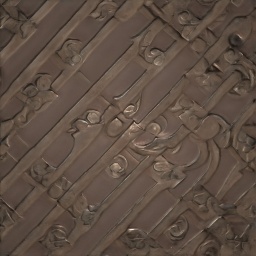} &
\includegraphics[width=\PBRColumnWidth]{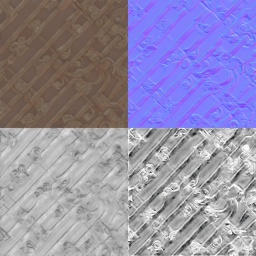} &
\includegraphics[width=\PBRColumnWidth]{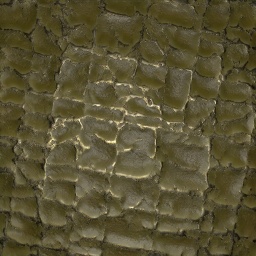} &
\includegraphics[width=\PBRColumnWidth]{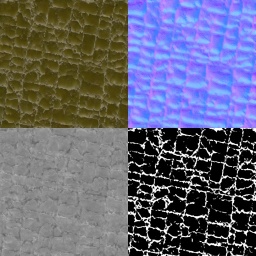} &
\includegraphics[width=\PBRColumnWidth]{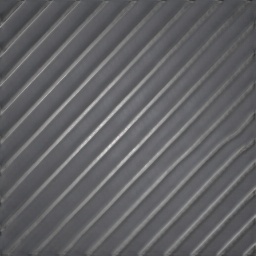} &
\includegraphics[width=\PBRColumnWidth]{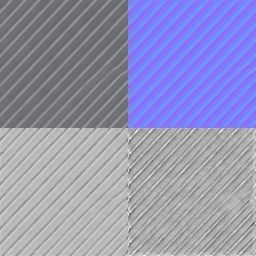} &
\includegraphics[width=\PBRColumnWidth]{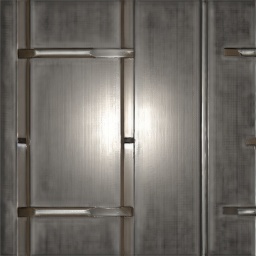} &
\includegraphics[width=\PBRColumnWidth]{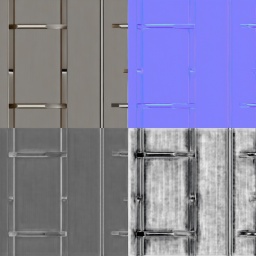} \\

&
\multicolumn{2}{c}{
  \begin{minipage}{3.1cm}
    \centering
    \scriptsize\textit{space cruiser panels, scifi} 
\end{minipage} 
} &
\multicolumn{2}{c}{
  \begin{minipage}{3.1cm}
    \centering
    \scriptsize\textit{wrought iron gate, ornate, outdoor} 
\end{minipage} 
} &
\multicolumn{2}{c}{
  \begin{minipage}{3.1cm}
    \centering
    \scriptsize\textit{golden metal wall, old} 
\end{minipage} 
} &
\multicolumn{2}{c}{
  \begin{minipage}{3.1cm}
    \centering
    \scriptsize\textit{anodized metal surface, industrial} 
\end{minipage} 
} &
\multicolumn{2}{c}{
  \begin{minipage}{3.1cm}
    \centering
    \scriptsize\textit{nickel plated hardware, smooth} 
\end{minipage} 
} \\
\noalign{\smallskip}

\raisebox{+0.7\height}{\rotatebox[origin=c]{90}{\small\textbf{Organic}}} &
\includegraphics[width=\PBRColumnWidth]{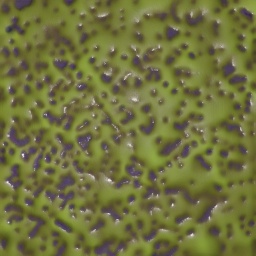} &
\includegraphics[width=\PBRColumnWidth]{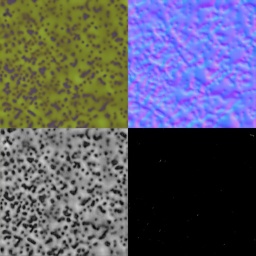} &
\includegraphics[width=\PBRColumnWidth]{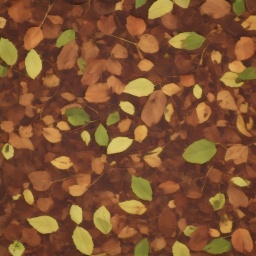} &
\includegraphics[width=\PBRColumnWidth]{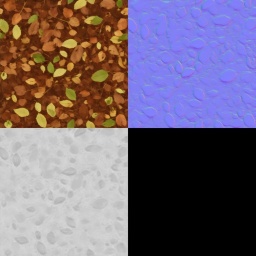} &
\includegraphics[width=\PBRColumnWidth]{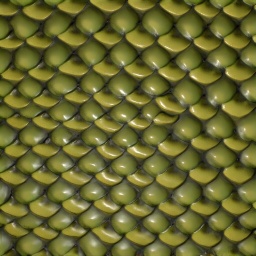} &
\includegraphics[width=\PBRColumnWidth]{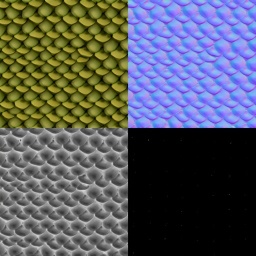} &
\includegraphics[width=\PBRColumnWidth]{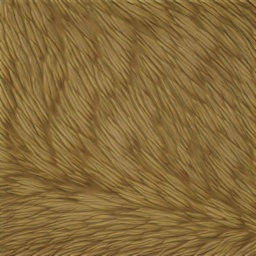} &
\includegraphics[width=\PBRColumnWidth]{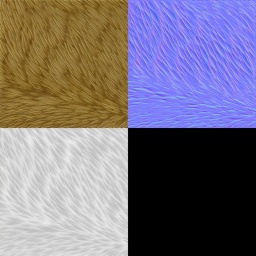} &
\includegraphics[width=\PBRColumnWidth]{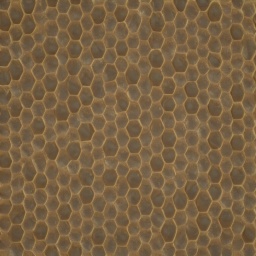} &
\includegraphics[width=\PBRColumnWidth]{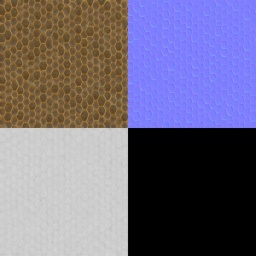} \\

&
\multicolumn{2}{c}{
  \begin{minipage}{3.1cm}
    \centering
    \scriptsize\textit{alien slime} 
\end{minipage} 
} &
\multicolumn{2}{c}{
  \begin{minipage}{3.1cm}
    \centering
    \scriptsize\textit{forest leaves, natural, autumn, dirt} 
\end{minipage} 
} &
\multicolumn{2}{c}{
  \begin{minipage}{3.1cm}
    \centering
    \scriptsize\textit{dragon scales} 
\end{minipage} 
} &
\multicolumn{2}{c}{
  \begin{minipage}{3.1cm}
    \centering
    \scriptsize\textit{stylized animal fur} 
\end{minipage} 
} &
\multicolumn{2}{c}{
  \begin{minipage}{3.1cm}
    \centering
    \scriptsize\textit{honeycomb structure, geometric, natural, beehive} 
\end{minipage} 
} \\
\noalign{\smallskip}

\raisebox{+0.7\height}{\rotatebox[origin=c]{90}{\small\textbf{Plastic}}} &
\includegraphics[width=\PBRColumnWidth]{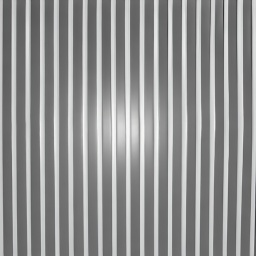} &
\includegraphics[width=\PBRColumnWidth]{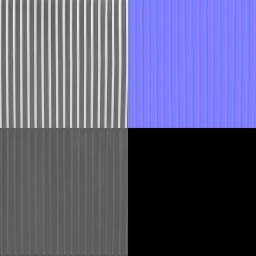} &
\includegraphics[width=\PBRColumnWidth]{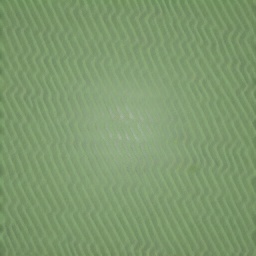} &
\includegraphics[width=\PBRColumnWidth]{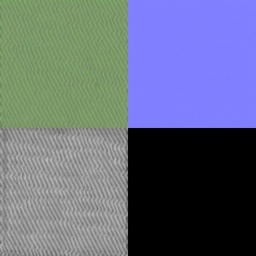} &

\includegraphics[width=\PBRColumnWidth]{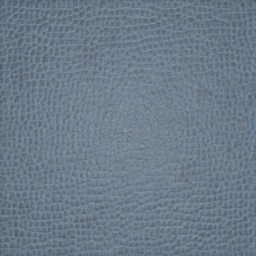} &
\includegraphics[width=\PBRColumnWidth]{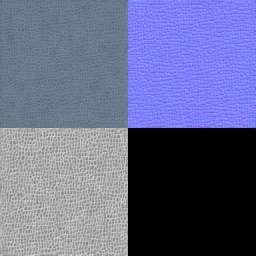} &
\includegraphics[width=\PBRColumnWidth]{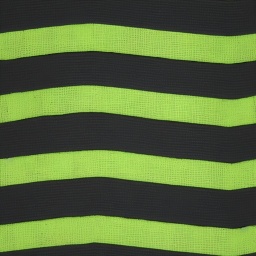} &
\includegraphics[width=\PBRColumnWidth]{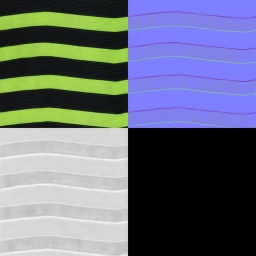} &
\includegraphics[width=\PBRColumnWidth]{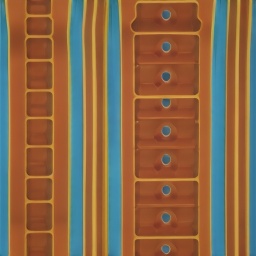} &
\includegraphics[width=\PBRColumnWidth]{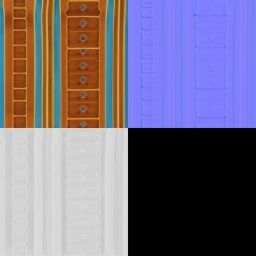} \\

&
\multicolumn{2}{c}{
  \begin{minipage}{3.1cm}
    \centering
    \scriptsize\textit{plastic pattern, synthetic} 
\end{minipage} 
} &
\multicolumn{2}{c}{
  \begin{minipage}{3.1cm}
    \centering
    \scriptsize\textit{yoga mat} 
\end{minipage} 
} &
\multicolumn{2}{c}{
  \begin{minipage}{3.1cm}
    \centering
    \scriptsize\textit{synthetic plastic, rough} 
\end{minipage} 
} &
\multicolumn{2}{c}{
  \begin{minipage}{3.1cm}
    \centering
    \scriptsize\textit{reflective safety vest, clothing} 
\end{minipage} 
} &
\multicolumn{2}{c}{
  \begin{minipage}{3.1cm}
    \centering
    \scriptsize\textit{childrens playground slide, colorful} 
\end{minipage} 
} \\
\noalign{\smallskip}

\raisebox{+0.7\height}{\rotatebox[origin=c]{90}{\small\textbf{Tile}}} &
\includegraphics[width=\PBRColumnWidth]{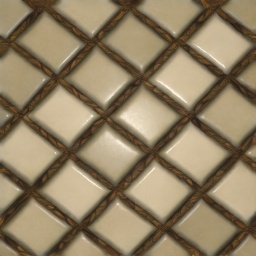} &
\includegraphics[width=\PBRColumnWidth]{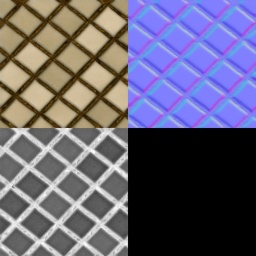}&
\includegraphics[width=\PBRColumnWidth]{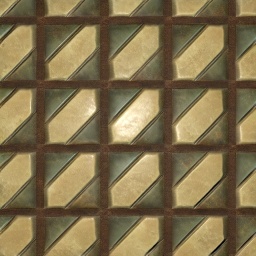} &
\includegraphics[width=\PBRColumnWidth]{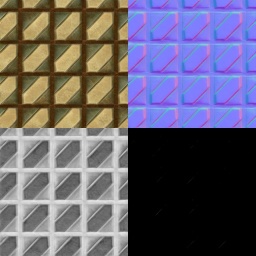} &
\includegraphics[width=\PBRColumnWidth]{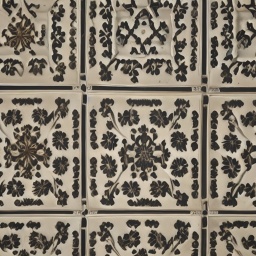} &
\includegraphics[width=\PBRColumnWidth]{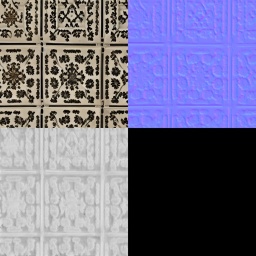} &
\includegraphics[width=\PBRColumnWidth]{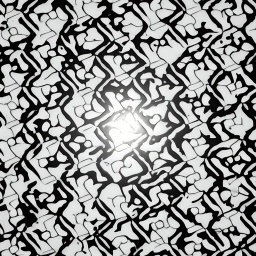} &
\includegraphics[width=\PBRColumnWidth]{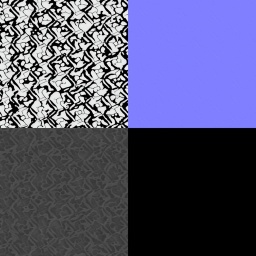} &
\includegraphics[width=\PBRColumnWidth]{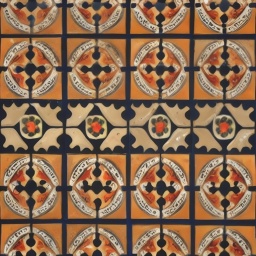} &
\includegraphics[width=\PBRColumnWidth]{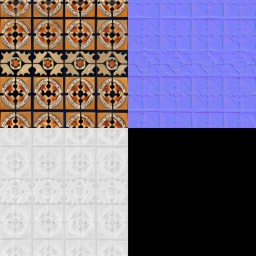} \\

&
\multicolumn{2}{c}{
  \begin{minipage}{3.1cm}
    \centering
    \scriptsize\textit{elegant, interior decoration} 
\end{minipage} 
} &
\multicolumn{2}{c}{
  \begin{minipage}{3.1cm}
    \centering
    \scriptsize\textit{art deco style tiles, vintage, indoor, decorative} 
\end{minipage} 
} &
\multicolumn{2}{c}{
  \begin{minipage}{3.1cm}
    \centering
    \scriptsize\textit{vintage ceiling tiles, indoor} 
\end{minipage} 
} &
\multicolumn{2}{c}{
  \begin{minipage}{3.1cm}
    \centering
    \scriptsize\textit{patterned bw vinyl, floors} 
\end{minipage} 
} &
\multicolumn{2}{c}{
  \begin{minipage}{3.1cm}
    \centering
    \scriptsize\textit{encaustic cement tiles, colorful, indoor, floor} 
\end{minipage} 
} \\
\noalign{\smallskip}

\raisebox{+0.7\height}{\rotatebox[origin=c]{90}{\small\textbf{Wall}}} &
\includegraphics[width=\PBRColumnWidth]{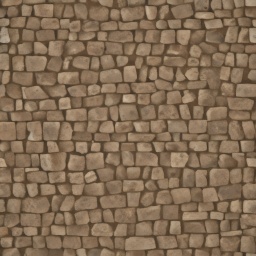} &
\includegraphics[width=\PBRColumnWidth]{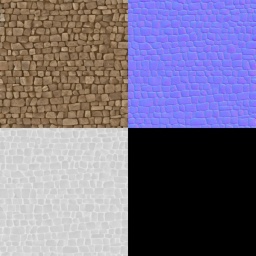} &
\includegraphics[width=\PBRColumnWidth]{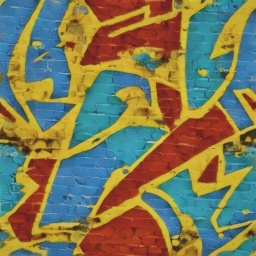} &
\includegraphics[width=\PBRColumnWidth]{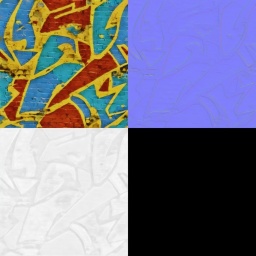} &
\includegraphics[width=\PBRColumnWidth]{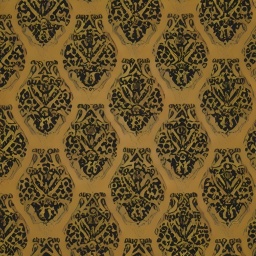} &
\includegraphics[width=\PBRColumnWidth]{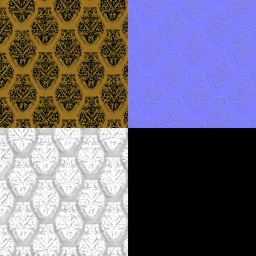} &
\includegraphics[width=\PBRColumnWidth]{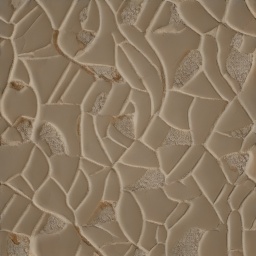} &
\includegraphics[width=\PBRColumnWidth]{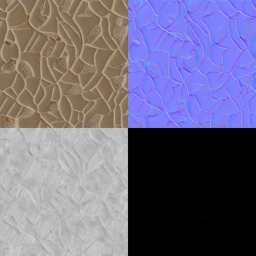} &
\includegraphics[width=\PBRColumnWidth]{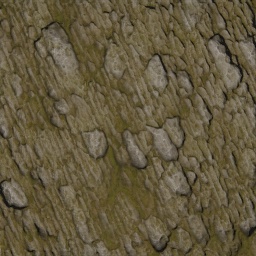} &
\includegraphics[width=\PBRColumnWidth]{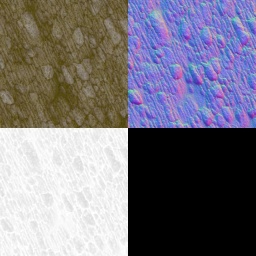} \\

&
\multicolumn{2}{c}{
  \begin{minipage}{3.1cm}
    \centering
    \scriptsize\textit{dry stone wall, natural, outdoor} 
\end{minipage} 
} &
\multicolumn{2}{c}{
  \begin{minipage}{3.1cm}
    \centering
    \scriptsize\textit{street art graffiti, colorful, urban} 
\end{minipage} 
} &
\multicolumn{2}{c}{
  \begin{minipage}{3.1cm}
    \centering
    \scriptsize\textit{victorian wallpaper, patterned, indoor, historic} 
\end{minipage} 
} &
\multicolumn{2}{c}{
  \begin{minipage}{3.1cm}
    \centering
    \scriptsize\textit{stucco finish, mediterranean} 
\end{minipage} 
} &
\multicolumn{2}{c}{
  \begin{minipage}{3.1cm}
    \centering
    \scriptsize\textit{cliff, outdoor} 
\end{minipage} 
} \\
\noalign{\smallskip}

\raisebox{+0.7\height}{\rotatebox[origin=c]{90}{\small\textbf{Wood}}} &
\includegraphics[width=\PBRColumnWidth]{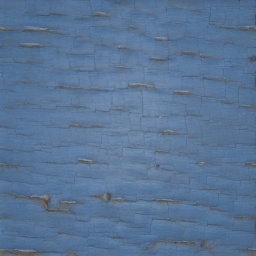} &
\includegraphics[width=\PBRColumnWidth]{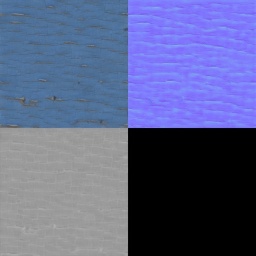} &
\includegraphics[width=\PBRColumnWidth]{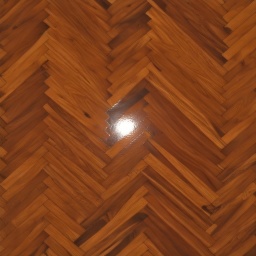} &
\includegraphics[width=\PBRColumnWidth]{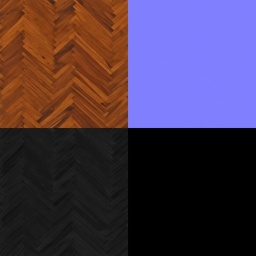} &
\includegraphics[width=\PBRColumnWidth]{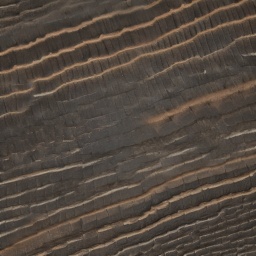} &
\includegraphics[width=\PBRColumnWidth]{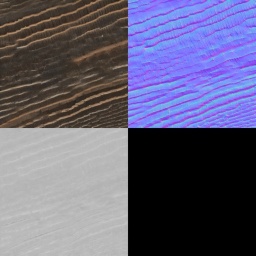} &
\includegraphics[width=\PBRColumnWidth]{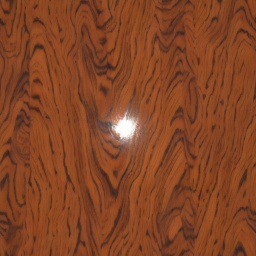} &
\includegraphics[width=\PBRColumnWidth]{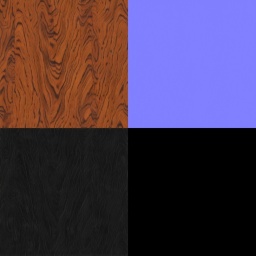} &
\includegraphics[width=\PBRColumnWidth]{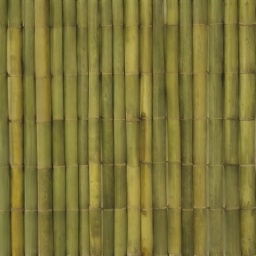} &
\includegraphics[width=\PBRColumnWidth]{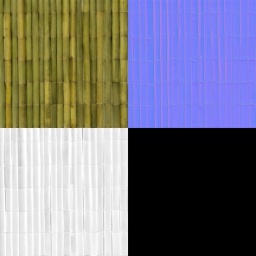} \\

&
\multicolumn{2}{c}{
  \begin{minipage}{3.1cm}
    \centering
    \scriptsize\textit{blue, worn painted wood siding, walls} 
\end{minipage} 
} &
\multicolumn{2}{c}{
  \begin{minipage}{3.1cm}
    \centering
    \scriptsize\textit{parquet wood flooring, geometric} 
\end{minipage} 
} &
\multicolumn{2}{c}{
  \begin{minipage}{3.1cm}
    \centering
    \scriptsize\textit{charcoal} 
\end{minipage} 
} &
\multicolumn{2}{c}{
  \begin{minipage}{3.1cm}
    \centering
    \scriptsize\textit{varnished walnut, glossy, indoor} 
\end{minipage} 
} &
\multicolumn{2}{c}{
  \begin{minipage}{3.1cm}
    \centering
    \scriptsize\textit{bamboo wall covering, eco-friendly} 
\end{minipage} 
} \\

\end{tabular}
\egroup

\caption{The generation results of DreamPBR under text-only conditions: We randomly sampled numerous materials with various types and wide tags, by the prompts, ``a PBR material of [type], [tags]". Not only can DreamPBR generate materials that match the descriptions, but also some out-of-domain materials are created as well such as \textbf{brick} of snow-covered bricks, \textbf{plastic} of a children's playground slide, and  \textbf{wall} of street art graffiti.} 
\label{fig:PBR} 
\Description{fig:PBR}
\end{figure*}

\newcommand{\directColumeFigWidth}{1.63cm} 

\begin{figure*}[htbp]
\centering 

\bgroup

\def\arraystretch{0.1} 
\setlength\tabcolsep{1pt} 
\begin{tabular}{
    m{\directColumeFigWidth}
    >{\raggedright\arraybackslash}m{\directColumeFigWidth}
    @{\hspace{0.5pt}}m{\directColumeFigWidth}
    @{\hspace{0.5pt}}m{\directColumeFigWidth}
    >{\raggedright\arraybackslash}m{\directColumeFigWidth}
    @{\hspace{0.5pt}}m{\directColumeFigWidth}
    @{\hspace{0.5pt}}m{\directColumeFigWidth}
    >{\raggedright\arraybackslash}m{\directColumeFigWidth}
    @{\hspace{0.5pt}}m{\directColumeFigWidth}
    @{\hspace{0.5pt}}m{\directColumeFigWidth}
    }
\multicolumn{1}{c}{Pattern} & \multicolumn{1}{c}{Prompt} & \multicolumn{1}{c}{Render} & \multicolumn{1}{c}{SVBRDF} & \multicolumn{1}{c}{Prompt} & \multicolumn{1}{c}{Render} & \multicolumn{1}{c}{SVBRDF} & \multicolumn{1}{c}{Prompt} & \multicolumn{1}{c}{Render} & \multicolumn{1}{c}{SVBRDF} \\

\includegraphics[width=\directColumeFigWidth]{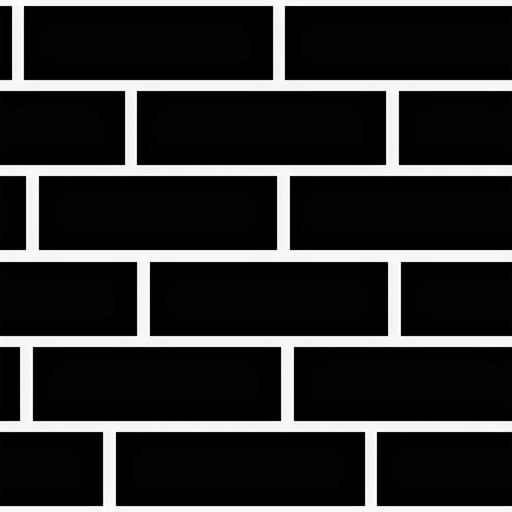} &
\scriptsize\textit{a PBR material of brick, narrow bricks, walls} & 
\includegraphics[width=\directColumeFigWidth]{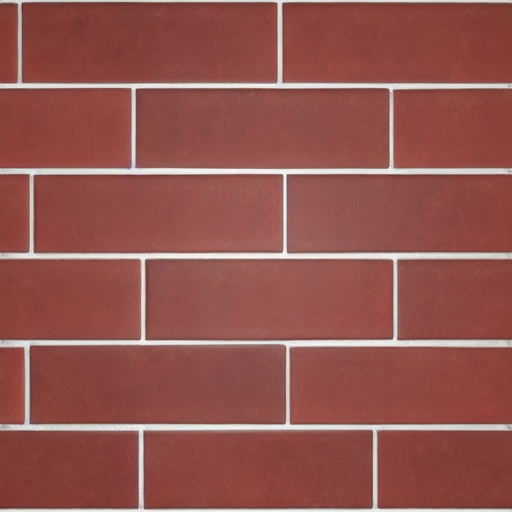} &
\includegraphics[width=\directColumeFigWidth]{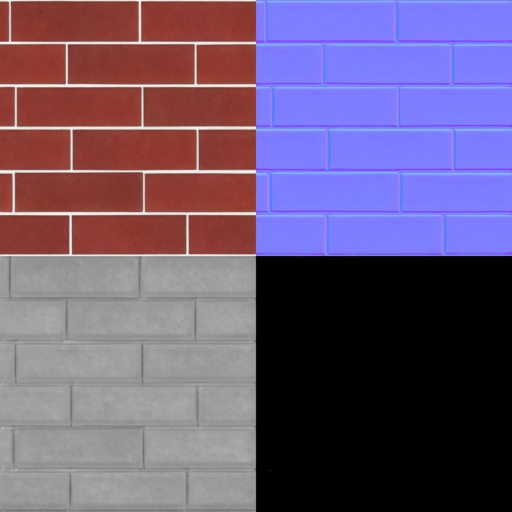} &

\scriptsize\textit{a PBR material of leather, smooth, white, clean} & 
\includegraphics[width=\directColumeFigWidth]{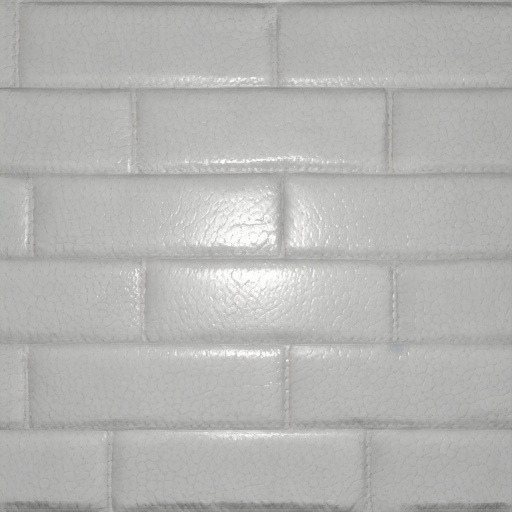} &
\includegraphics[width=\directColumeFigWidth]{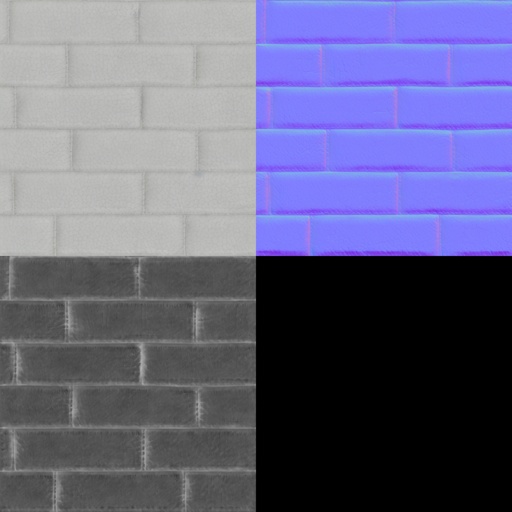} &
\scriptsize\textit{a PBR material of metal, ornate celtic gold} & 
\includegraphics[width=\directColumeFigWidth]{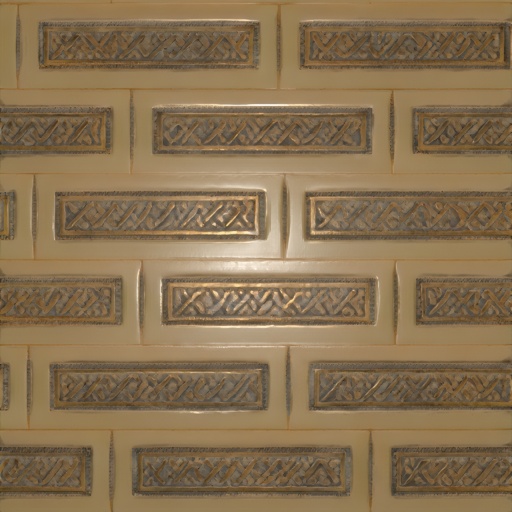} &
\includegraphics[width=\directColumeFigWidth]{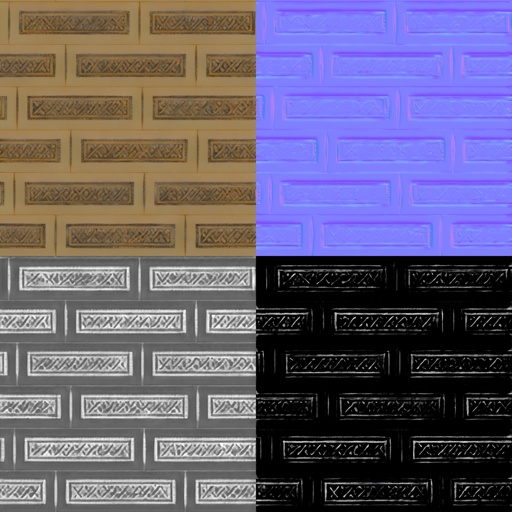} \\
&
\scriptsize\textit{a PBR material of fabric, plush toy fur, soft, indoor} & 
\includegraphics[width=\directColumeFigWidth]{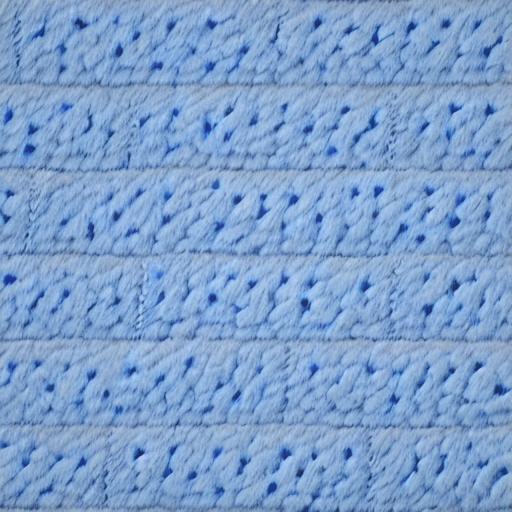} &
\includegraphics[width=\directColumeFigWidth]{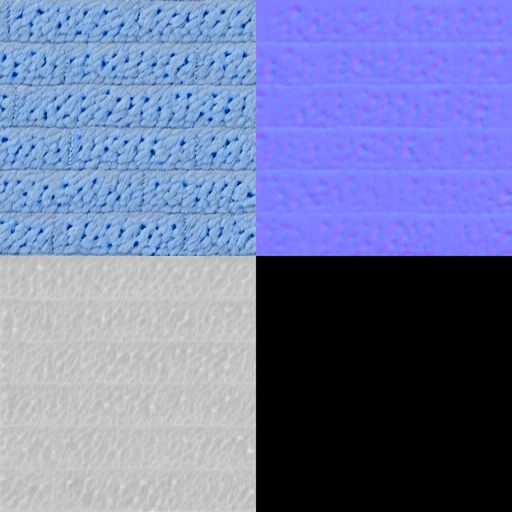} &
\scriptsize\textit{a PBR material of plastic, synthetic turf blades, green, sports} & 
\includegraphics[width=\directColumeFigWidth]{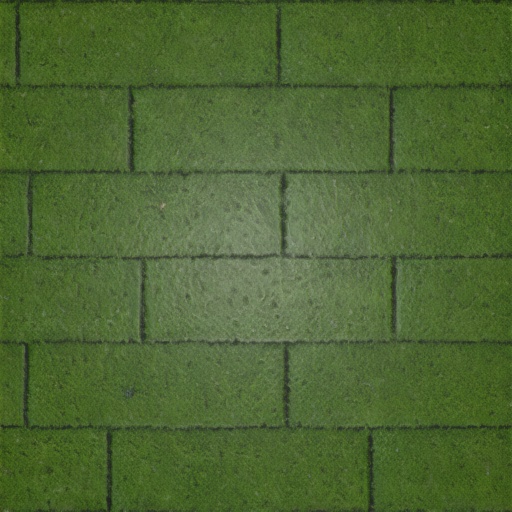} &
\includegraphics[width=\directColumeFigWidth]{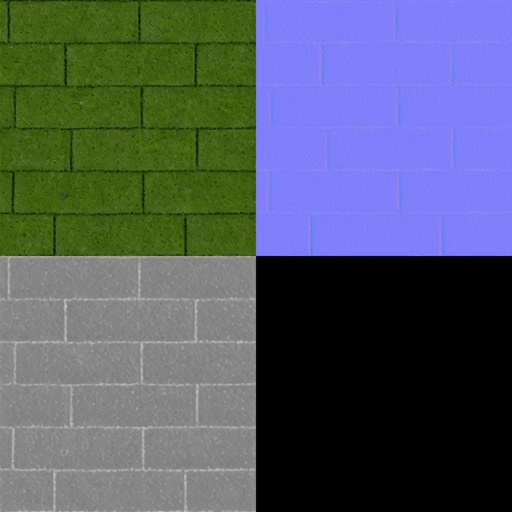} &
\scriptsize\textit{a PBR material of tile, glass mosaic art, translucent, decorative} & 
\includegraphics[width=\directColumeFigWidth]{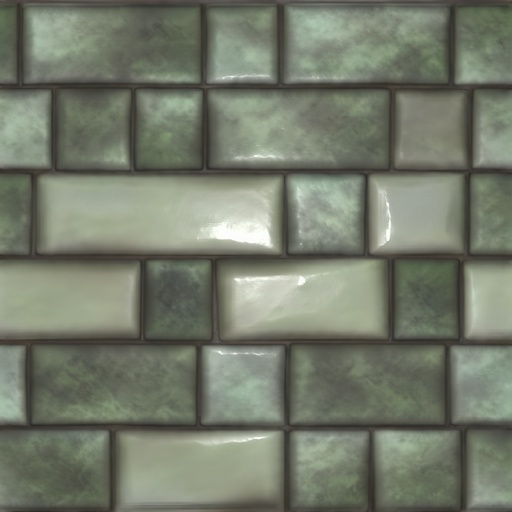} &
\includegraphics[width=\directColumeFigWidth]{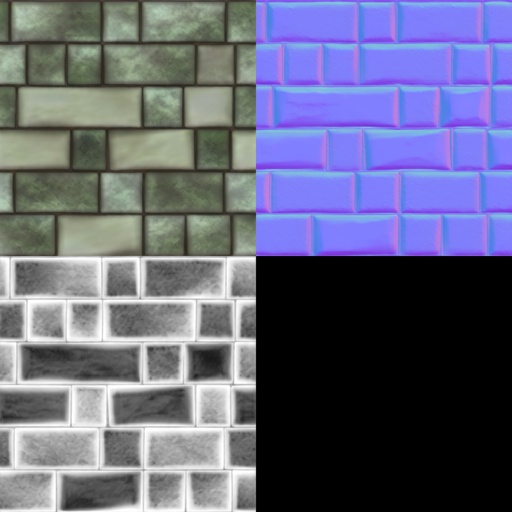} \\

\includegraphics[width=\directColumeFigWidth]{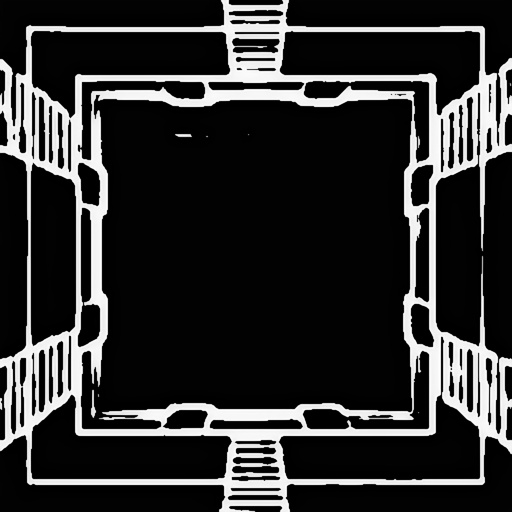} &
\scriptsize\textit{a PBR material of ground, marble floor tiles, polished, indoor, luxury} &
\includegraphics[width=\directColumeFigWidth]{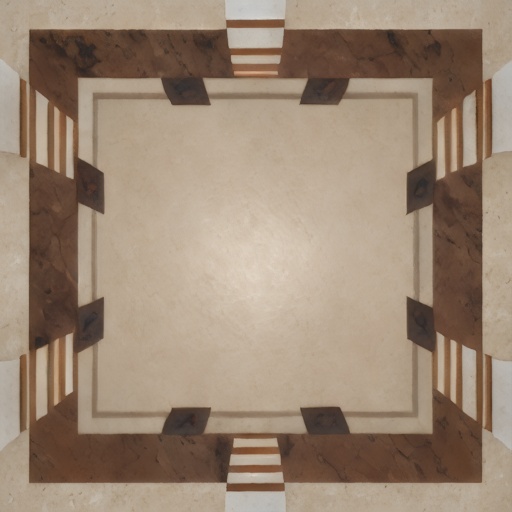} &
\includegraphics[width=\directColumeFigWidth]{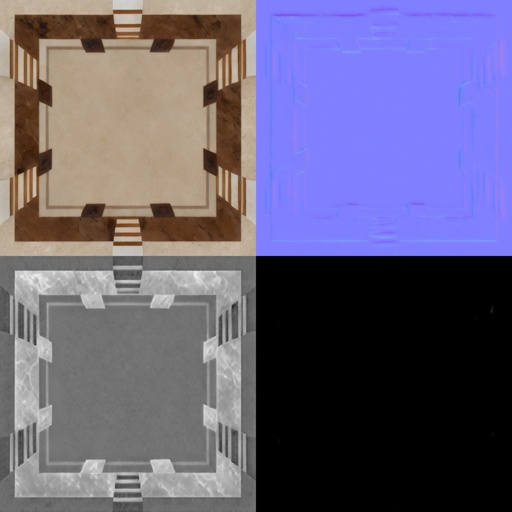} &
\scriptsize\textit{a PBR material of fabric, dirty carpet, carpet, textile, faded, floor} &
\includegraphics[width=\directColumeFigWidth]{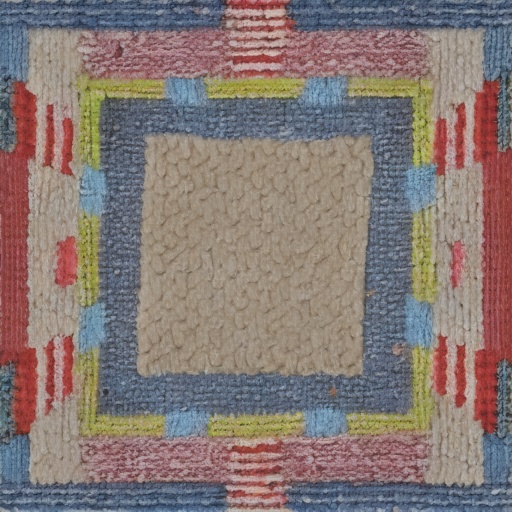} &
\includegraphics[width=\directColumeFigWidth]{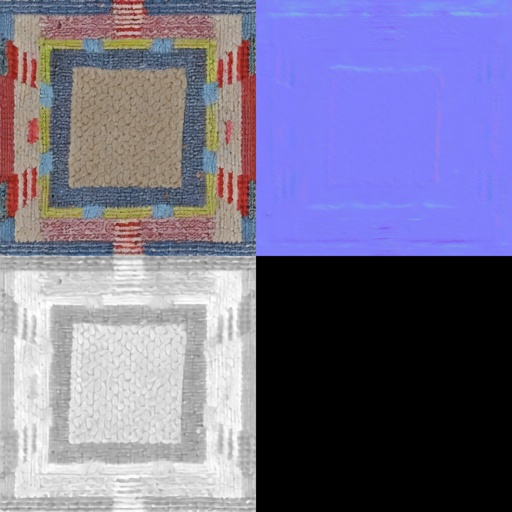} &
\scriptsize\textit{a PBR material of wood, oak flooring, classic, indoor} &
\includegraphics[width=\directColumeFigWidth]{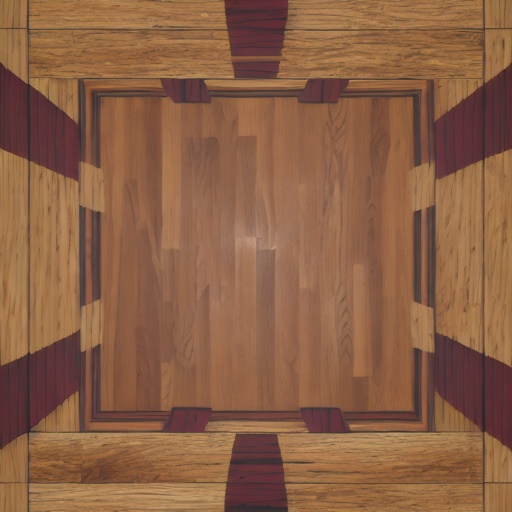} &
\includegraphics[width=\directColumeFigWidth]{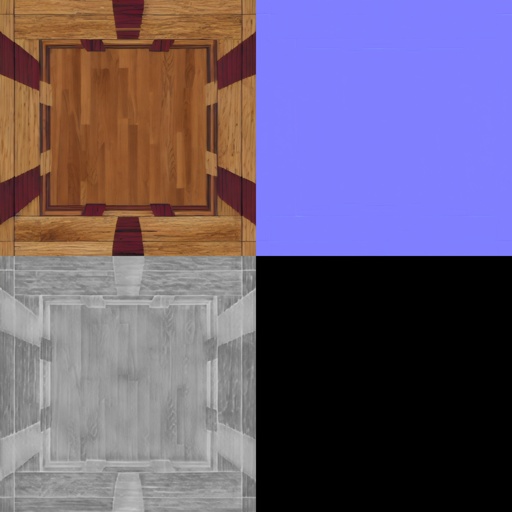} \\
&
\scriptsize\textit{a PBR material of leather, fabric leather, clean, seat, chair, couch} &
\includegraphics[width=\directColumeFigWidth]{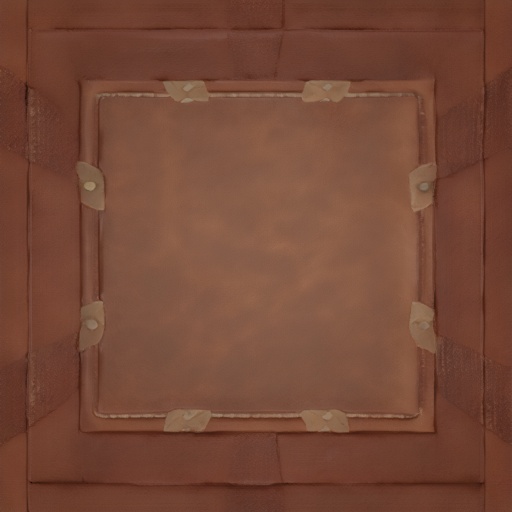} &
\includegraphics[width=\directColumeFigWidth]{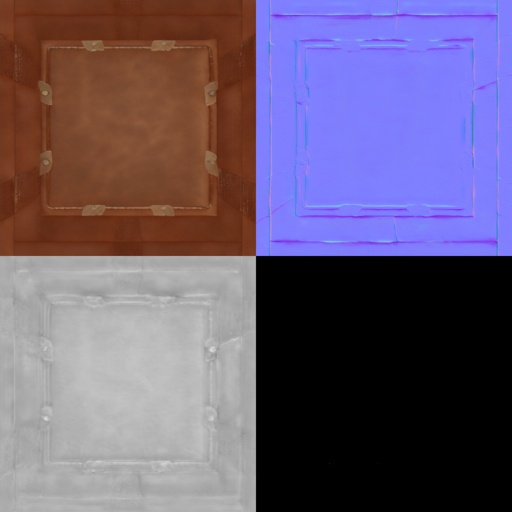} &
\scriptsize\textit{a PBR material of plastic, yoga mat} &
\includegraphics[width=\directColumeFigWidth]{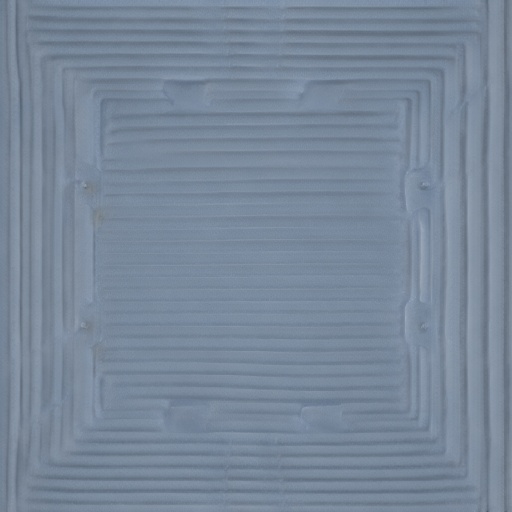} &
\includegraphics[width=\directColumeFigWidth]{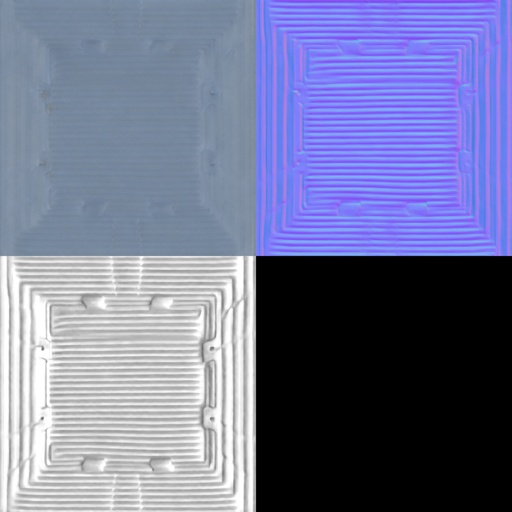} &
\scriptsize\textit{a PBR material of fabric, hand woven carpet, artisan, indoor} &
\includegraphics[width=\directColumeFigWidth]{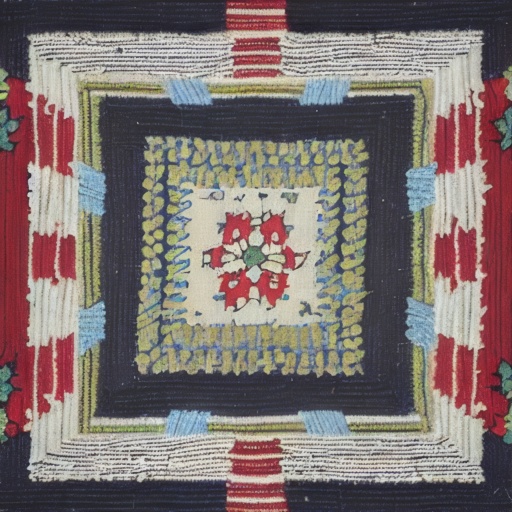} &
\includegraphics[width=\directColumeFigWidth]{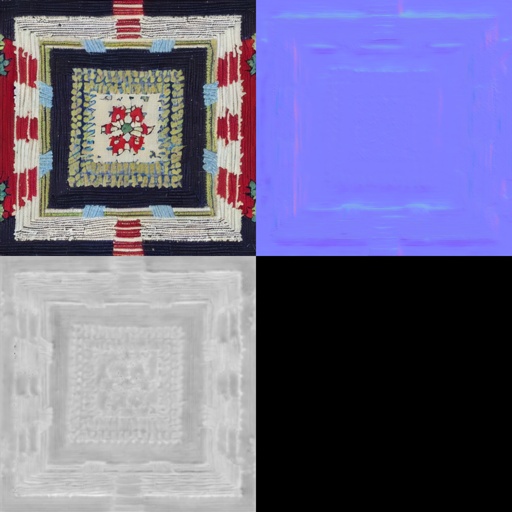} \\

\includegraphics[width=\directColumeFigWidth]{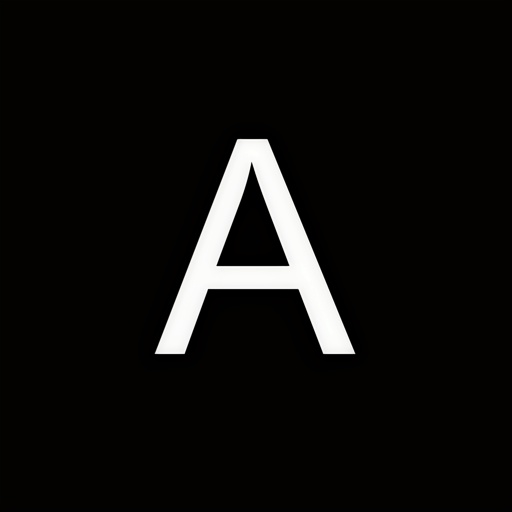} &
\scriptsize\textit{a PBR material of tile, bathroom floor tiles, non-slip, indoor} &
\includegraphics[width=\directColumeFigWidth]{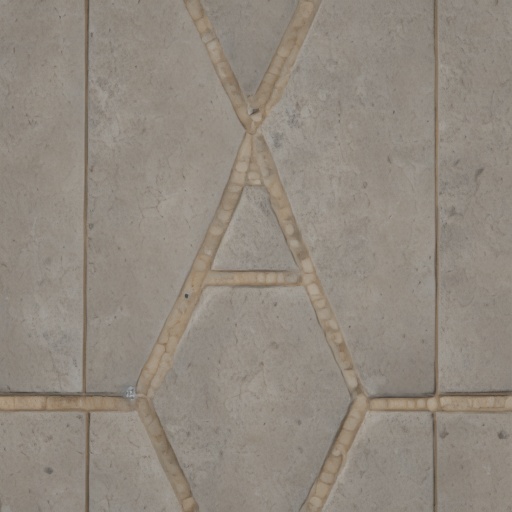} &
\includegraphics[width=\directColumeFigWidth]{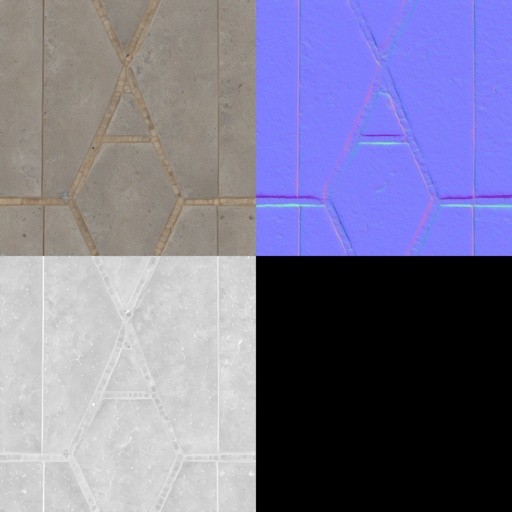} &
\scriptsize\textit{a PBR material of tile, slate walkway tiles, rugged, outdoor} &
\includegraphics[width=\directColumeFigWidth]{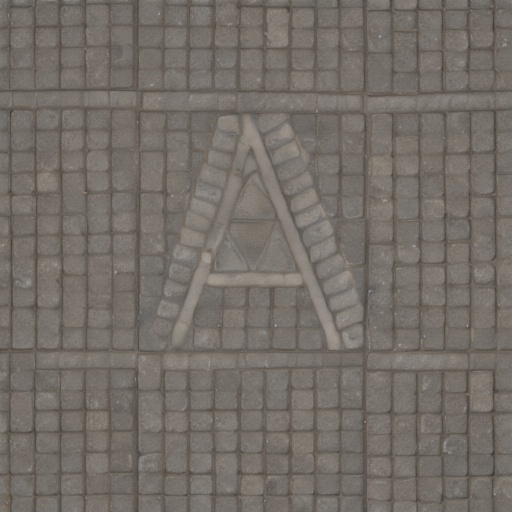} &
\includegraphics[width=\directColumeFigWidth]{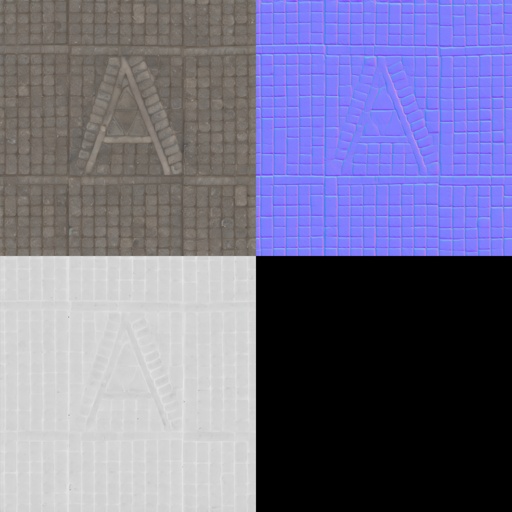} &
\scriptsize\textit{a PBR material of tile, art deco style tiles, vintage, indoor, decorative} &
\includegraphics[width=\directColumeFigWidth]{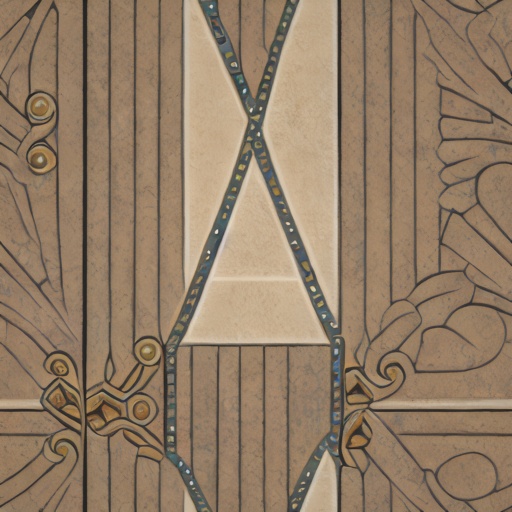} &
\includegraphics[width=\directColumeFigWidth]{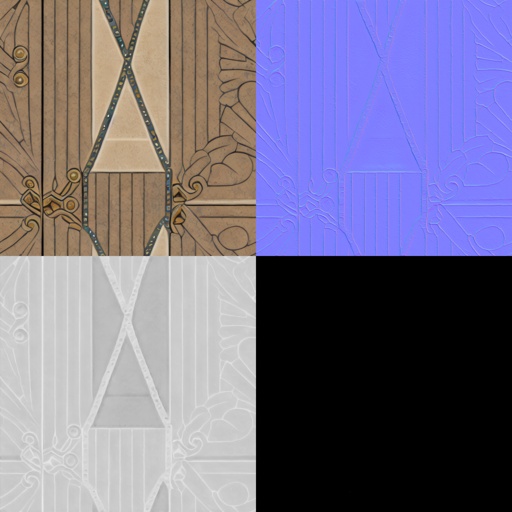} \\
&
\scriptsize\textit{a PBR material of wall, tiled bathroom wall, moisture-resistant, indoor} &
\includegraphics[width=\directColumeFigWidth]{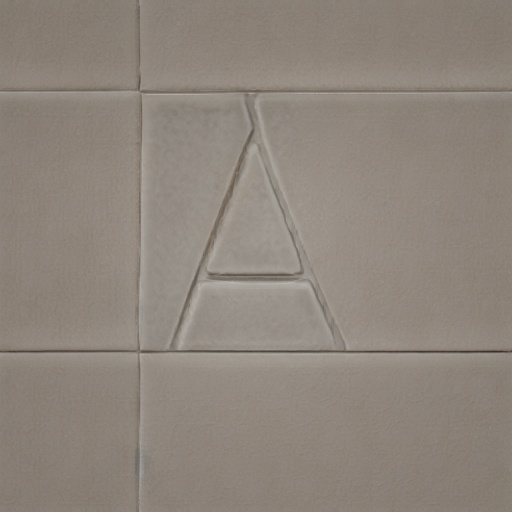} &
\includegraphics[width=\directColumeFigWidth]{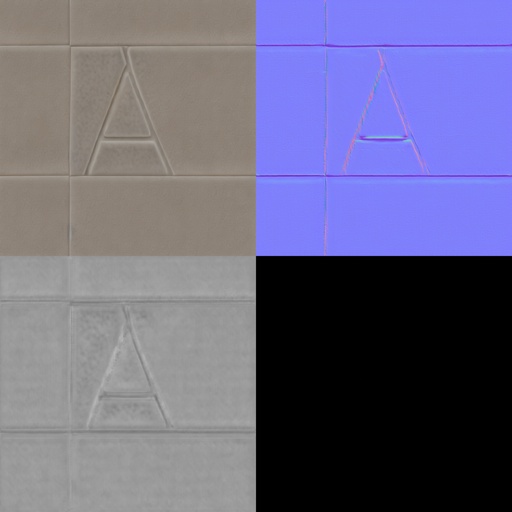} &
\scriptsize\textit{a PBR material of metal, scratched scuffed metal} &
\includegraphics[width=\directColumeFigWidth]{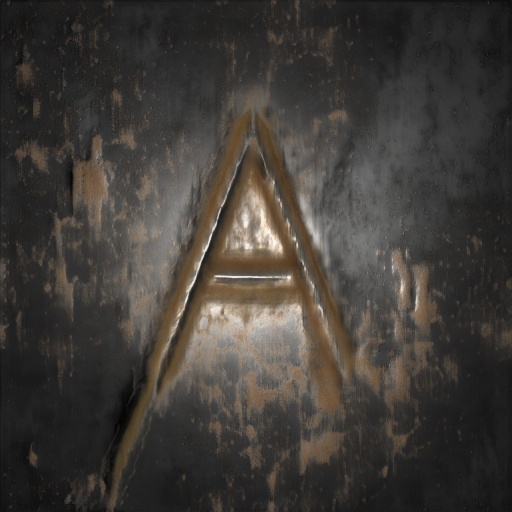} &
\includegraphics[width=\directColumeFigWidth]{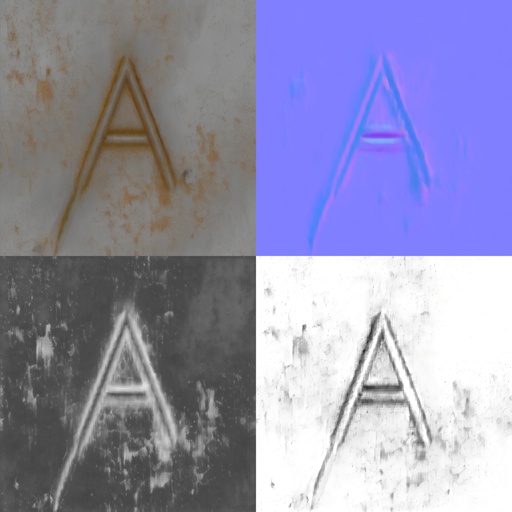} &
\scriptsize\textit{a PBR material of brick, sewer brick, walls} &
\includegraphics[width=\directColumeFigWidth]{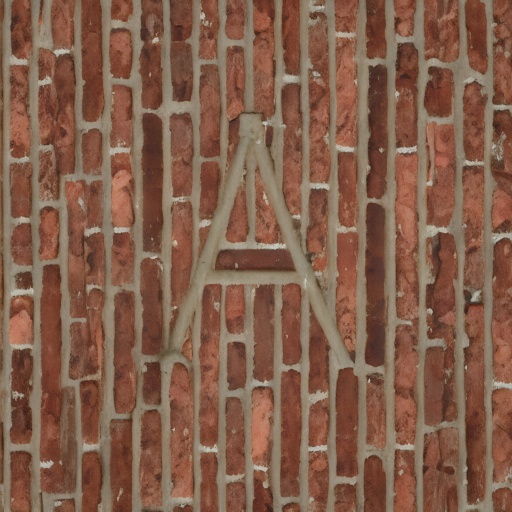} &
\includegraphics[width=\directColumeFigWidth]{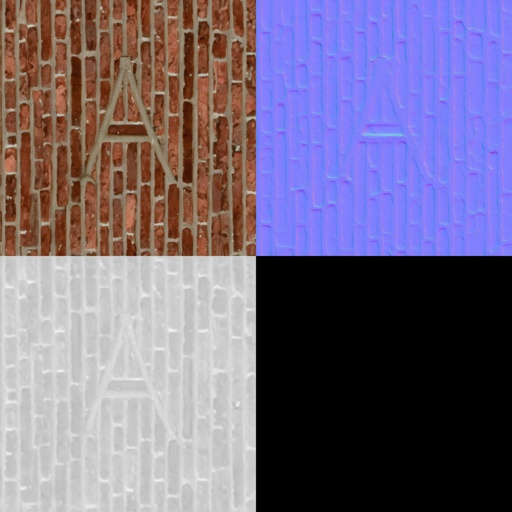} \\

\includegraphics[width=\directColumeFigWidth]{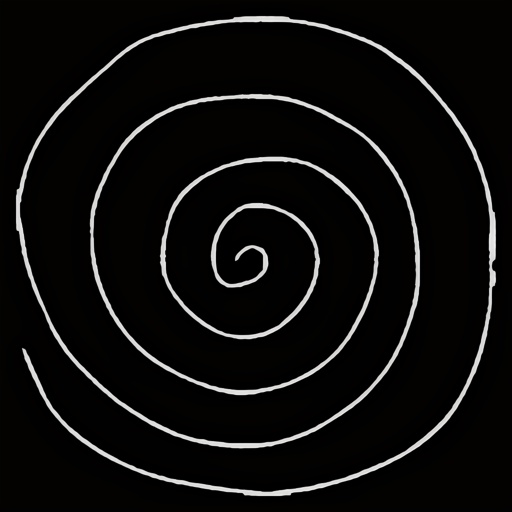} &

\scriptsize\textit{a PBR material of brick, brick floor, outdoor, clean, man made} &
\includegraphics[width=\directColumeFigWidth]{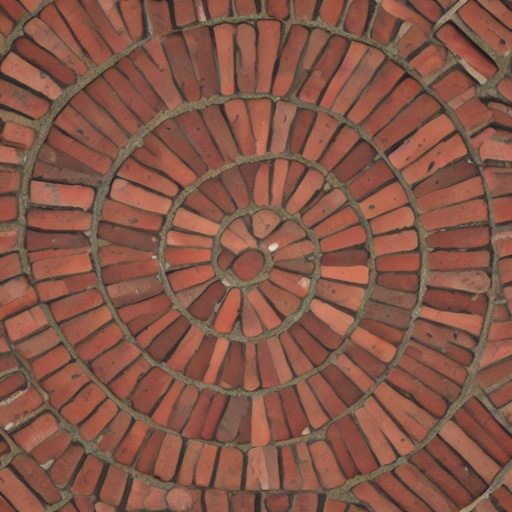} &
\includegraphics[width=\directColumeFigWidth]{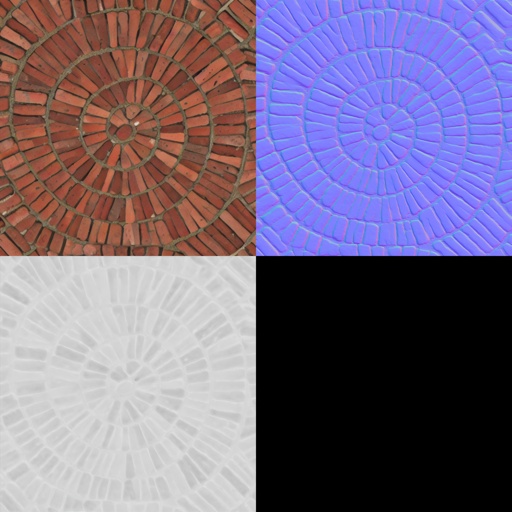} &
\scriptsize\textit{a PBR material of metal, chrome car detailing, reflective, car trim} &
\includegraphics[width=\directColumeFigWidth]{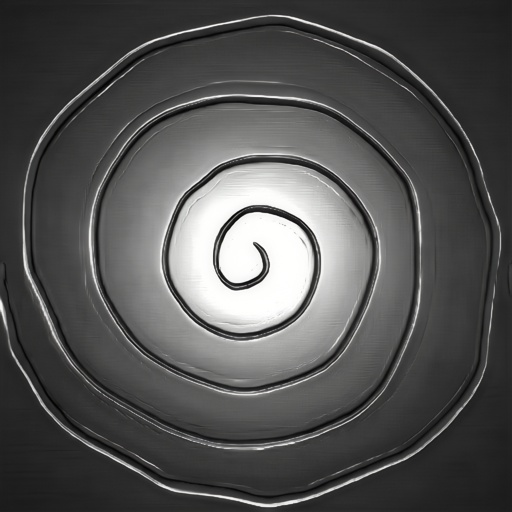} &
\includegraphics[width=\directColumeFigWidth]{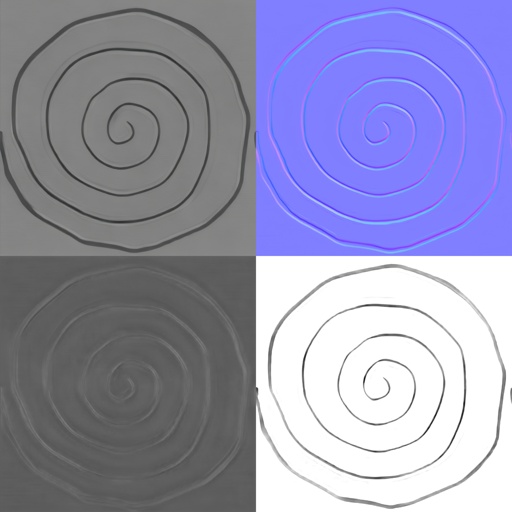} &
\scriptsize\textit{a PBR material of wood, burnt wood finish, charred, artistic, decor} &
\includegraphics[width=\directColumeFigWidth]{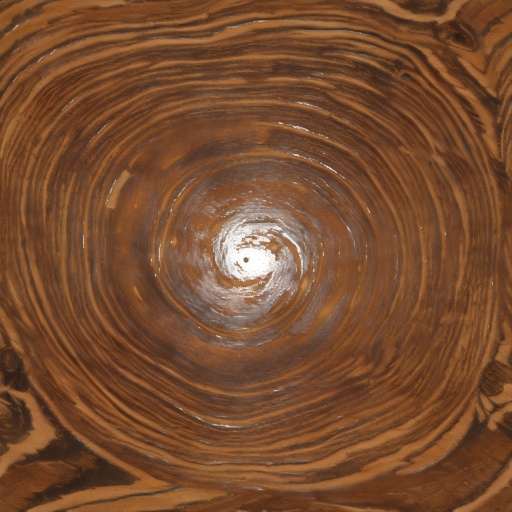} &
\includegraphics[width=\directColumeFigWidth]{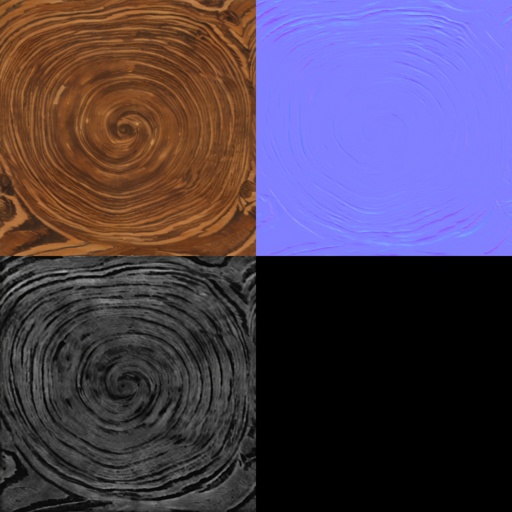} \\
&
\scriptsize\textit{a PBR material of tile, patterned bw vinyl, floors} &
\includegraphics[width=\directColumeFigWidth]{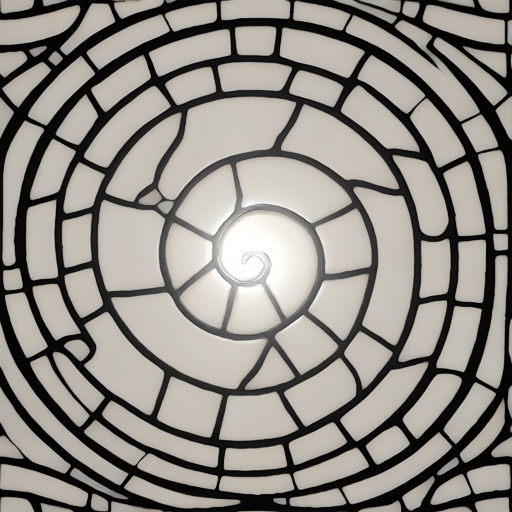} &
\includegraphics[width=\directColumeFigWidth]{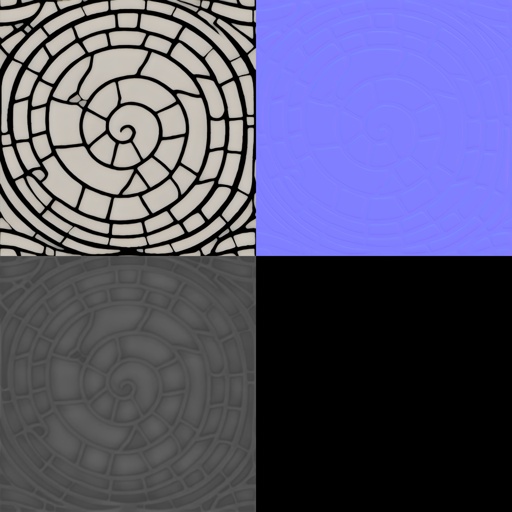} &
\scriptsize\textit{a PBR material of wall, victorian wallpaper, patterned, indoor, historic} &
\includegraphics[width=\directColumeFigWidth]{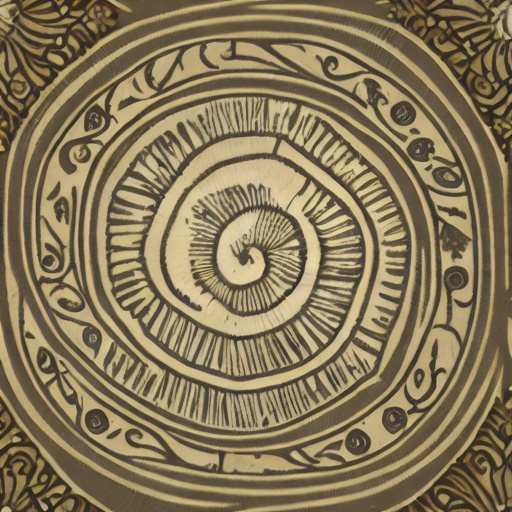} &
\includegraphics[width=\directColumeFigWidth]{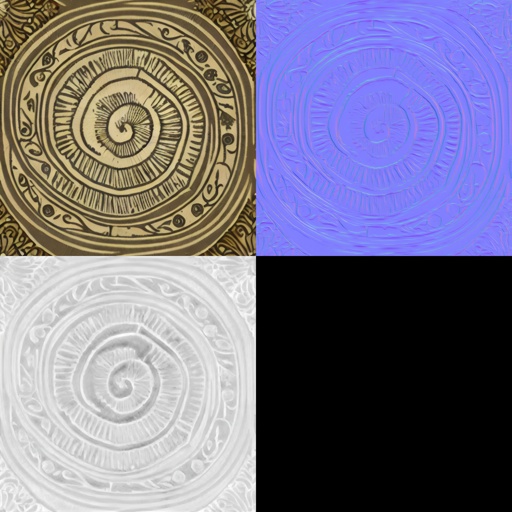} &
\scriptsize\textit{a PBR material of fabric, hand woven carpet, artisan, indoor, carpet} &
\includegraphics[width=\directColumeFigWidth]{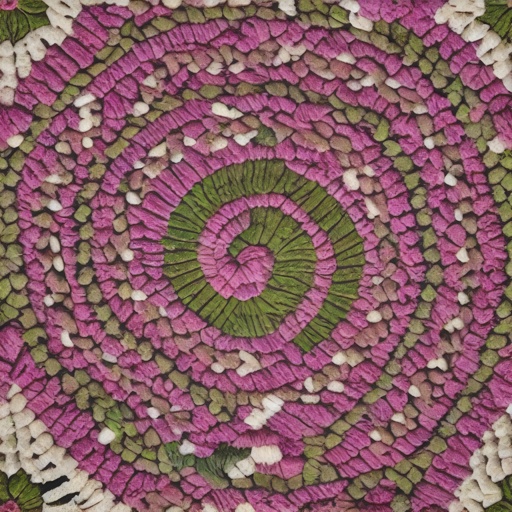} &
\includegraphics[width=\directColumeFigWidth]{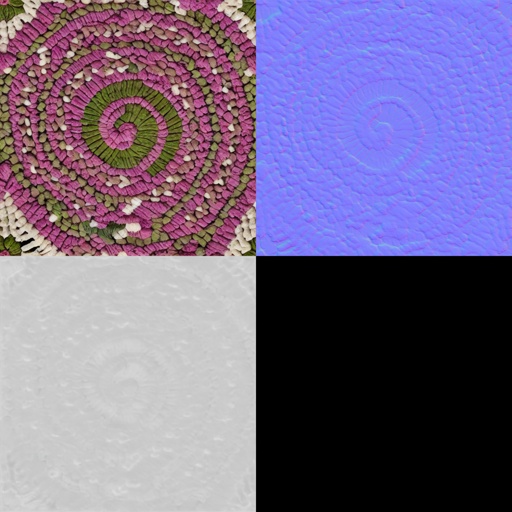} \\

\includegraphics[width=\directColumeFigWidth]{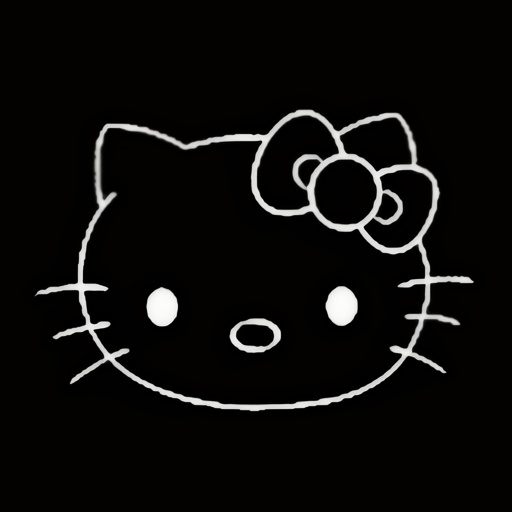} &
\scriptsize\textit{a PBR material of wall, Hello Kitty sticker wallpaper, colorful, indoor, nursery, easy-apply} &
\includegraphics[width=\directColumeFigWidth]{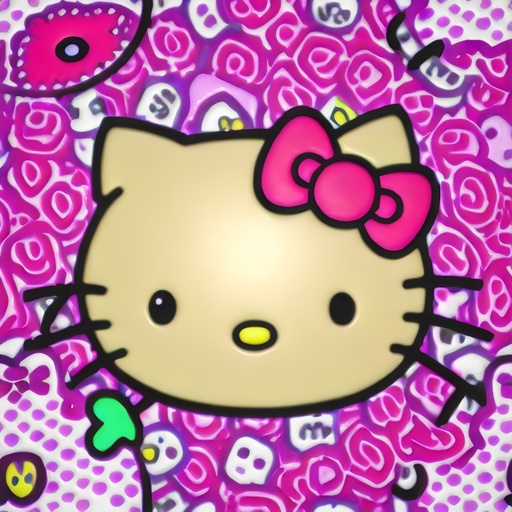} &
\includegraphics[width=\directColumeFigWidth]{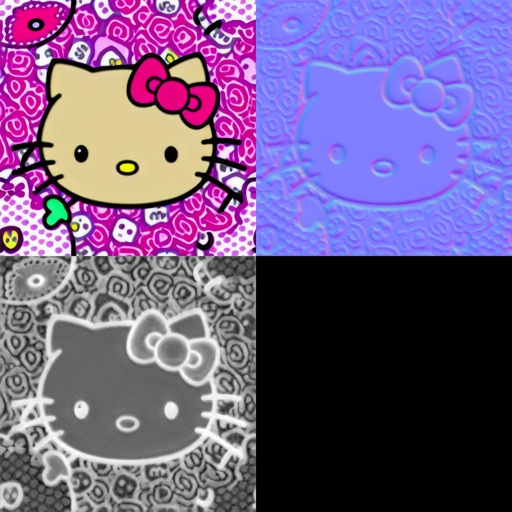} &
\scriptsize\textit{a PBR material of wall, street art graffiti, colorful, outdoor, urban} &
\includegraphics[width=\directColumeFigWidth]{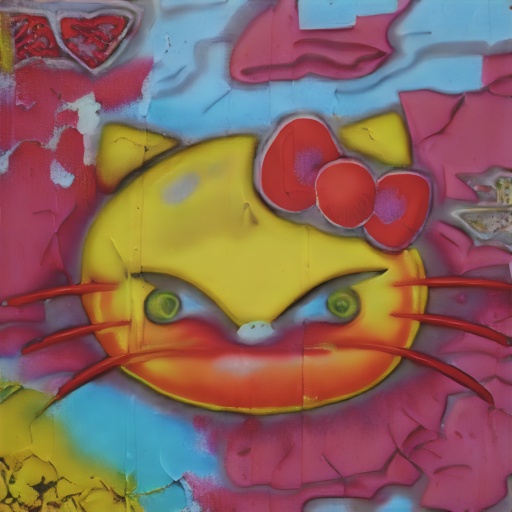} &
\includegraphics[width=\directColumeFigWidth]{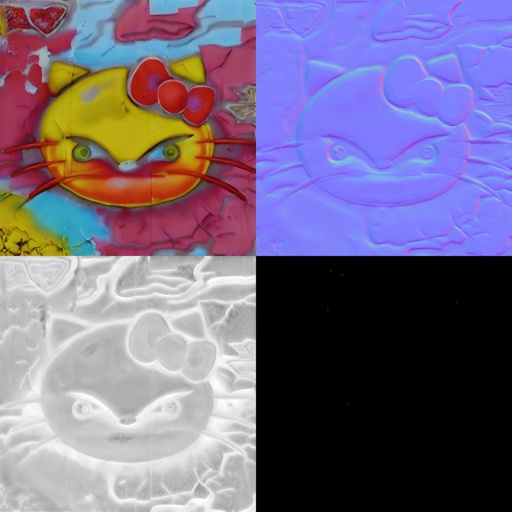} &
\scriptsize\textit{a PBR material of brick, multi-colored street bricks, vibrant, outdoor} &
\includegraphics[width=\directColumeFigWidth]{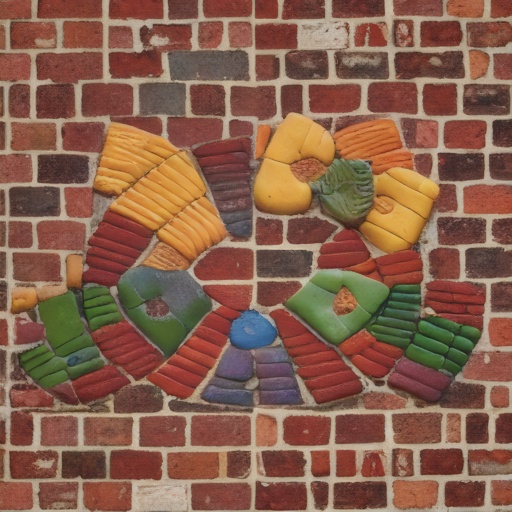} &
\includegraphics[width=\directColumeFigWidth]{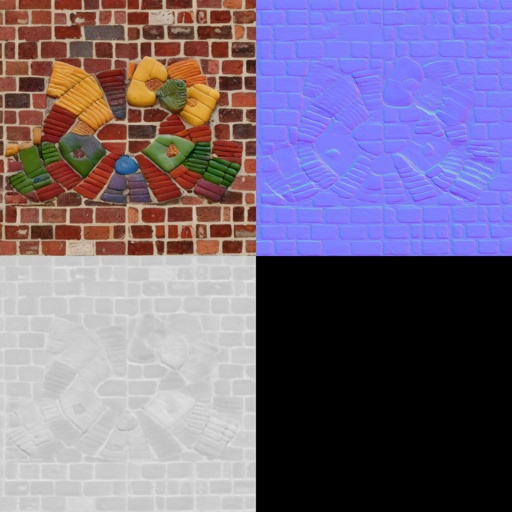} \\
&
\scriptsize\textit{a PBR material of fabric, embroidered linen, delicate, indoor, tablecloth} &
\includegraphics[width=\directColumeFigWidth]{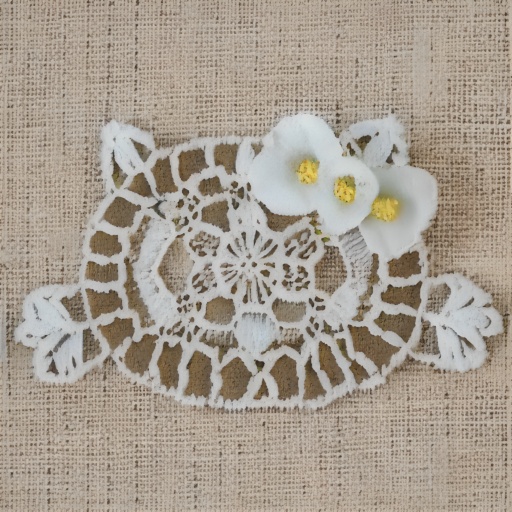} &
\includegraphics[width=\directColumeFigWidth]{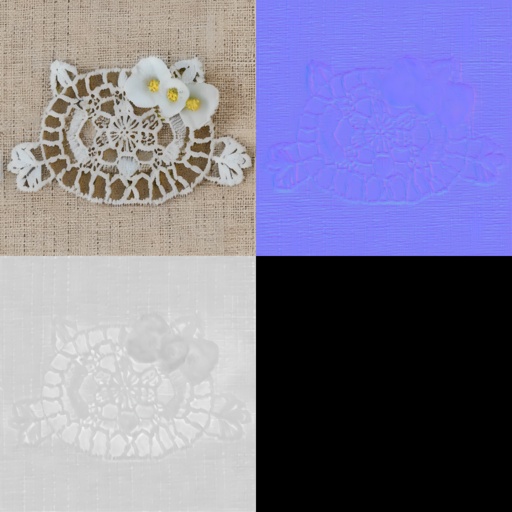} &
\scriptsize\textit{a PBR material of metal, metal plate, scifi} &
\includegraphics[width=\directColumeFigWidth]{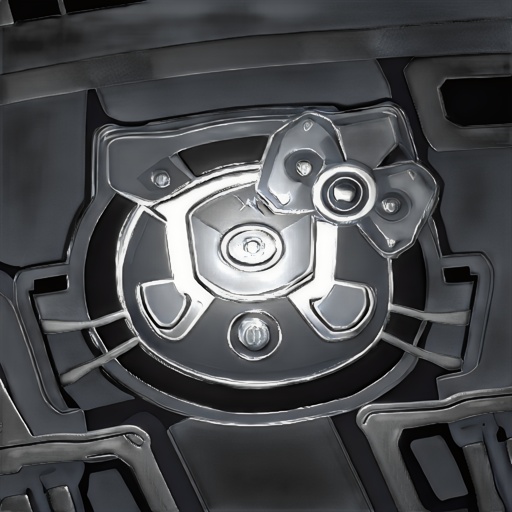} &
\includegraphics[width=\directColumeFigWidth]{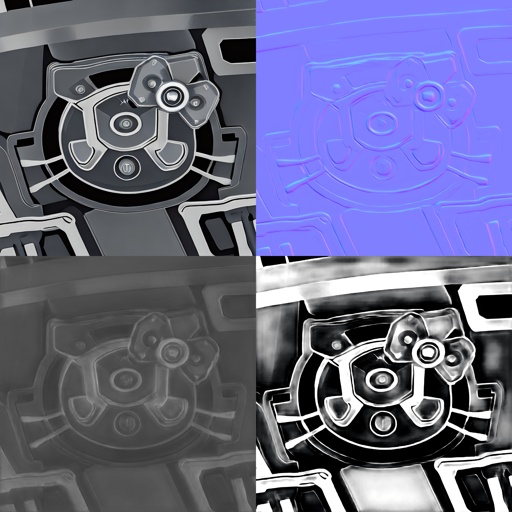} &

\scriptsize\textit{a PBR material of fabric, carpet, Hello Kitty outdoor picnic mat, durable, foldable} &
\includegraphics[width=\directColumeFigWidth]{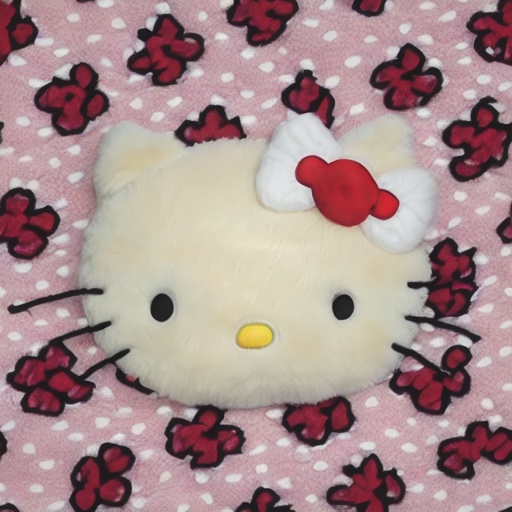} &
\includegraphics[width=\directColumeFigWidth]{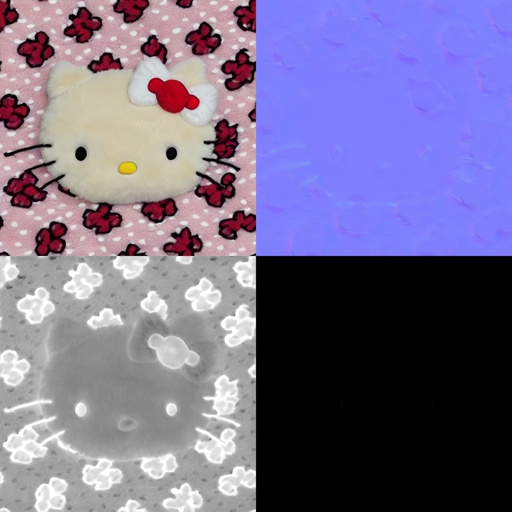} \\

\includegraphics[width=\directColumeFigWidth]{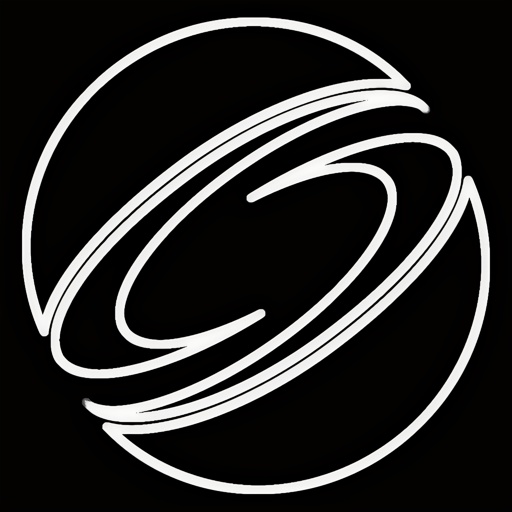} &
\scriptsize\textit{a PBR material of ground, marble floor tiles, polished, indoor, luxury} &
\includegraphics[width=\directColumeFigWidth]{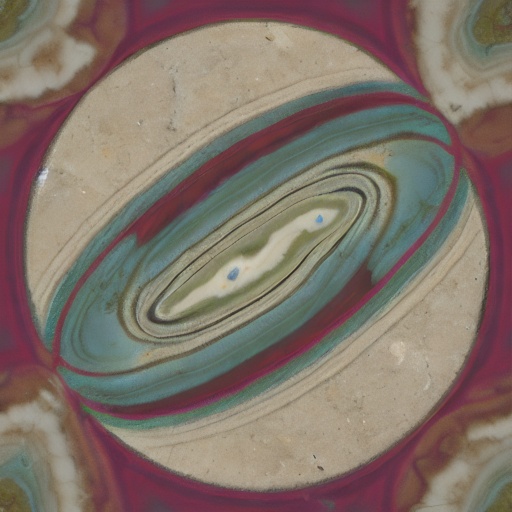} &
\includegraphics[width=\directColumeFigWidth]{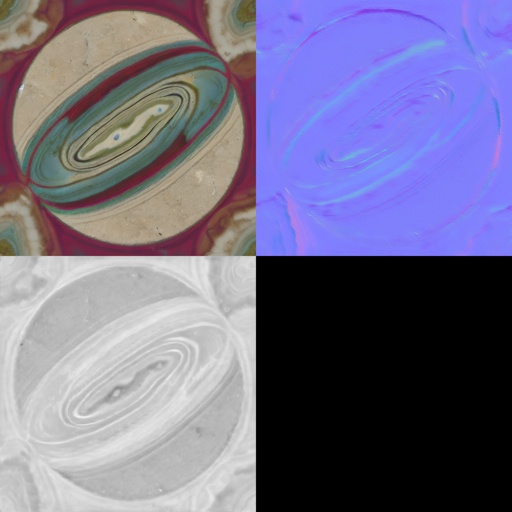} &
\scriptsize\textit{a PBR material of leather, black motorcycle jacket, tough, clothing, jacket} &
\includegraphics[width=\directColumeFigWidth]{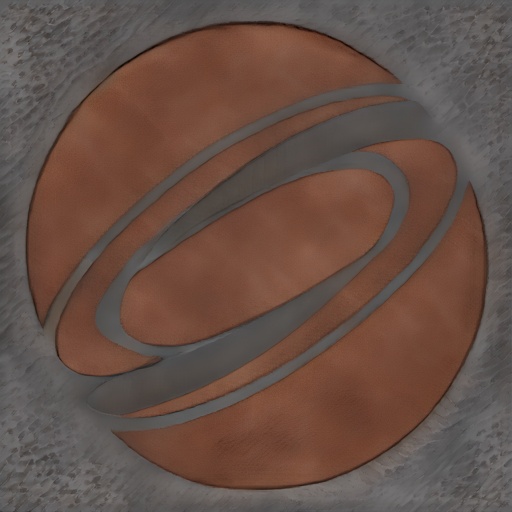} &
\includegraphics[width=\directColumeFigWidth]{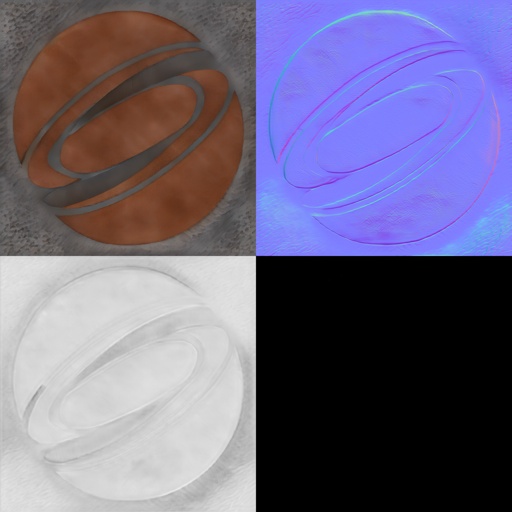} &
\scriptsize\textit{a PBR material of ground, forest leaves, natural, leaves, autumn} &
\includegraphics[width=\directColumeFigWidth]{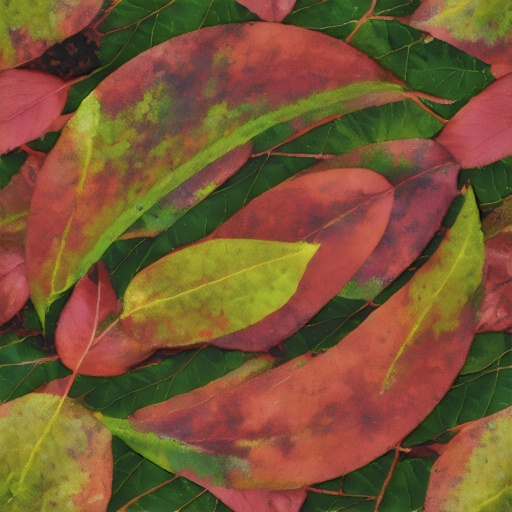} &
\includegraphics[width=\directColumeFigWidth]{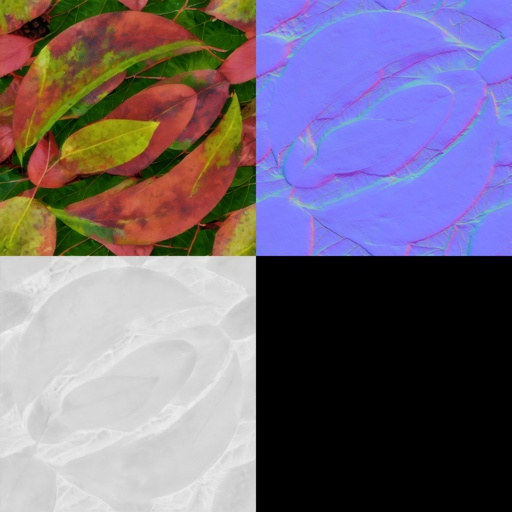} \\
&
\scriptsize\textit{a PBR material of metal, colored metal plate, scifi} &
\includegraphics[width=\directColumeFigWidth]{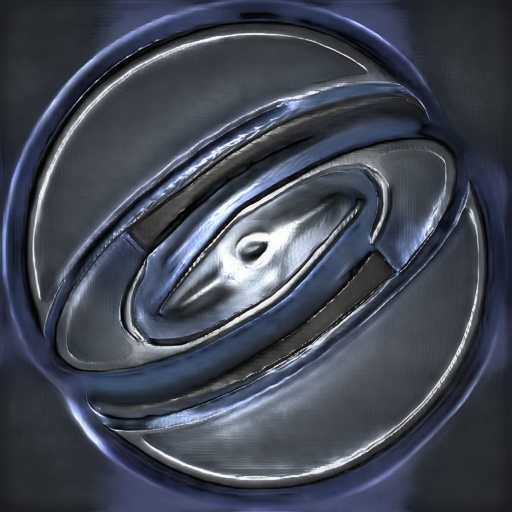} &
\includegraphics[width=\directColumeFigWidth]{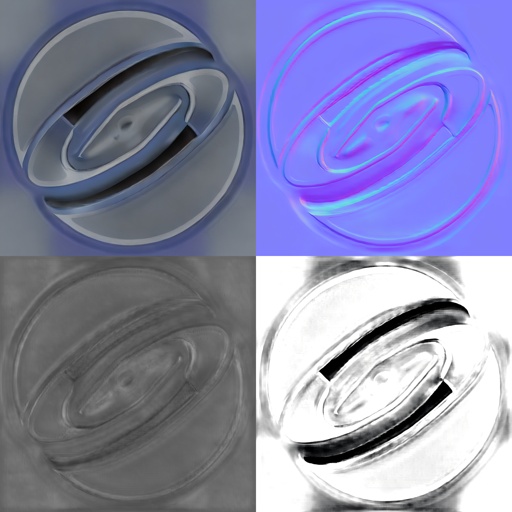} &
\scriptsize\textit{a PBR material of brick, stenciled brick floor, man made, worn, paving, dry, terracotta} &
\includegraphics[width=\directColumeFigWidth]{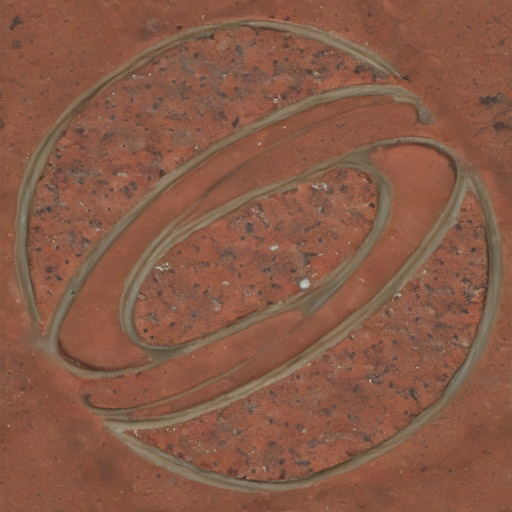} &
\includegraphics[width=\directColumeFigWidth]{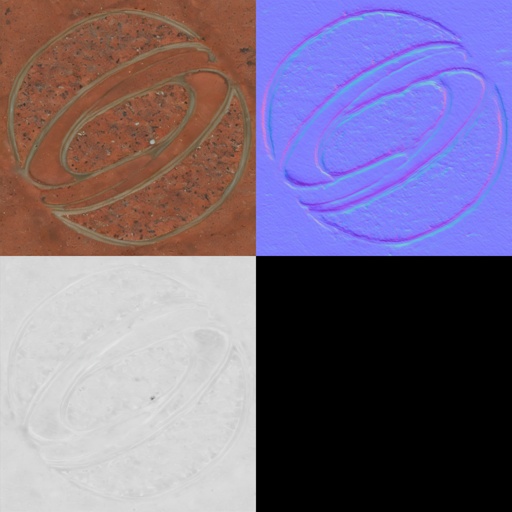} &
\scriptsize\textit{a PBR material of fabric, loose tablecloth} &
\includegraphics[width=\directColumeFigWidth]{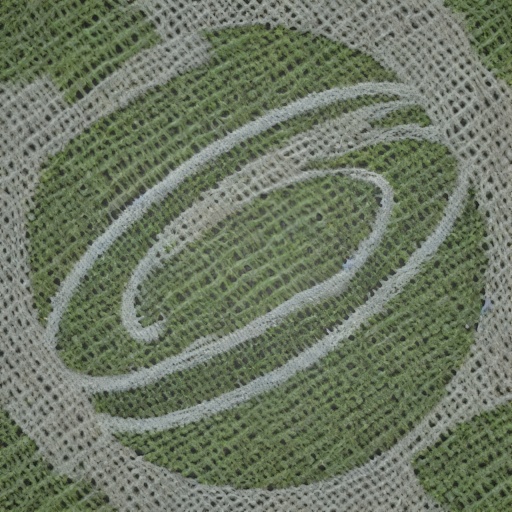} &
\includegraphics[width=\directColumeFigWidth]{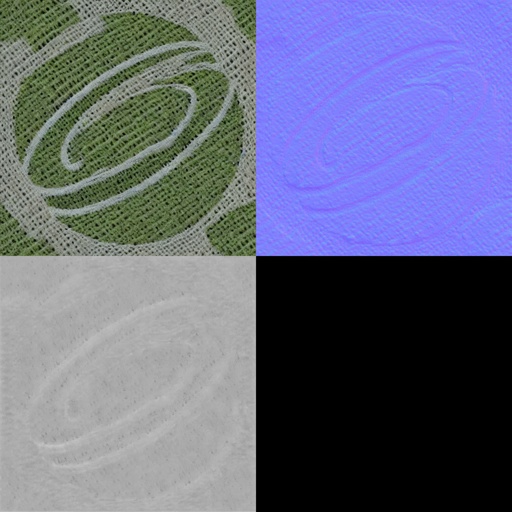} \\

\end{tabular}
\egroup

\caption{Pixel Control's results with the same pattern but different materials. The binary images in the first column are control conditions of different sketches and the generated materials are on their right with certain patterns same as our given images, following their material properties such as the edge of bricks and the growth rings of wood.} 
\label{fig:pixel_control_1} 
\Description{fig:pixel_control_1}
\end{figure*}

\newcommand{\myColumnWidth}{1.9cm} 

\begin{figure*}[htbp]
\centering 

\bgroup
\def\arraystretch{0.2}
\setlength\tabcolsep{1.6pt}
\begin{tabular}{
    >{\raggedright\arraybackslash}m{\myColumnWidth} 
    m{\myColumnWidth}
    @{\hspace{0.8pt}}m{\myColumnWidth}
    m{\myColumnWidth}
    @{\hspace{0.8pt}}m{\myColumnWidth}
    m{\myColumnWidth}
    @{\hspace{0.8pt}}m{\myColumnWidth}
    m{\myColumnWidth}
    @{\hspace{0.8pt}}m{\myColumnWidth}
    }
\multicolumn{1}{c}{Prompt} & \multicolumn{1}{c}{Render} & \multicolumn{1}{c}{SVBRDF} & \multicolumn{1}{c}{Render} & \multicolumn{1}{c}{SVBRDF} &\multicolumn{1}{c}{Render} & \multicolumn{1}{c}{SVBRDF} &\multicolumn{1}{c}{Render} & \multicolumn{1}{c}{SVBRDF} \\
\scriptsize\textit{a PBR material of metal, space cruiser panels} & 
\includegraphics[width=\myColumnWidth]{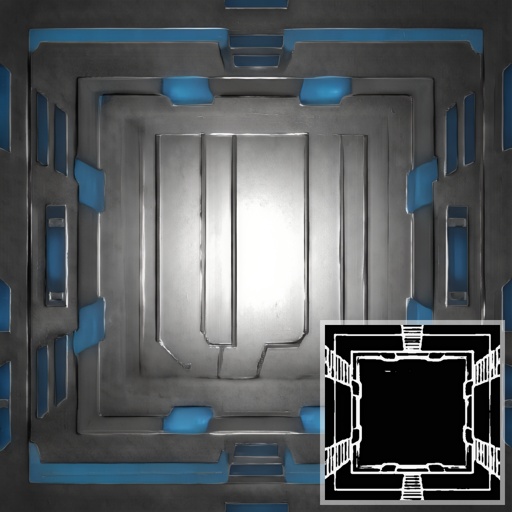} &
\includegraphics[width=\myColumnWidth]{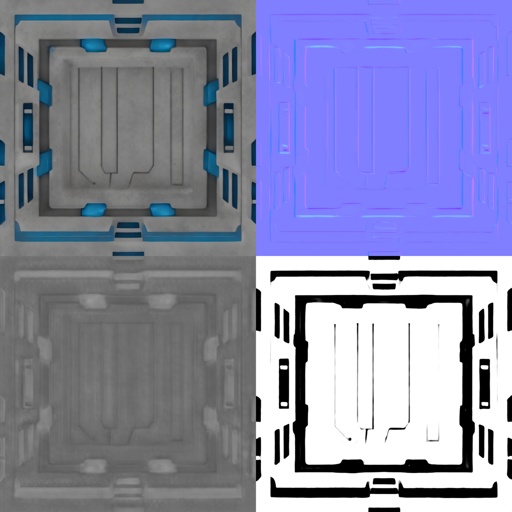} &
\includegraphics[width=\myColumnWidth]{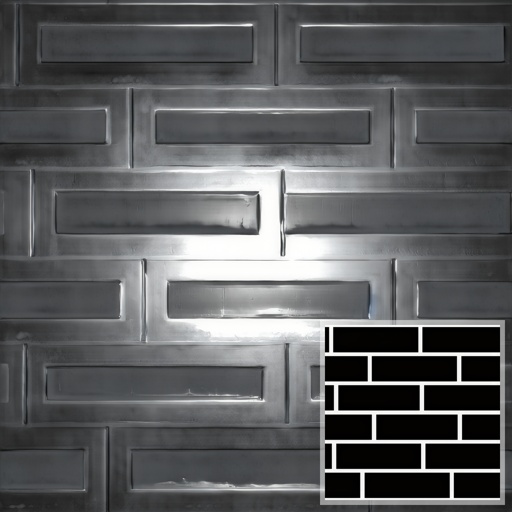} &
\includegraphics[width=\myColumnWidth]{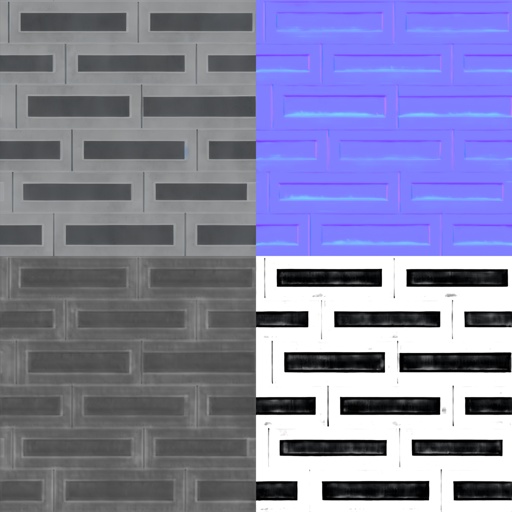} &
\includegraphics[width=\myColumnWidth]{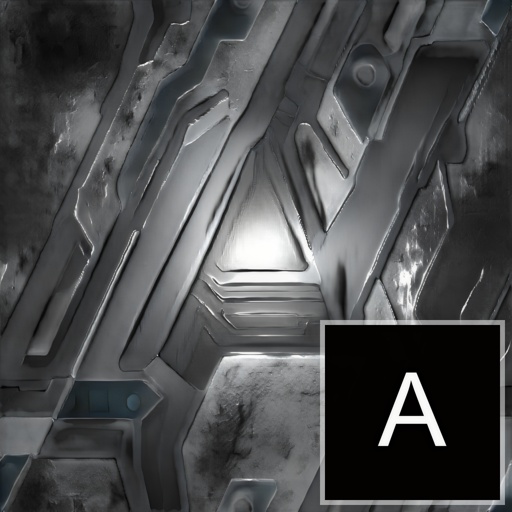} &
\includegraphics[width=\myColumnWidth]{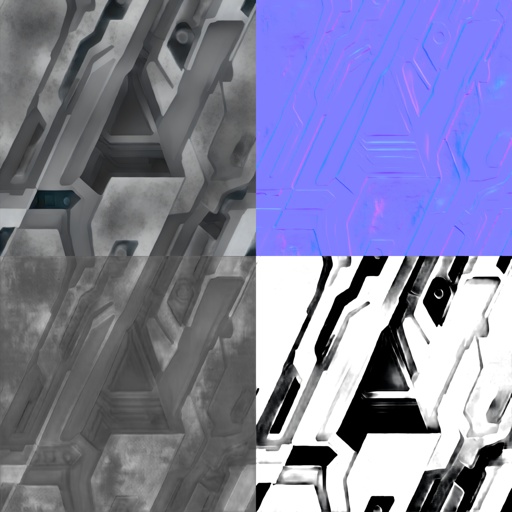} &
\includegraphics[width=\myColumnWidth]{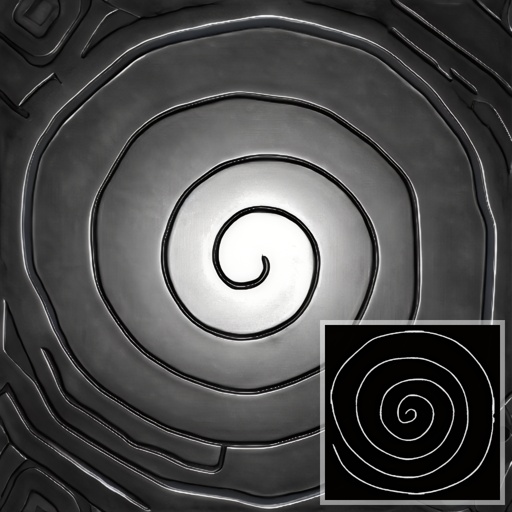} &
\includegraphics[width=\myColumnWidth]{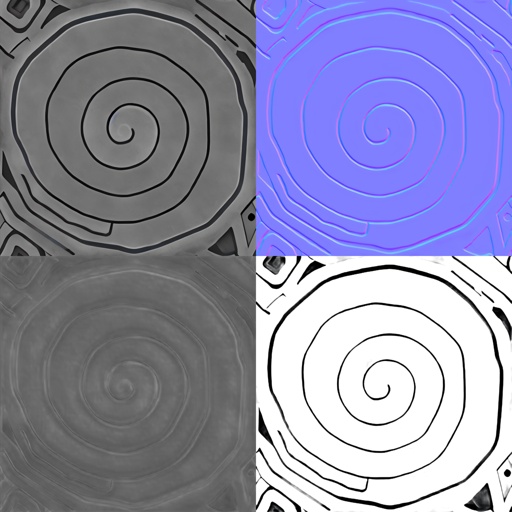} \\
\scriptsize\textit{a PBR material of wall, street art graffiti, colorful, outdoor, urban} & 
\includegraphics[width=\myColumnWidth]{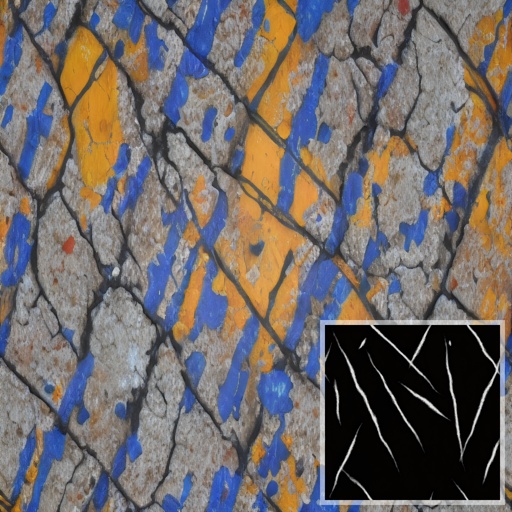} &
\includegraphics[width=\myColumnWidth]{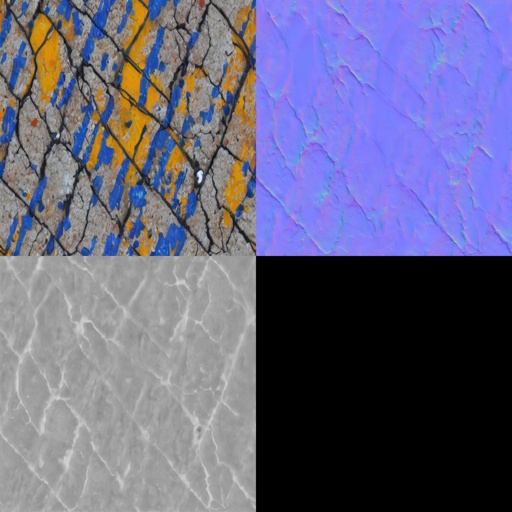} &
\includegraphics[width=\myColumnWidth]{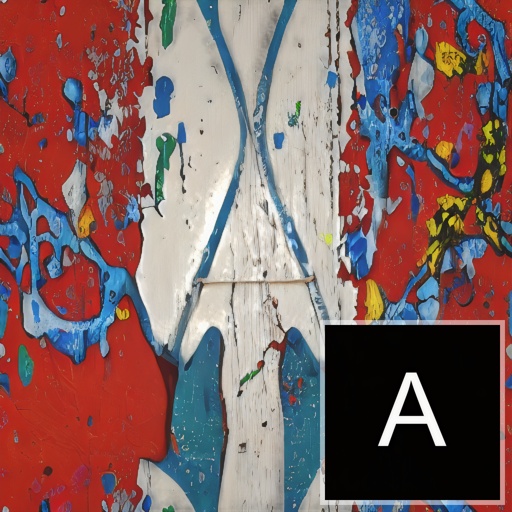} &
\includegraphics[width=\myColumnWidth]{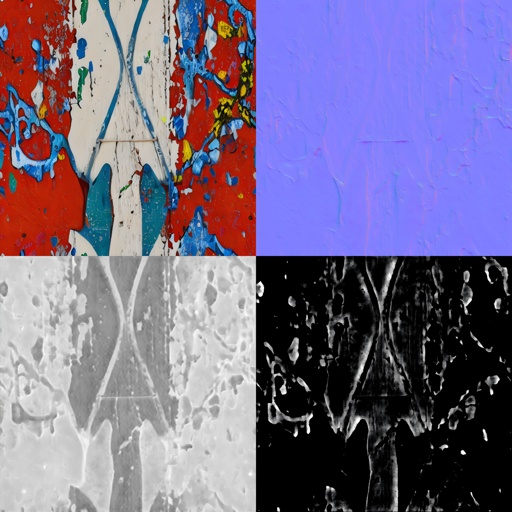} &
\includegraphics[width=\myColumnWidth]{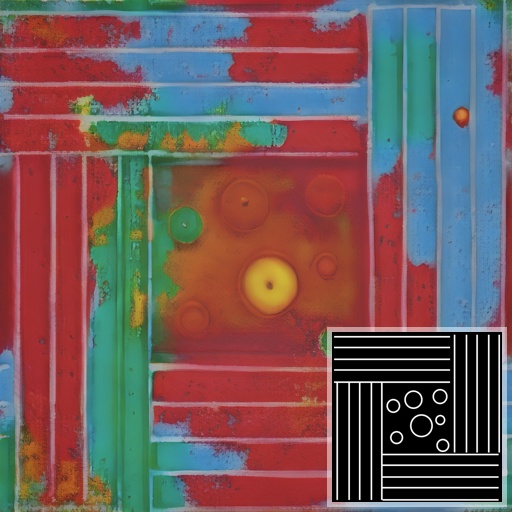} &
\includegraphics[width=\myColumnWidth]{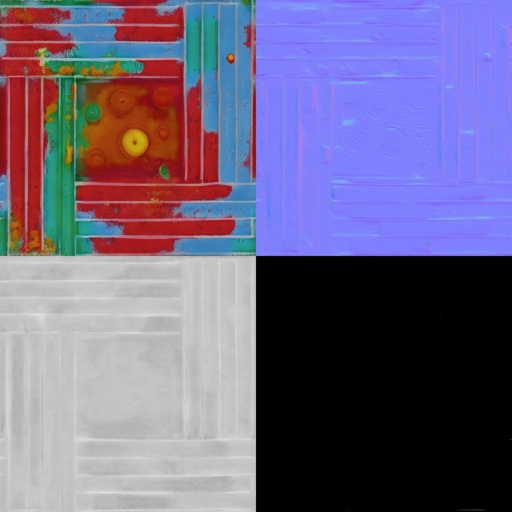} &
\includegraphics[width=\myColumnWidth]{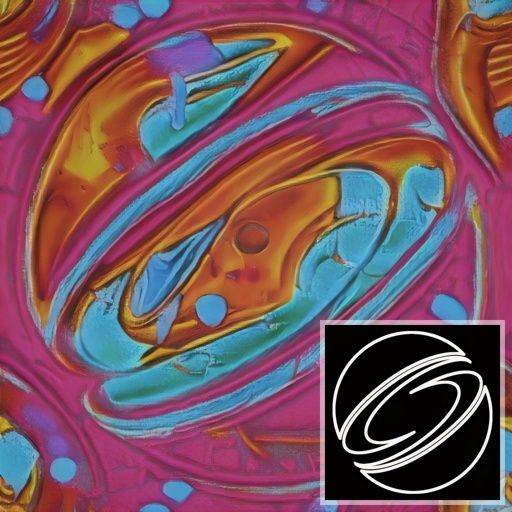} &
\includegraphics[width=\myColumnWidth]{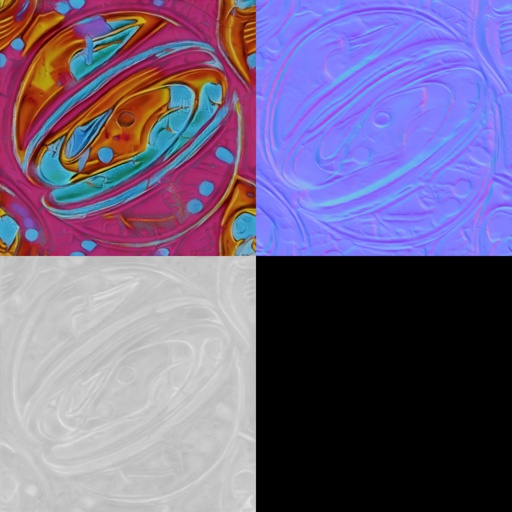} \\
\scriptsize\textit{a PBR material of tiles, encaustic cement tiles, colorful, indoor, floor} & 
\includegraphics[width=\myColumnWidth]{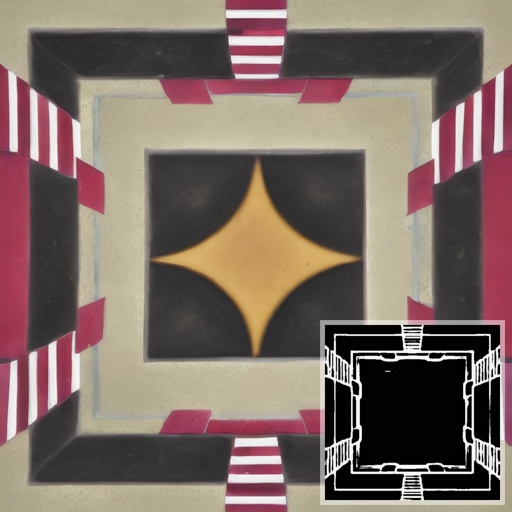} &
\includegraphics[width=\myColumnWidth]{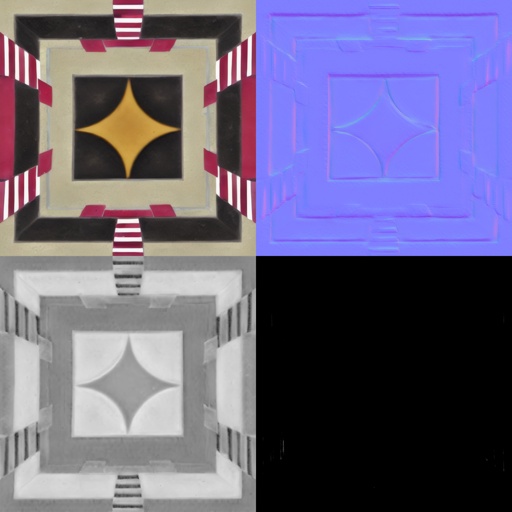} &
\includegraphics[width=\myColumnWidth]{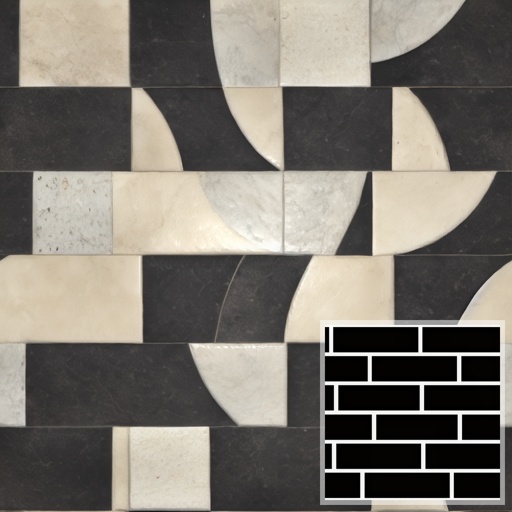} &
\includegraphics[width=\myColumnWidth]{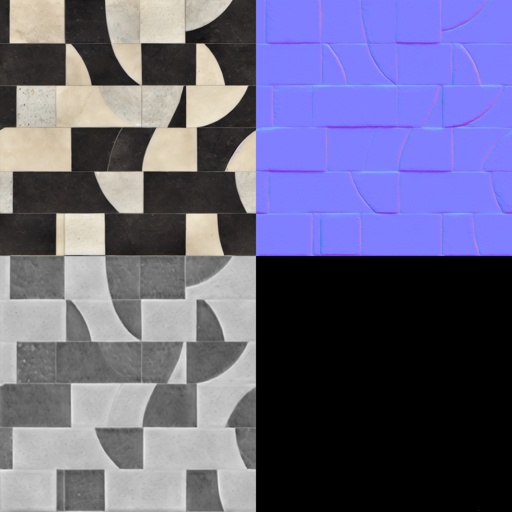} &
\includegraphics[width=\myColumnWidth]{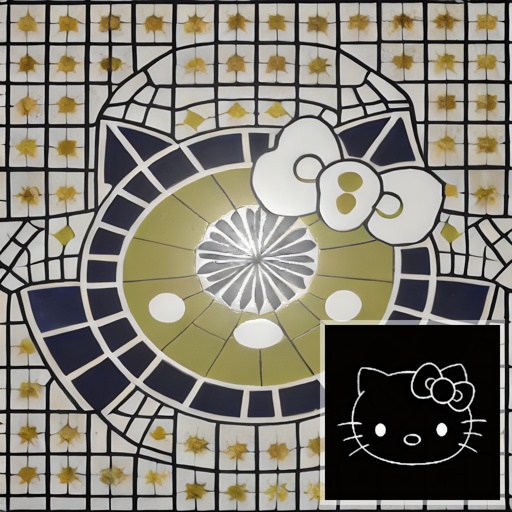} &
\includegraphics[width=\myColumnWidth]{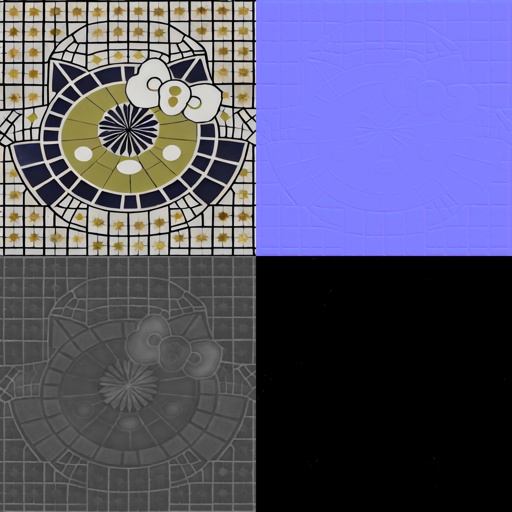} &
\includegraphics[width=\myColumnWidth]{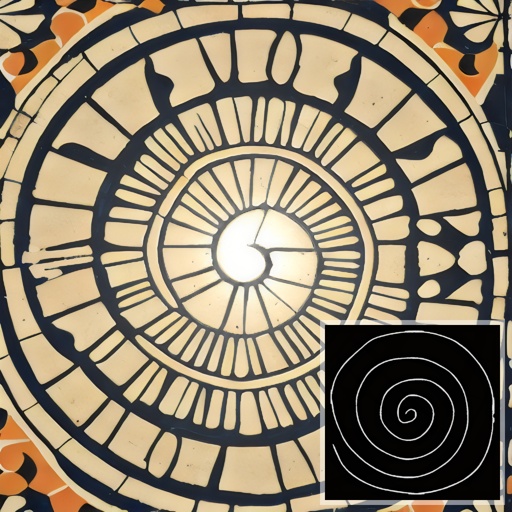} &
\includegraphics[width=\myColumnWidth]{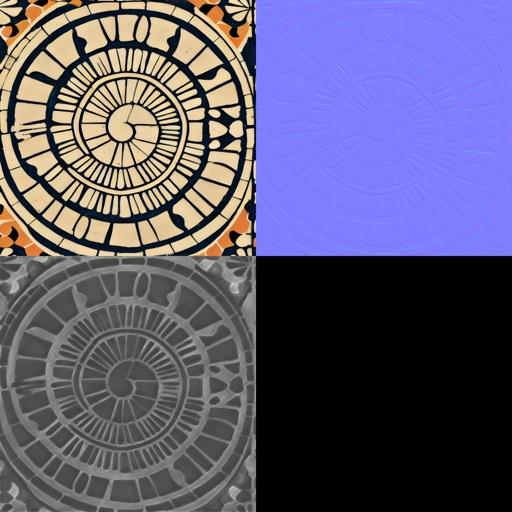} \\
\end{tabular}
\egroup

\caption{Pixel Control's results with the same material but different patterns. The first column shows the descriptions of materials and the control patterns are in the lower right corner of the rendering images. Like \autoref{fig:pixel_control_1}, the results have also shown great consistency of materials and patterns.} 

\label{fig:pixel_control_2} 
\Description{fig:pixel_control_2}
\end{figure*}

\newcommand{\StyleColumnWidth}{1.8cm} 
\newcommand{\StyleTextColumnWidth}{3.6cm} 
\begin{figure*}[htbp]
\centering 

\bgroup

\def\arraystretch{0.1} 
\setlength\tabcolsep{1pt} 
\begin{tabular}{
    m{\StyleColumnWidth}
    m{\StyleColumnWidth}
    @{\hspace{0.5pt}}m{\StyleColumnWidth}
    m{\StyleColumnWidth}
    @{\hspace{0.5pt}}m{\StyleColumnWidth}
    m{\StyleColumnWidth}
    @{\hspace{0.5pt}}m{\StyleColumnWidth}
    m{\StyleColumnWidth}
    @{\hspace{0.5pt}}m{\StyleColumnWidth}
    }
    
\multicolumn{1}{c}{Style} & \multicolumn{1}{c}{Render} & \multicolumn{1}{c}{SVBRDF} & \multicolumn{1}{c}{Render} & \multicolumn{1}{c}{SVBRDF} & \multicolumn{1}{c}{Render} & \multicolumn{1}{c}{SVBRDF} & \multicolumn{1}{c}{Render} & \multicolumn{1}{c}{SVBRDF}\\

\includegraphics[width=\StyleColumnWidth]{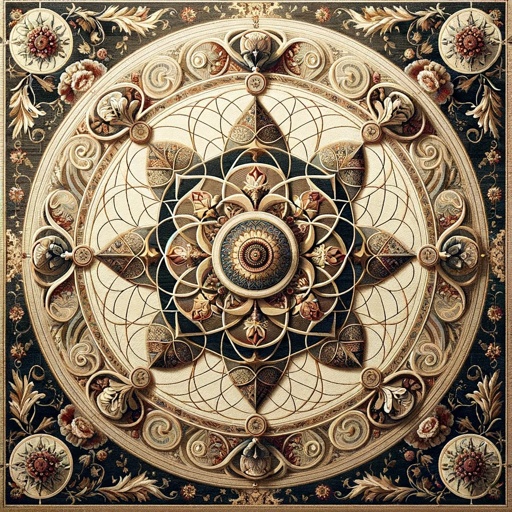} &
\includegraphics[width=\StyleColumnWidth]{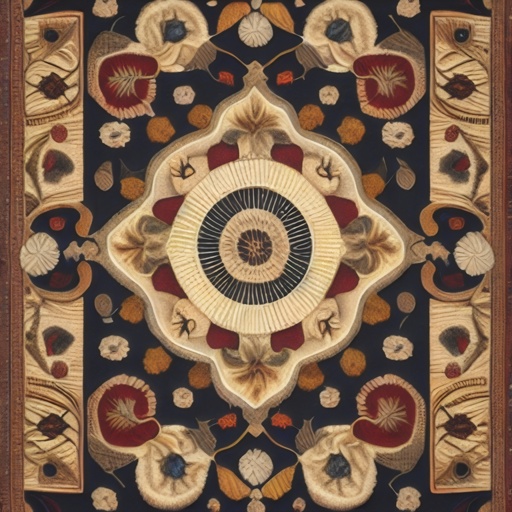} &
\includegraphics[width=\StyleColumnWidth]{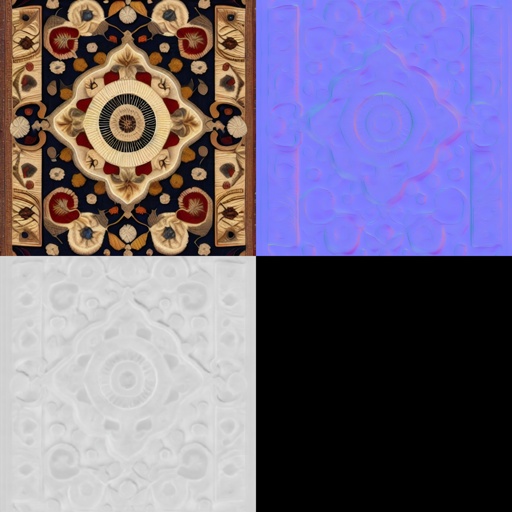} &
\includegraphics[width=\StyleColumnWidth]{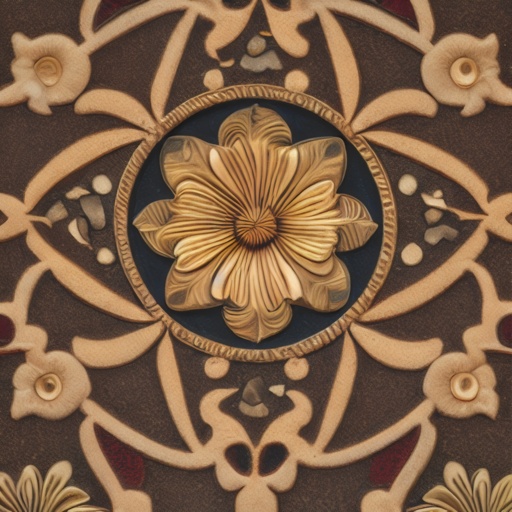} &
\includegraphics[width=\StyleColumnWidth]{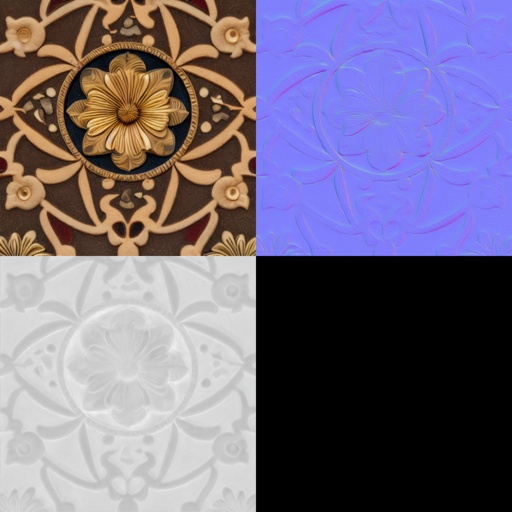} &
\includegraphics[width=\StyleColumnWidth]{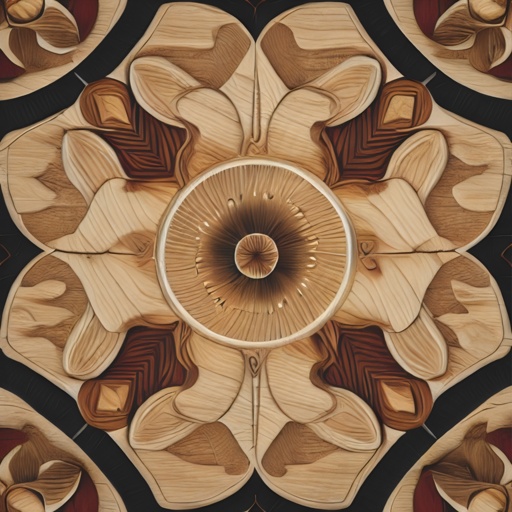} &
\includegraphics[width=\StyleColumnWidth]{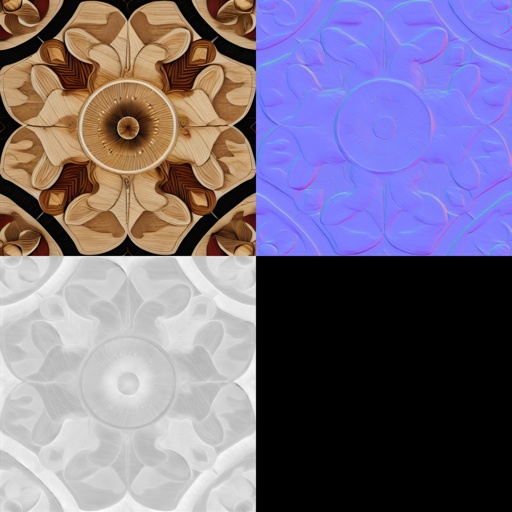} &
\includegraphics[width=\StyleColumnWidth]{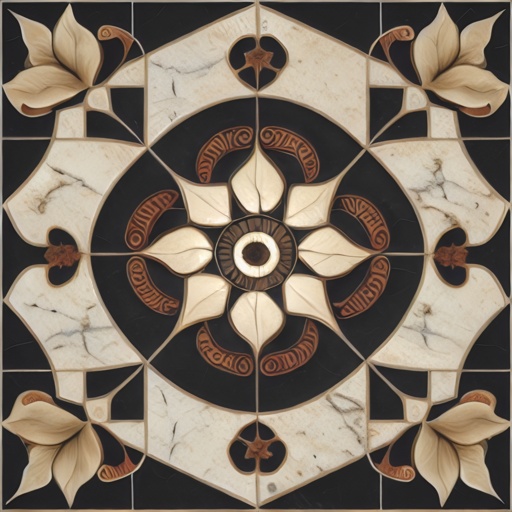} &
\includegraphics[width=\StyleColumnWidth]{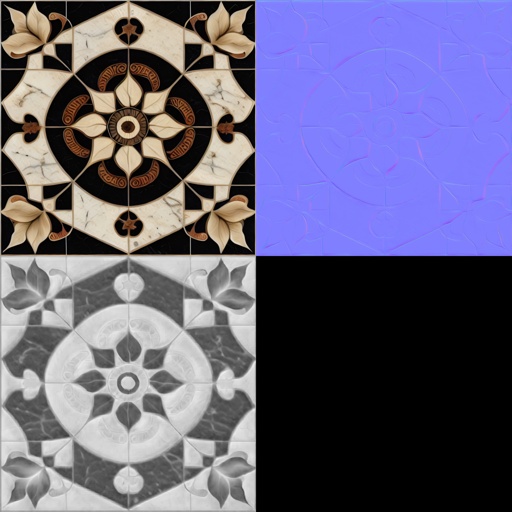} \\

&
\multicolumn{2}{c}{
  \begin{minipage}{\StyleTextColumnWidth}
    \centering
    \scriptsize\textit{a PBR material of fabric, carpet} 
\end{minipage} 
} &
\multicolumn{2}{c}{
  \begin{minipage}{\StyleTextColumnWidth}
    \centering
    \scriptsize\textit{a PBR material of ground, stone, outdoor} 
\end{minipage} 
} &
\multicolumn{2}{c}{
  \begin{minipage}{\StyleTextColumnWidth}
    \centering
    \scriptsize\textit{a PBR material of wood} 
\end{minipage} 
} &
\multicolumn{2}{c}{
  \begin{minipage}{\StyleTextColumnWidth}
    \centering
    \scriptsize\textit{a PBR material of tile, marble} 
\end{minipage} 
} \\
\noalign{\smallskip}

\includegraphics[width=\StyleColumnWidth]{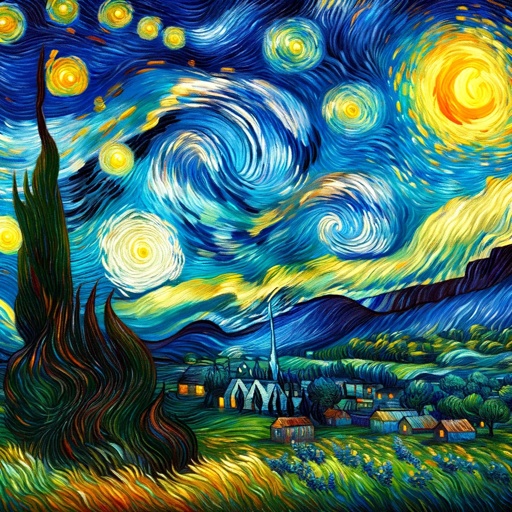} &
\includegraphics[width=\StyleColumnWidth]{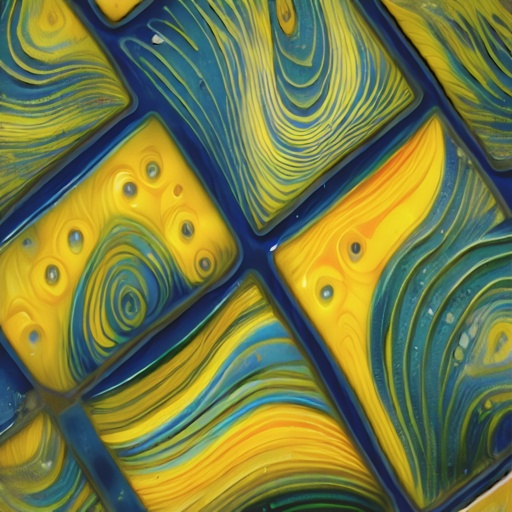} &
\includegraphics[width=\StyleColumnWidth]{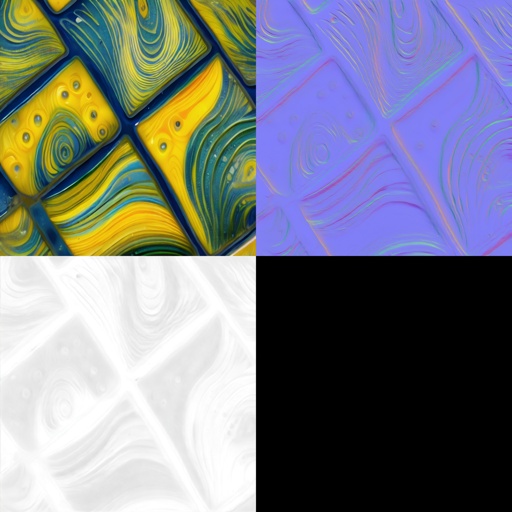} &
\includegraphics[width=\StyleColumnWidth]{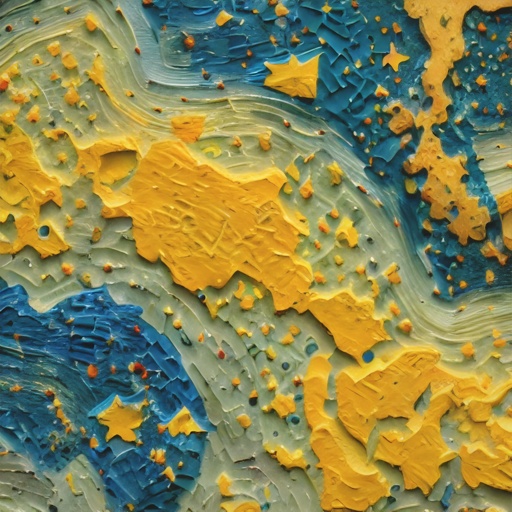} &
\includegraphics[width=\StyleColumnWidth]{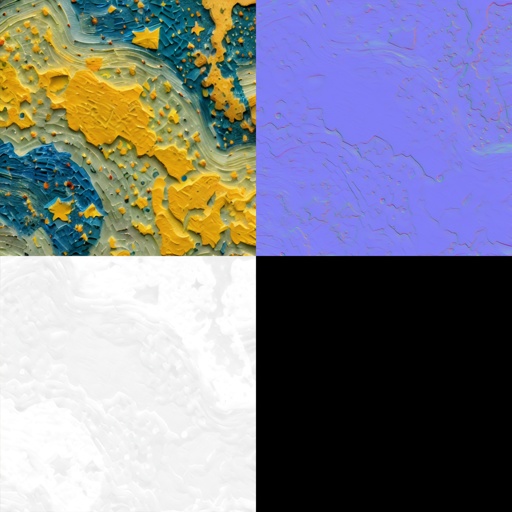} &
\includegraphics[width=\StyleColumnWidth]{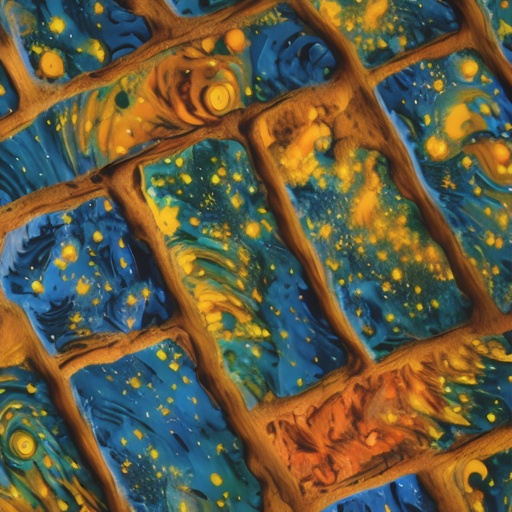} &
\includegraphics[width=\StyleColumnWidth]{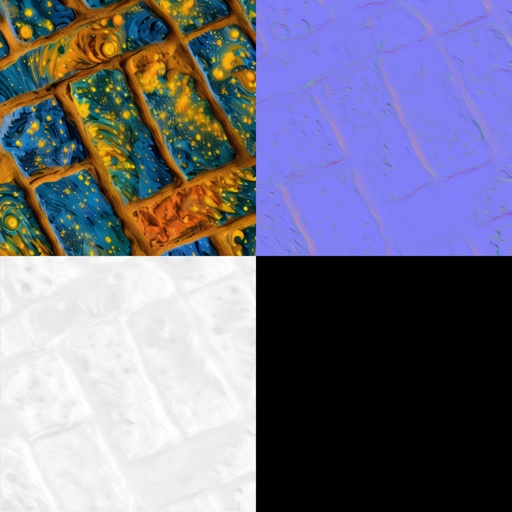} &
\includegraphics[width=\StyleColumnWidth]{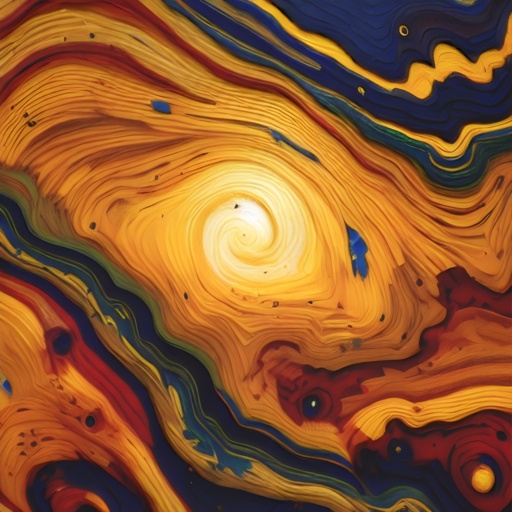} &
\includegraphics[width=\StyleColumnWidth]{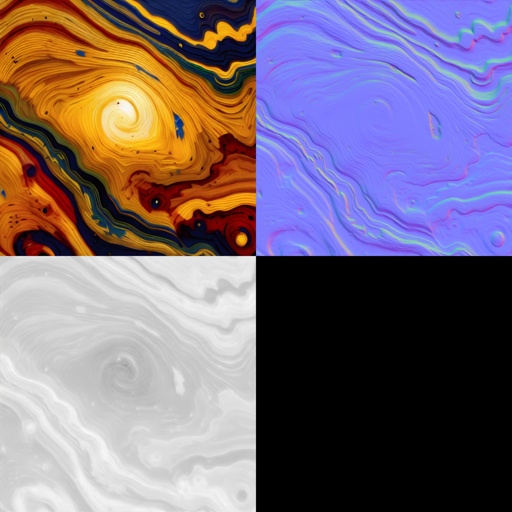} \\

&
\multicolumn{2}{c}{
  \begin{minipage}{\StyleTextColumnWidth}
    \centering
    \scriptsize\textit{a PBR material of tile, encaustic cement} 
\end{minipage} 
} &
\multicolumn{2}{c}{
  \begin{minipage}{\StyleTextColumnWidth}
    \centering
    \scriptsize\textit{a PBR material of wall, concrete wall, outdoor, cracked, man made, rough, painted} 
\end{minipage} 
} &
\multicolumn{2}{c}{
  \begin{minipage}{\StyleTextColumnWidth}
    \centering
    \scriptsize\textit{a PBR material of brick, street brick, outdoor} 
\end{minipage} 
} &
\multicolumn{2}{c}{
  \begin{minipage}{\StyleTextColumnWidth}
    \centering
    \scriptsize\textit{a PBR material of wood, varnished walnut, painted, artistic} 
\end{minipage} 
} \\
\noalign{\smallskip}

\includegraphics[width=\StyleColumnWidth]{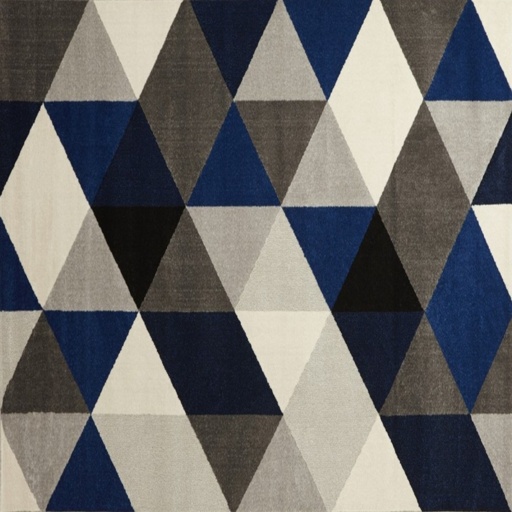} &
\includegraphics[width=\StyleColumnWidth]{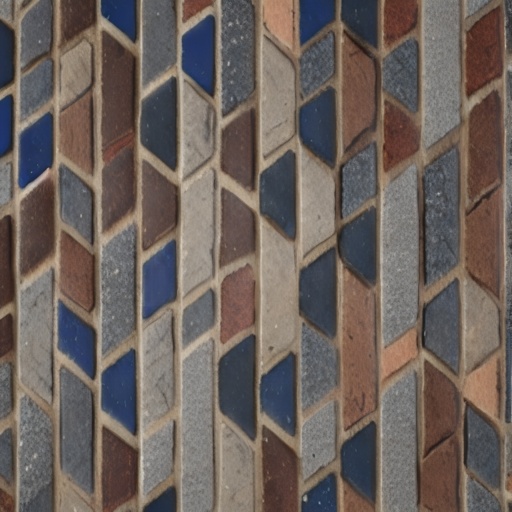} &
\includegraphics[width=\StyleColumnWidth]{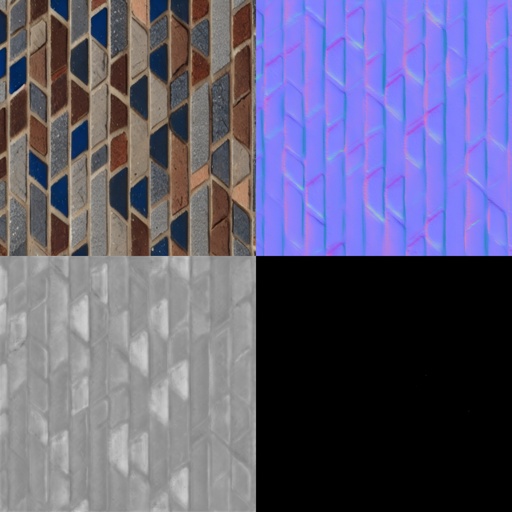} &
\includegraphics[width=\StyleColumnWidth]{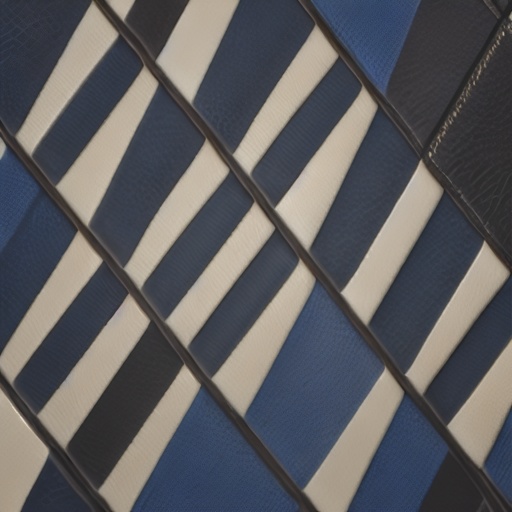} &
\includegraphics[width=\StyleColumnWidth]{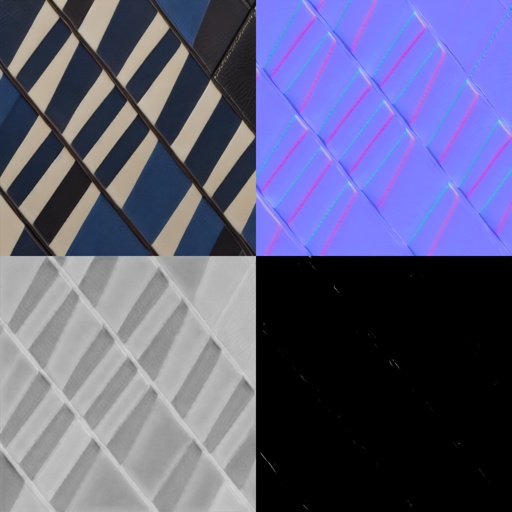} &
\includegraphics[width=\StyleColumnWidth]{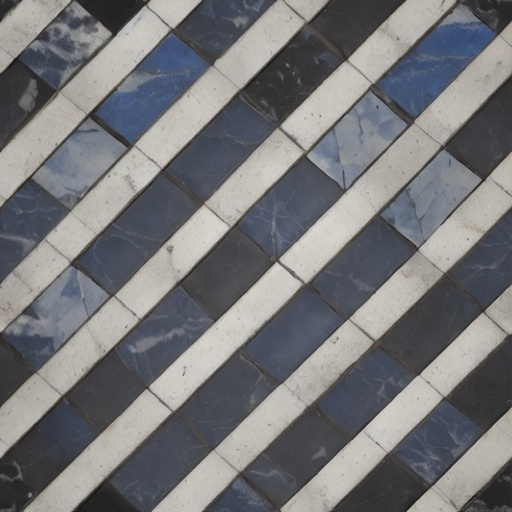} &
\includegraphics[width=\StyleColumnWidth]{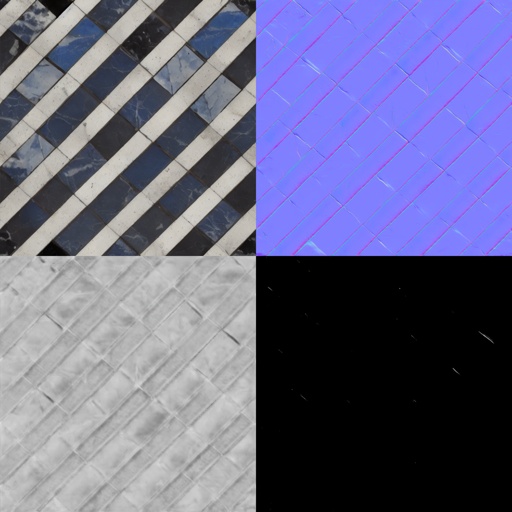} &
\includegraphics[width=\StyleColumnWidth]{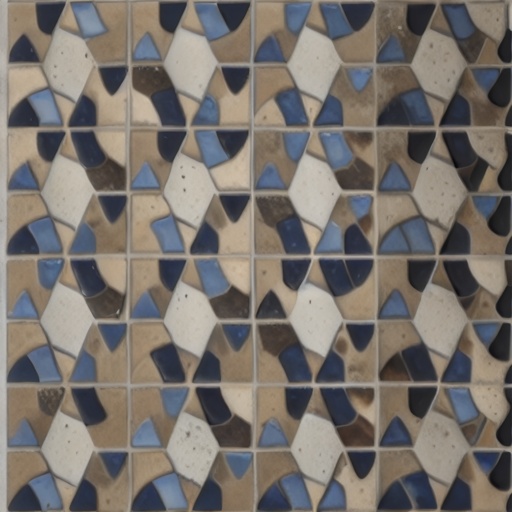} &
\includegraphics[width=\StyleColumnWidth]{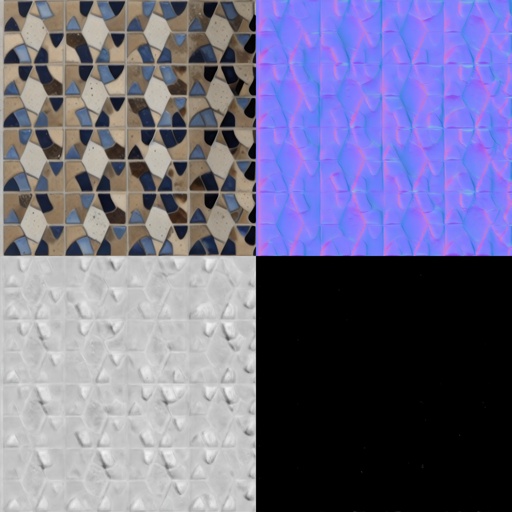} \\

&
\multicolumn{2}{c}{
  \begin{minipage}{\StyleTextColumnWidth}
    \centering
    \scriptsize\textit{a PBR material of brick, street brick, outdoor, art} 
\end{minipage} 
} &
\multicolumn{2}{c}{
  \begin{minipage}{\StyleTextColumnWidth}
    \centering
    \scriptsize\textit{a PBR material of leather} 
\end{minipage} 
} &
\multicolumn{2}{c}{
  \begin{minipage}{\StyleTextColumnWidth}
    \centering
    \scriptsize\textit{a PBR material of ground, sidewalk} 
\end{minipage} 
} &
\multicolumn{2}{c}{
  \begin{minipage}{\StyleTextColumnWidth}
    \centering
    \scriptsize\textit{a PBR material of tile, encaustic cement} 
\end{minipage} 
} \\
\noalign{\smallskip}

\includegraphics[width=\StyleColumnWidth]{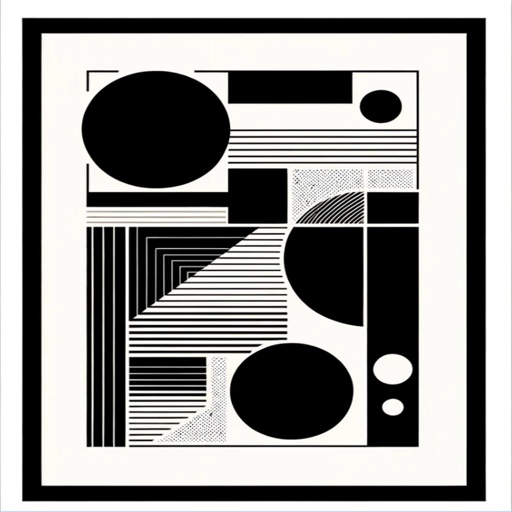} &
\includegraphics[width=\StyleColumnWidth]{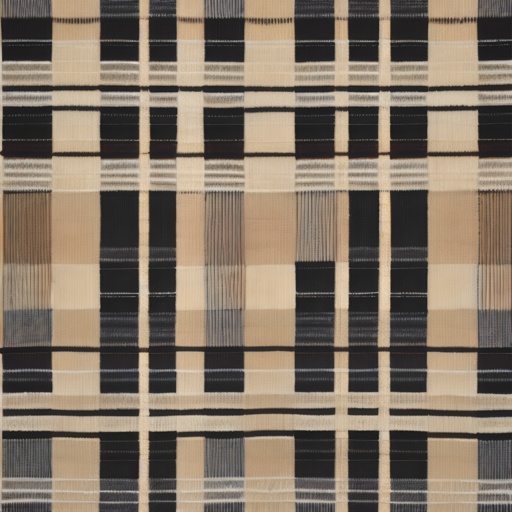} &
\includegraphics[width=\StyleColumnWidth]{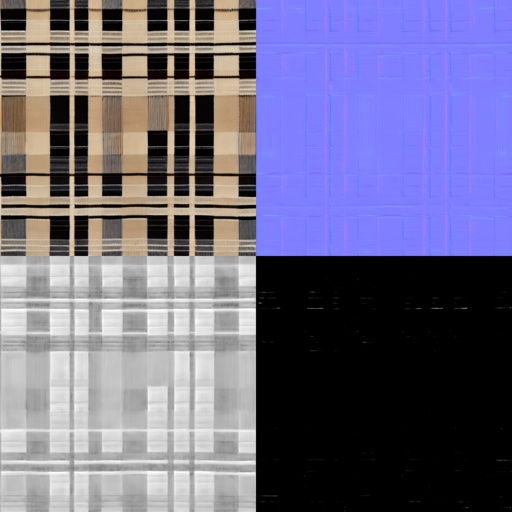} &
\includegraphics[width=\StyleColumnWidth]{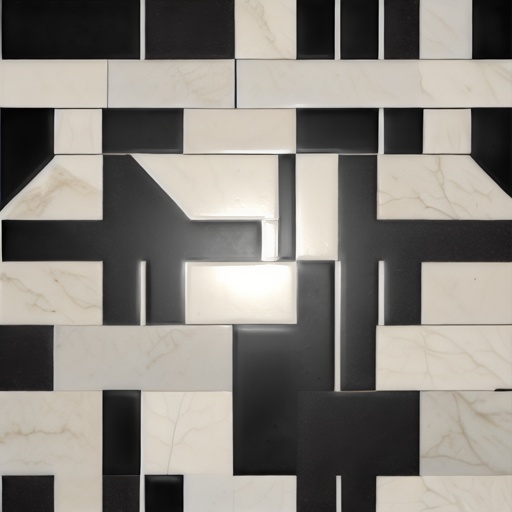} &
\includegraphics[width=\StyleColumnWidth]{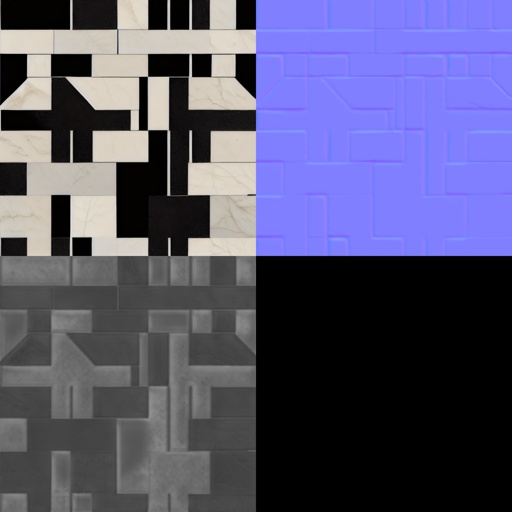} &
\includegraphics[width=\StyleColumnWidth]{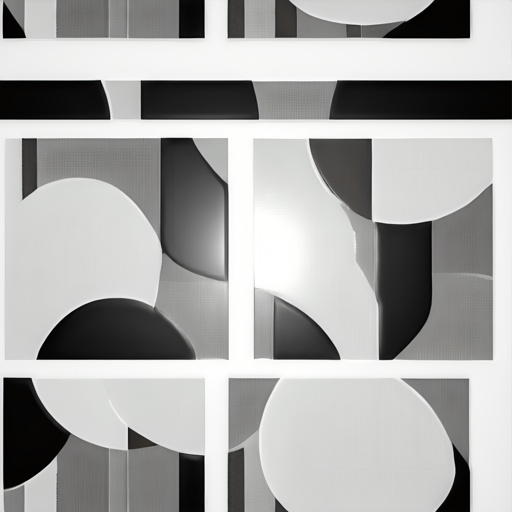} &
\includegraphics[width=\StyleColumnWidth]{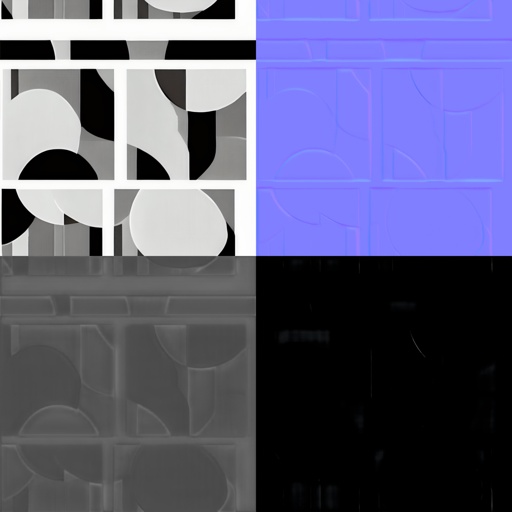} &
\includegraphics[width=\StyleColumnWidth]{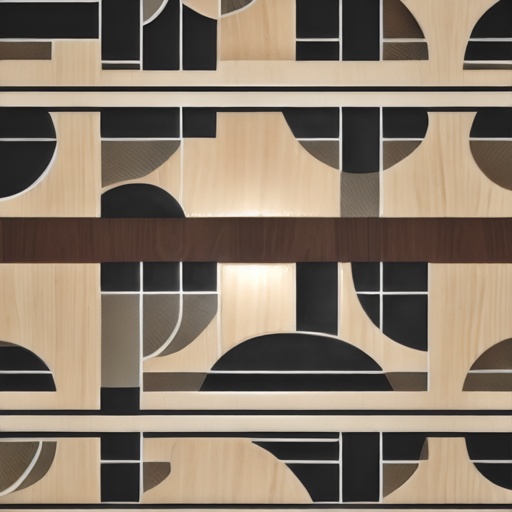} &
\includegraphics[width=\StyleColumnWidth]{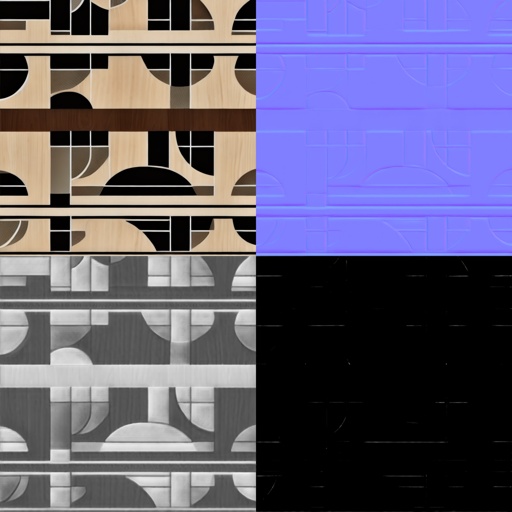} \\

&
\multicolumn{2}{c}{
  \begin{minipage}{\StyleTextColumnWidth}
    \centering
    \scriptsize\textit{a PBR material of fabric, hand woven carpet} 
\end{minipage} 
} &
\multicolumn{2}{c}{
  \begin{minipage}{\StyleTextColumnWidth}
    \centering
    \scriptsize\textit{a PBR material of tile, marble} 
\end{minipage} 
} &
\multicolumn{2}{c}{
  \begin{minipage}{\StyleTextColumnWidth}
    \centering
    \scriptsize\textit{a PBR material of wall, wallpaper} 
\end{minipage} 
} &
\multicolumn{2}{c}{
  \begin{minipage}{\StyleTextColumnWidth}
    \centering
    \scriptsize\textit{a PBR material of wood, synthetic wood, painted} 
\end{minipage} 
} \\

\end{tabular}
\egroup
\caption{Style Control's results with the same style but different materials. The styled images are given in the first column, and each description of the material is below the image, which provides users with more artistic ways to design textures.} % 图形的标题
\label{fig:style_control} 
\Description{fig:style_control}
\end{figure*}

\newcommand{\SeedWidth}{2.6cm} 
\begin{figure}[htbp]
\centering 

\bgroup

\def\arraystretch{0.2}
\setlength\tabcolsep{0.5pt} 

\begin{tabular}{
    m{\SeedWidth}
    m{\SeedWidth}
    m{\SeedWidth}
    }

\multicolumn{3}{c}{
  \begin{minipage}{7.8cm}
    \centering
    \small\textit{a PBR material of wood} 
\end{minipage} 
} \\

\includegraphics[width=\SeedWidth]{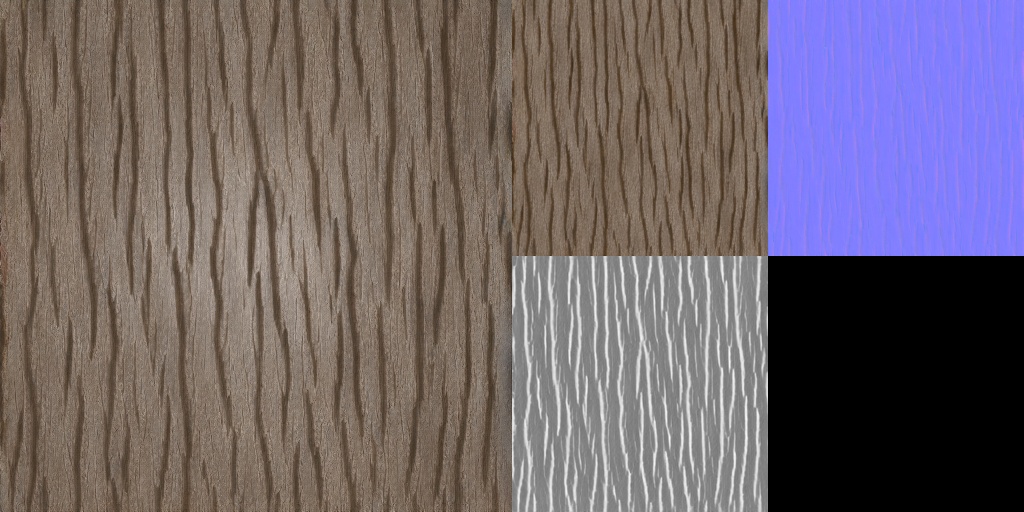} &
\includegraphics[width=\SeedWidth]{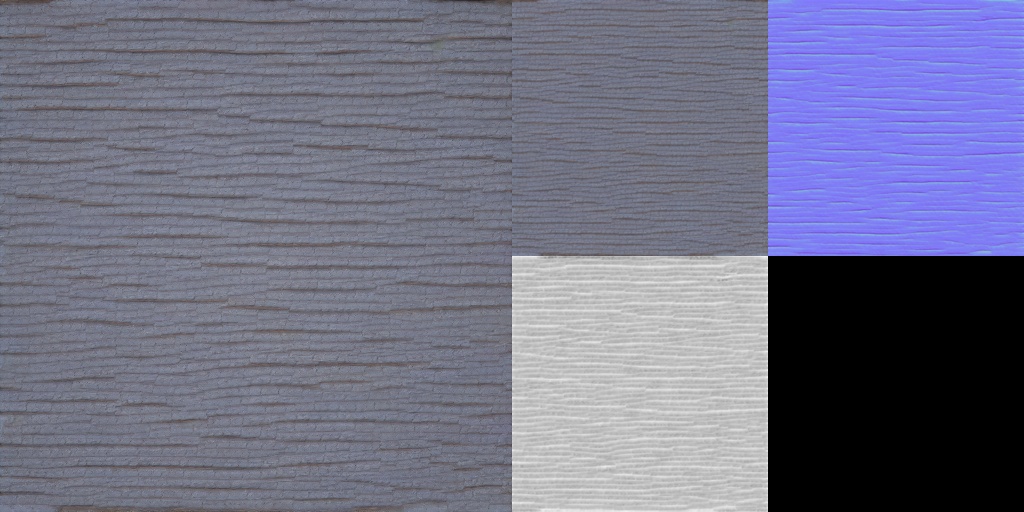} &
\includegraphics[width=\SeedWidth]{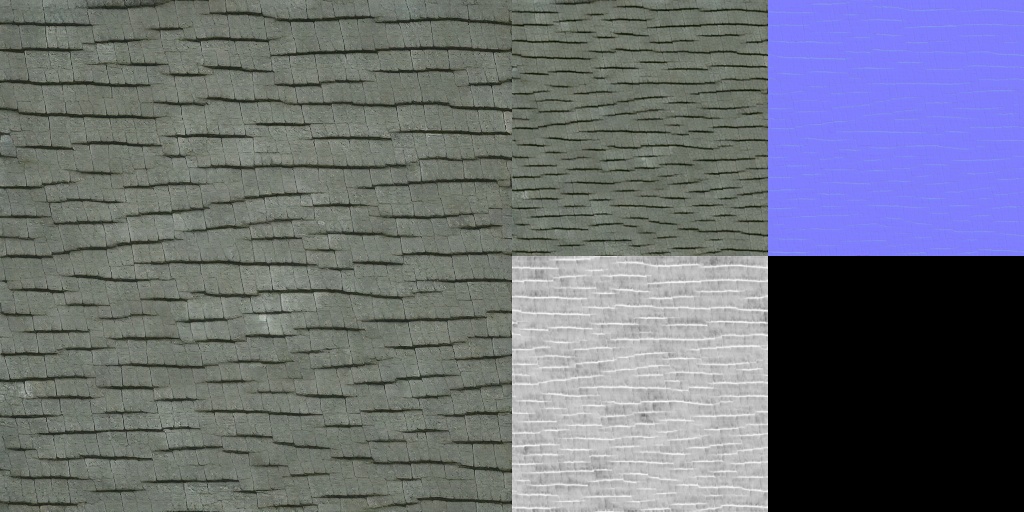} \\

\includegraphics[width=\SeedWidth]{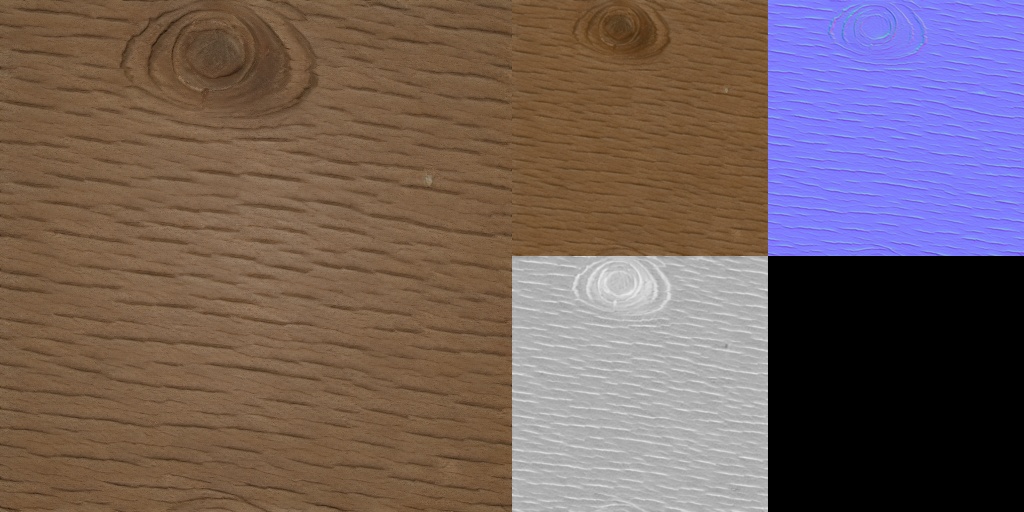} &
\includegraphics[width=\SeedWidth]{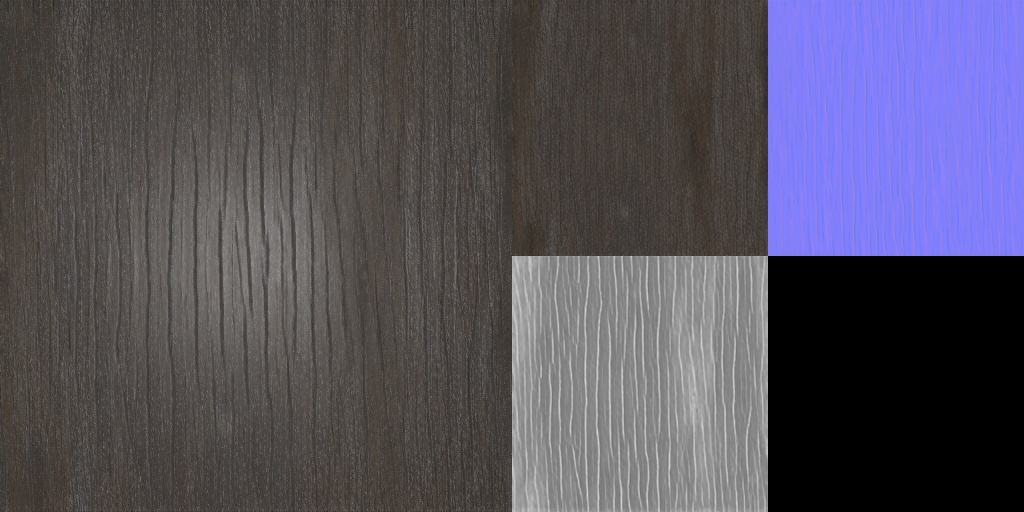} &
\includegraphics[width=\SeedWidth]{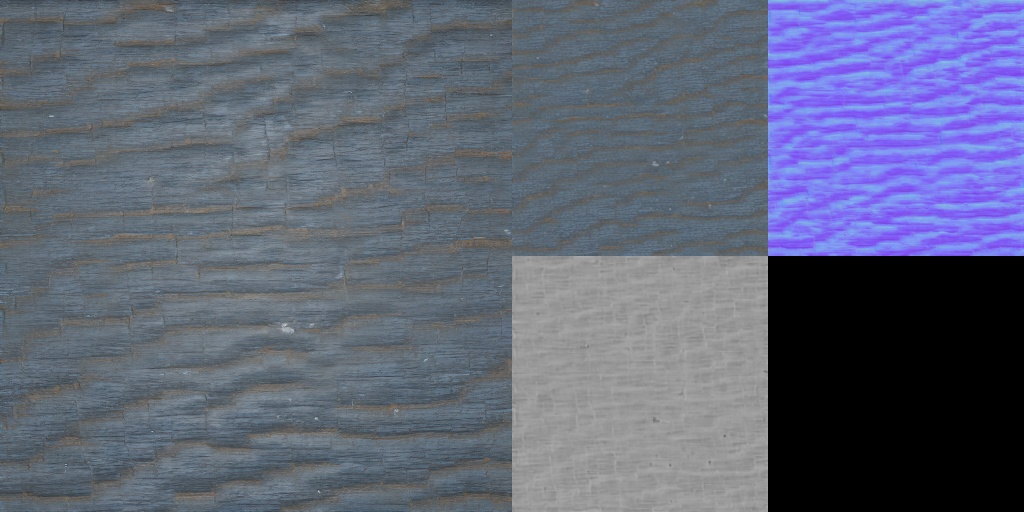} \\

\includegraphics[width=\SeedWidth]{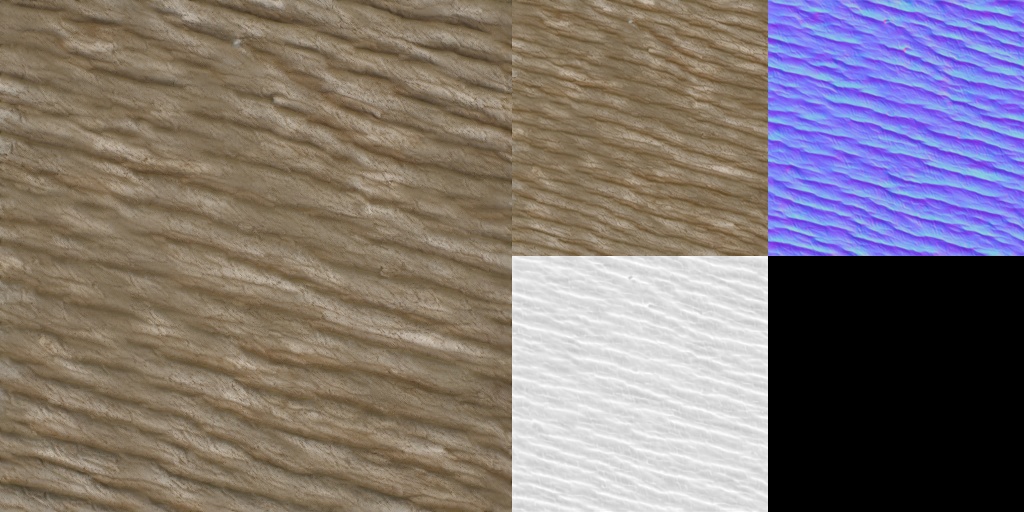} &
\includegraphics[width=\SeedWidth]{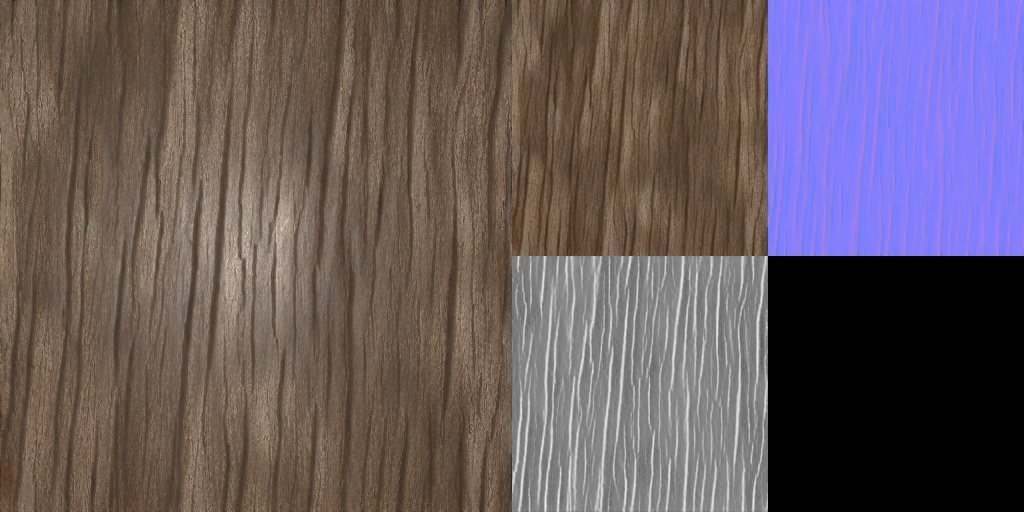} &
\includegraphics[width=\SeedWidth]{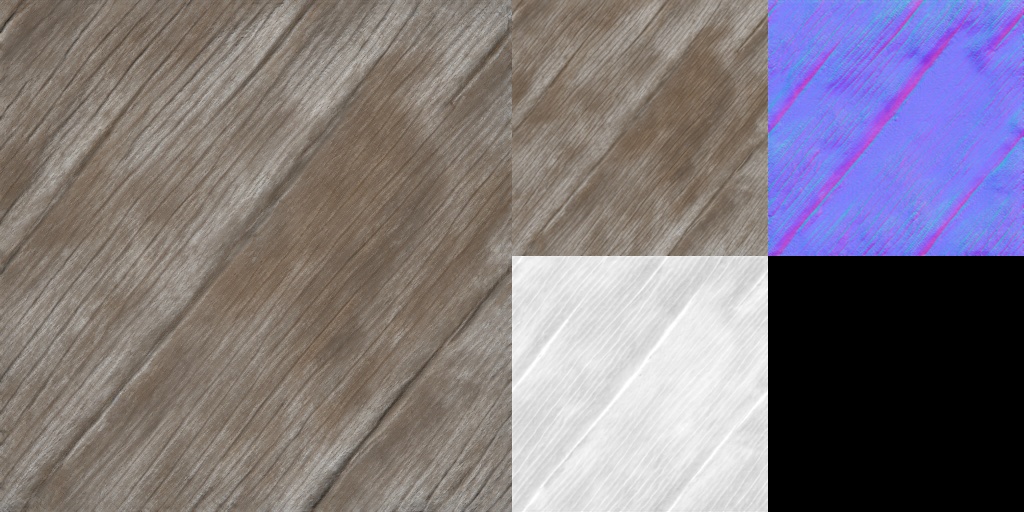} \\

\\
\vspace{5mm}

\\

\multicolumn{3}{c}{
  \begin{minipage}{7.8cm}
    \centering
    \small\textit{a PBR material of tile, encaustic cement tiles, indoor, floor} 
\end{minipage} 
} \\
\includegraphics[width=\SeedWidth]{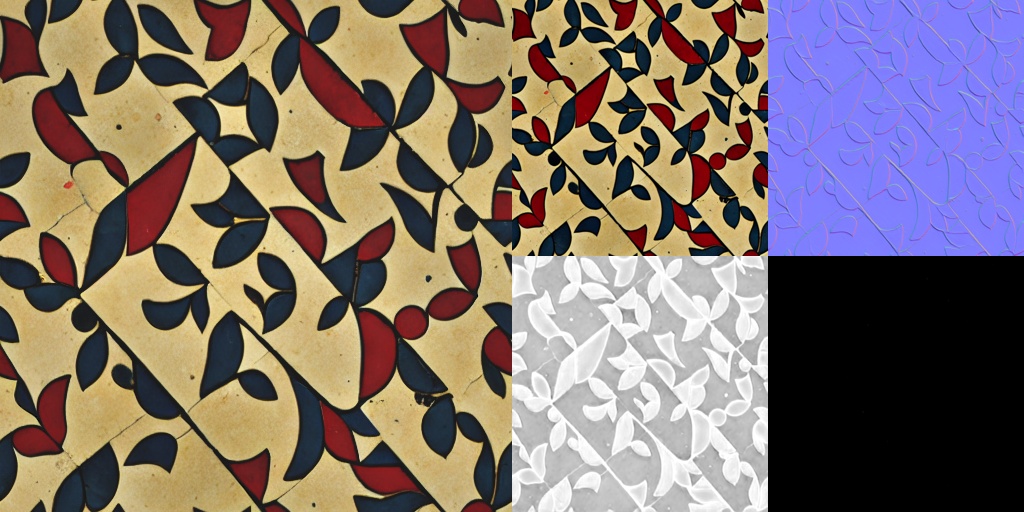} &
\includegraphics[width=\SeedWidth]{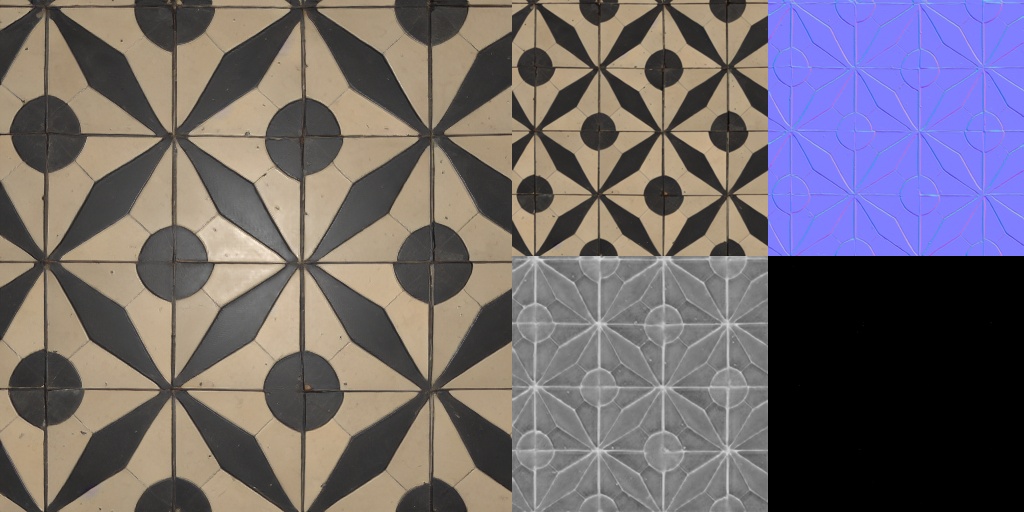} &
\includegraphics[width=\SeedWidth]{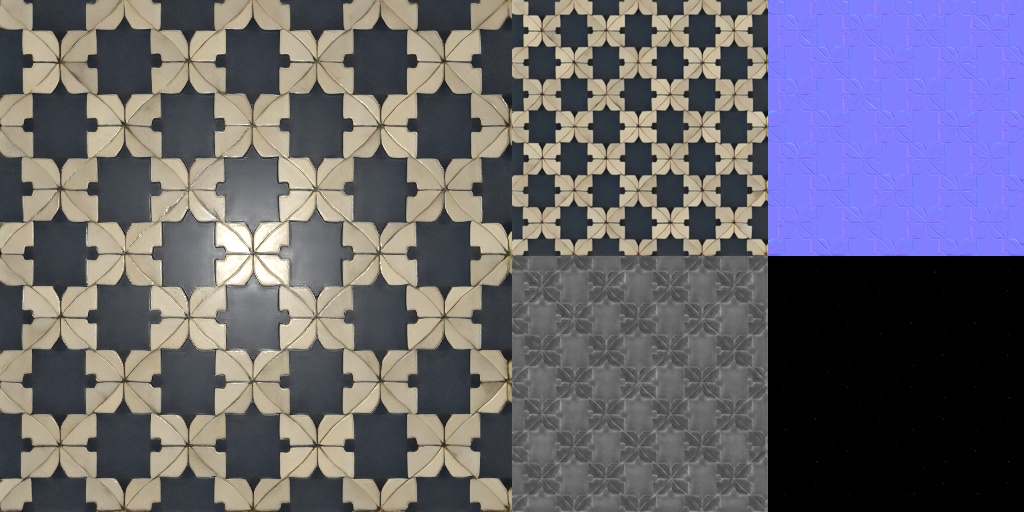} \\

\includegraphics[width=\SeedWidth]{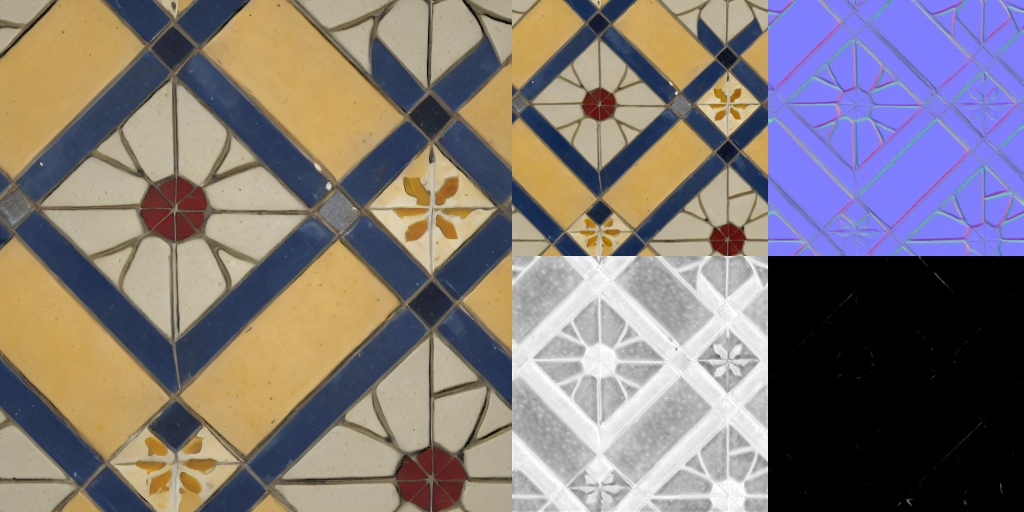} &
\includegraphics[width=\SeedWidth]{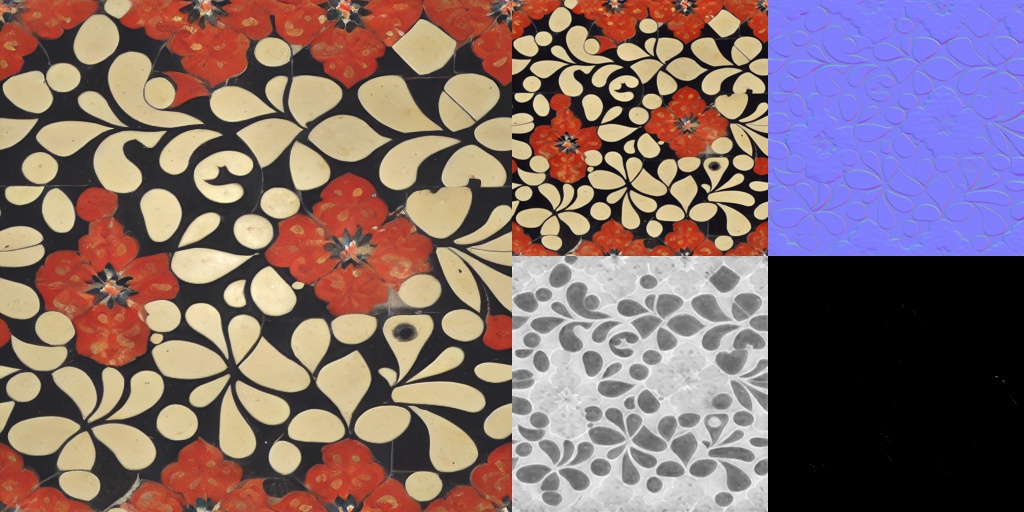} &
\includegraphics[width=\SeedWidth]{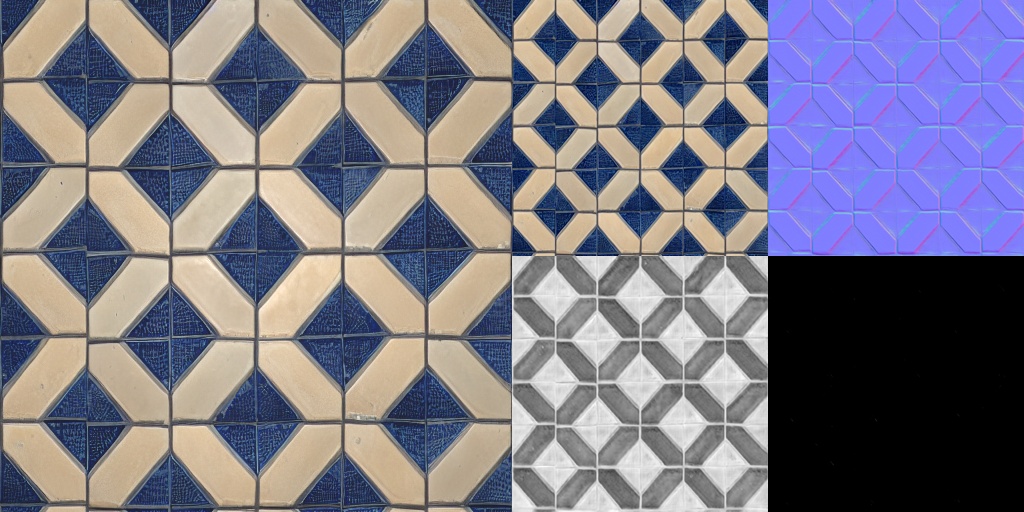} \\

\includegraphics[width=\SeedWidth]{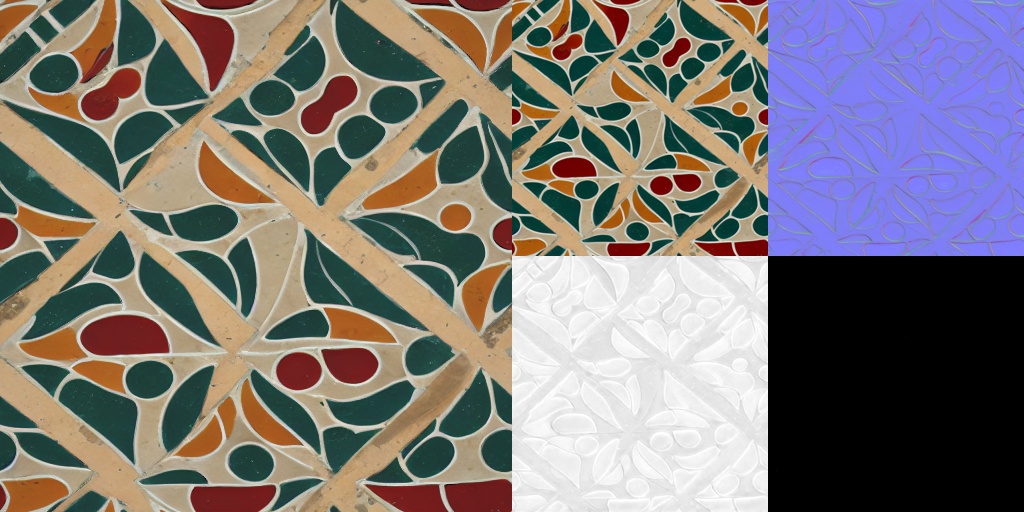} &
\includegraphics[width=\SeedWidth]{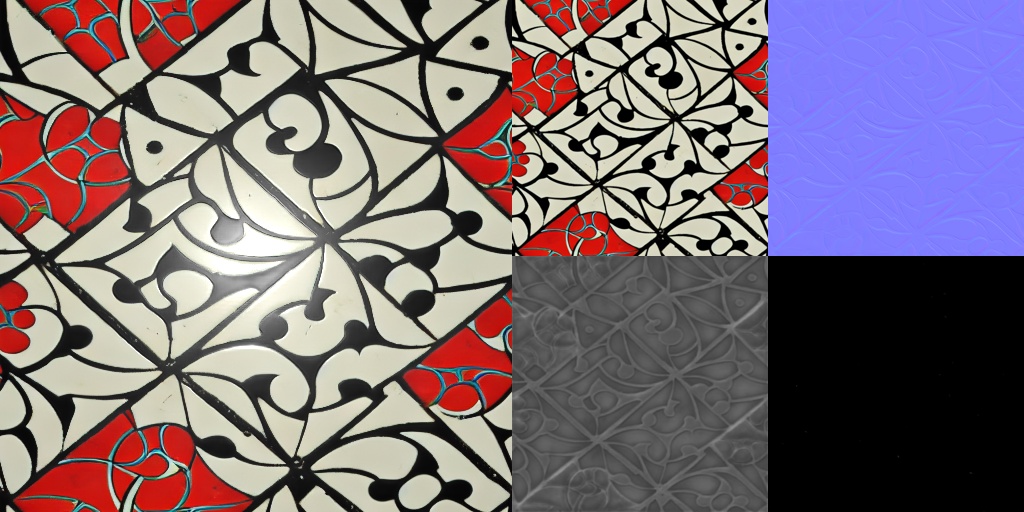} &
\includegraphics[width=\SeedWidth]{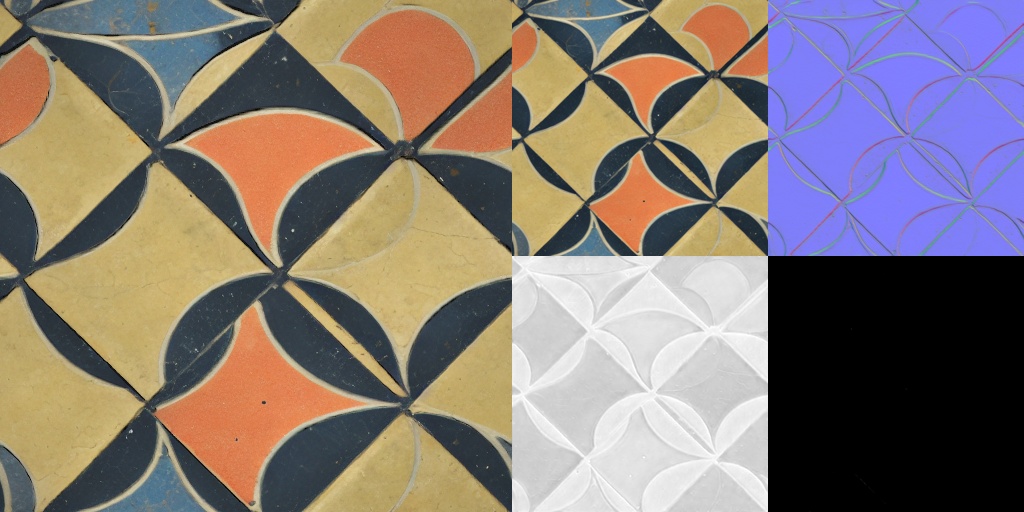} \\

\end{tabular}

\egroup
\caption{Diverse sampling results under the same prompts. We evaluate the diversity with the same basic description (top) and the same detailed description (bottom) but different random seeds. Both of them show quite different patterns and textures although the same prompt is used.} 
\label{fig:seed} 
\Description{fig:seed}
\end{figure}

\newcommand{\SeamlessWidth}{2cm} 
\begin{figure}[htbp]
\centering 

\bgroup

\def\arraystretch{0.5} 
\setlength\tabcolsep{0.5pt} 
\begin{tabular}{
    m{\SeamlessWidth}
    m{\SeamlessWidth}
    m{\SeamlessWidth}
    m{\SeamlessWidth}
    }
    
\multicolumn{1}{c}{Output} & \multicolumn{1}{c}{Expansion} & \multicolumn{1}{c}{Output} & \multicolumn{1}{c}{Expansion} \\

\noalign{\smallskip}

\includegraphics[width=\SeamlessWidth]{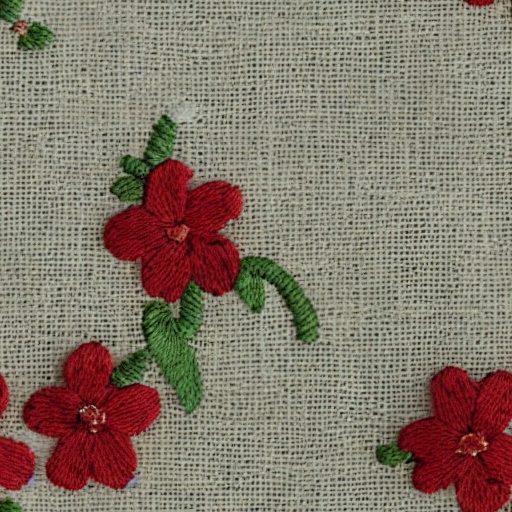} &
\includegraphics[width=\SeamlessWidth]{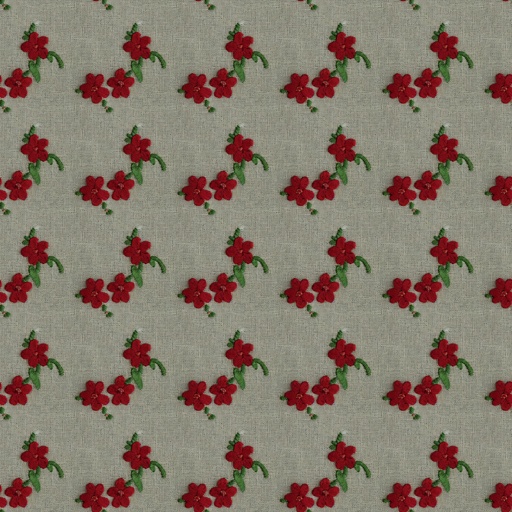} &
\includegraphics[width=\SeamlessWidth]{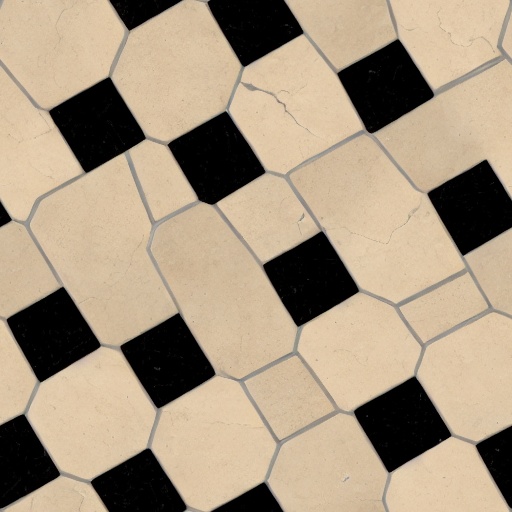} &
\includegraphics[width=\SeamlessWidth]{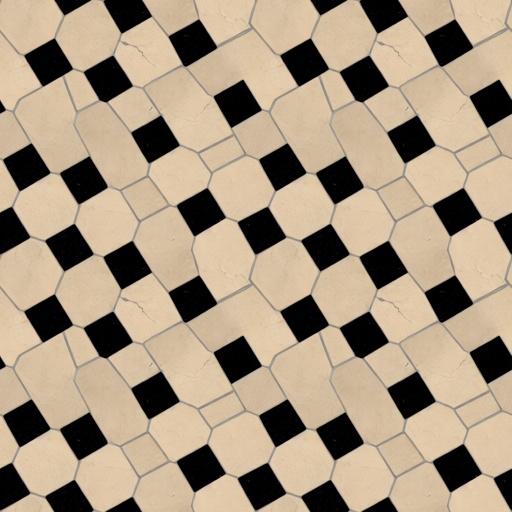} \\

\includegraphics[width=\SeamlessWidth]{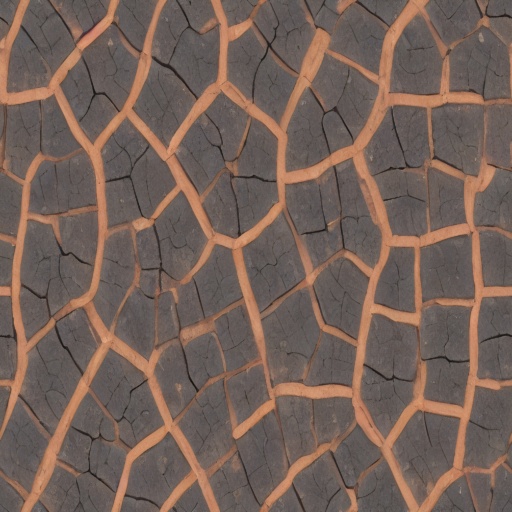} &
\includegraphics[width=\SeamlessWidth]{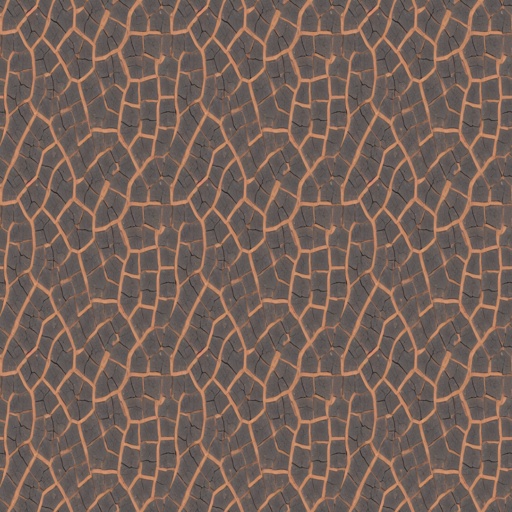} &
\includegraphics[width=\SeamlessWidth]{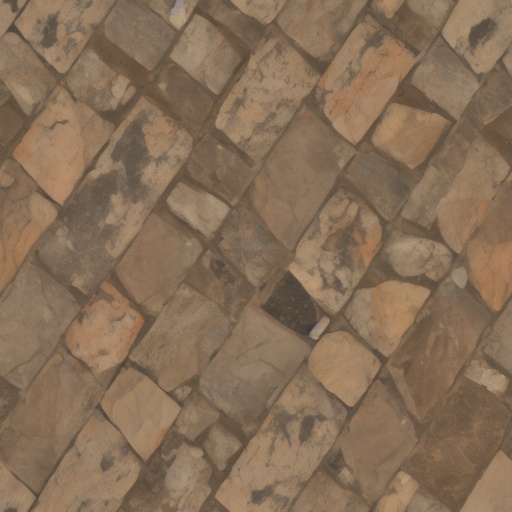} &
\includegraphics[width=\SeamlessWidth]{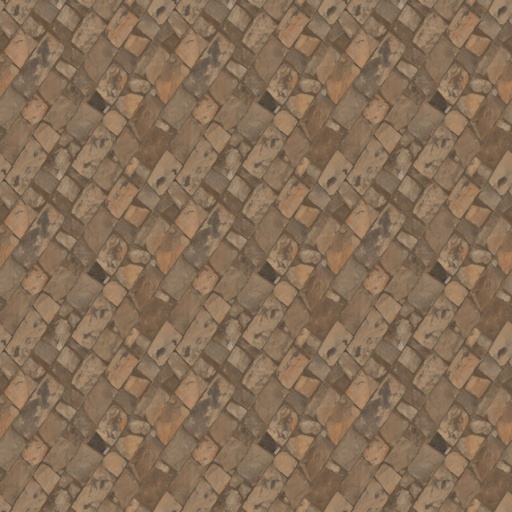} \\

\includegraphics[width=\SeamlessWidth]{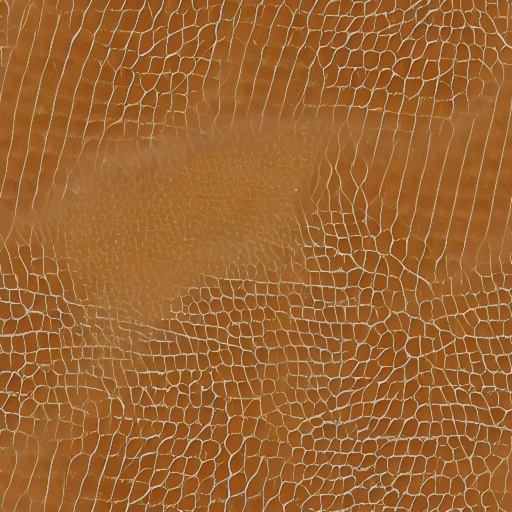} &
\includegraphics[width=\SeamlessWidth]{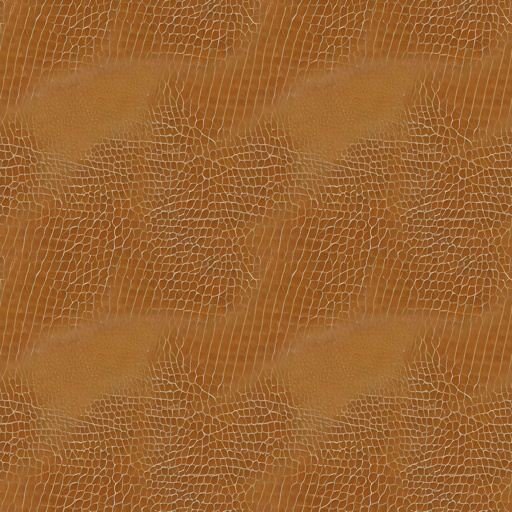} &
\includegraphics[width=\SeamlessWidth]{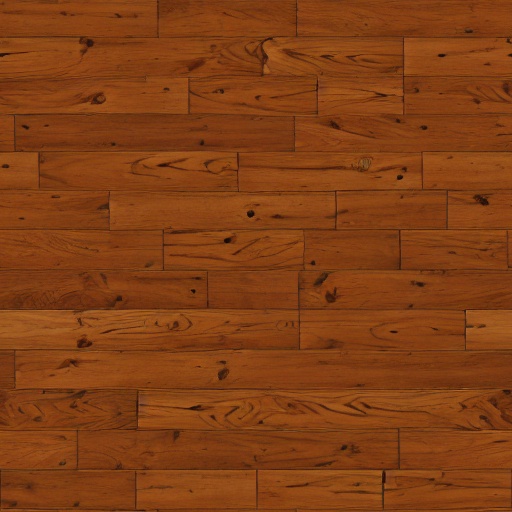} &
\includegraphics[width=\SeamlessWidth]{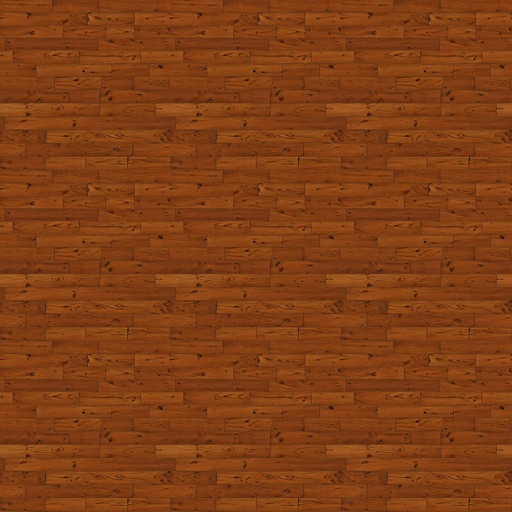} \\

\includegraphics[width=\SeamlessWidth]{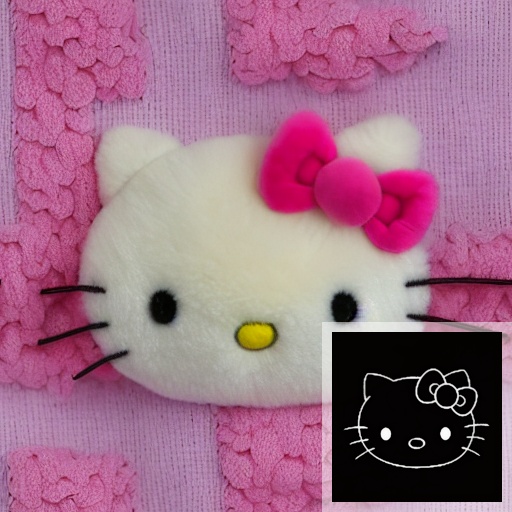} &
\includegraphics[width=\SeamlessWidth]{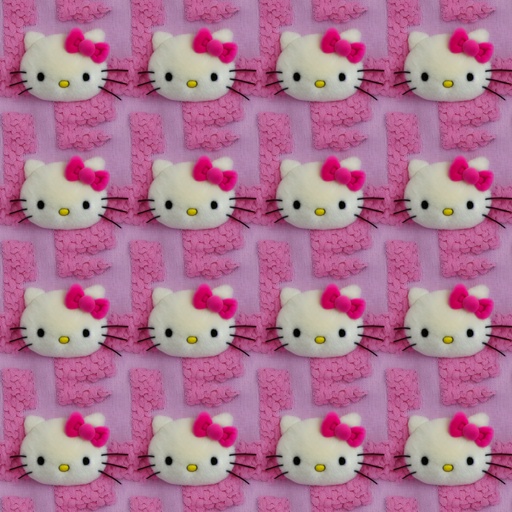} &
\includegraphics[width=\SeamlessWidth]{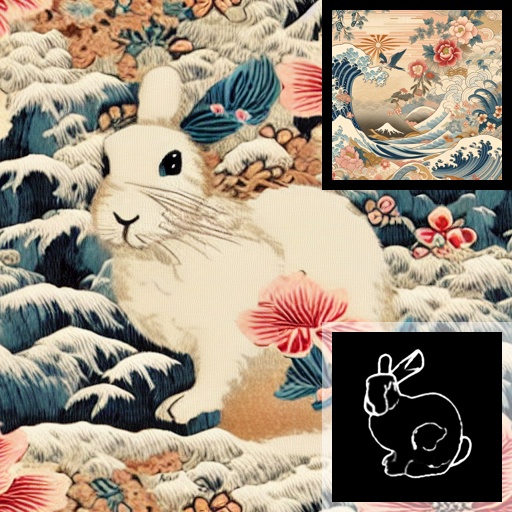} &
\includegraphics[width=\SeamlessWidth]{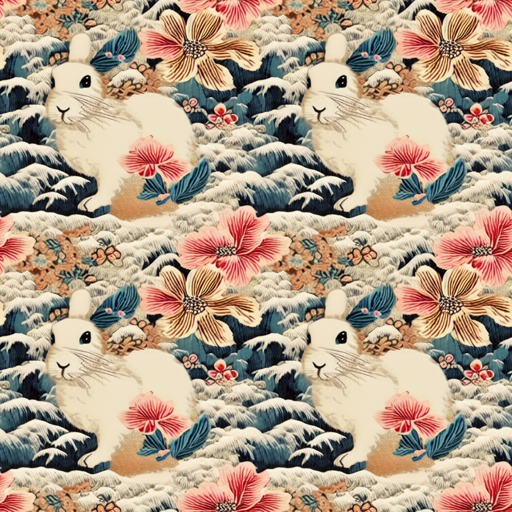} \\

\end{tabular}
\egroup

\caption{Splicing results of our tileable textures. The first 3 rows show the tileable results generated under only text descriptions and the last row shows the results with Pixel Control and Style Control, whose control conditions are attached to the edge of images. All the textures we generated in this figure show high tile abilities without artifacts.}
\label{fig:seamless} 
\Description{fig:seamless}
\end{figure}

\newcommand{\inpaintingWidth}{2.5cm} 
\begin{figure}[htbp]
\centering 

\bgroup

\def\arraystretch{0.5} 
\setlength\tabcolsep{0.5pt}

\begin{tabular}{
  m{\inpaintingWidth}
  m{\inpaintingWidth}
  m{\inpaintingWidth}
}

\multicolumn{1}{c}{Input} & \multicolumn{1}{c}{Yellow flower} & \multicolumn{1}{c}{Red flower} \\

\includegraphics[width=\inpaintingWidth]{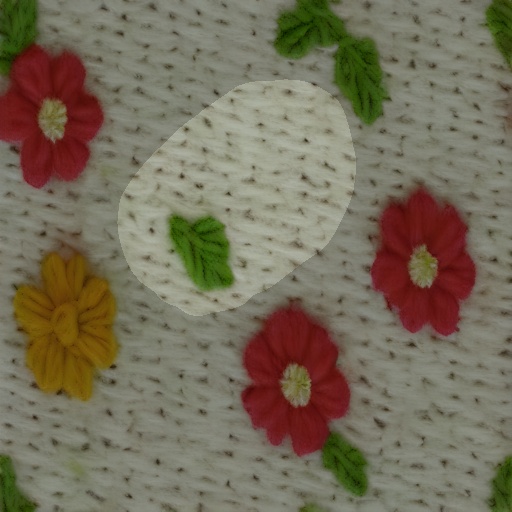} &
\includegraphics[width=\inpaintingWidth]{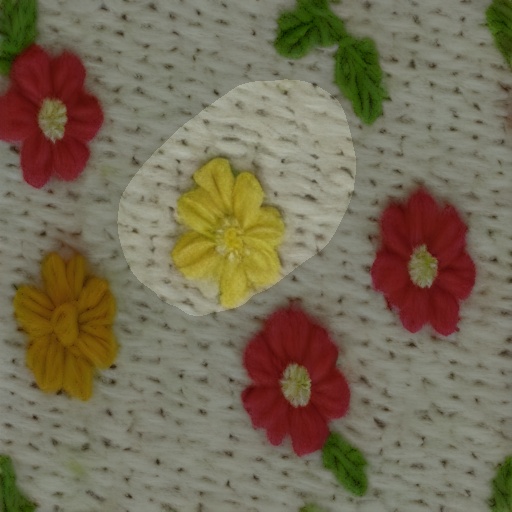} &
\includegraphics[width=\inpaintingWidth]{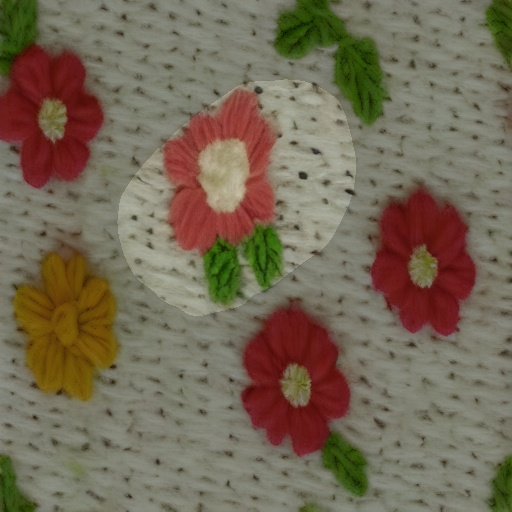} \\

\multicolumn{1}{c}{Blue flower} & \multicolumn{1}{c}{Cyan flower} & \multicolumn{1}{c}{Purple flower} \\

\includegraphics[width=\inpaintingWidth]{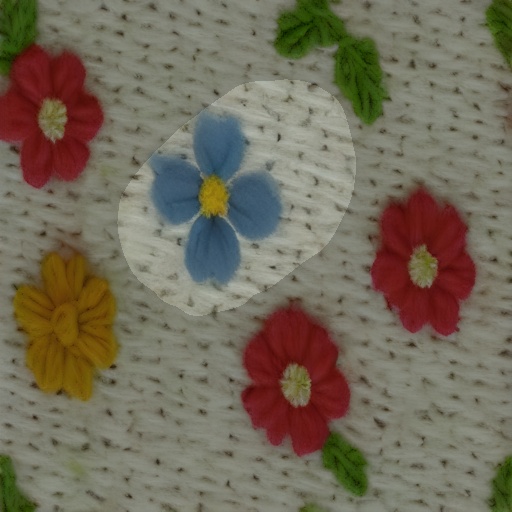} &
\includegraphics[width=\inpaintingWidth]{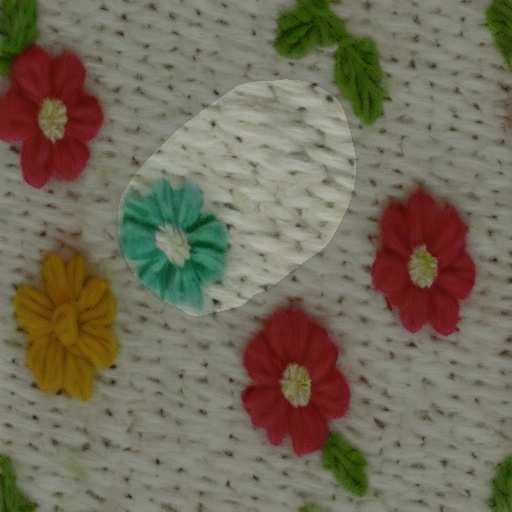} &
\includegraphics[width=\inpaintingWidth]{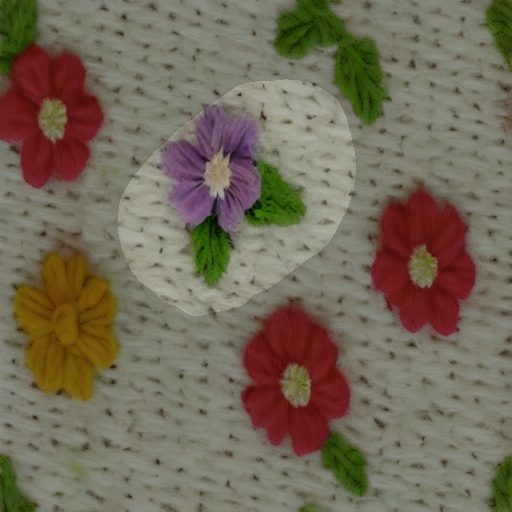} \\

\multicolumn{1}{c}{Pink flower} & \multicolumn{1}{c}{Leaf} & \multicolumn{1}{c}{Grass} \\
\includegraphics[width=\inpaintingWidth]{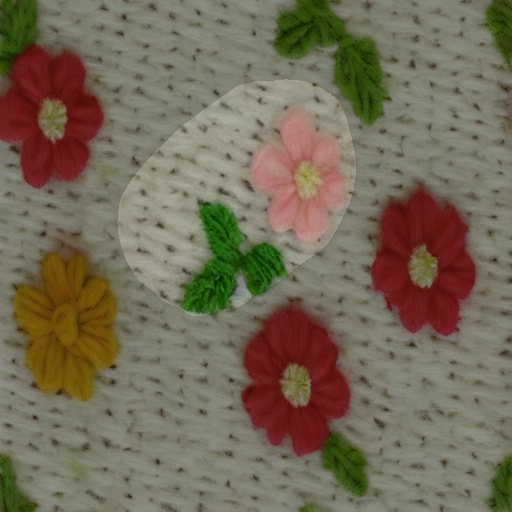} &
\includegraphics[width=\inpaintingWidth]{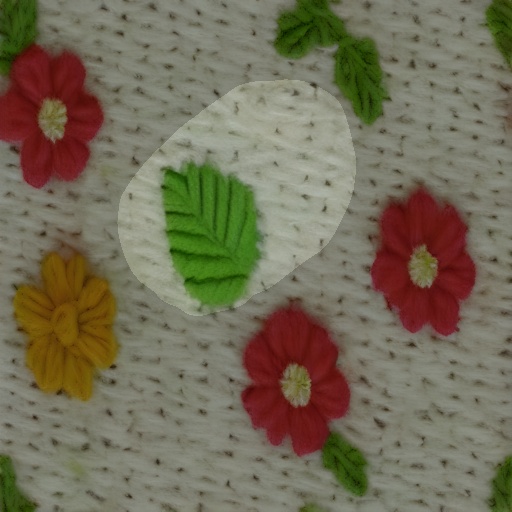} &
\includegraphics[width=\inpaintingWidth]{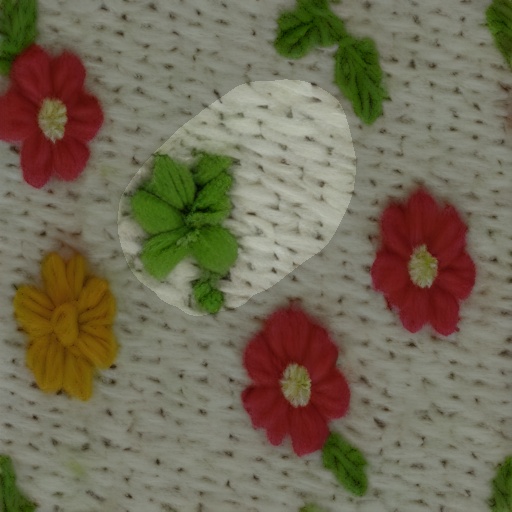} \\

\end{tabular}

\egroup

\caption{Inpainting results. The original texture is shown in the upper left corner, which is generated by the prompt ``a PBR material of the fabric, floral cotton fabric''. By different tags (above each image) and regions to be inpainted (bright areas in each image), we can manipulate user-specified areas in the textures according to preferences such as changing a leaf to colorful flowers.} % 图形的标题
\label{fig:inpainting} 
\Description{fig:inpainting}
\end{figure}

\newcommand{\multimodalColumnWidth}{1.65cm} 
\begin{figure}[htbp]
\centering 

\bgroup

\def\arraystretch{0.1} 
\setlength\tabcolsep{0.5pt} 
\begin{tabular}{
    m{\multimodalColumnWidth}
    m{\multimodalColumnWidth}
    m{\multimodalColumnWidth}
    m{\multimodalColumnWidth}
    m{\multimodalColumnWidth}
    }
    
\multicolumn{1}{c}{Prompt} & \multicolumn{1}{c}{Style} & \multicolumn{1}{c}{Pixel} & \multicolumn{1}{c}{Render} & \multicolumn{1}{c}{SVBRDF} \\

\multicolumn{1}{c}{
  \begin{minipage}{\multimodalColumnWidth}
    \centering
    \scriptsize\textit{a PBR material of tiles, marble} 
\end{minipage} 
} &
\includegraphics[width=\multimodalColumnWidth]{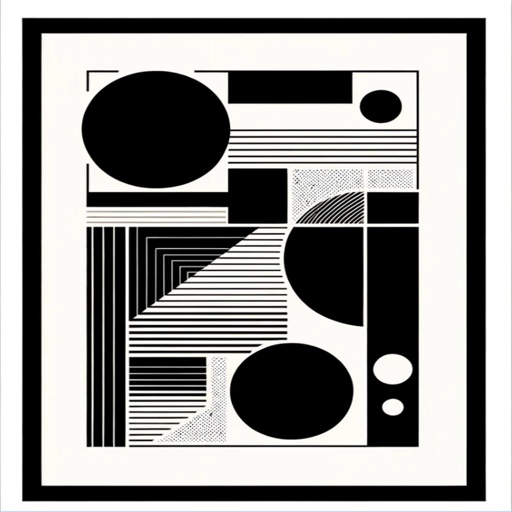} &
\includegraphics[width=\multimodalColumnWidth]{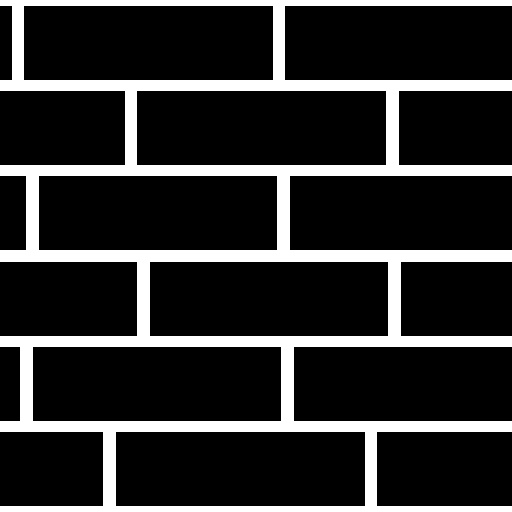} &
\includegraphics[width=\multimodalColumnWidth]{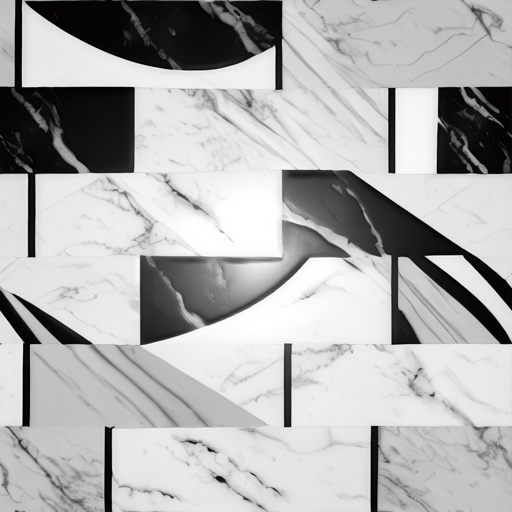} &
\includegraphics[width=\multimodalColumnWidth]{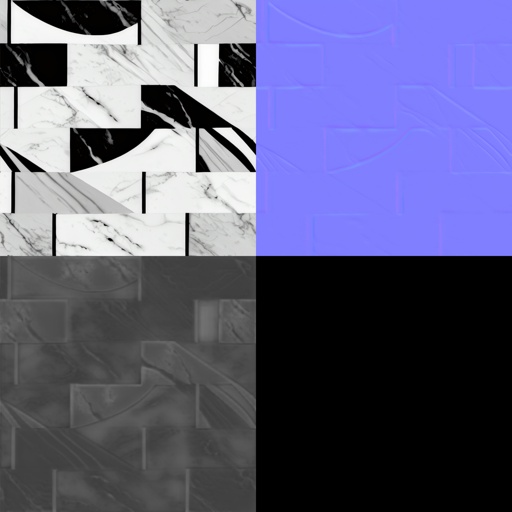} \\

\multicolumn{1}{c}{
  \begin{minipage}{\multimodalColumnWidth}
    \centering
    \scriptsize\textit{a PBR material of wood, indoor} 
\end{minipage} 
} &
\includegraphics[width=\multimodalColumnWidth]{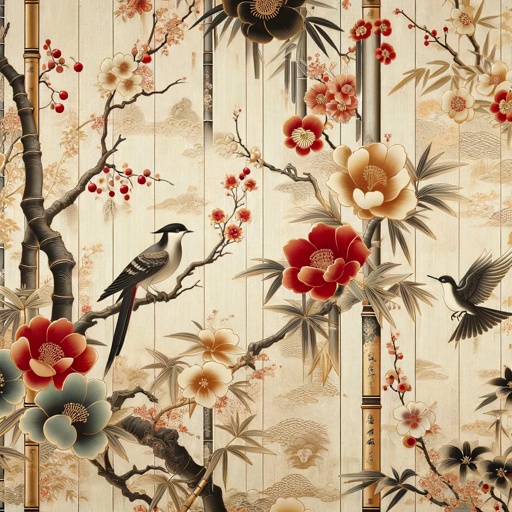} &
\includegraphics[width=\multimodalColumnWidth]{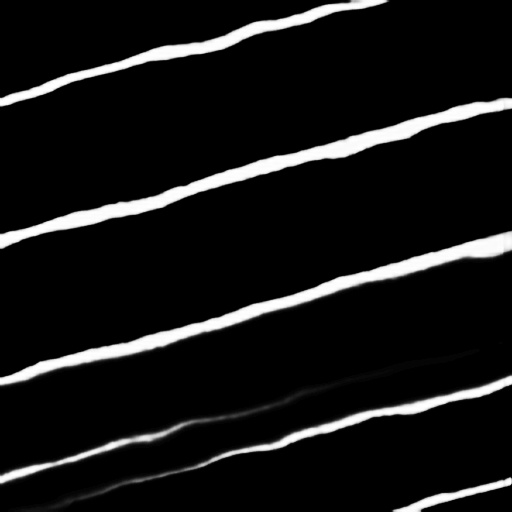} &
\includegraphics[width=\multimodalColumnWidth]{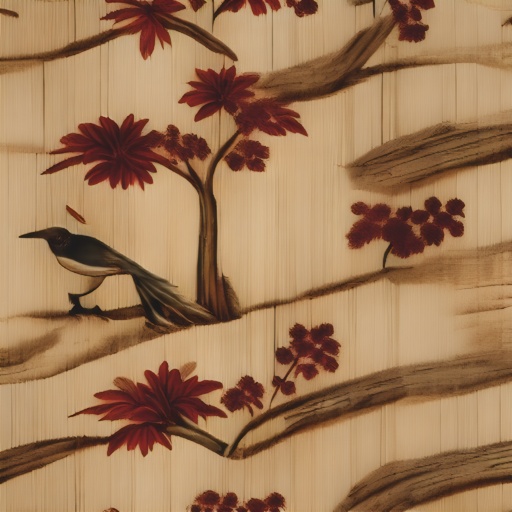} &
\includegraphics[width=\multimodalColumnWidth]{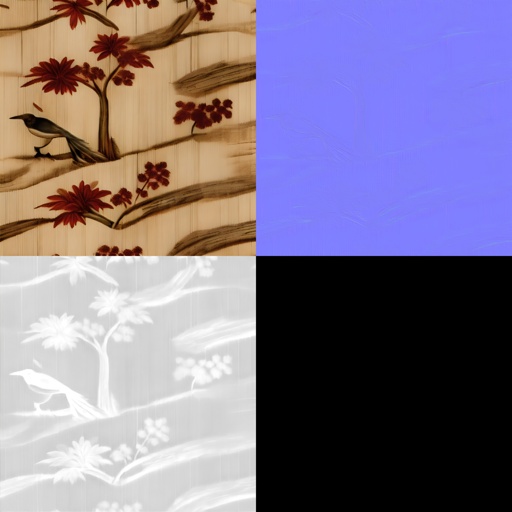} \\

\multicolumn{1}{c}{
  \begin{minipage}{\multimodalColumnWidth}
    \centering
    \scriptsize\textit{a PBR material of tiles, art deco style tiles, vintage, indoor, decorative} 
\end{minipage} 
} &
\includegraphics[width=\multimodalColumnWidth]{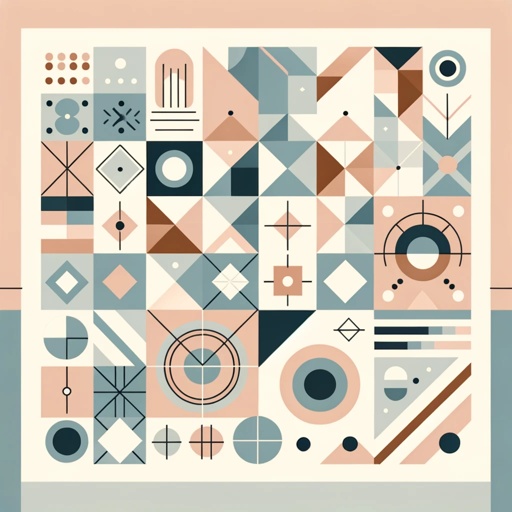} &
\includegraphics[width=\multimodalColumnWidth]{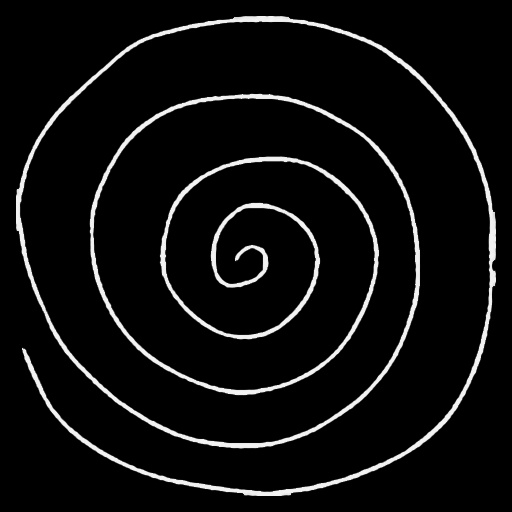} &
\includegraphics[width=\multimodalColumnWidth]{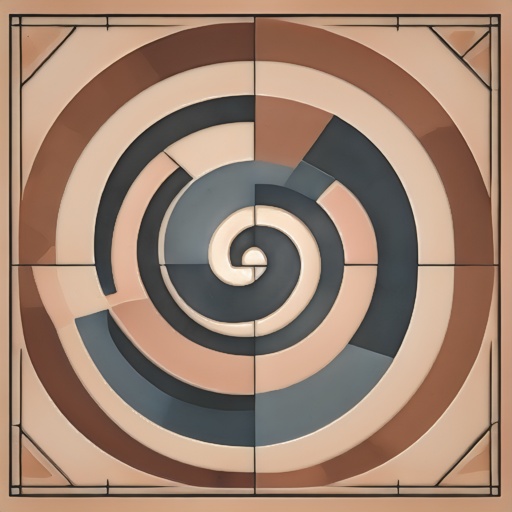} &
\includegraphics[width=\multimodalColumnWidth]{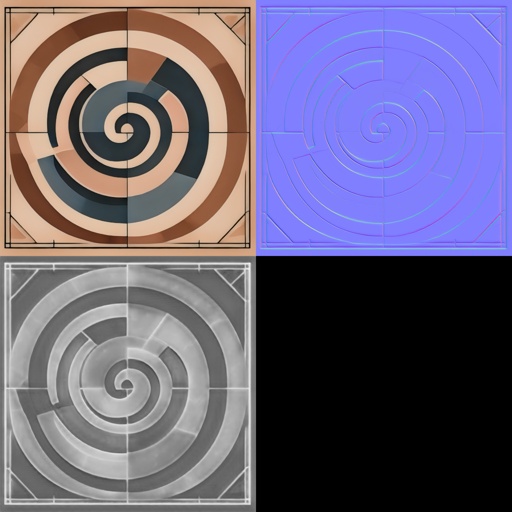} \\

\multicolumn{1}{c}{
  \begin{minipage}{\multimodalColumnWidth}
    \centering
    \scriptsize\textit{a PBR material of fabric, patchwork quilt, colorful, indoor, bedding} 
\end{minipage} 
} &
\includegraphics[width=\multimodalColumnWidth]{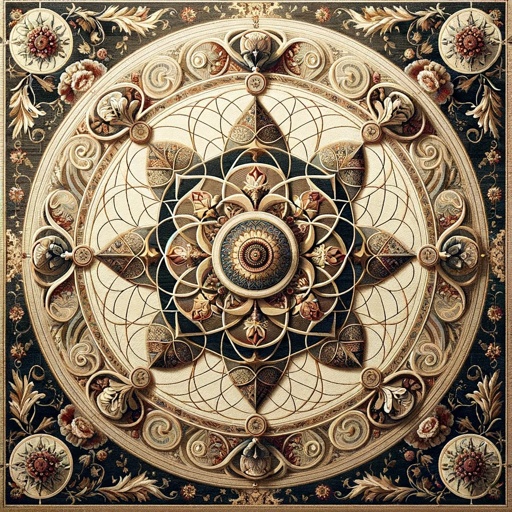} &
\includegraphics[width=\multimodalColumnWidth]{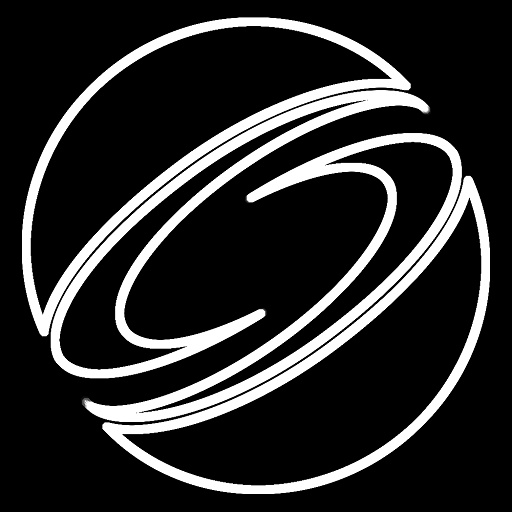} &
\includegraphics[width=\multimodalColumnWidth]{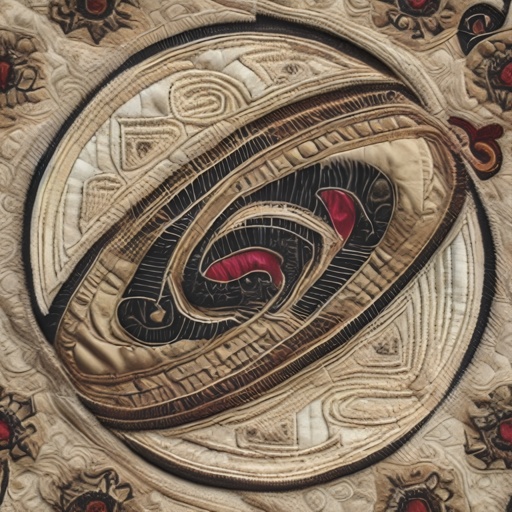} &
\includegraphics[width=\multimodalColumnWidth]{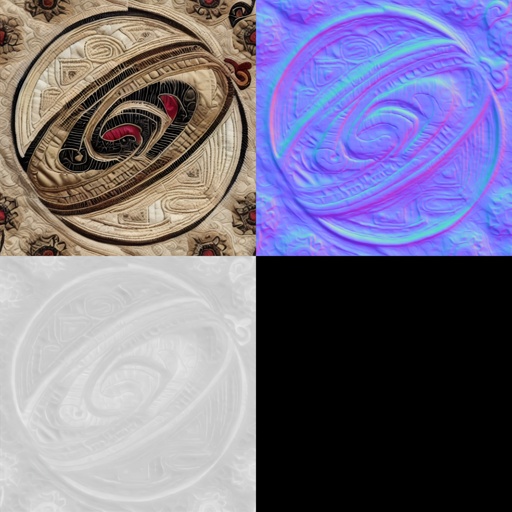} \\

\multicolumn{1}{c}{
  \begin{minipage}{\multimodalColumnWidth}
    \centering
    \scriptsize\textit{a PBR material of fabric, hand woven carpet, cute bunny, artisan, indoor} 
\end{minipage} 
} &
\includegraphics[width=\multimodalColumnWidth]{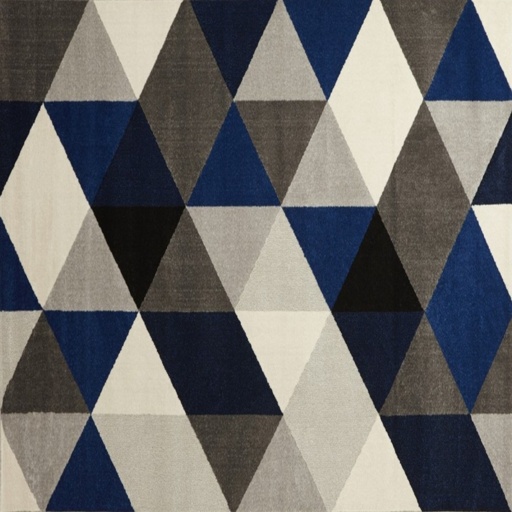} &
\includegraphics[width=\multimodalColumnWidth]{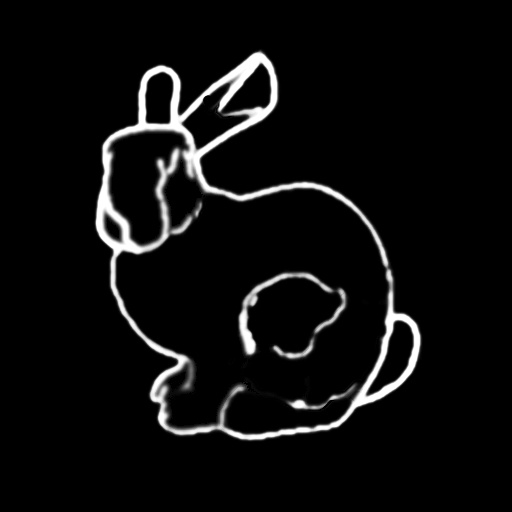} &
\includegraphics[width=\multimodalColumnWidth]{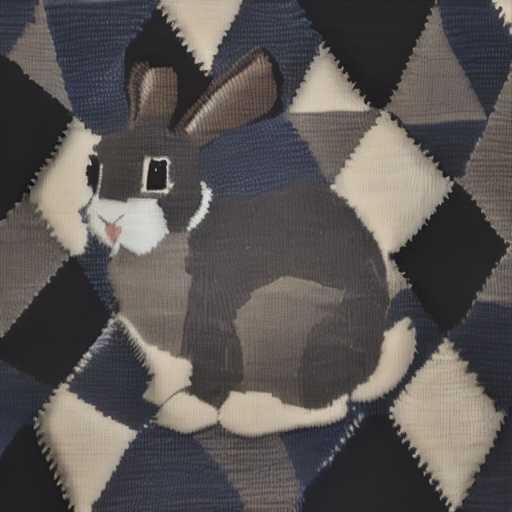} &
\includegraphics[width=\multimodalColumnWidth]{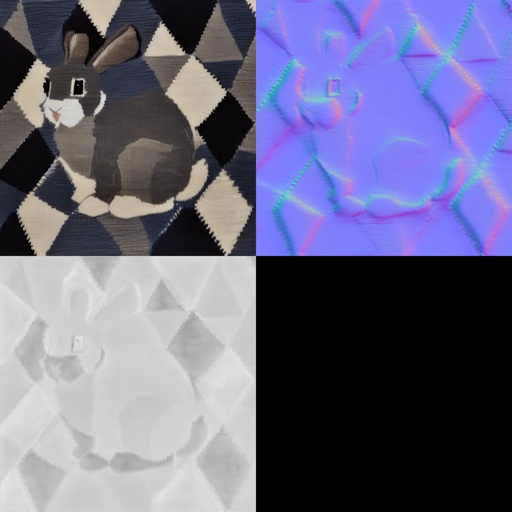} \\

\end{tabular}
\egroup
\caption{Results of combined control. We combine descriptions of materials (the first column), styled images (the second column), and binary images (the third column) to control the generated textures. Under the descriptions of materials, the generated results have both the given pattern and style incorporated into them.} 
\label{fig:multiModal} 
\Description{fig:multiModal}
\end{figure}

\begin{figure}[htbp]
    \centering
    \includegraphics[width=\linewidth]{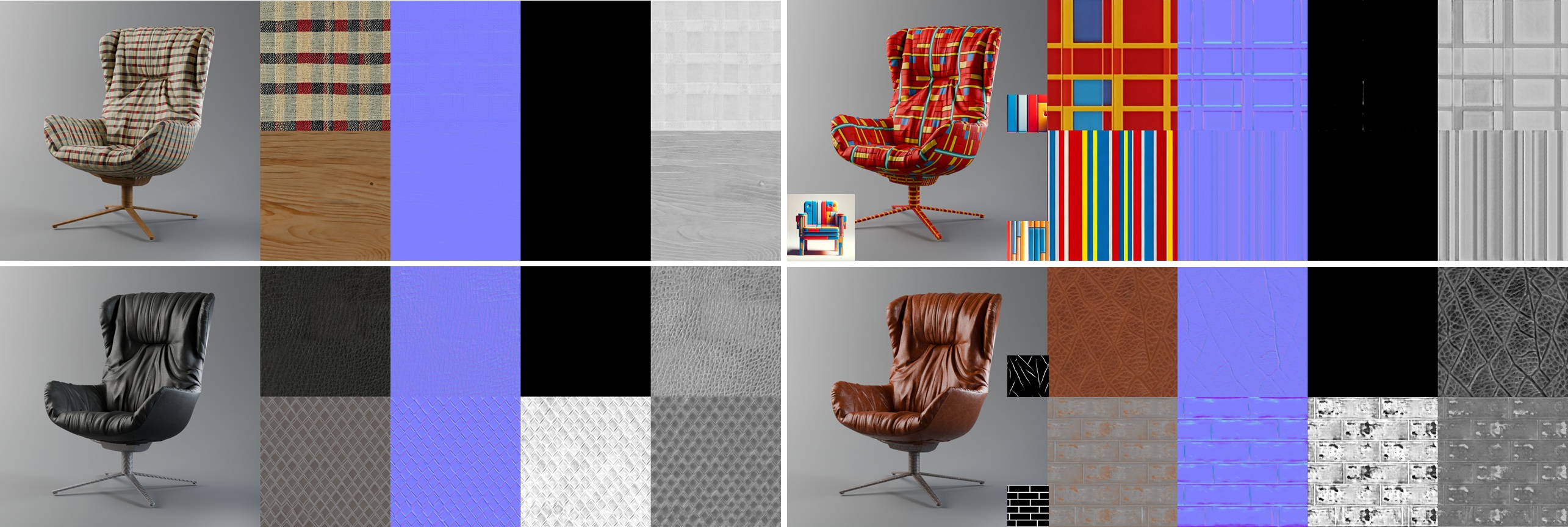}
    \caption{Results of ShapeControl. we leverage a large language model(LLM) to describe the segmented chair legs and chair back, which is used to generate their textures with the help of DreamPBR. Besides text-only descriptions(left two figures), additional user-specific controls such as RGB images (upper right) and binary images (lower right) are allowed for personalized design as well.}
    \label{fig:shape}
    \Description{fig:shape}
\end{figure}
\newcommand{\MaterialganWidth}{2.6cm} 
\begin{figure}[htbp]
\centering 

\bgroup

\def\arraystretch{0.5} 
\setlength\tabcolsep{0.4pt}

\begin{tabular}{
    m{0.4cm}
    m{\MaterialganWidth}
    m{\MaterialganWidth}
    m{\MaterialganWidth}
    }
\multicolumn{1}{c}{} & \multicolumn{1}{c}{MaterialGAN} & \multicolumn{1}{c}{TileGen} & \multicolumn{1}{c}{Ours} \\

\multirow{4}{*}[-1.7cm]{\rotatebox[origin=c]{90}{Stone}} &
\includegraphics[width=\MaterialganWidth]{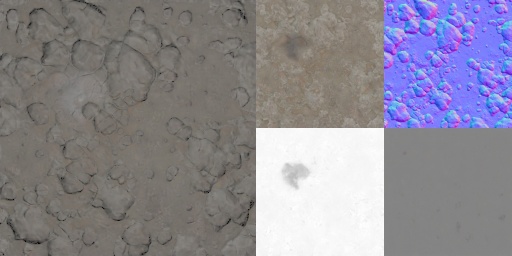} &
\includegraphics[width=\MaterialganWidth]{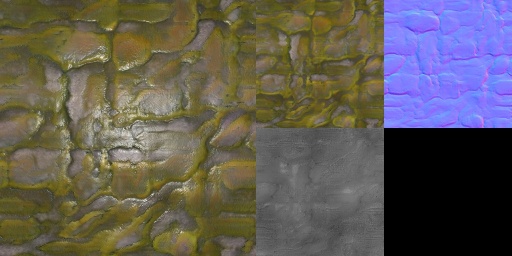} &
\includegraphics[width=\MaterialganWidth]{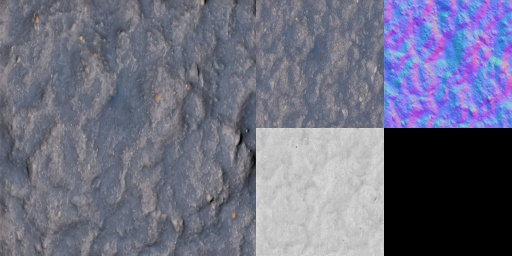} \\

&
\includegraphics[width=\MaterialganWidth]{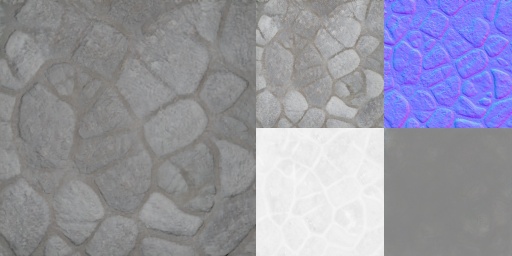} &
\includegraphics[width=\MaterialganWidth]{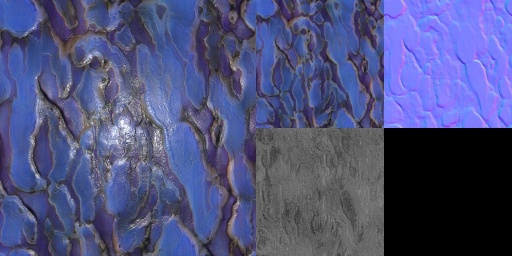} &
\includegraphics[width=\MaterialganWidth]{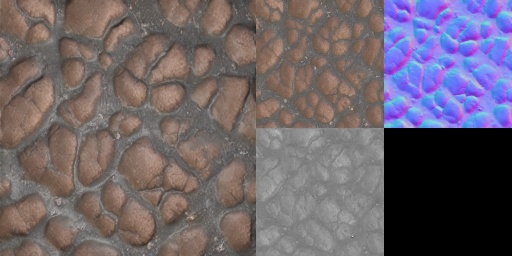} \\

&
\includegraphics[width=\MaterialganWidth]{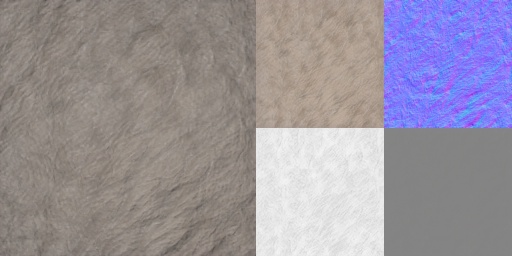} &
\includegraphics[width=\MaterialganWidth]{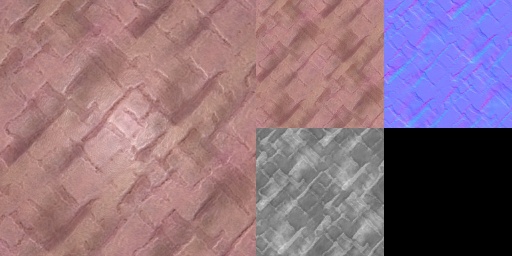} &
\includegraphics[width=\MaterialganWidth]{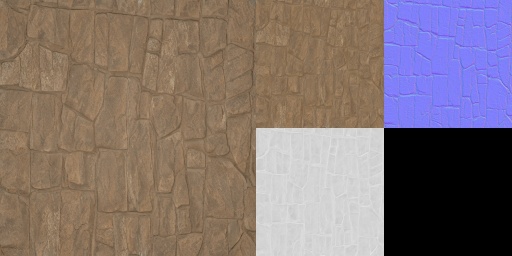} \\

&
\includegraphics[width=\MaterialganWidth]{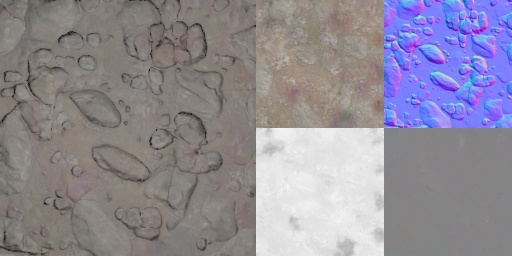} &
\includegraphics[width=\MaterialganWidth]{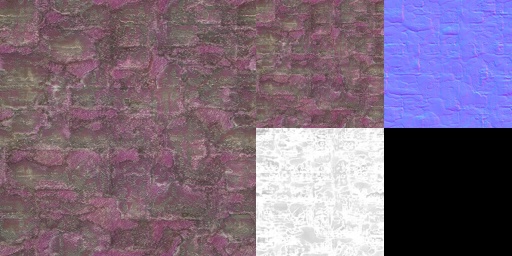} &
\includegraphics[width=\MaterialganWidth]{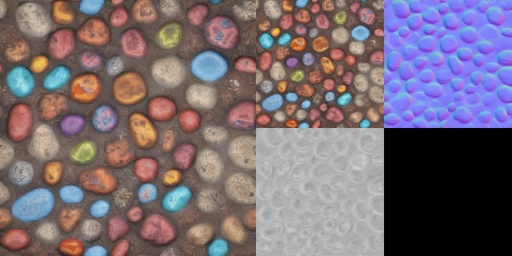} \\

\\
\vspace{5mm}
\multirow{4}{*}[-1cm]{\rotatebox[origin=c]{90}{Metal}} &
\includegraphics[width=\MaterialganWidth]{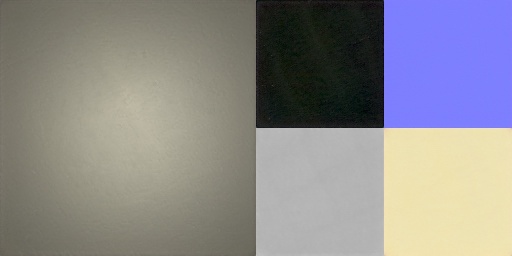} &
\includegraphics[width=\MaterialganWidth]{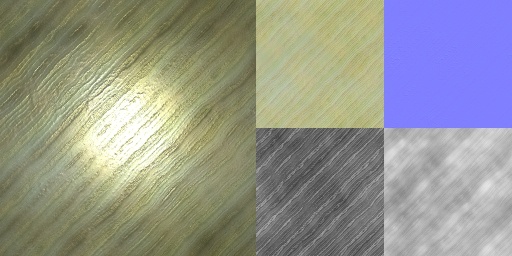} &
\includegraphics[width=\MaterialganWidth]{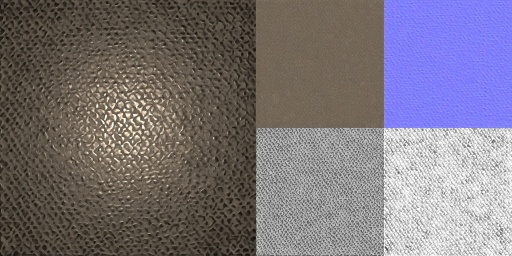} \\
&
\includegraphics[width=\MaterialganWidth]{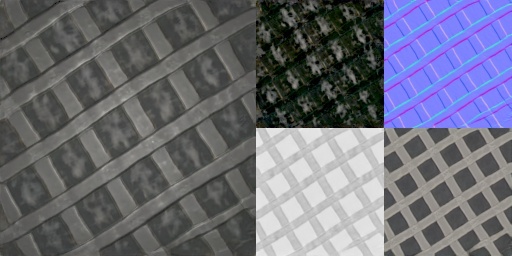} &
\includegraphics[width=\MaterialganWidth]{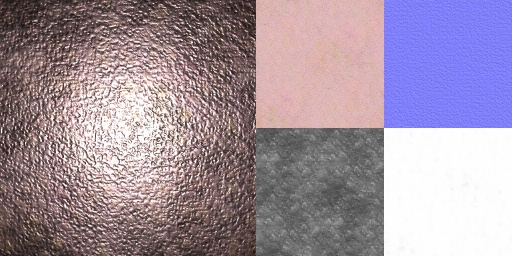} &
\includegraphics[width=\MaterialganWidth]{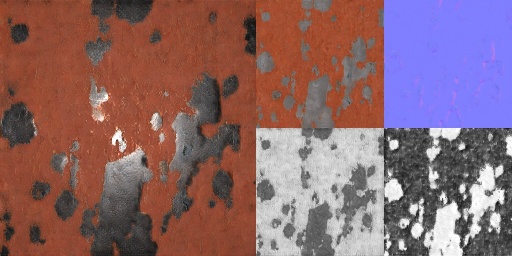} \\
&
\includegraphics[width=\MaterialganWidth]{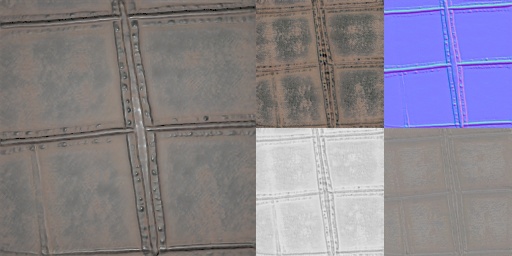} &
\includegraphics[width=\MaterialganWidth]{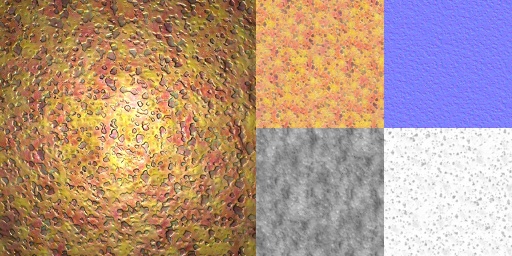} &
\includegraphics[width=\MaterialganWidth]{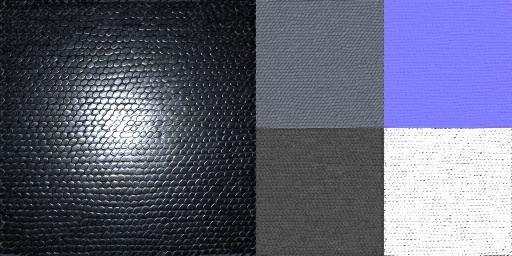} \\
&
\includegraphics[width=\MaterialganWidth]{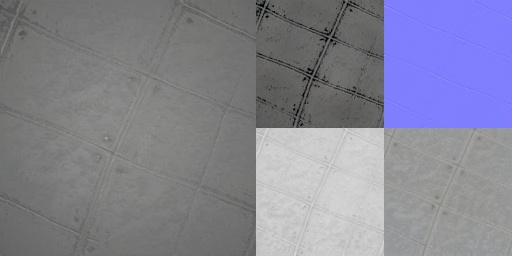} &
\includegraphics[width=\MaterialganWidth]{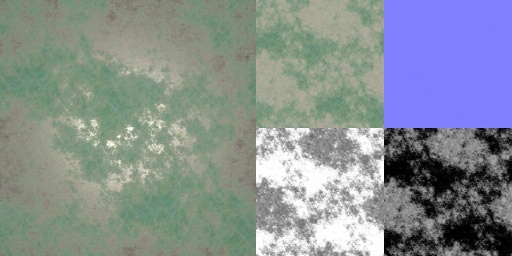} &
\includegraphics[width=\MaterialganWidth]{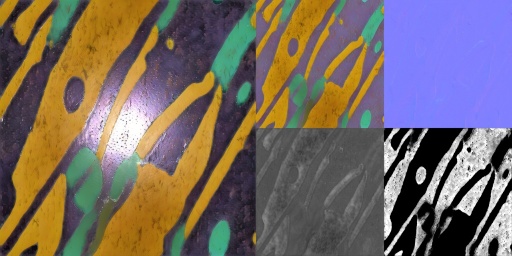} \\

\end{tabular}

\egroup
\caption{Qualitative comparison on texture generation. We randomly sample several materials of stone and metal by \citet{guo2020materialgan} and \citet{zhou2022tilegen}, which are used to be compared with ours. Importantly, we can generate out-of-domain textures as shown in the fourth row and the last row of ours, which is beyond the capabilities of GAN-based methods.} 
\label{fig:materialgan_1} 
\Description{fig:materialgan}
\end{figure}

\newcommand{\TilegenWidth}{2.6cm} 
\begin{figure}[htbp]
\centering 

\bgroup

\def\arraystretch{0.5} 
\setlength\tabcolsep{0.5pt} 

\begin{tabular}{
    m{0.3cm}
    m{\TilegenWidth}
    m{\TilegenWidth}
    m{\TilegenWidth}
    }

\raisebox{+0.8\height}{\rotatebox[origin=c]{90}{\small{Ours}}} &

\includegraphics[width=\TilegenWidth]{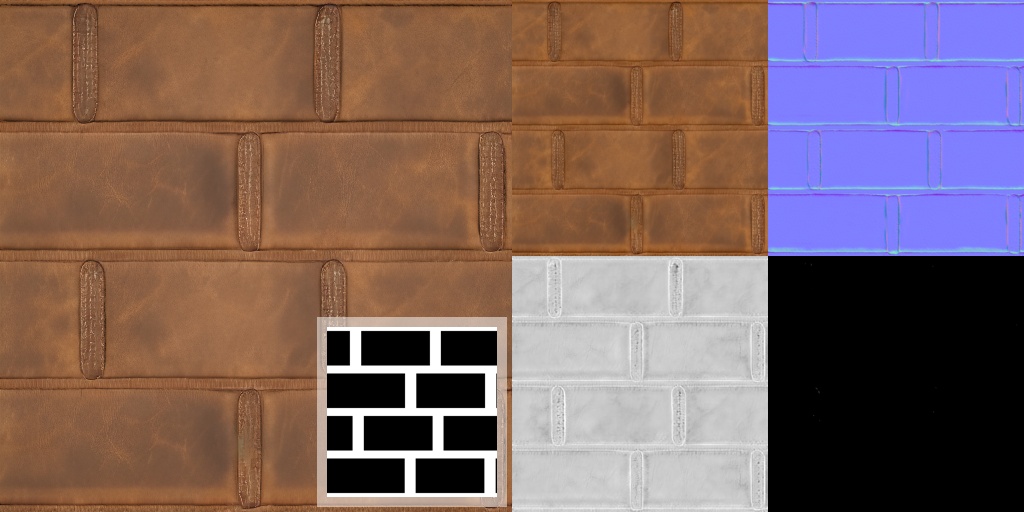} &
\includegraphics[width=\TilegenWidth]{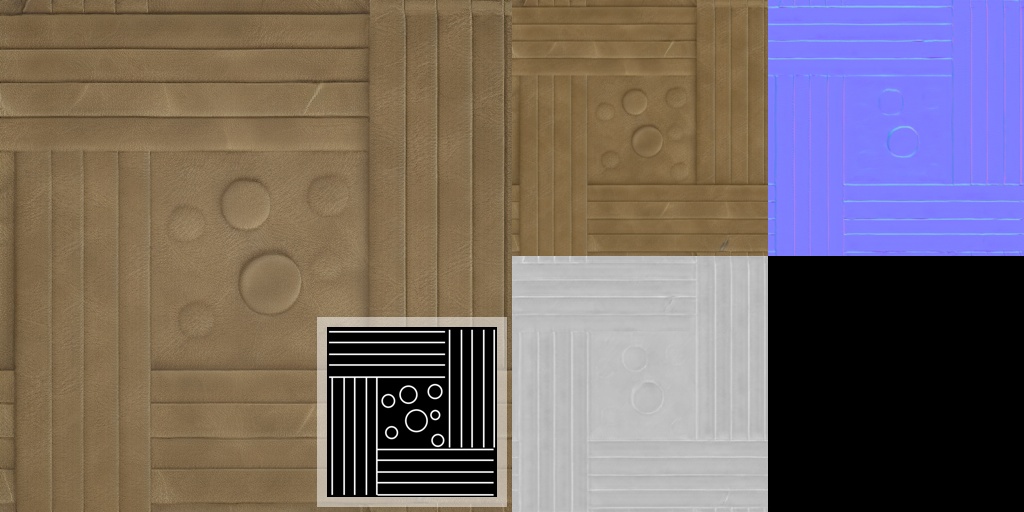} &
\includegraphics[width=\TilegenWidth]{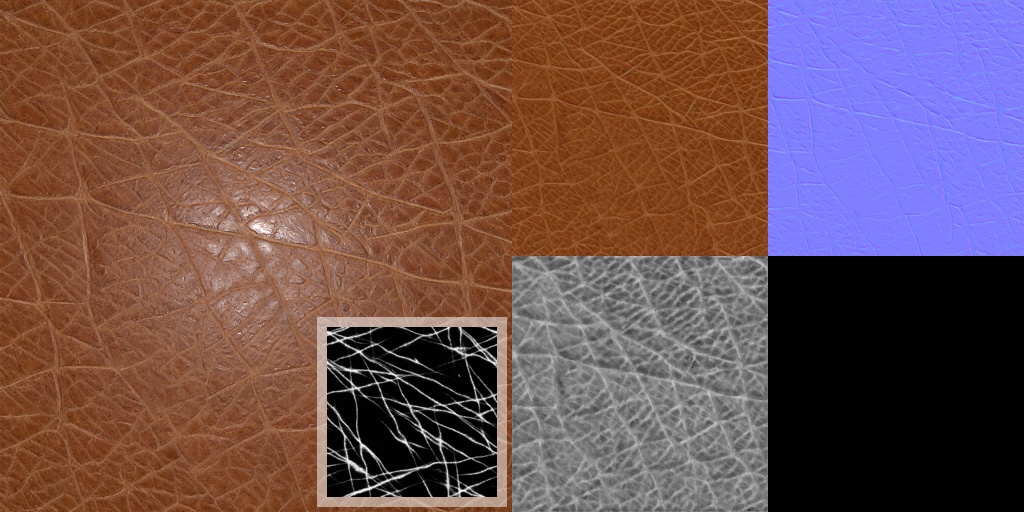} \\

\raisebox{+0.8\height}{\rotatebox[origin=c]{90}{\small{TileGen}}} &
\includegraphics[width=\TilegenWidth]{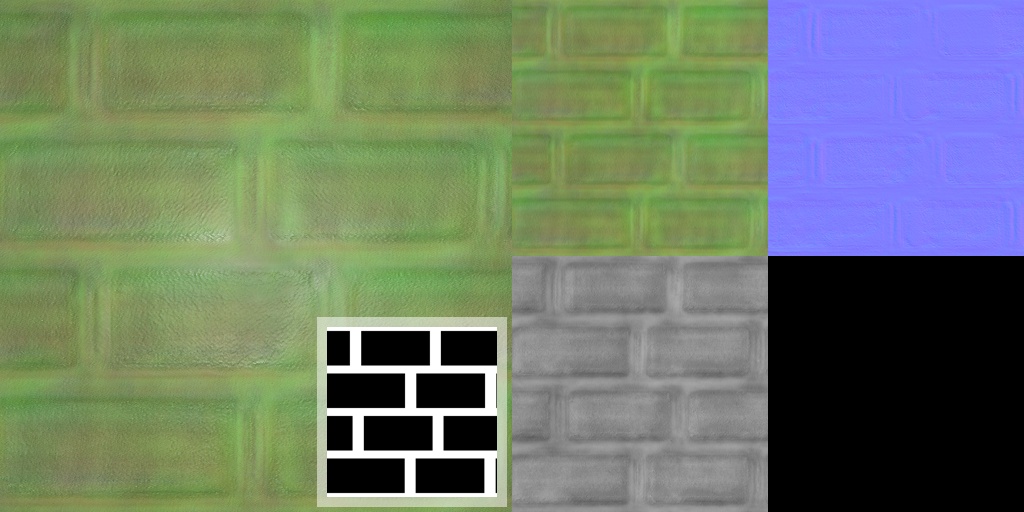} &
\includegraphics[width=\TilegenWidth]{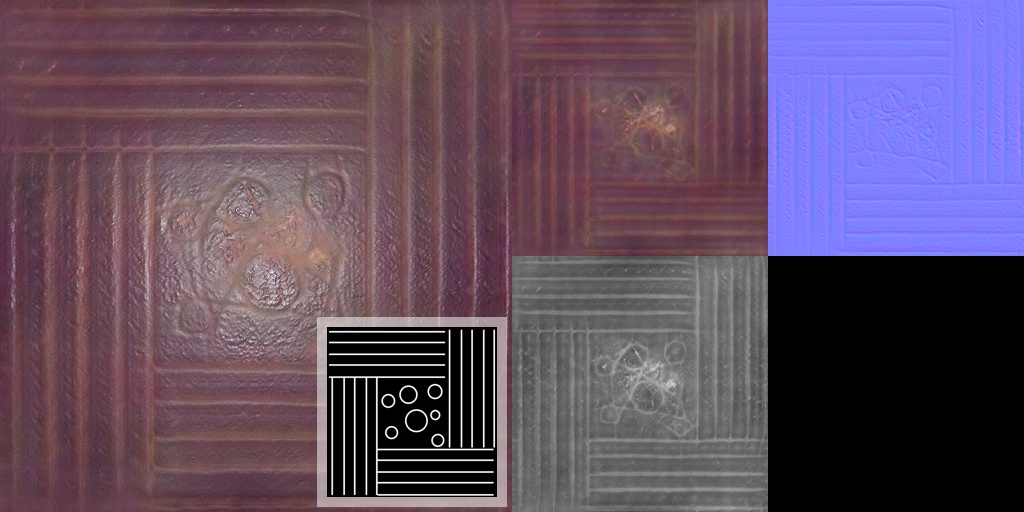} &
\includegraphics[width=\TilegenWidth]{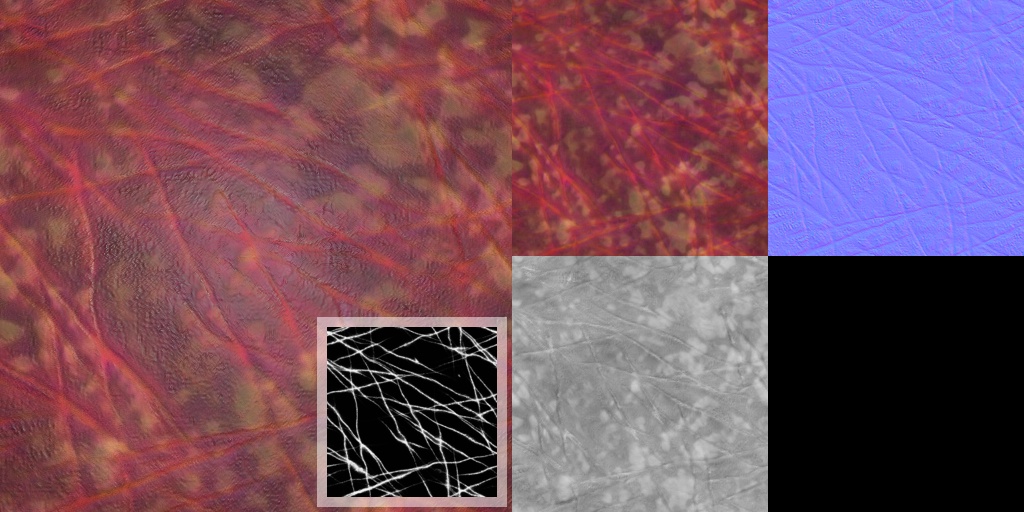} \\

\raisebox{+0.8\height}{\rotatebox[origin=c]{90}{\small{Ours}}} &
\includegraphics[width=\TilegenWidth]{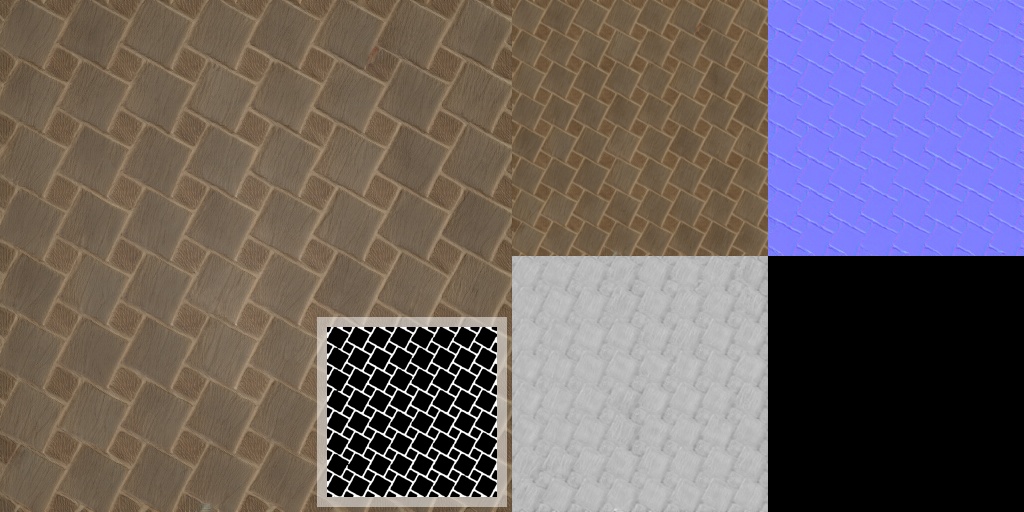} &
\includegraphics[width=\TilegenWidth]{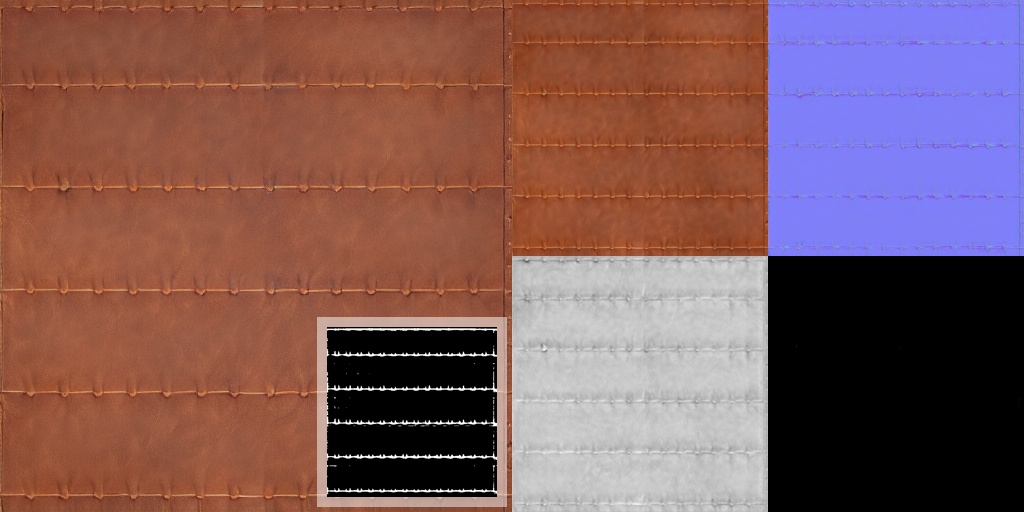} &

\includegraphics[width=\TilegenWidth]{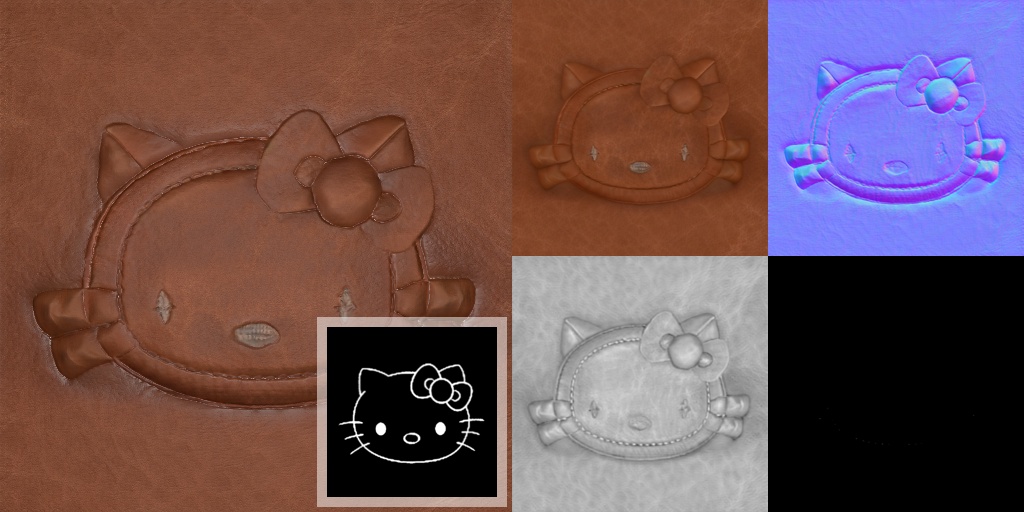} \\
\
\raisebox{+0.8\height}{\rotatebox[origin=c]{90}{\small{TileGen}}} &
\includegraphics[width=\TilegenWidth]{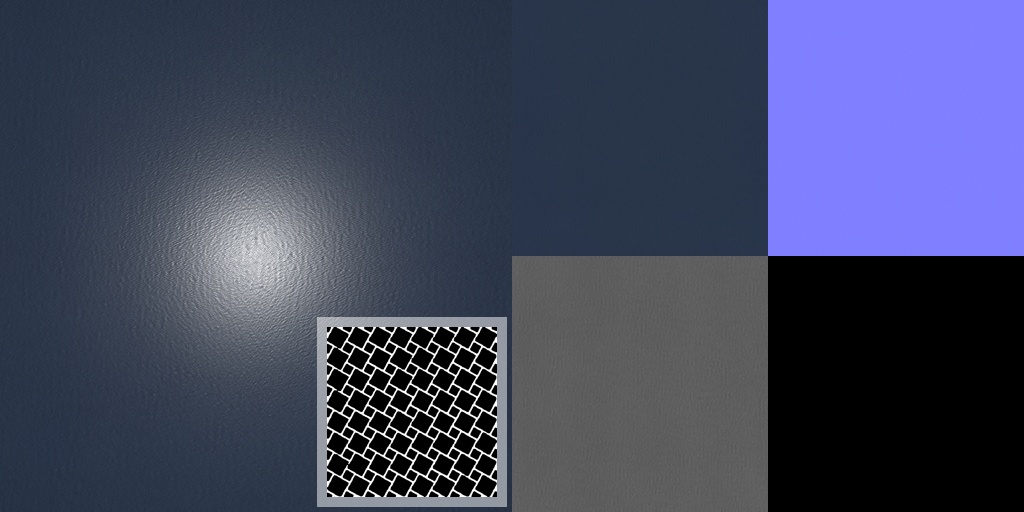} &
\includegraphics[width=\TilegenWidth]{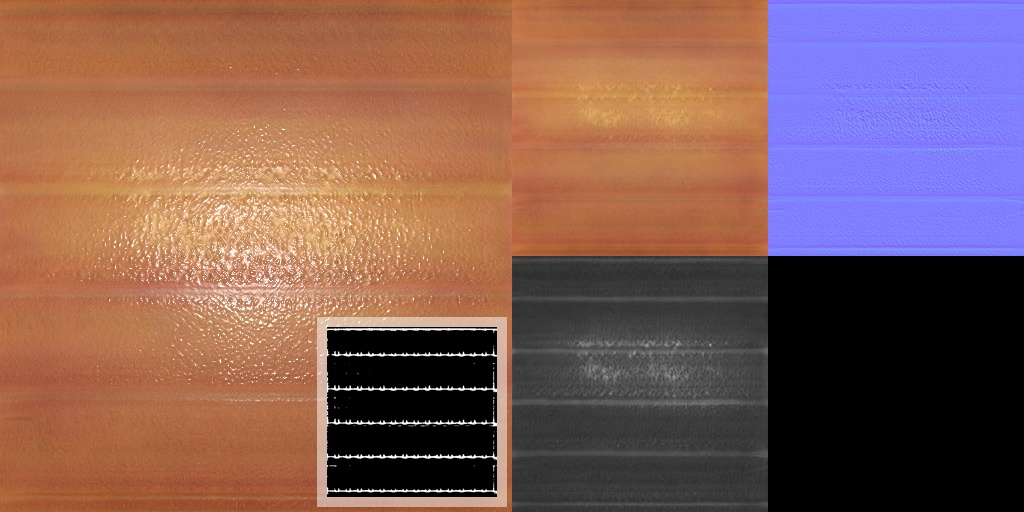} &
\includegraphics[width=\TilegenWidth]{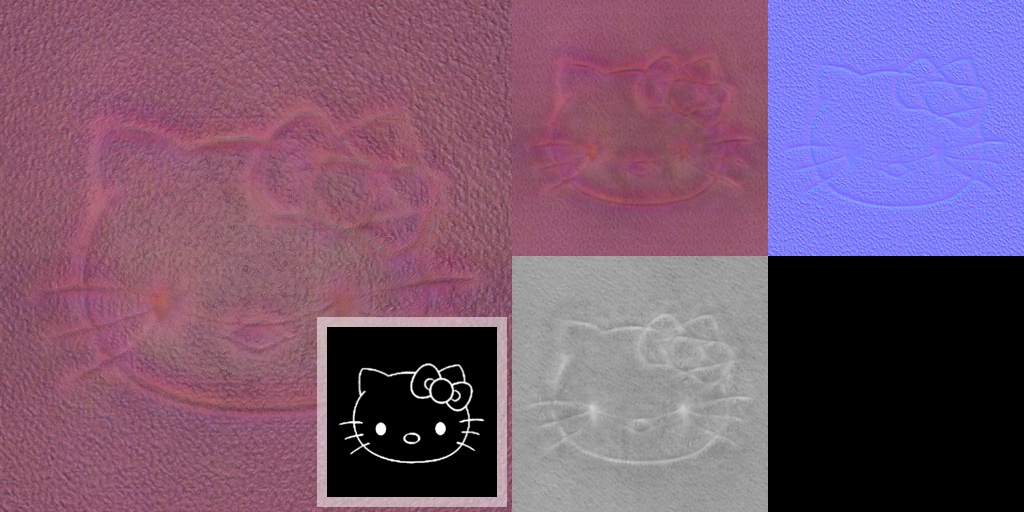} \\

\raisebox{+0.8\height}{\rotatebox[origin=c]{90}{\small{Ours}}} &
\includegraphics[width=\TilegenWidth]{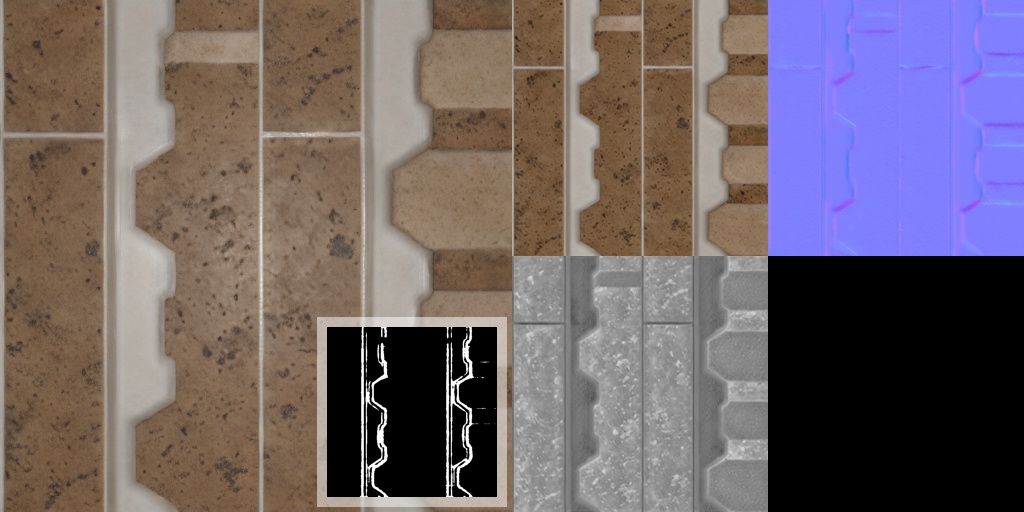} &
\includegraphics[width=\TilegenWidth]{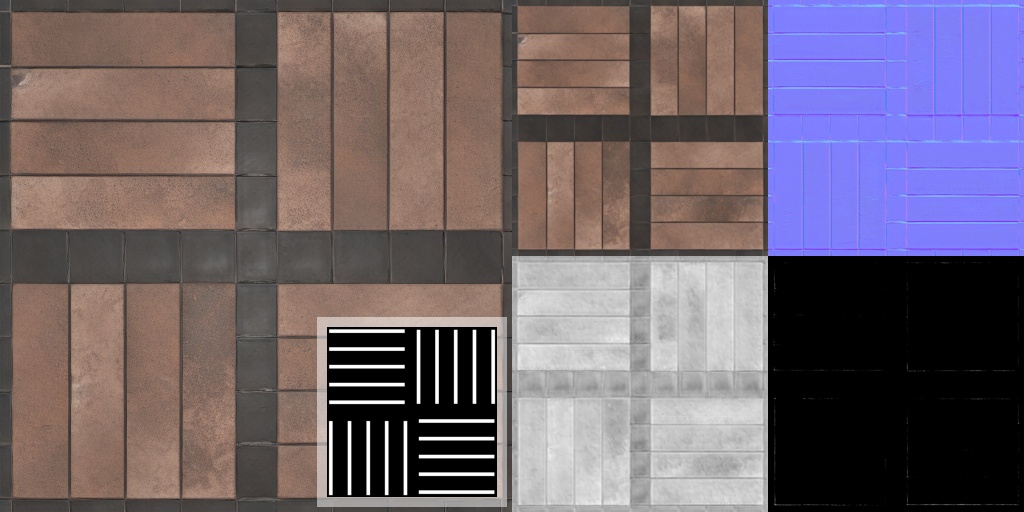} &
\includegraphics[width=\TilegenWidth]{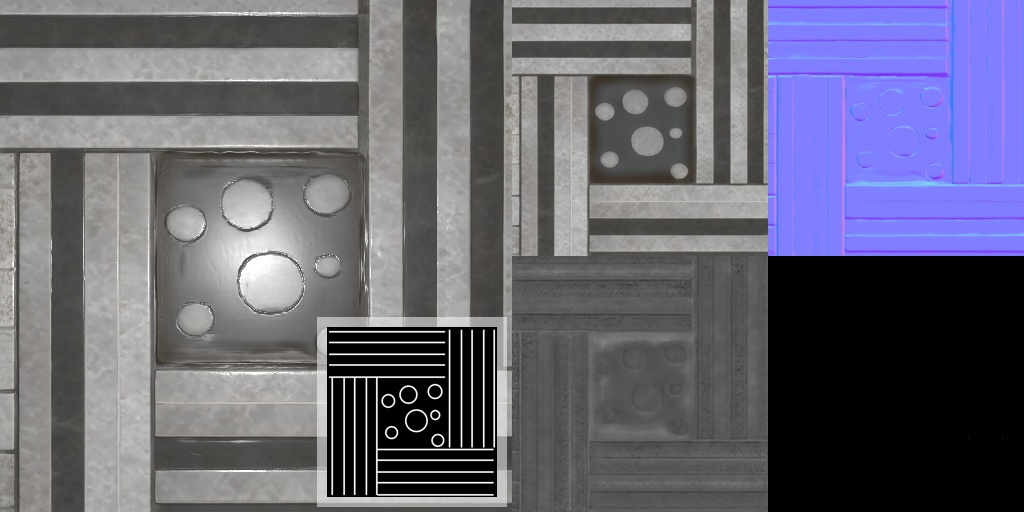} \\

\raisebox{+0.8\height}{\rotatebox[origin=c]{90}{\small{TileGen}}} &
\includegraphics[width=\TilegenWidth]{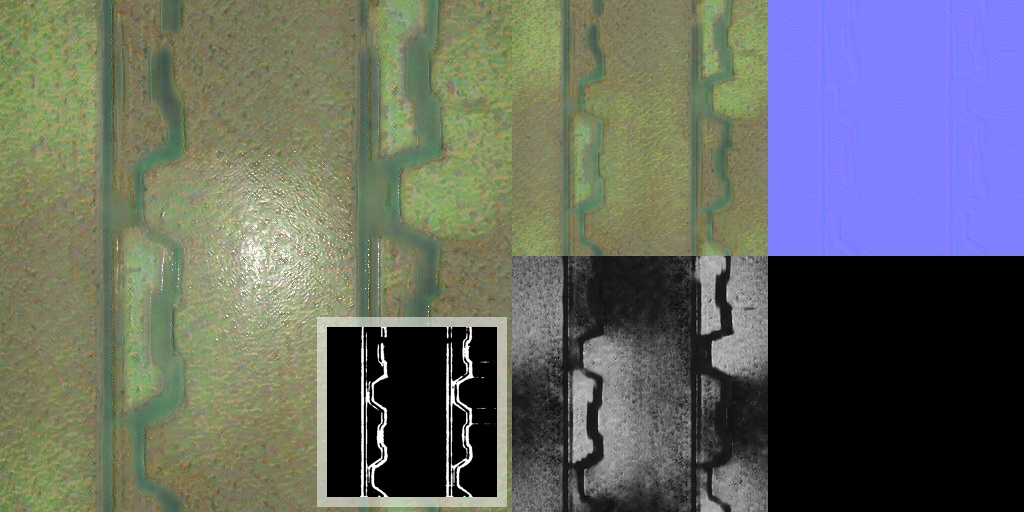} &
\includegraphics[width=\TilegenWidth]{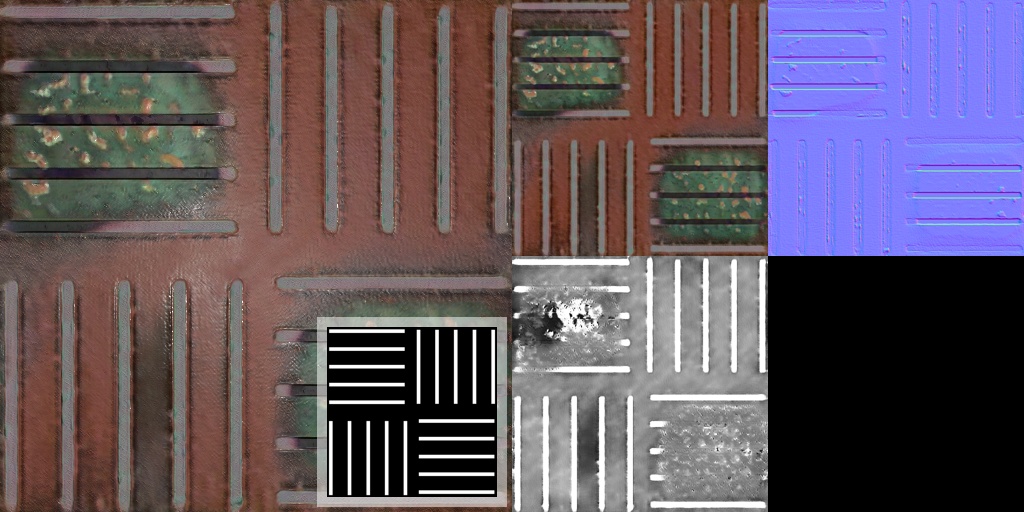} &
\includegraphics[width=\TilegenWidth]{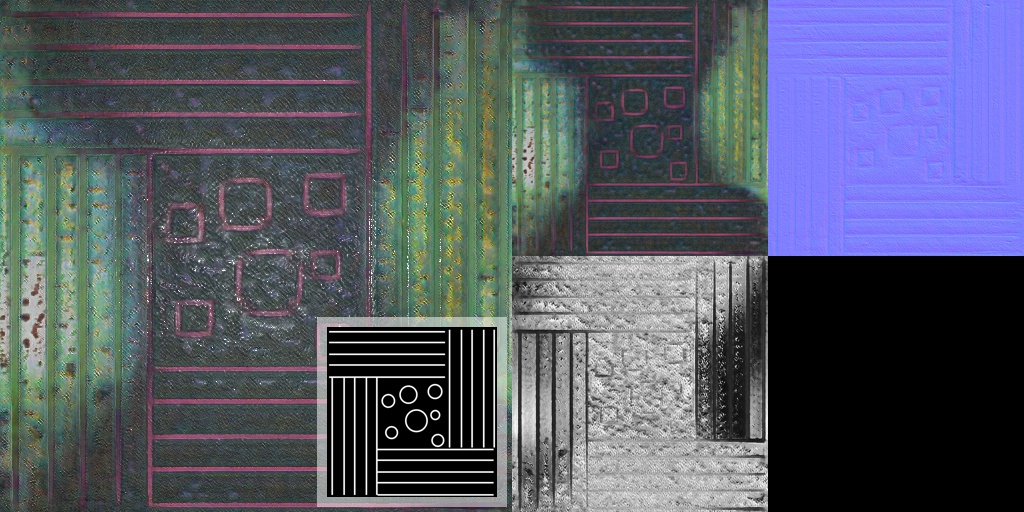} \\

\raisebox{+0.8\height}{\rotatebox[origin=c]{90}{\small{Ours}}} &
\includegraphics[width=\TilegenWidth]{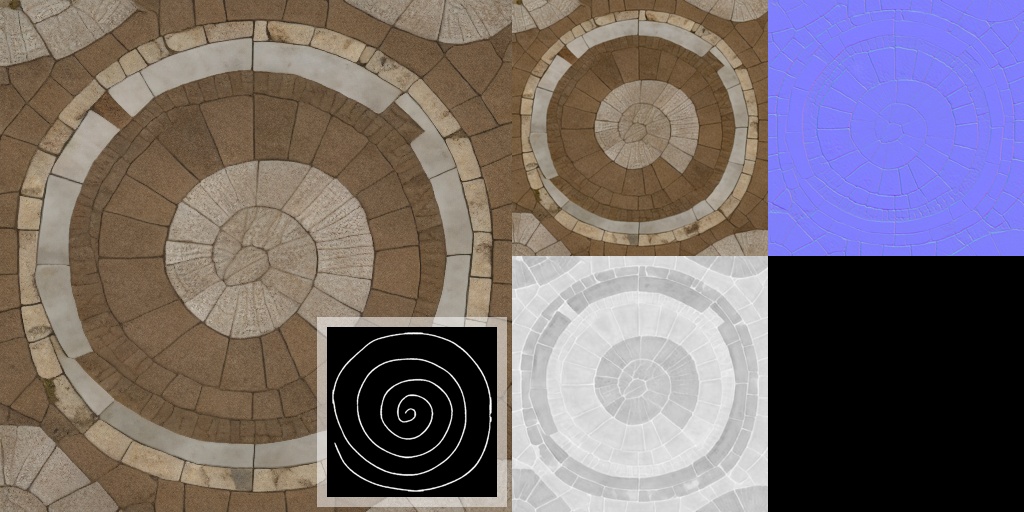} &
\includegraphics[width=\TilegenWidth]{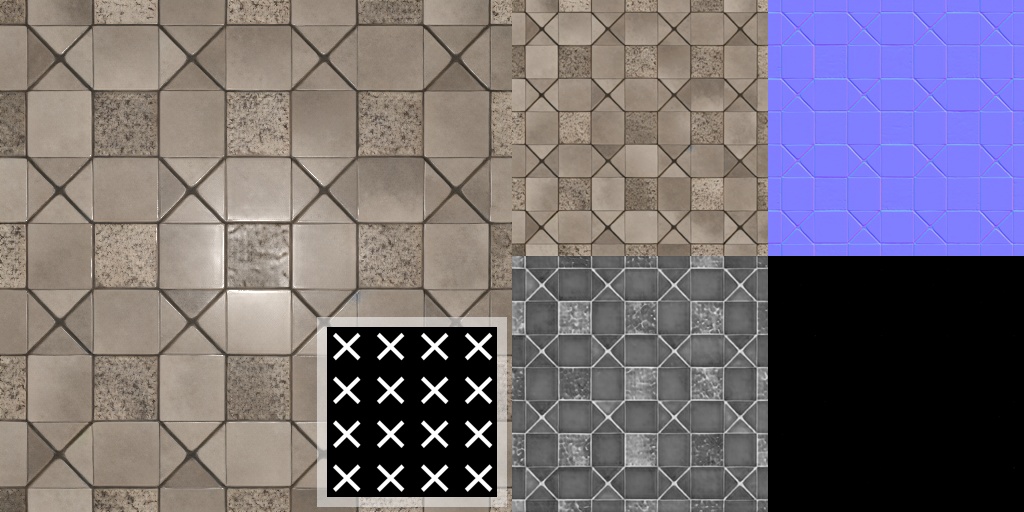} &
\includegraphics[width=\TilegenWidth]{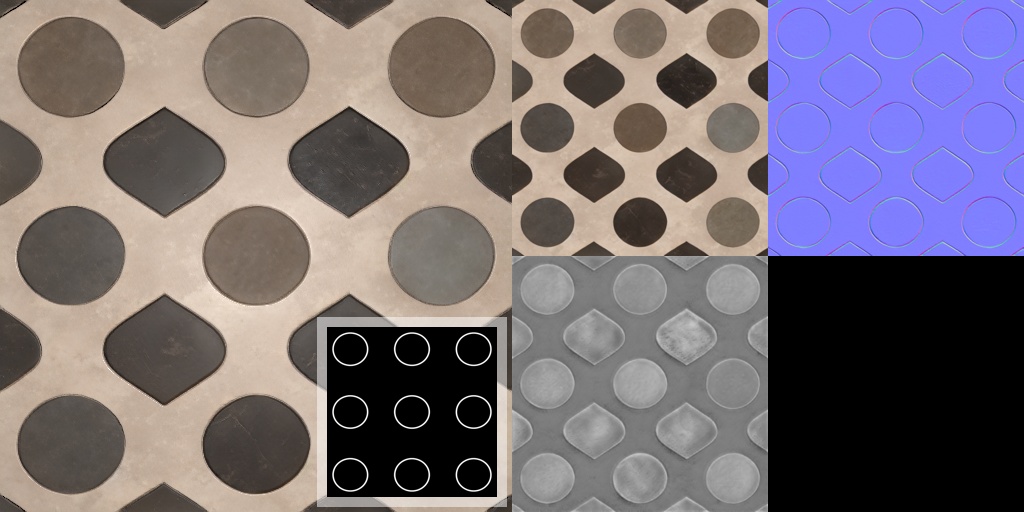} \\

\raisebox{+0.8\height}{\rotatebox[origin=c]{90}{\small{TileGen}}} &
\includegraphics[width=\TilegenWidth]{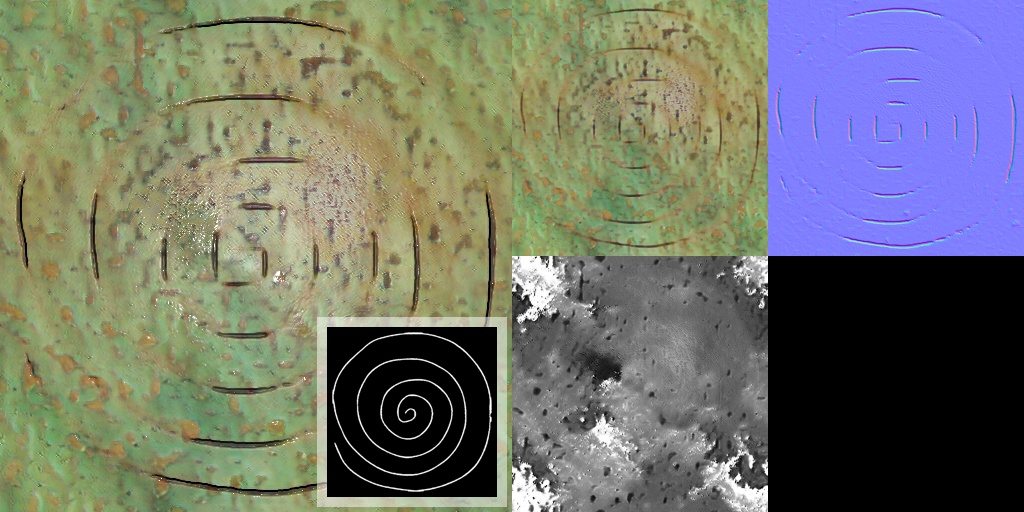} &
\includegraphics[width=\TilegenWidth]{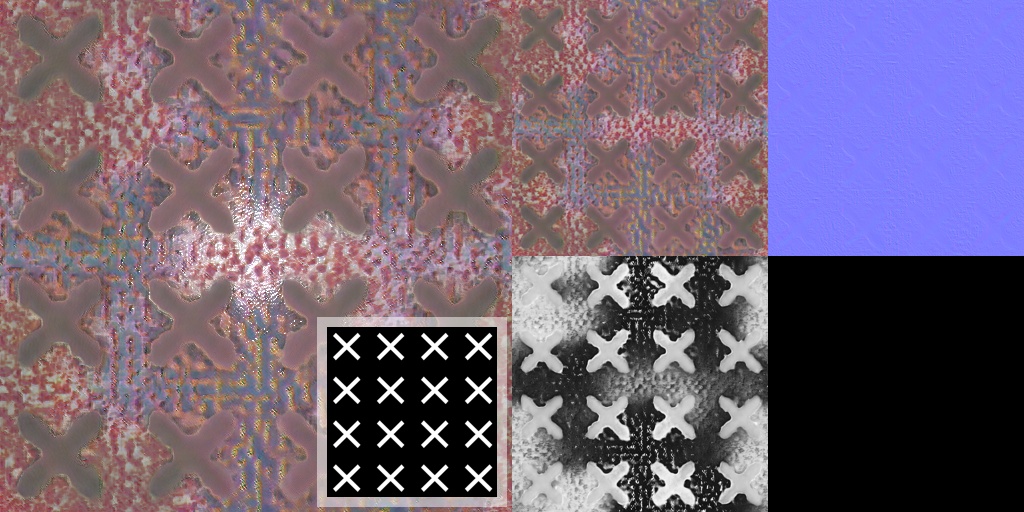} &
\includegraphics[width=\TilegenWidth]{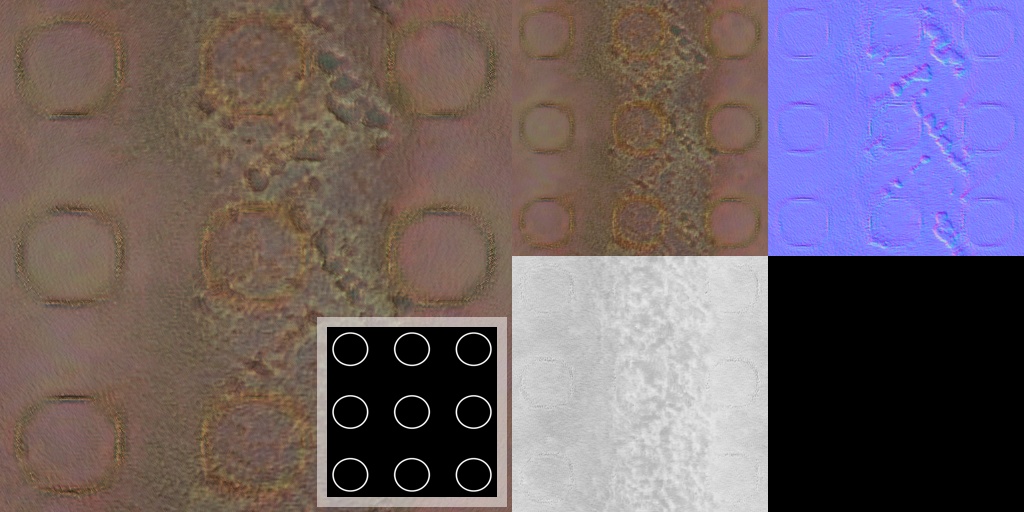} \\

\end{tabular}

\egroup
\caption{Qualitative comparison on pixel guidance. We compare our PixelControl against \citet{zhou2022tilegen} in leather (first 4 rows) and tile (last 4 rows). In some cases, TileGen fails to generate textures that match given patterns such as (column 1, row 4), (column 2, row 4), and (column 1, row 8). There are some artifacts in results to fit given patterns from \citet{zhou2022tilegen} such as (column 1, row 2), (column 2, row 6), (column 3, row 6), and (column 3, row 8). In contrast, ours show better consistency to pattern based on their natural materials.} 
\label{fig:tilegencon} 
\Description{fig:tilegen_con}
\end{figure}

\newcommand{\decoderWidth}{1.35cm} 
\begin{figure}[htbp]
\centering 

\bgroup

\def\arraystretch{0.5}
\setlength\tabcolsep{0.01cm}
\begin{tabular}{
    m{0.3cm}
    m{\decoderWidth}
    m{\decoderWidth}
    m{\decoderWidth}
    m{\decoderWidth}
    m{\decoderWidth}
    m{\decoderWidth}
    }
    
\multicolumn{1}{c}{} & \multicolumn{2}{c}{Render} & \multicolumn{1}{c}{SVBRDF} & \multicolumn{2}{c}{Render} & \multicolumn{1}{c}{SVBRDF} \\

\raisebox{+0.8\height}{\rotatebox[origin=c]{90}{\small{Reference}}} &
\includegraphics[width=\decoderWidth]{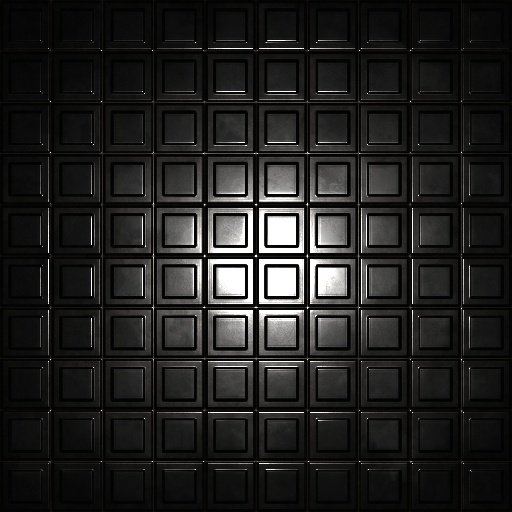} &
\includegraphics[width=\decoderWidth]{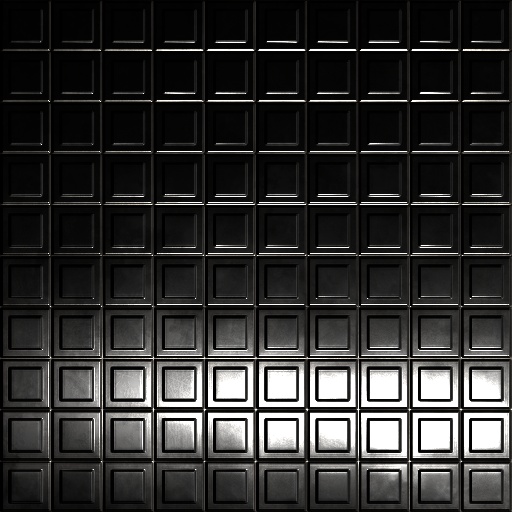} &
\includegraphics[width=\decoderWidth]{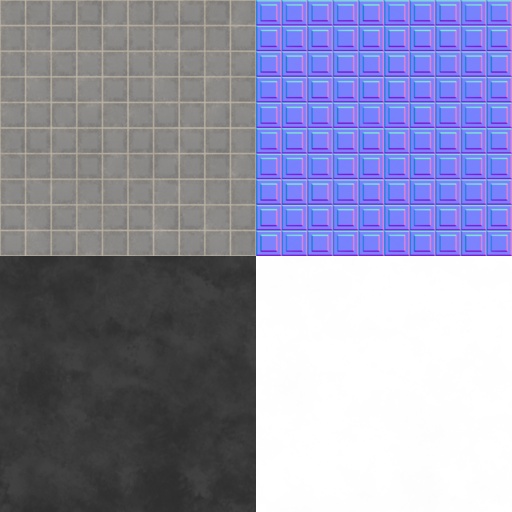} &
\includegraphics[width=\decoderWidth]{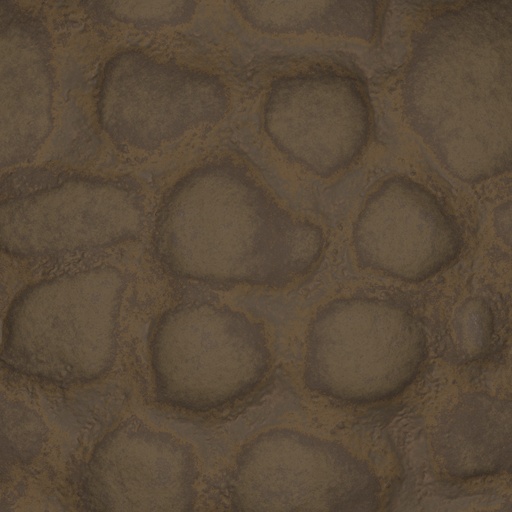} &
\includegraphics[width=\decoderWidth]{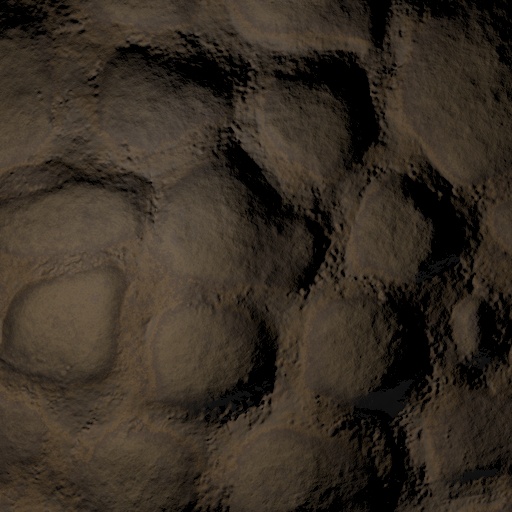} &
\includegraphics[width=\decoderWidth]{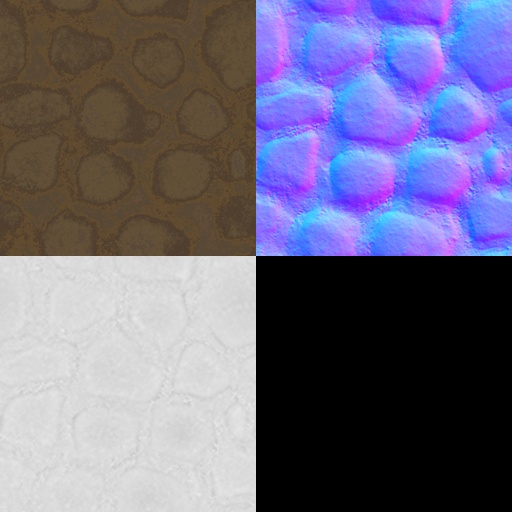} \\

\raisebox{+0.7\height}{\rotatebox[origin=c]{90}{\small{w/o $\mathcal{L}_\text{render}$}}} &
\includegraphics[width=\decoderWidth]{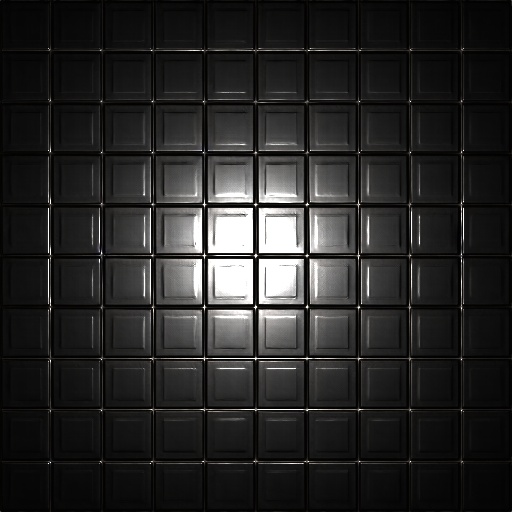} &
\includegraphics[width=\decoderWidth]{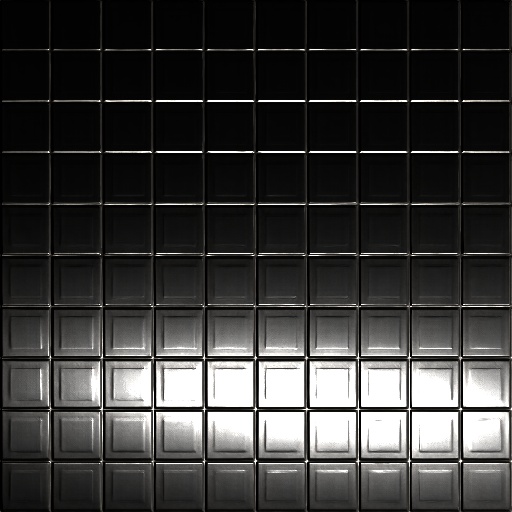} &
\includegraphics[width=\decoderWidth]{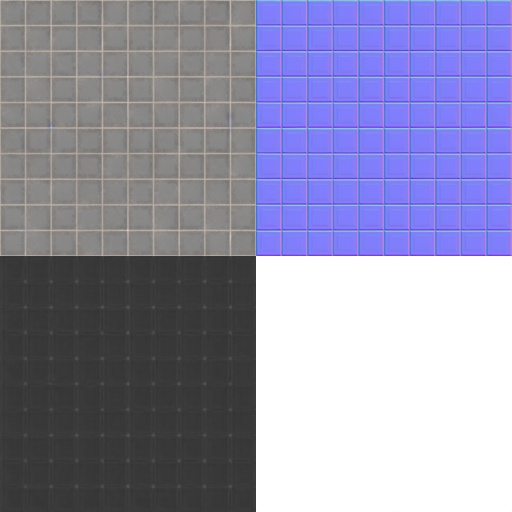} &
\includegraphics[width=\decoderWidth]{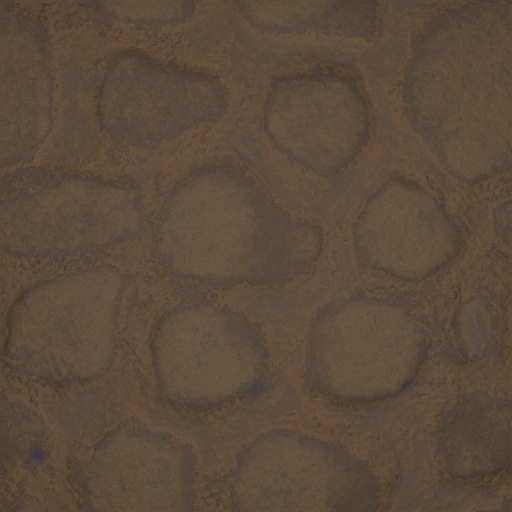} &
\includegraphics[width=\decoderWidth]{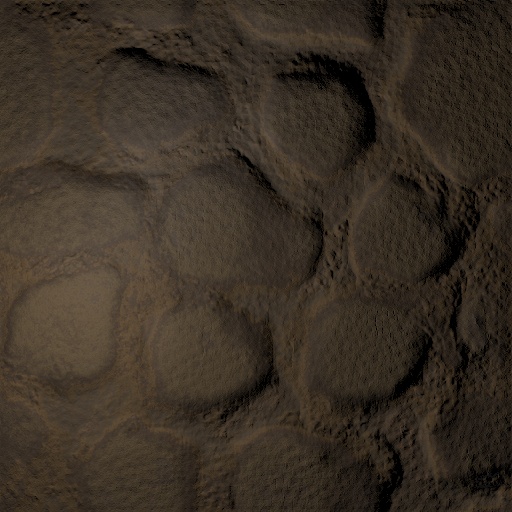} &
\includegraphics[width=\decoderWidth]{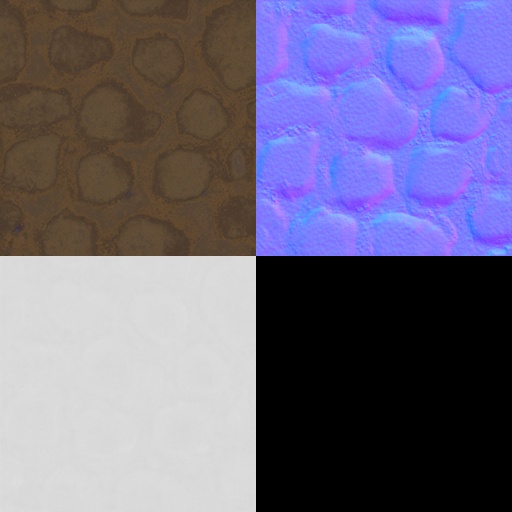} \\

\raisebox{+0.3\height}{\rotatebox[origin=c]{90}{\small{Ours}}} &
\includegraphics[width=\decoderWidth]{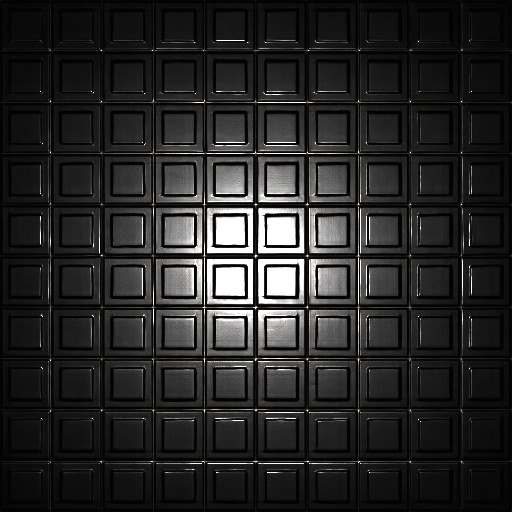} &
\includegraphics[width=\decoderWidth]{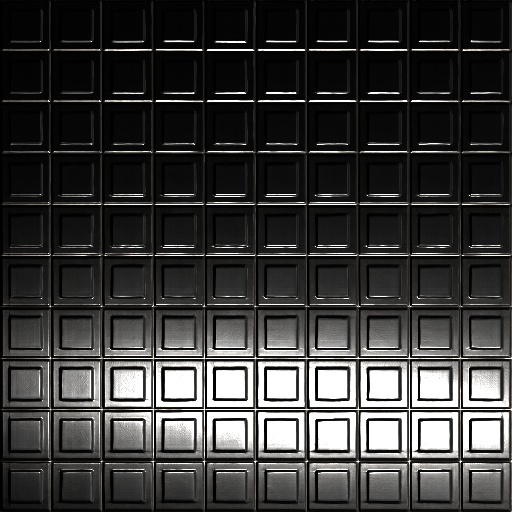} &
\includegraphics[width=\decoderWidth]{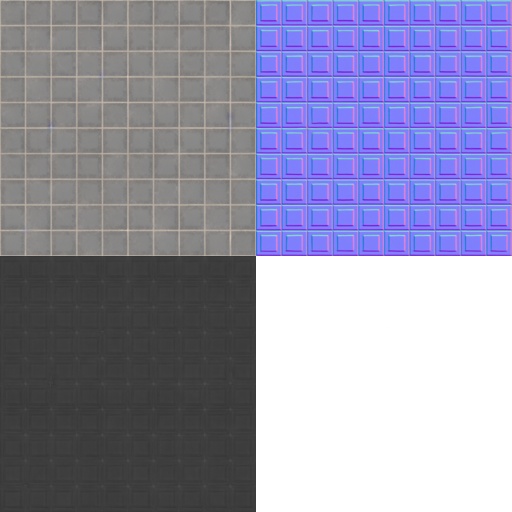} &
\includegraphics[width=\decoderWidth]{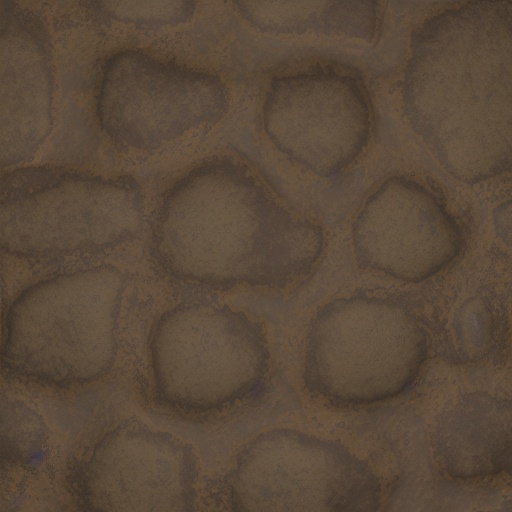} &
\includegraphics[width=\decoderWidth]{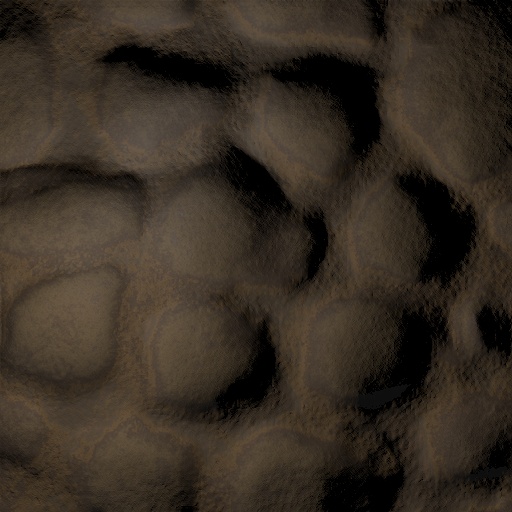} &
\includegraphics[width=\decoderWidth]{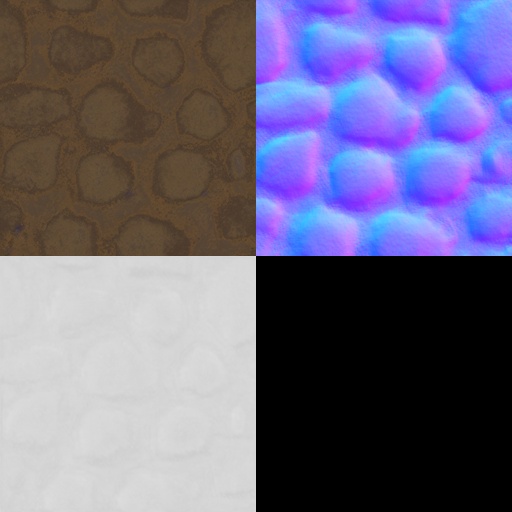} \\

\end{tabular}
% }

\egroup

\vspace{0.5cm} 
\bgroup
\small
\begin{tabular}{c|ccccc}
\multirow{2}{*}{}  & \multicolumn{1}{c|}{LPIPS} & \multicolumn{4}{c}{RMSE} \\
                    &  \multicolumn{1}{c|}{Render} & Albedo & Metallic & Normal & Roughness \\ \hline
 w/o $\mathcal{L}_\text{render}$  & 0.107           & 0.0361  & 0.0126 & 0.0542 & 0.0406\\
 Ours(w/ $\mathcal{L}_\text{render}$)  & 0.101      & 0.0357 & 0.0086 & 0.0531 & 0.0365\\
\end{tabular}

\egroup

\caption{Qualitative and quantitative comparison on PBR decoder with rendering loss. We trained two PBR decoders with and without rendering loss. With rendering loss, the decoded textures show better consistency in both rendering images and SVBRDF (especially for normal maps) visually and achieve the lowest LPIPS of rendering images and RMSE of SVBRDF maps as illustrated in the table.} 
\label{fig:decoder}
\Description{fig:decoder}
\end{figure}

\newcommand{\SuperResWidth}{1.65cm} 
\begin{figure}[htbp]
\centering 

\bgroup

\def\arraystretch{0.5} 
\setlength\tabcolsep{0.01cm} 
\begin{tabular}{
    m{\SuperResWidth}
    m{\SuperResWidth}
    m{\SuperResWidth}
    m{\SuperResWidth}
    m{\SuperResWidth}
    }
    
\multicolumn{5}{c}{Reference} \\
\multicolumn{5}{c}{\includegraphics[width=8.29cm]{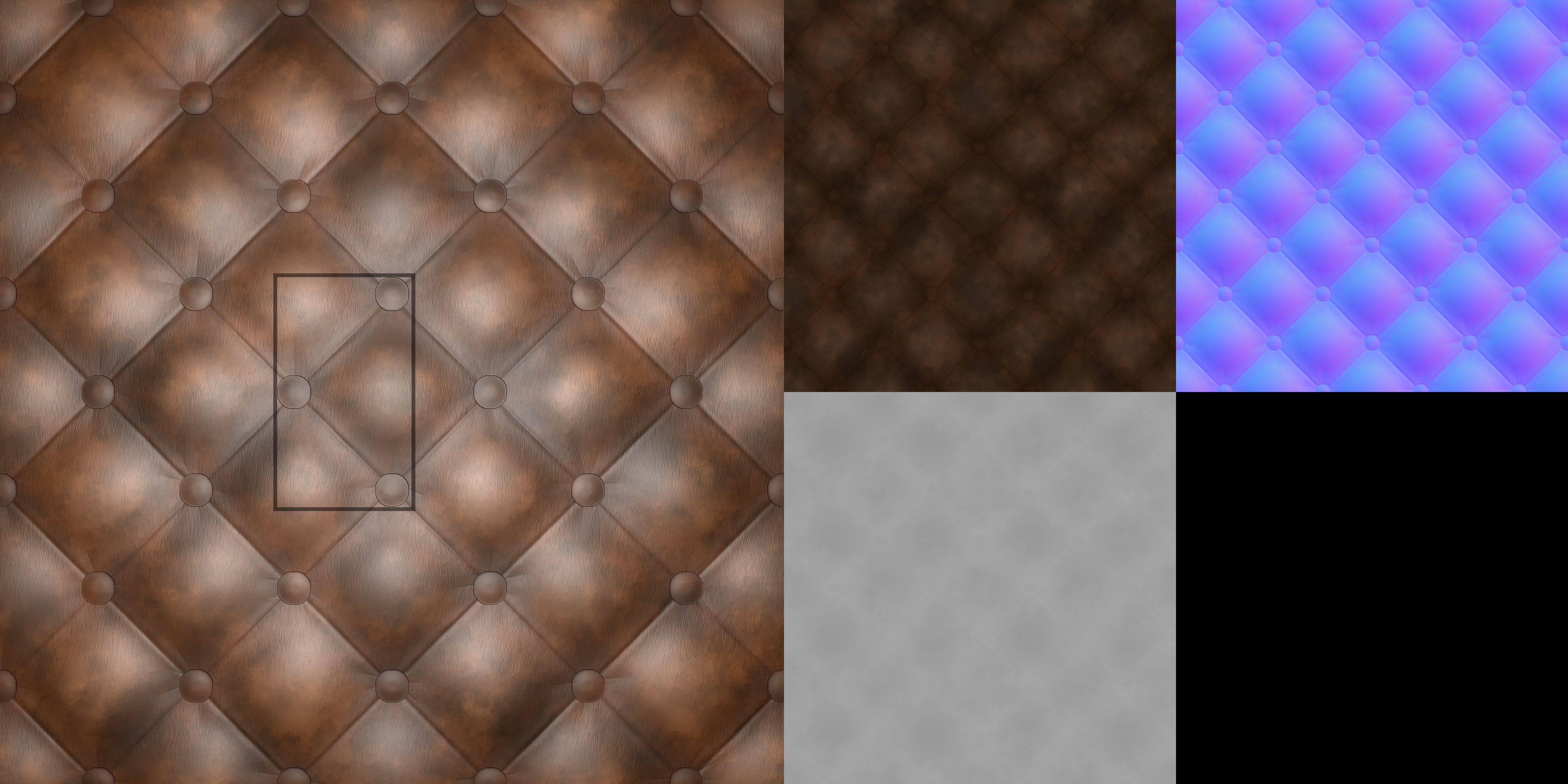}} \\

\multicolumn{1}{c}{\begin{minipage}{\SuperResWidth} \centering \scriptsize {Reference}\end{minipage} } & \multicolumn{1}{c}{\begin{minipage}{\SuperResWidth} \centering \scriptsize {Low Res.}\end{minipage} } & \multicolumn{1}{c}{\begin{minipage}{\SuperResWidth} \centering \scriptsize {Pretrained}\end{minipage} } & \multicolumn{1}{c}{\begin{minipage}{\SuperResWidth} \centering \scriptsize {w/o $\mathcal{L}_\text{render}$}\end{minipage} } & \multicolumn{1}{c}{\begin{minipage}{\SuperResWidth} \centering \scriptsize {Ours(w/ $\mathcal{L}_\text{render}$)}\end{minipage} } \\

\includegraphics[width=\SuperResWidth]{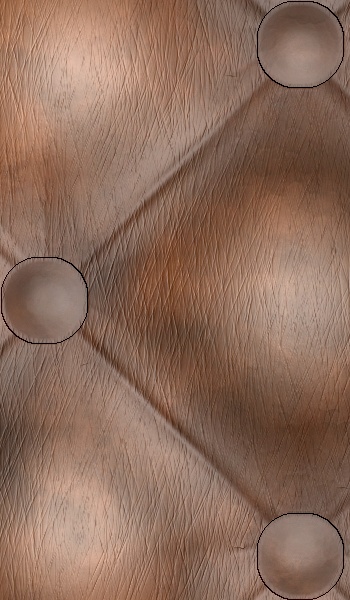} &
\includegraphics[width=\SuperResWidth]{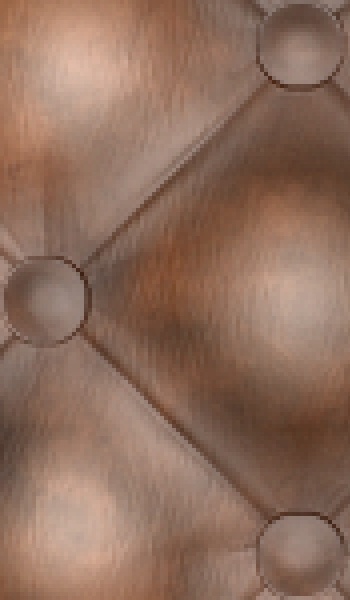} &
\includegraphics[width=\SuperResWidth]{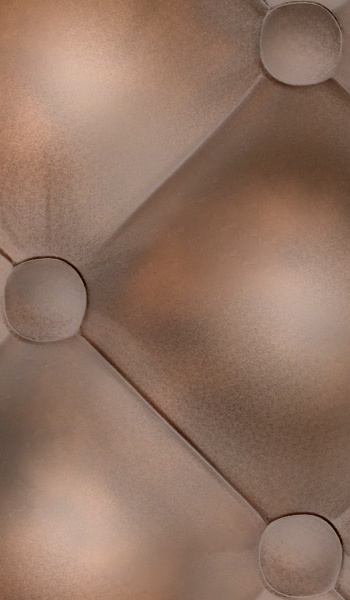} &
\includegraphics[width=\SuperResWidth]{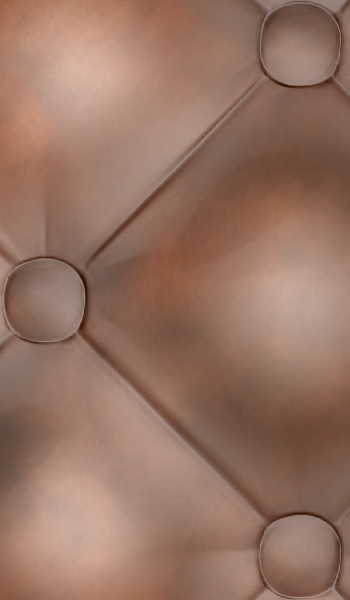} &
\includegraphics[width=\SuperResWidth]{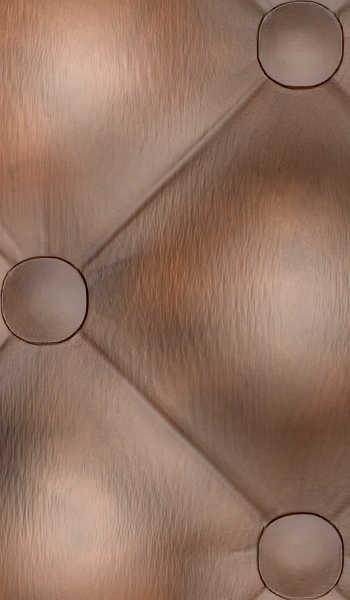} \\

\end{tabular}

\egroup

\vspace{0.5cm} 
\bgroup
\small
\begin{tabular}{c|ccccc}
\multirow{2}{*}{}  & \multicolumn{1}{c|}{LPIPS} & \multicolumn{4}{c}{RMSE} \\
    &  \multicolumn{1}{c|}{Render} & Albedo & Metal. & Normal & Rough. \\ \hline
 Pretrained  & 0.450 & 0.0272 & 0.0816 &  0.0598 &  0.0588 \\
 w/o $\mathcal{L}_\text{render}$  & 0.342 & 0.0248  & 0.0652 & 0.0474 & 0.0451\\
 Ours(w/ $\mathcal{L}_\text{render}$)  & 0.321 & 0.0211 & 0.0643 & 0.0398 & 0.0445\\
\end{tabular}

\egroup

\caption{Qualitative and quantitative comparison on Super-Resolution module. We show local zoom rendering results with high-resolution textures($2048\times 2048$), low-resolution textures($512\times 512$), and textures from three super-resolution modules\cite{wang2021realesrgan} with different training strategies: pre-trained one, fine-tuned one without $\mathcal{L}_\text{render}$ and fine-tuned one with $\mathcal{L}_\text{render}$(ours). Our final methods produce details and textures more consistent with ground truth compared with incomplete methods as shown in five images at the bottom of the figure. With fine-tuning our datasets and $\mathcal{L}_\text{render}$, the table below presents our final models giving the lowest LPIPS between rendering images with super-resolution textures and rendering images with real high-resolution textures, and lowest RMSE between super-resolution SVBRDF maps and ground truth.} 
\label{fig:SR} 
\Description{fig:SR}
\end{figure}

\newcommand{\HighlightWidth}{1.35cm} 
\begin{figure}[htbp]
\centering 

\bgroup
\small
\def\arraystretch{0.5} 
\setlength\tabcolsep{0.01cm}

\begin{subfigure}[b]{8.2cm}

\begin{tabular}{
    m{\HighlightWidth}
    m{\HighlightWidth}
    m{\HighlightWidth}
    m{\HighlightWidth}
    m{\HighlightWidth}
    m{\HighlightWidth}
    }
\multicolumn{1}{c}{\begin{minipage}{\HighlightWidth} \centering \small{w/o HA}\end{minipage} } & \multicolumn{1}{c}{\begin{minipage}{\HighlightWidth} \centering \small{w/ HA}\end{minipage} } &  \multicolumn{1}{c}{\begin{minipage}{\HighlightWidth} \centering \small{w/o HA}\end{minipage} } & \multicolumn{1}{c}{\begin{minipage}{\HighlightWidth} \centering \small{w/ HA}\end{minipage} } & \multicolumn{1}{c}{\begin{minipage}{\HighlightWidth} \centering \small{w/o HA}\end{minipage} } & \multicolumn{1}{c}{\begin{minipage}{\HighlightWidth} \centering \small{w/ HA}\end{minipage} } \\

\includegraphics[width=\HighlightWidth]{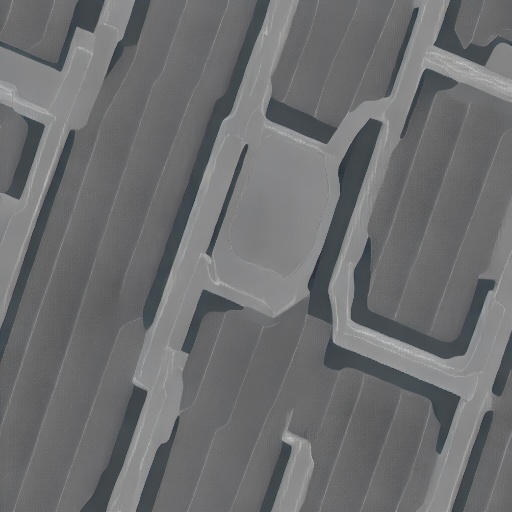} &
\includegraphics[width=\HighlightWidth]{src/imgs/ablation/highlight/1.jpg} &
\includegraphics[width=\HighlightWidth]{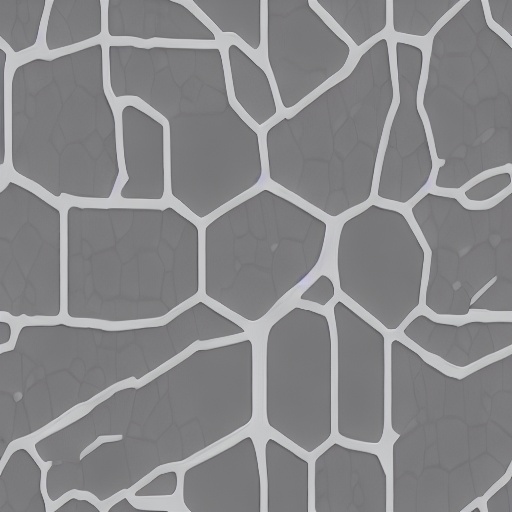} &
\includegraphics[width=\HighlightWidth]{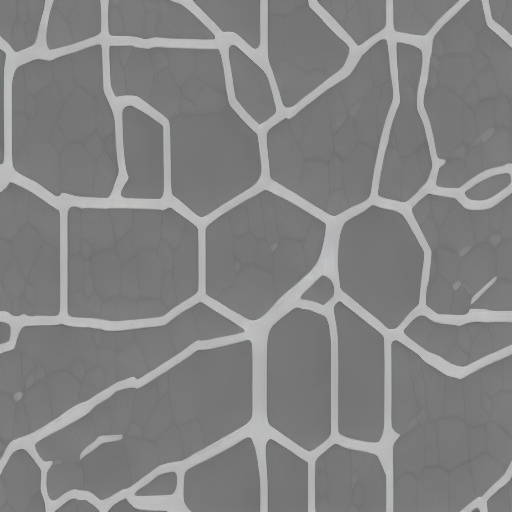} &
\includegraphics[width=\HighlightWidth]{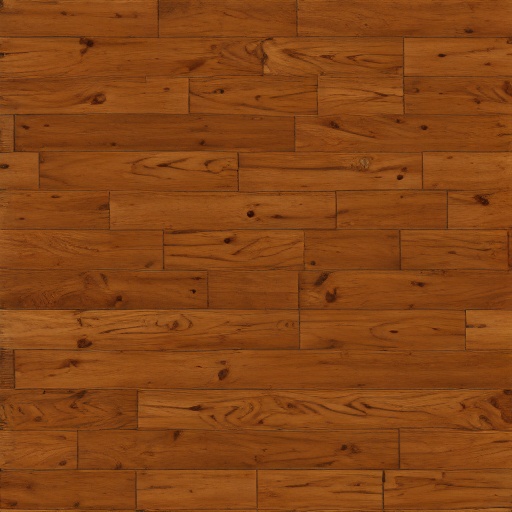} &
\includegraphics[width=\HighlightWidth]{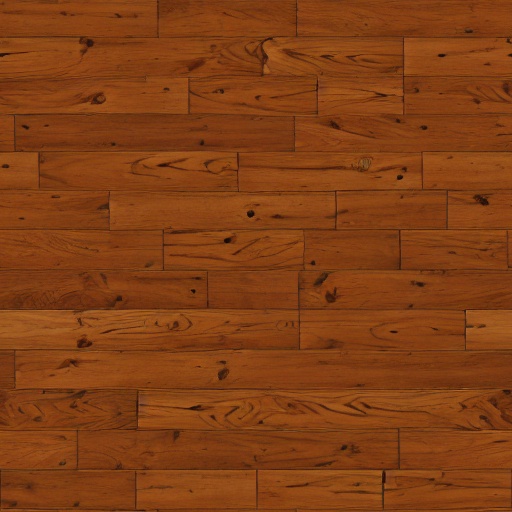} \\

\includegraphics[width=\HighlightWidth]{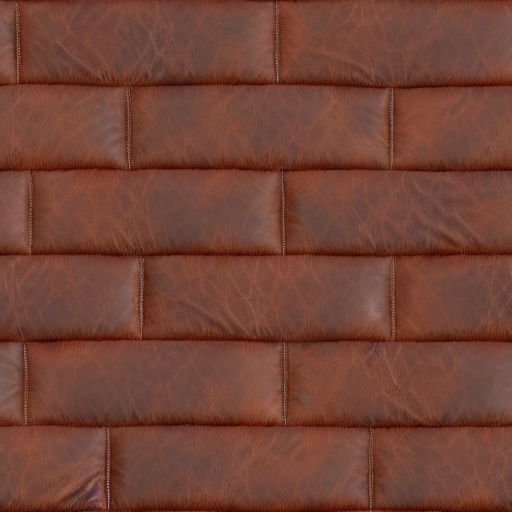} &
\includegraphics[width=\HighlightWidth]{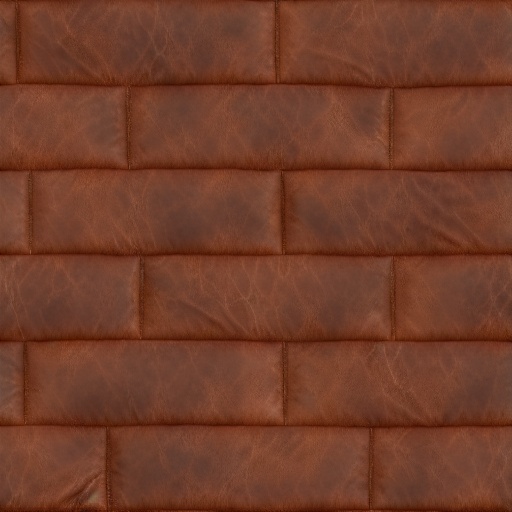} &

\includegraphics[width=\HighlightWidth]{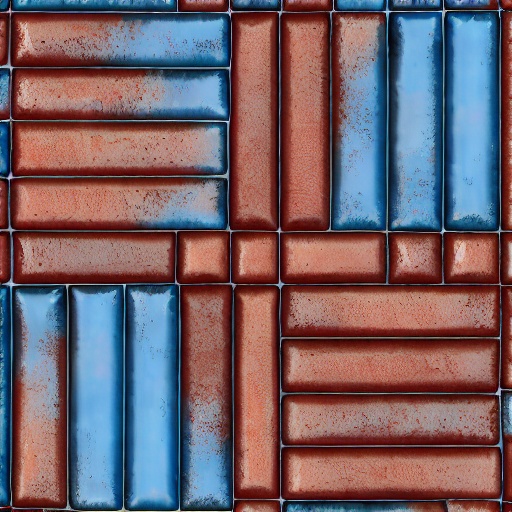} &
\includegraphics[width=\HighlightWidth]{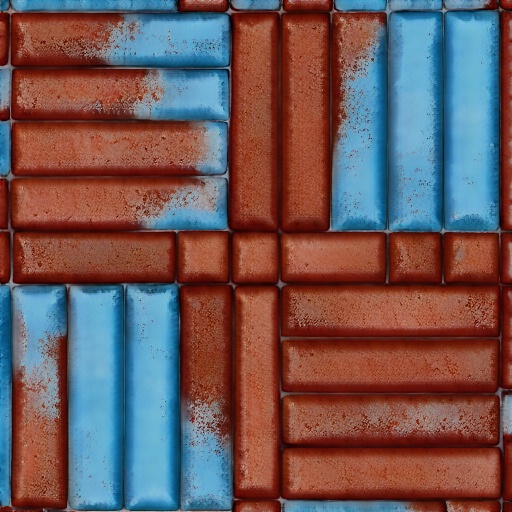} &

\includegraphics[width=\HighlightWidth]{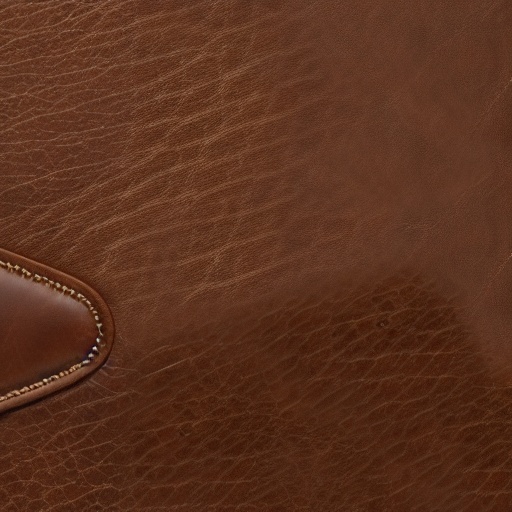} &
\includegraphics[width=\HighlightWidth]{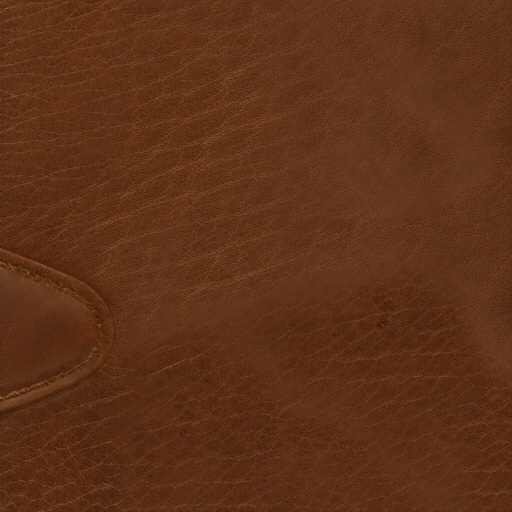} \\

\end{tabular}

\caption{Generated cases}
\label{fig:sub1}
\end{subfigure}
\begin{subfigure}[b]{8.2cm}

\begin{tabular}{
    m{\HighlightWidth}
    m{\HighlightWidth}
    m{\HighlightWidth}
    m{\HighlightWidth}
    m{\HighlightWidth}
    m{\HighlightWidth}
    }

\multicolumn{1}{c}{Input} & \multicolumn{1}{c}{w/o HA} &  \multicolumn{1}{c}{w/ HA} & \multicolumn{1}{c}{Input} & \multicolumn{1}{c}{w/o HA} &  \multicolumn{1}{c}{w/ HA} \\
\includegraphics[width=\HighlightWidth]{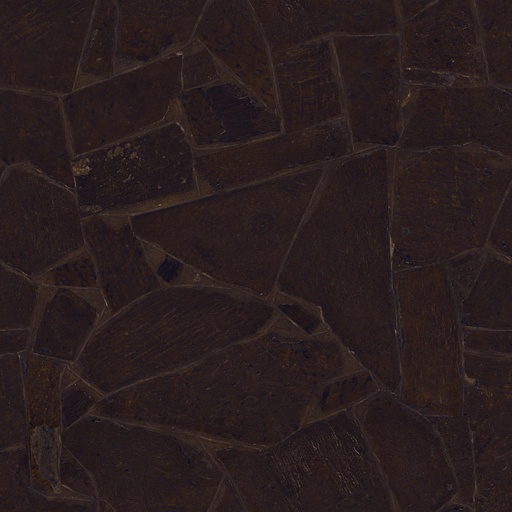} &
\includegraphics[width=\HighlightWidth]{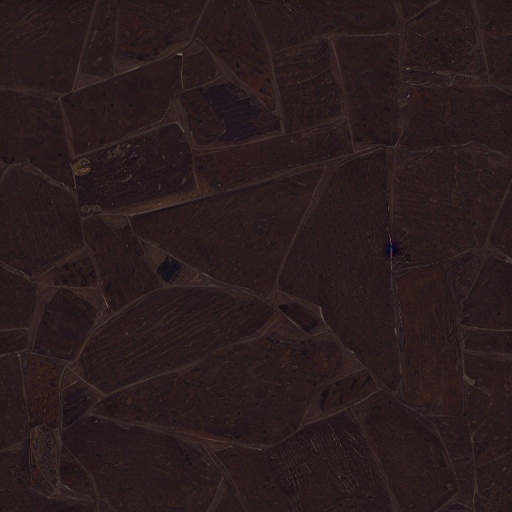} &
\includegraphics[width=\HighlightWidth]{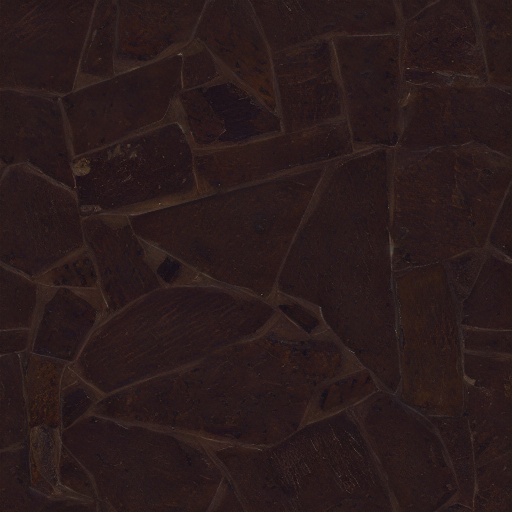} &
\includegraphics[width=\HighlightWidth]{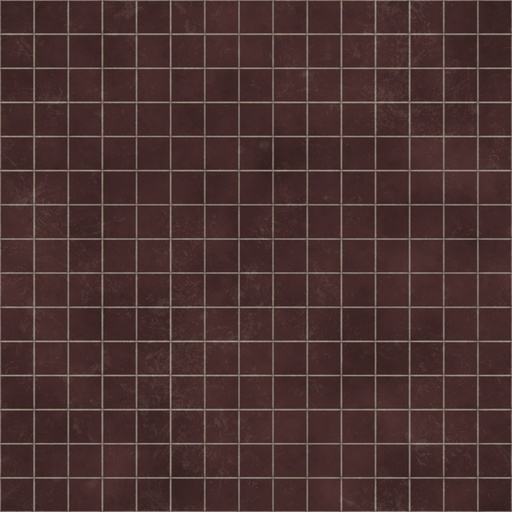} &
\includegraphics[width=\HighlightWidth]{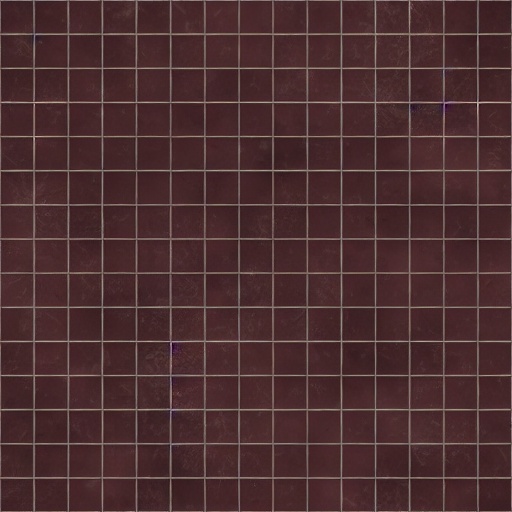} &
\includegraphics[width=\HighlightWidth]{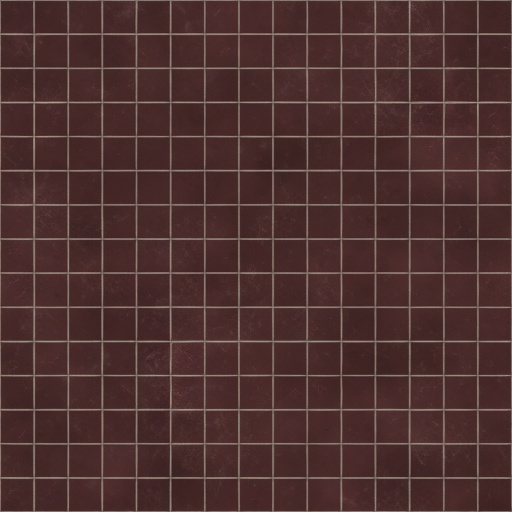} \\

\end{tabular}

\begin{tabular}{
    m{1cm}
    m{1cm}
    m{1cm}
    m{1cm}
    m{1cm}
    m{1cm}
    m{1cm}
    m{1cm}
    }
\multicolumn{1}{c}{Input}  &\multicolumn{1}{c}{Refer.} &   \multicolumn{1}{c}{w/o HA} &    \multicolumn{1}{c}{w/ HA} &  \multicolumn{1}{c}{Input}  &   \multicolumn{1}{c}{Refer.} & \multicolumn{1}{c}{w/o HA} &  \multicolumn{1}{c}{w/ HA} \\
\includegraphics[width=1cm]{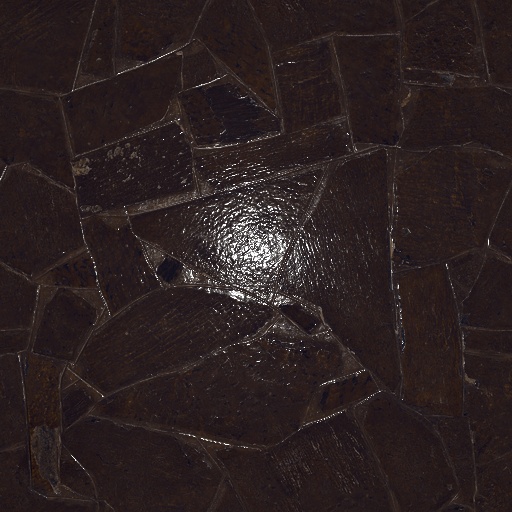} &
\includegraphics[width=1cm]{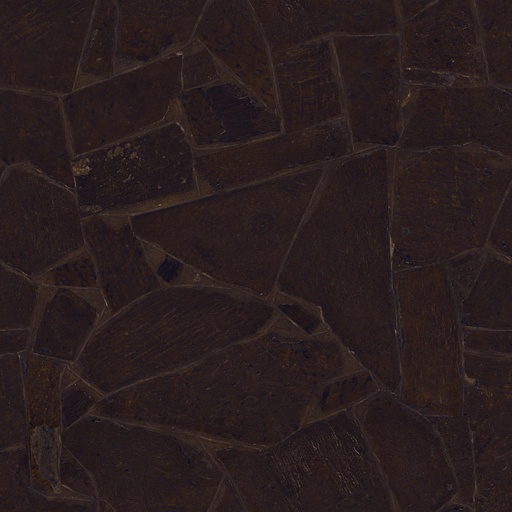} &
\includegraphics[width=1cm]{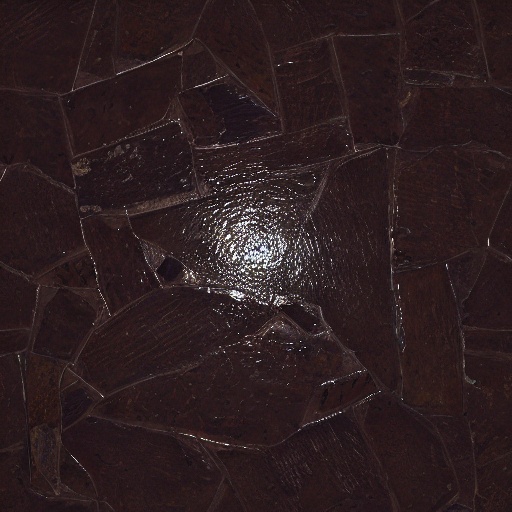} &
\includegraphics[width=1cm]{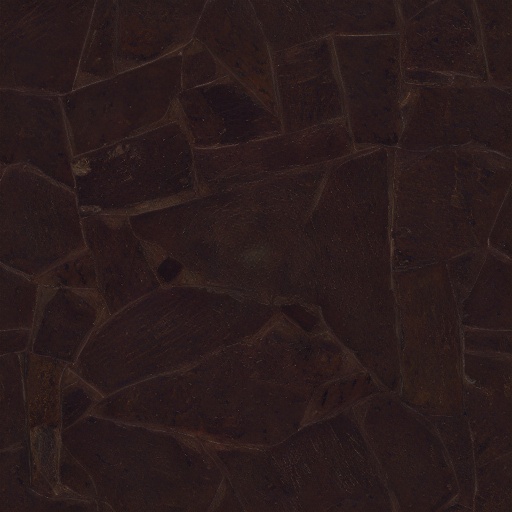} &
\includegraphics[width=1cm]{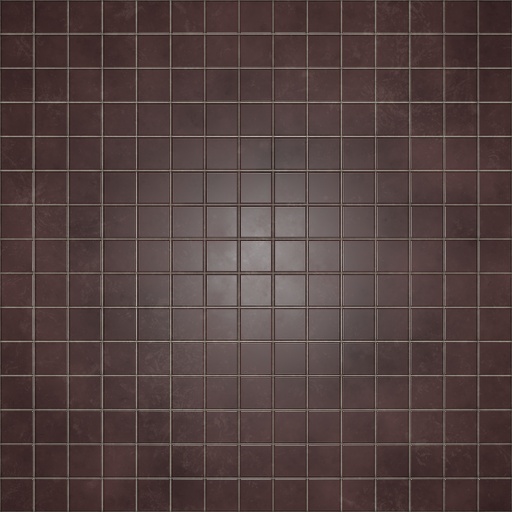} &
\includegraphics[width=1cm]{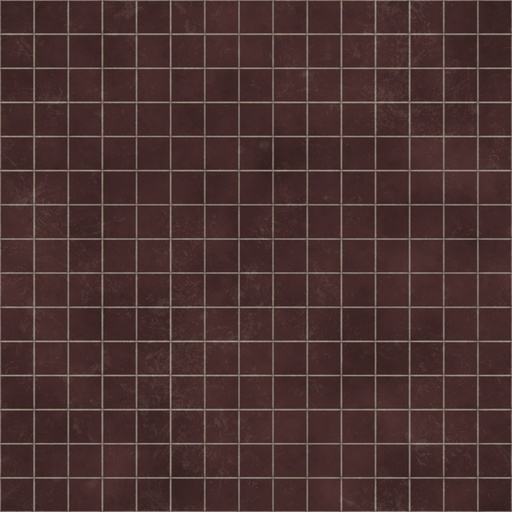} &
\includegraphics[width=1cm]{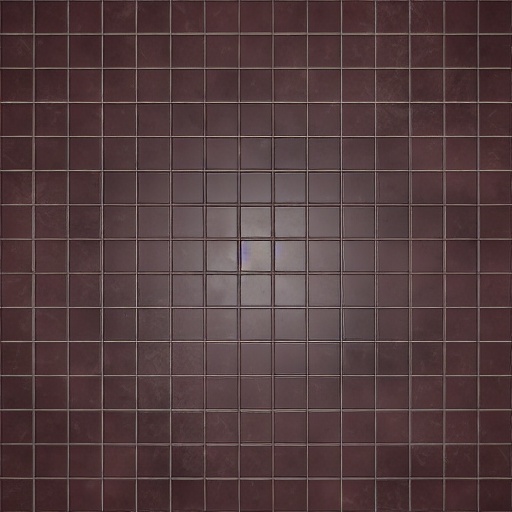} &
\includegraphics[width=1cm]{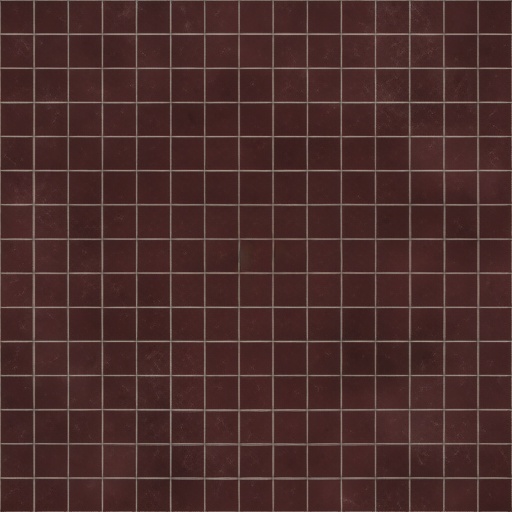} \\

\end{tabular}

\caption{Test cases in dataset}
\label{fig:sub2}
\end{subfigure}

\egroup

\vspace{0.5cm} 

\begin{tabular}{c|ccc|ccc}
& \multicolumn{3}{c|}{highlight inputs} & \multicolumn{3}{c}{non-highlight inputs} \\
& L1 & PSNR & LPIPS & L1 & PSNR & LPIPS \\ \hline
w/o HA & 0.0409 & 25.7460 & 0.1928 & 0.0201 & 33.2621 & 0.1220 \\
w/ HA & 0.0211 & 32.6578 & 0.1452 & 0.0202 & 33.2904 & 0.1241
\end{tabular}

\caption{Qualitative and quantitative comparison on highlight-aware module (HA). As shown in \autoref{fig:sub1}, the original albedo decoder from VAE may generate data with highlight and the highlight-aware module can de-highlight them to make better generation results. As shown in \autoref{fig:sub2}, For those inputs without highlight, there is a tiny difference regardless of whether the highlight module is used or not. When inputting images with the highlight, we can de-highlight them with our highlight-aware module. For images with and without highlights, the table below numerically reaches the same conclusion by comparing the L1, PSNR, and LPIPS between real albedo maps and de-highlighted albedo maps.} 
\label{fig:highlight_decoder} 
\Description{fig:highlight_decoder}
\end{figure}

\newcommand{\AblationPixelWidth}{1.3cm} 
\begin{figure}[htbp]
\centering

\bgroup

\def\arraystretch{0.1}
\setlength\tabcolsep{0.25pt} 
\begin{tabular}{
    m{0.3cm}
    m{\AblationPixelWidth}
    m{\AblationPixelWidth}
    m{\AblationPixelWidth}
    @{\hspace{5pt}}
    m{\AblationPixelWidth}
    m{\AblationPixelWidth}
    m{\AblationPixelWidth}
    }

\raisebox{+0.8\height}{\rotatebox[origin=c]{90}{\small{w/o ft}}} &
\includegraphics[width=\AblationPixelWidth]{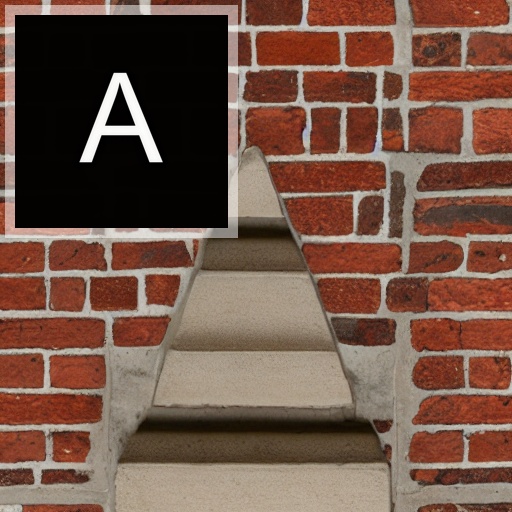} &
\includegraphics[width=\AblationPixelWidth]{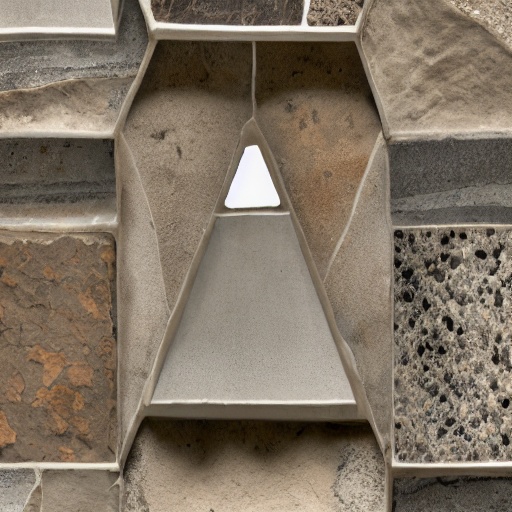} &
\includegraphics[width=\AblationPixelWidth]{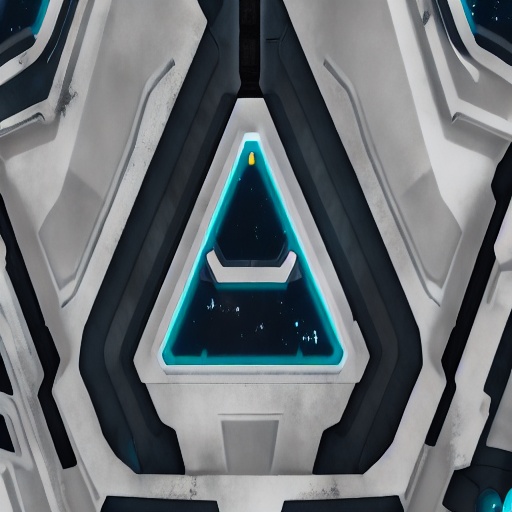} &
\includegraphics[width=\AblationPixelWidth]{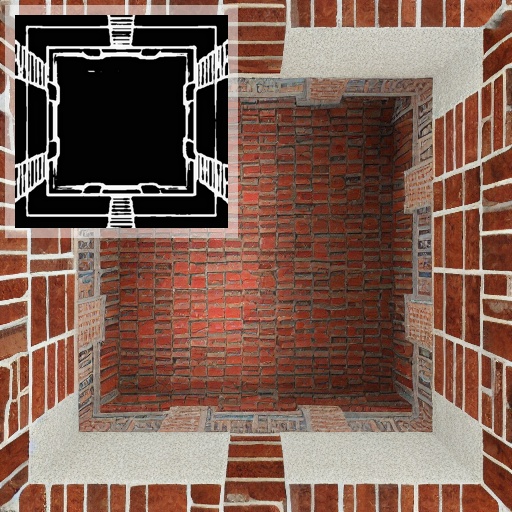} &
\includegraphics[width=\AblationPixelWidth]{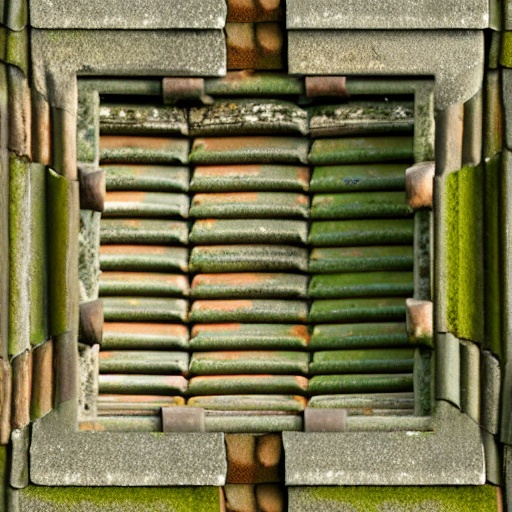} &
\includegraphics[width=\AblationPixelWidth]{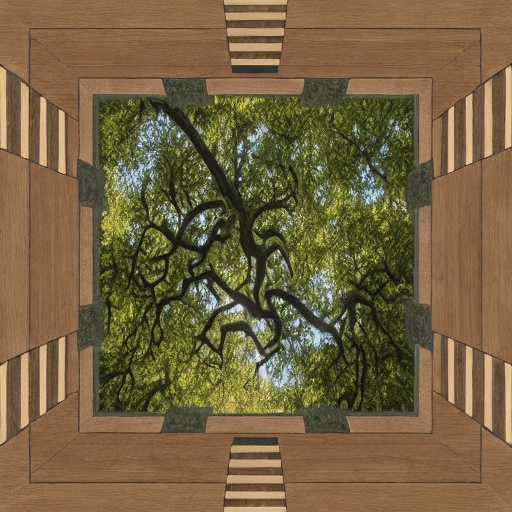}\\

\raisebox{+0.8\height}{\rotatebox[origin=c]{90}{\small{Ours}}} &
\includegraphics[width=\AblationPixelWidth]{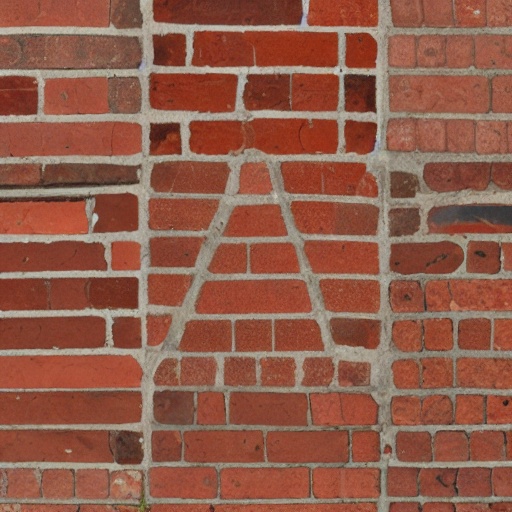} &
\includegraphics[width=\AblationPixelWidth]{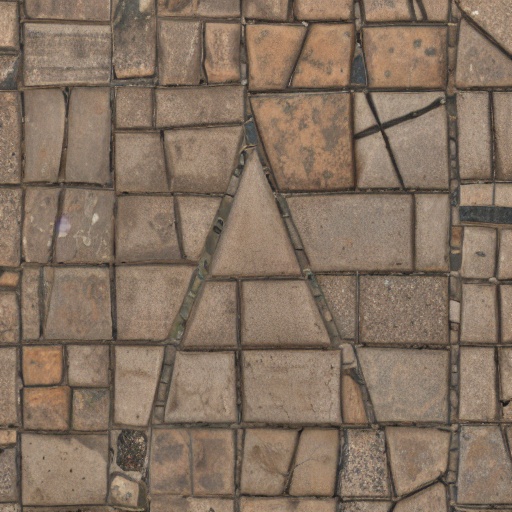} &
\includegraphics[width=\AblationPixelWidth]{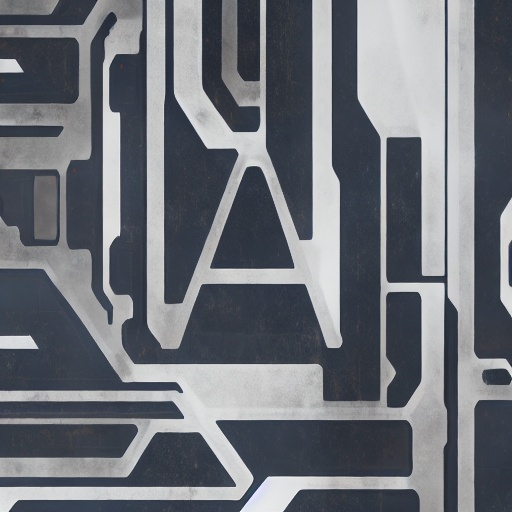} &
\includegraphics[width=\AblationPixelWidth]{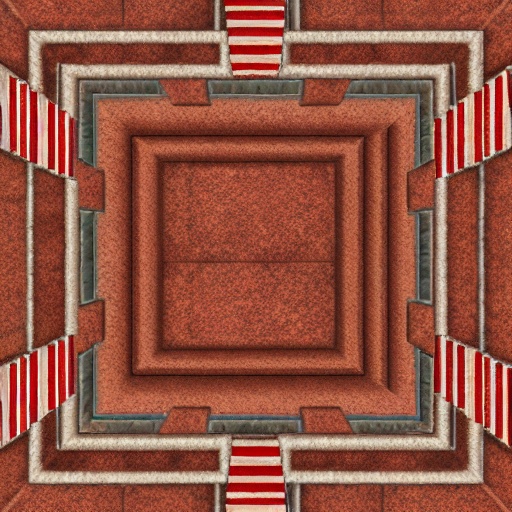} &
\includegraphics[width=\AblationPixelWidth]{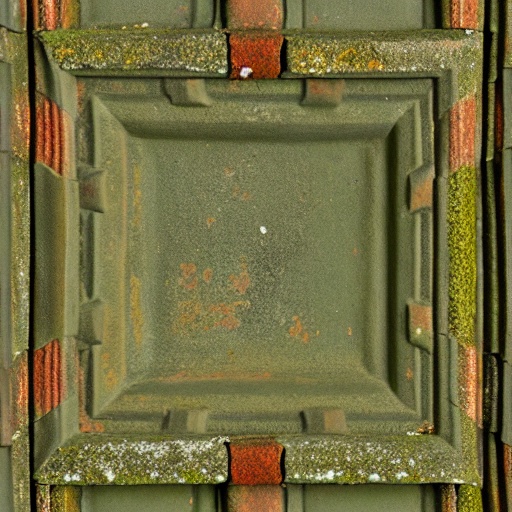} &
\includegraphics[width=\AblationPixelWidth]{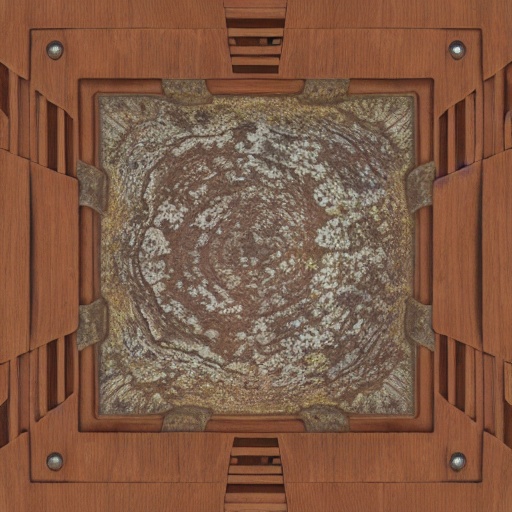}\\

\end{tabular}
\egroup
\caption{Qualitative comparison on fine-tuning ControlNet. We evaluate the pre-trained ControlNet and our fine-tuned version based on the pre-trained one to employ pixel-guidance generation. The textures from pre-trained ControlNet (w/o ft) are more like natural images rather than textures.} 
\label{fig:ablation_pixel} 
\Description{fig:ablation_pixel}
\end{figure}

\subsubsection{Dataset Generation}
\label{ssec:result:dataset}

Our dataset comprises a total of 711 PBR materials, each including four 2$k$ texture maps: albedo, normal, metallic, and roughness, along with corresponding textual labels. The data are sourced from PolyHaven \footnote{\href{https://polyhaven.com/}{https://polyhaven.com/}} and freePBR \footnote{\href{https://freepbr.com/}{https://freepbr.com/}}. We categorized the data into ten types manually: Brick (58), Fabric (60), Ground (99), Leather (45), Metal (130), Organic (45), Plastic (40), Tile (75), Wall (69), and Wood (90).

The input text prompt $\mathcal{T}$ is in the format of ``a PBR material of [type], [name], [tags]" during the finetuning of material-LDM, where `type' refers to the type of material, `name' (title name) and `tags' for each material are given by the website. These tags are randomly retained at a ratio of $30\%-100\%$ during training. To address the issue of uneven distribution in the original data, we selected high-quality and representative data within categories of large volumes and randomly duplicated existing data for categories with smaller data volumes, which helps to balance the sample sizes across all categories, ensuring more uniform training data distribution.

For the 2$k$ textures we obtained, we perform horizontal flipping, vertical flipping, random rotation, and multi-scale cropping and adjust the direction of the normal maps accordingly, eventually resizing them to $512 \times 512$ pixels as our training data. 
After augmenting textures, we render each of them with randomly sampled viewpoints and lightings by \citet{Laine2020diffrast}. The rendering images are also used to train our highlight-aware albedo decoder.

Concerning the paired data for training ControlNet, we utilized Pidinet ~\cite{su2021pixel} to extract sketches from the albedo maps as mentioned in \Cref{sssec:method:PixelControl}.

\subsubsection{Other Details}
DreamPBR was trained on quadruple Nvidia RTX 3090 GPUs. During the training of Material LDM, we employ Adam as our optimizer with a base learning rate of $1.6 \times 10^{-3}$ and closed learning rate scaling. Starting with the stable-diffusion-v1-5 checkpoint for 9000 epochs, we finetuned it for approximately 10 days. For the training of the PBR decoder, we set the base learning rate to $4.5 \times 10^{-6}$ and enabled scale\_lr, taking 4 days total in which the output channels of the decoder were set to 8, with albedo and normal having three channels each, and metallic and roughness being single-channel. For the highlight-aware albedo decoder, we set the base learning rate to $4.5 \times 10^{-6}$ and enabled scale\_lr, taking 2 days total in which the output channels of the decoder were set to 3. We incorporate rendering loss as detailed in Section \Cref{ssec:method:decoder} during the training process above.

During the training of the Rendering-aware super-resolution module, we initially utilized the preset weights from Real-ESRGAN ~\cite{wang2021realesrgan} and finetuned four super-resolution modules specifically for albedo, normal, metallic, and roughness textures. These modules were finetuned using the learning rates of $1 \times 10^{-4}$ and 10000 total iter. Furthermore, we combined the training of all four modules in a model to render the result of each module during training and incorporated rendering loss.

To enhance image control performance, we set the learning rate to $1 \times 10^{-5}$ for training ControlNet, which requires about 2 days to complete. For Style Control, we directly utilize the ip-adapter\_sd15 checkpoint along with our finetuned checkpoint, as we have observed satisfactory results.

\subsection{Generation Results}

DreamPBR is capable of generating realistic or magic materials with only descriptions. To demonstrate the ability to synthesize wide materials,
we obtain a mount of descriptions of materials in by LLM for each type, which is used to sample materials with DreamPBR. The generated textures are enhanced by the super-resolution module and are then rendered as shown in \autoref{fig:PBR}. 
In our sampled 400 textures, they show high consistency with text and the mean of CLIP Score between rendering images and given prompts is \textbf{30.198}.
Besides the consistency of text and images, the diversity of results is quite important for text-driven generative models as well. As demonstrated in \autoref{fig:seed}, we further sample several textures with the same prompt but different random seeds, DreamPBR succeeds in producing diverse textures that follow the descriptions we specify.

\subsubsection{Tileable texture generation}
Although the users would introduce various controls, we can generate seamless tileable textures all the time, which allows users to apply the generated textures in different scales and different scenes. In \autoref{fig:seamless}, we present several tileable textures from direct and guided generation with their splicing results, showing the effectiveness of circular padding in our method as mentioned in \Cref{ssec:method:ldm}.

\subsubsection{Results of Pixel Control}

By finetuning an additional ControlNet, DreamPBR is able to generate textures according to given patterns. In practice, a designer could decide on a pattern in advance, and then try different materials. It may also be the other way around. For those two situations, DreamPBR ensures reasonable textures for certain patterns or materials as demonstrated in \autoref{fig:pixel_control_1} and \autoref{fig:pixel_control_2}.

With additional control of binary images, inpainting is also a usual method for users to obtain specified results so we present several inpainting results in \autoref{fig:inpainting} to replace a region in texture with another region users describe.

\subsubsection{Results of Style Control}
A styled image expresses more easily for a person than only text like \citet{su2021pixel} does. To do so, we evaluate the adaptation of \citet{su2021pixel} for our Style Control. Specifically, we obtain several styled images online and present the generation results under different styles from images as shown in \autoref{fig:style_control}. \autoref{fig:multiModal} illustrates the situation in that users would like to combine Style Control with Pixel Control, which enables users to generate the results they want more freely.

\subsubsection{Results of Shape Control}
% 3D geometric control (dialogue model, TMT, SDS optimization)
With the ability to generate various textures, DreamPBR can be extended to non-planar objects such as chairs. Specifically by giving a segmented object, we can utilize dialogue with a large language model to get different descriptions of each region. For more specified objects, a more direct way is to be in conjunction with cropped areas from exemplar images used with pixel control and style control. Thanks to the tileable features, the results from our pipeline of Shape Control are shown in \autoref{fig:shape}.

\subsection{Comparative Experiments}
Leveraging the state-of-the-art generative model, StableDiffusion, DreamPBR is very competitive with previous methods for materials generation.
We compare the results generated from DreamPBR of different materials against MaterialGAN ~\cite{guo2020materialgan} and TileGen ~\cite{zhou2022tilegen} in \autoref{fig:materialgan_1}.  Notably, there are only two categories provided in the competing methods so our results are generated by giving prompts, ``a PBR material of ground, stone'' and ``a PBR material of metal''. The comparison shows that DreamPBR can generate textures following the distribution of realistic data from datasets like GAN-based methods as well as magic textures from prior information for 2D images.

Moreover, we compare our Pixel Control with those of TileGen in generation with sketches guidance. The comparison results are shown in \autoref{fig:tilegencon}, in which we demonstrate different generation results of TileGen and ours with the same binary masks. DreamPBR surpasses TileGen in sketches-driven generation and shows fewer artifacts and more precise controls than previous research on material generation like TileGen.

\subsection{Ablation Study}

The training of DreamPBR consists of some alternative modules and additional loss functions. In this section, we focus on evaluating the effect of each of the designs. To evaluate them, we randomly selected 100 textures from our obtained data that were not used in the whole training stage.

\subsubsection{PBR Decoder}
When the PBR Decoder is trained, we introduce $\mathcal{L}_\text{render}$ to solve the regression problem from images rendered with random lights and viewpoints, which enforces that the decoded textures are realistic after being rendered. It reduces the search space of output values compared to the one that rendering images is not used.
We trained two PBR decoders with and without $\mathcal{L}_\text{render}$, and evaluated their effectiveness of them by comparing the outputs with reference textures. \autoref{fig:decoder} presents the comparison results, in which our rendering-aware decoder is capable of achieving more realistic results in rendered results and more consistent results in generated textures.

\subsubsection{Super-Resolution Module}

Although the super-resolution models originally show great results in natural images, we finetune it again with our material data and employ a novel rendering loss $\mathcal{L}_\text{render}$ from the level of perception. In practice, we finetune super-resolution modules for each component of textures based on the pre-trained Real-ESRGAN as our baseline. With four single modules(albedo, metallic, normal, and roughness), we jointly finetune them and introduce the $\mathcal{L}_\text{render}$ by rendering four textures after super-resolution to image space.
The comparison results are shown in \autoref{fig:SR}. Similar to the training of PBR Decoder, the finetuning super-reso\-lution modules with $\mathcal{L}_\text{render}$ contributes to better results.

\subsubsection{Highlight-aware decoder}
As mentioned in \Cref{ssec:method:highlight}, we introduce a highlight-aware albedo decoder to remove the potential highlights in generated RGB images.
For a good de-highlight module, there are two key points to be taken into account: 1) effectively removing the highlights in images, and 2) leaving them unchanged for those without highlights. In practice, only training on rendered images potentially affects the decoded albedo(without highlights), so we finetune the highlight-aware decoder by randomly choosing rendered images from different lights or pure albedo maps. Furthermore, we compare the outputs of the highlight-aware decoder with the ones of the initially pre-trained decoder in \autoref{fig:highlight_decoder}, suggesting that our decoder addresses the issues of two key points above.

\subsubsection{Pixel Control}
To realize the sketch-guidance control, we embed a pre-trained ControlNet in DreamPBR. However, different from the IP-Adapter for Style Control focuses on incorporating semantics of images in clip space independent of training data, the initial ControlNet leads to domain shift, from the albedo domain back to the image domain, in our experiments. To address this problem, we finetuned the ControlNet with our sketch-albedo pairs as mentioned above. The comparison of ControlNet before and after being finetuned is shown in \autoref{fig:ablation_pixel}.

\subsection{Limitations}

Despite the promising capabilities of DreamPBR in generating high-quality and diverse material textures, our method encounters certain limitations that merit further exploration and improvement. We employ normal maps to reveal surface details in textures. However, using normal maps without displacement maps leaves self-occlusion ignored when rendering them with those textures, which makes the rendering results unrealistic. In addition, although a more lengthy description contributes to a more detailed texture that the user wants, it is also complex work for users to produce such a detailed description like ``a PBR material of the wall, concrete wall, outdoor, cracked, man-made, rough, painted...".
% !TeX root = ../main.tex
\section{Conclusions and Future work}
\label{sec:conclusions}
In this paper, we propose DreamPBR, a novel diffusion-based generative framework for creating physically-based material textures. Our methods do not rely on large data sets as image generation does but transfer their original prior information to desired textures. Given text descriptions and other optional multi-modal conditions, we can generate textures that are highly consistent with text descriptions and the other conditions such as styles of RGB images and patterns of binary images. By using DreamPBR, one can create planar textures freely according to their imagination. Specifically, we start with finetuning diffusion models for albedo generation and then decompose albedo to other SVBRDFs(normal, metallic, and roughness) by our highlight-aware decoder and PBR Decoder. For higher-resolution textures, we easily introduce an additional loss function in rendering images to our super-resolution module and bring significant improvement visually. With the properties above, DreamPBR can also produce some textures for simple geometries by dialogue with LLM. 

For future work, although DreamPBR currently targets planar textures, it could be extended to complex geometries with further development of retopology. Additionally, because of our effective PBR Decoder and highlight-aware decoder, DreamPBR has the potential to be used in SVBRDF estimation.
Lastly, there are inevitably problems such as limited resolution and time-consuming inference when utilizing diffusion models, which is also a challenging problem in the future.

%%%%%%%%%%%%%%%%%%%%%%%%%%%%%%%%%%%%%%%%%%%%%%%%%%%%%%%%%%%%%%%%%%%
%\nocite{*}
\bibliographystyle{ACM-Reference-Format}
\bibliography{src/reference}

%%%%%%%%%%%%%%%%%%%%%%%%%%%%%%%%%%%%%%%%%%%%%%%%%%%%%%%%%%%%%%%%%%%

%\newpage
%\input{src/7_supplement.tex}

\end{document}